\documentclass[preprint,12pt,authoryear]{elsarticle}
\usepackage[colorlinks,linkcolor=blue]{hyperref}
\usepackage{natbib,lineno,hyperref,amsmath,graphicx,epstopdf,epsfig,multirow,booktabs,subfigure,amsfonts,amssymb}
\modulolinenumbers[5]
\usepackage{multicol}
\journal{Remote Sensing of Environment}
\usepackage{geometry}
\usepackage[ruled]{algorithm2e}
\newcommand{\secref}[1]{Section~\ref{#1}}
\newcommand{\figref}[1]{Fig.~\ref{#1}}
\newcommand{\tabref}[1]{Table~\ref{#1}}
\newcommand{\equref}[1]{Equation~\ref{#1}}

\usepackage{booktabs}
\usepackage{amssymb}
\usepackage{amsmath}
\usepackage{color}
\usepackage[T1]{fontenc}
\usepackage[ruled]{algorithm2e}
\usepackage{graphicx} 
\usepackage[figuresright]{rotating}
\usepackage{makecell}
\usepackage{multirow}
\usepackage[export]{adjustbox}
\usepackage{appendix}
\usepackage{float}
\usepackage{url}
\usepackage{stfloats}

\begin{document}

\begin{frontmatter}



\title{Semi-supervised Road Updating Network (SRUNet): A Deep Learning Method for Road Updating from Remote Sensing Imagery and Historical Vector Maps}


\author[1]{Xin Chen}

\author[1]{Anzhu Yu}

\author[1]{Qun Sun\corref{mycorrespondingauthor}}

\author[1]{Wenyue Guo}

\author[1]{Qing Xu}

\author[1]{Bowei Wen}


\address[1]{PLA Strategic Support Force Information Engineering University, Zhengzhou, 450001, China}

\cortext[mycorrespondingauthor]{Corresponding author at: PLA Strategic Support Force Information Engineering University, Zhengzhou, 450001, China} 

\begin{abstract}

A road is the skeleton of a city and is a fundamental and important geographical component. Currently, many countries have built geo-information databases and gathered large amounts of geographic data. However, with the extensive construction of infrastructure and rapid expansion of cities, automatic updating of road data is imperative to maintain the high quality of current basic geographic information.
Currently, supervised road extraction methods are widely used in full-volume update patterns; however, these require large amounts of annotated data to obtain ideal results and suffer from cross-domain issues, thus limiting their application in large-scale road updating processes. The change-detection task corresponds to the incremental update mode. However, obtaining bi-phase images for the same area is difficult, and complex post-processing methods are required to update the existing databases.
To solve these problems, we proposed a road detection method based on semi-supervised learning (SRUNet) specifically for road-updating applications; in this approach, historical road information was fused with the latest images to directly obtain the latest state of the road.
Considering that the texture of a road is complex, a multi-branch network, named the Map Encoding Branch (MEB) was proposed for representation learning, where the Boundary Enhancement Module (BEM) was used to improve the accuracy of boundary prediction, and the Residual Refinement Module (RRM) was used to optimize the prediction results. 
Further, to fully utilize the limited amount of label information and to enhance the prediction accuracy on unlabeled images, we utilized the mean teacher framework as the basic semi-supervised learning framework and introduced Regional Contrast (ReCo) in our work to improve the model capacity for distinguishing between the characteristics of roads and background elements.
We applied our method to two datasets: the DeepGlobe public dataset and our self-constructed dataset from Zhengzhou and Nanjing. The experimental results showed that SRUNet has clearer edges and more continuous road fragments. The intersection-over-union (IoU) metrics for these two datasets were 75\% and 74\%, respectively; for Zhengzhou and Nanjing, they reached 75.14\% and 72.96\%, respectively., Our model can effectively improve the performance of a model with fewer labels. Overall, the proposed SRUNet can provide stable, up-to-date, and reliable prediction results for a wide range of road renewal tasks.

\end{abstract}




\end{frontmatter}


\section{Introduction}\label{intro}
With the rapid acceleration of urbanization, the population, environment, and infrastructure of urban built-up areas are undergoing dramatic changes\citep{CHENSurvey2022}. Geographic information can provide managers with the status of urban regions and is crucial for geographic information system (GIS)  applications, such as mapping and urban planning\citep{jiao_survey_2021}. The road network is one of the most diverse and critical components of geospatial information. The road network describes a city’s framework and supports navigation studies. Therefore, detection has gained considerable attention over the past decade. 

Because  a road network is a core geographic information element, automatic road detection and update methods are urgently needed\citep{zu_roadrouter_2020}. Various data are utilized in the process of road updating. Frequently utilized data include the GPS trajectory and volunteer information applications, such as OpenStreetMap (OSM). However, owing to the heterogeneity and instability of open-source data, a complicated update process is required to obtain reliable data\citep{Chen2022GANetAG}. Considering the rapid growth of remote sensing, fast-updated, high-resolution aerial imagery has shown remarkable promise for road detection\citep{LuCross-domain2021296}.

For road-updating using remote sensing images, two common patterns are generally used: full-volume update and incremental update. They are mainly implemented separately using road extraction  and change-detection techniques. Both tasks can be considered binary semantic-segmentation problems at the pixel level. Where, 1) the road surface extraction method concentrates on accurately separating roads from non-road regions, and trains a road extraction model using historical map data and pre-change images. After the pretraining process, the model is finetuned with post-change images to generate updated results \citep{Ghaffarian2019PostDisasterBD}. 2) Change detection attempts to detect changed areas by comparing bi-temporal aerial images or the latest aerial images with a geographic database, which requires many labeled images to complete the training process. 

Inspired by the idea of previous research, We hope to construct a method to update the historical states based on the latest images, requiring as little manual work as possible. Road updating is our goal, and semi-supervision is a technical method.From this perspective, our work is similar to road extraction, which directly obtains road surface information from the latest images based on historical road data. However, unlike the supervised semantic segmentation method, our work attempts to construct a strong model with only a small number of manual annotations to improve the automation level and generality of the method in practical applications. Unlike the change detection task, our work does not extract the change areas but directly obtains accurate prediction results for the entire learning area based on unreliable labels; thus, the results are accurate, complete, and easy to process in practical use.

There are two main challenges in our work: one is designing a road detection network with a good generalization performance, and the other is making full use of historical information and limited high-precision labeled data.

The first challenge essentially involves designing a road extraction model. Road extraction methods can be divided into traditional and deep learning-based methods. Traditional methods are primarily algorithm-driven and rely on the design of road features. Key points and line segments, which are crucial parts of road networks, are extracted from imagery using various road features such as geometry, topology, and texture. In general, the performance of traditional methods relies heavily on the effectiveness and robustness of the designed feature descriptor. However, limited conditions are considered for handcrafted features, which restricts the use of conventional techniques. Road detection research has benefited greatly from the use of deep learning techniques. Based on the output format, these methods can be classified into two types: road-area extraction and road-graph extraction. Convolutional neural network (CNN)-based methods are typical road extraction methods focusing on the classification of road and non-road pixels. \cite{zhang_road_2018} proposed a u-net-based architecture to extract the road binary map. Many studies have made improvements on this basis, such as enlarging the receptive field, introducing multiscale features, and adding geometric supervision information. \cite{deng_spd-linknet_2021} and \cite{tan_remote_2021} integrated strip pooling into the model, to capture the long-distance dependent information of road features. DFANet has a feature aggregation module, which fuses shallow, intermediate, and deep features, embedded into its backbone \citep{WangHigh-resolution2021}. Many knowledge-based methods, including geometry, structure, and multisource auxiliary knowledge, have been proposed for combining the available knowledge prior to the learning process. The ScRoadExtractor combines both the buffer properties of road areas and the texture information of superpixels. Using only scribble labels,  ScRoadExtractor achieves a state-of-the-art (SOTA) performance \citep{wei_scribble-based_2022}. \cite{Girard2021PolygonalBE} designed a frame field from the perspective of the loss function and used edge information as a constraint to obtain the precise boundary information. Road graph extraction attempts to construct road networks with vertices and lines that consider the location of key points and road structure, subsequently maintaining better topology consistency. VecRoad is a classical point-based iterative graph exploration scheme that can enhance road connectivity and maintain precise alignment between the graph and a real road \citep{Tan2020VecRoadPI}. Sat2Graph combines the advantages of CNN and graph-based methods into a unified framework and proposes a novel encoding scheme \citep{He2020Sat2GraphRG}. RNGDet proposes a novel approach based on imitation learning and produces a network graph directly \citep{Xu2022RNGDetRN}. Deep learning methods demonstrate that the data-driven strategy produces SOTA results and is appropriate for various remote-sensing applications. 
However, most of the existing road extraction techniques use a supervised architecture, thus making them susceptible to cross-domain problems and deficient in highly accurate labeled data.

The second challenge is that data-driven methods depend on the number of samples for learning abundant representations, which hampers their performance when faced with fewer labels or new situations. For instance, the data distribution of regions to be updated is typically different from that of the training samples, leading to a mis-classification problem. Additionally, insufficient labeled data increase the difficulty of the model in distinguishing different category pixels, thereby reducing the ’capacity of the model for representation. To address this issue, a semi-supervised learning approach was developed. As a branch of machine learning that aims to combine supervised and unsupervised learning, semi-supervised learning methods attempt to improve performance by associating labeled and unlabeled information. 
There are three strategies for semi-supervised learning to solve the problems of consistency regularization: pseudo-labeling and contrastive learning \citep{Wu2013SemisupervisedDA}. Consistency regulation assumes that data disturbances do not influence the position of decision boundaries and establish a consistency loss function to unify the predictions of the original and augmented data. Pseudo labeling tries to iteratively amplify a labeled dataset by adding high-confidence pseudo-labels to the training dataset. Contrastive learning constructs contrastive pairs to guide the model, focusing on typical characteristics.
The outperformance of contrastive learning in downstream tasks has recently attracted considerable attention. Early studies mainly combined consistency regularization and pseudo-labeling to enrich samples, which showed limited performance on unlabeled data. Examples include Mean Teachers \citep{Tarvainen2017MeanTA}, mix matches \citep{Berthelot2019MixMatchAH}, fix matches \citep{Sohn2020FixMatchSS}, and flex matches \citep{Zhang2021FlexMatchBS}, etc. Recently, some studies have attempted to fully utilize the unlabeled data via contrastive learning. The memory bank retains the representations of contrastive pairs to approach the distribution of all samples. MoCo adds a dynamic update queue to further improve the memory bank \citep{He2019MomentumCF}. The sampling process is then replaced with a momentum-update mechanism. Although, SimCLR offers new ideas for creating negative samples from a unique perspective, further highlighting the value of data transformations \citep{Liu2021SimCLSAS}; these techniques focused more on image classification problems than on semantic segmentation. Regional Contrast (ReCo) \citep{Liu2021BootstrappingSS} addressed this issue by proposing a representation method for each category at the pixel level. By focusing more on high-frequency regions with active query sampling, the model can learn more disentangled features and obtain clearer object boundaries. ReCo achieved a SOTA performance on various segmentation datasets and was superior to other methods in terms of computational efficiency. We borrowed the idea of ReCo and made it more suitable for remote sensing images with some modifications.

In this study, we present a road-updating architecture. To cope with the insufficiency of annotated data, we explored a semi-supervised deep learning technique for road detection, which can fully utilize the few accurate labels provided. We recommend the use of geometry information and a light-refinement module to enhance the accuracy and comprehensiveness of the results. Using only up-to-date remote-sensing images and a historical database, our updating method can produce up-to-date road information in a self-learning manner. The main contributions of this study are as follows:
\begin{enumerate}[1)]
\item A semi-supervised road updating method is proposed. The overall updating procedure only requires the most recent imagery and a minimal amount of labeling samples, making it a highly automatic architecture requiring little human involvement. We introduce ReCo to the training procedure, which combines pixel-level contrastive learning with semi-supervised learning. This enhances the representational capability of the encoder and makes our model robust to various circumstances. Our approach is fairly simple to implement and produces reasonable results with few labels.

\item With the help of the historical map patch and spatial refine module, the network can preserve location details and generate a more comprehensive output. The entire architecture, comprising a prediction model and an auxiliary teacher model, has a mean-teacher-like structure. Two strategies are designed for the prediction network. First, the dual-branch encoder encodes the map patch and image patch simultaneously and fuses the multi-source feature map with a spatial and channel attention module. Second, an extra module refines the prediction result by learning the residual details. In this manner, our method shows better performance in boundary areas and has better completeness.

\item In this paper, abundant experiments have been conducted on two datasets: the public dataset DeepGlobe, and a self-constructed large-scale urban road dataset. The experimental results indicate that the proposed method is superior to some SOTA methods, and that the IoU indicator reaches approximately 75\% on the self-constructed dataset. Further, the visualization results show that our proposed method can better deal with details and maintain the integrity of the road. These findings indicate that our method can be well applied for practical large-scale urban road datasets.
\end{enumerate}

\section{Methodology}
\begin{figure*}[!htb]
	\centering
		\includegraphics[scale = 0.25]{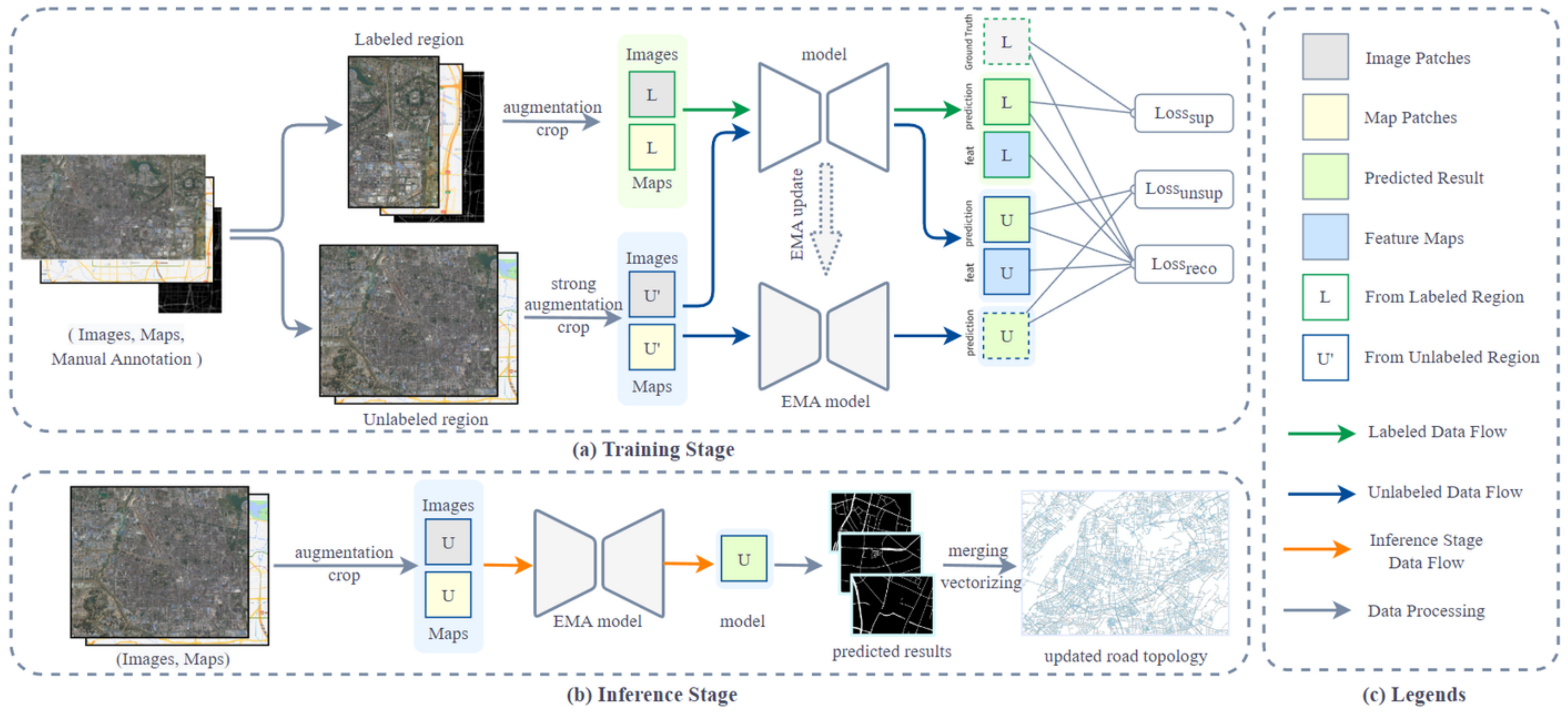}
	  \caption{Flowchart of the proposed road updating architecture. (a) shows the training stage, where labeled data flow is marked in green, and unlabeled data flow is marked in blue. (b) shows the inference stage, where the test data flow is marked in yellow. A map is a three-channel image, and can be obtained from any source (e.g., OSM, Google Map etc.).} \label{fig:overall}
\end{figure*}

This section provides a detailed introduction to the proposed road update framework and its implementation, including the overall framework, network model, and training. \figref{fig:overall} illustrates the proposed road update model framework in this paper, which mainly comprises a semi-supervised learning paradigm, a prediction network and an objective loss function. Both labeled and unlabeled data are trained in an end-to-end framework, and the model can obtain prediction results directly from unlabeled data.

\subsection{Overall architecture\label{2.1}}
The problem of road updating can be described as that in obtaining reliable predictions over an entire image with limited manual annotation, given the new time-phased imagery and historical mapping data. In other words, full use of historical mapping data and manual annotation information is necessary to improve the ability of a model to detect roads.

This study proposes a new road update model. This model uses new temporal remote sensing images and historical map data as data sources, partial or small amounts of annotated information as supervised signals, and learns roads in unlabeled images based on the semi-supervised learning framework of Mean Teacher \citep{Tarvainen2017MeanTA}. The training and prediction processes of the model are shown in Figure 1, where semi-supervised learning is mainly based on Mean Teacher, and the prediction model is mainly based on DeeplabV3. The proposed road-update framework comprises three main parts.

First, a new framework for updating roads is constructed . Using Mean Teacher semi-supervised learning as the basic framework for training, a small amount of labeled data is passed through the student network to obtain the model output and calculate the supervised learning losses. Unlabeled data are used to calculate consistency losses using the teacher network prediction results as labels for the student network prediction results. The student and teacher models share a network structure, and the teacher model weight is updated through the empirical moving average (EMA), which allows the model to effectively use information from the training process to improve model stability and prediction performance. The training and prediction methods for the model are explained in detail in \secref{2.2}.

Second, a new road extraction network is proposed. The image and historical map features are jointly learned using a two-branch structure to aggregate information from multiple data sources. Simultaneously, an additional edge-enhancement module is added to the backbone network to enhance the ability of the model to perceive edge details. The details of network implementation are explained in detail in \secref{2.3}.

Finally, this paper introduces the Regional Contrast (ReCo) loss function to further improve the  focus and classification ability of the model for road categories. By performing contrastive learning on confusing pixels, ReCo can alleviate degradation of the model performance caused by the problem of imbalance in foreground and background pixels to a certain extent. By actively sampling instead of calculating all pixels, ReCo can also enable the model to obtain better representation capabilities while improving the prediction efficiency. The objective function during the training process is detailed in \secref{2.4}.

\subsection{The simi-supervised road updating scheme }\label{2.2}
To solve the two problems of making full use of historical map data from existing databases and making the full use of the small amounts of manual annotation, semi-supervised learning is considered to combine the ideas of supervised learning and unsupervised learning, which can establish a classifier with strong representation ability and generalization performance. Thus, it is suitable for the actual data update process, where the amount of unlabeled image data exceeds the labeled training data, and for label acquisition situations where the labor cost is high.

Some scholars use historical mapping data directly as a supervised signal to train the road extraction model, but this is not applicable when the level of detail and accuracy of the historical mapping data and image representation differ significantly, and the wrong supervised signal degrades the model performance. In this paper, historical mapping information is regarded as a description of different perspectives of the same area, which can provide some auxiliary information for image feature learning and interpretation; therefore, we use historical mapping information as input data in the prediction model together with images. Notably, maps are available from various sources, not only directly from map tiles but also from vector data with rasterization. This provides flexibility and ease in obtaining historical maps.

For labeled information, the prediction model is trained using a supervised signal, because manual labelling is accurate, reliable, and provides accurate category information. To improve the generalization ability of the model for unlabeled data, a semi-supervised learning paradigm was used to predict the pseudo-labels for unlabeled data based on the distribution of labeled data features. Considering the smoothing assumption that data perturbations should not change the model prediction results, a consistency loss is constructed on unlabeled data to enhance the generalization ability of the model on unlabeled data. Based on the clustering assumption that samples of the same category are as close as possible and that samples of different categories are as distant as possible, contrast learning is introduced to construct a contrast learning loss on different feature categories to improve the  ability of the model to extract and classify the features of different categories.

As shown in \figref{fig:overall}, we present a road-update model based on the mean-teacher semi-supervised learning framework. The Mean Teacher model generates both noisy student and stable teacher model predictions and uses a sliding average method to update the teacher network using the student network. In this study, the student and teacher models share the same prediction network, namely, the road extraction network.

\textbf{Training Phase}: To begin with, the data of the area to be updated are obtained and divided into two groups: a small area with manual tagging as labeled data $L=\{L_{img_i}, L_{map_i}, L_{lab_i}\}_{i=1}^N$ and a large area without manual tagging as unlabeled data $U=\{U_{img_i}, U_{map_i}\}_{i=1}^N$, where $L_{img_i}, L_{map_i}, L_{lab_i}$ are the image blocks, map blocks, and real labels corresponding to the labeled data, and $U_{img_i}, U_{map_i}$ are the image blocks and map blocks corresponding to unlabeled data, respectively. 
First, for the labeled data stream branch, the image $L_{img}$ and map blocks $L_{map}$ are input into the student network. The prediction results and pixel feature vectors are the two outputs, and the prediction results are then used with true labels to calculate the supervised loss $Loss_{sup}$. 
Second, for the unlabeled data branch, the unlabeled data are first fed into the teacher model, and the output is used as a pseudolabel. Subsequently, the unlabeled data are perturbed, fed into the student network, and the pseudo-label and student network predictions are used to calculate unlabeled loss $Loss_{unsup}$. 
Next, to calculate the contrast loss of different categories, the ReCo \citep{Liu2021BootstrappingSS} loss function is introduced, and the proposed active sampling strategy is used to calculate the contrastive learning loss for difficult pixels in the labeled and unlabeled data. ReCo loss is described in detail in \secref{2.4}.
Finally, back-propagation updates the student model weights and teacher model weights using the EMA method, that is, $W'_T = decay \times W_T +(1-decay)\times W_S$  where $W_T$ and $W'_T$ are the teacher model weights before and after the update, respectively, and $W_S$  is the student model weight for the current iteration. The network weights are iteratively updated until the model converges, and the weights reaching the highest iteration of IoU on the test set are saved.

\textbf{Inference Phase}: \figref{fig:overall} (b) shows the flowchart of the inference phase. To obtain the model prediction results, the images are chunked, fed into the prediction network, and stitched together to obtain road detection results for the entire image. In practical applications, the newest imagery constitutes the unlabeled subset, and a small number of manual annotations constitute the labeled subset. The prediction results are progressively optimized with the iterations and learning of a few epochs. After vectorization and consistency processing, the updated data are stored in a geodatabase.

\begin{algorithm} 
    \renewcommand{\thealgocf}{}
    \caption{\texttt{Training phase of our proposed SRUNet framework }}
    \KwIn{Annotated samples $\{L_{img_i}, L_{map_i}, L_{lab_i}\}^N_{i=1}$ and unlabeled samples $\{U_{img_i}, U_{img_i}\}^N_{i=1}$}
    \textbf{Models and Parameters}: student network $M_S$, teacher network $M_T$. \\
    \For {$n \leftarrow 1$ $\mathbf{to}$ $T$}
    {            
        \textbf{Get a batch of training samples:}  $\{L_{img_i}, L_{map_i}, L_{lab_i}\}^N_{i=1}$  and  $\{U_{img_i}, U_{img_i}\}^N_{i=1}$\\
        $U_{pred} = M_T(U_{img_i}, U_{map_i})$ \\
        \textbf{Augmentation: } $U'_{img_i}, U'_{map_i} = aug (U_{img_i}, U_{map_i})$ \\
        $L_{pred}, L_{cls} = M_T(L_{img_i}, L_{map_i})$ \\
        $U'_{pred}, U'_{cls} = M_T(U'_{img_i}, U'_{map_i})$ \\
        \textbf{Calculate the training loss: } $Loss_{sup}, Loss_{unsup}, Loss_{str}$ \\
        \textbf{Update:} Update $M_S$ with Adam optimizer,  Update $M_T$ with EMA moethod: $W'_T = decay \times W_T +(1-decay)\times W_S$ \\
    }
\end{algorithm}
\subsection{SRUNet: a multi-branch road extraction network\label{2.3}}
\begin{figure*}[htb]
	\centering
		\includegraphics[scale = 0.3]{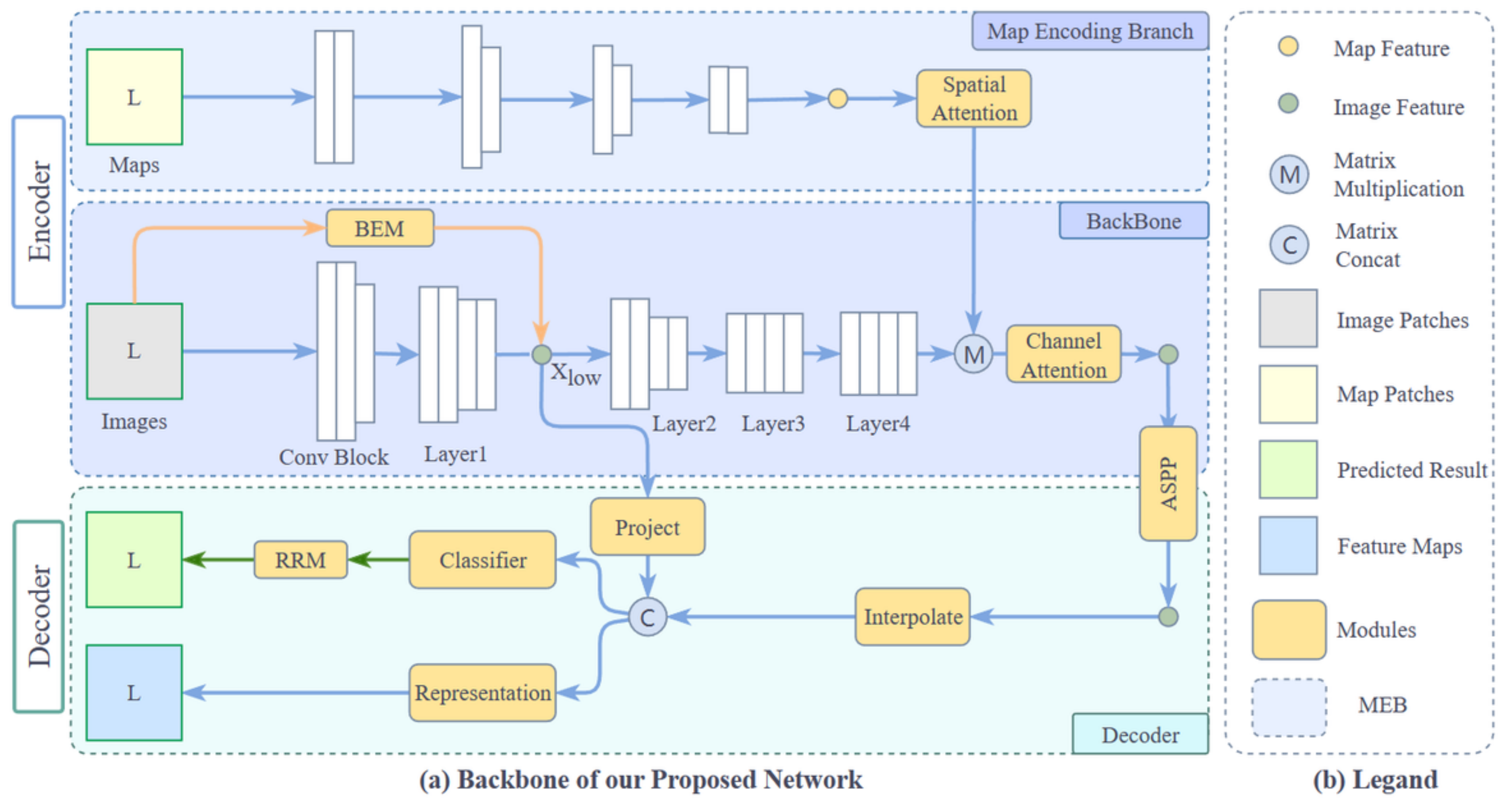}
	  \caption{The segmentation network of SRUNet. The blue line represents the data flow of the backbone, and the yellow and green lines represent the flow of BEM and RRM, respectively. }\label{fig:backbone}
\end{figure*}
Road networks exhibit typical features in remote sensing images. In terms of geometry, roads are long, thin, and continuous. In terms of the number of pixels, they occupy a relatively small portion of the entire image, and in terms of textural features, they are easily obscured by their surroundings. Therefore, unlike semantic segmentation problems, road extraction requires special attention to short- and long-distance relationships. To address this problem, the Map Encoding Branch (MEB) is added to the baseline DeeplabV3 to form the backbone network, and two lightweight modules, namely, the Boundary Enhancement Module (BEM) and Residual Refinement Module (RRM), are designed to help recover spatial details, as shown in \figref{fig:backbone}, where the blue lines represent the backbone of the network and the yellow and green lines represent BEM and RRM, respectively. The two coding branches of the backbone network extract multi-scale features from images and map blocks and use a simple decoding structure to restore spatial details and generate prediction results. The BEM is a lightweight feature-enhancement module. RRM uses a U-shaped shallow convolution to refine the 2-channel prediction results. The details of the implementation are presented in this section.
\subsubsection{Basic Backbone}\label{2.3.1}
Existing research indicates that expanding the convolutional kernel field and introducing multiscale features can improve the continuity and completeness of road extraction results. DeeplabV3, a classic fully convolutional network (FCN)-based semantic segmentation model, uses deep CNN (DCNN) with empty convolutions as an encoder and pyramid pooling modules to further extract multi-scale information, and obtain SOTA results on multiple publicly available semantic segmentation datasets. Therefore, DeeplabV3 has good multiscale feature extraction capabilities. Here, it is used as a base model and part of the model is migrated to form backbones.

\textbf{Backbone}:The backbone is mainly responsible for extracting multi-level features and providing outputs of the same prediction results as those in the original dimension. According to the general semantic segmentation process, a network can be divided into three stages: encoding, center bridge, and decoding. The encoding stage is a dual-branch structure that encodes image features and map-block information. The center bridge stage is the bridge between the encoding and decoding parts, including dual-branch feature fusion and further extraction of multiscale features. The decoding stage recovers spatial details from the high-dimensional feature map to the original spatial resolution through two upsampling and one convolution processes.

\textbf{Encoding stage}: ResNet can eliminate the network degradation caused by increasing network depth; the residual module further improves the feature extraction capability of the model, thus the pretrained ResNet-101 is transferred to our model. The feature encoding part of this study can be divided into two branches: image-encoding and map-encoding. The image-encoding branch comprises one convolution block and four stacked residential convolution layers. The convolution block comprises the first layer of ResNet-101 without the AvgPooling and Fully Connected parts. The other four residential convolution layers correspond to layers 1-4 of ResNet-101. The last layer is changed to a dilated convolution for further expanding the receptive field of the network. After the first layer, the image block becomes the $\frac{1}{4}$ size of the original image. After the remaining four layers, the feature map is $\frac{1}{4}, \frac{1}{8}, \frac{1}{16}, \frac{1}{16}$ of the original image size. For the map-encoding branch, considering the relatively low complexity of the map features, four shallow convolution layers are used to encode the features, where the first three layers have the same parameters: kernel size = 7 and stride = 2. The parameters of the fourth convolutional layer are kernel = 3 and stride = 2. With the processing of each layer, the image size is reduced by half, the final number of channels is 64, and the size of the original map is $\frac{1}{16}$ of the feature map $x_e^4$.

\textbf{Center bridge stage}: As a bridge between encoding and decoding, the center bridge mainly includes two parts: feature map fusion and atrous spatial pyramid pooling (ASPP). The former aims to eliminate the semantic gap between image and map features, whereas the latter can further capture multiscale information from fused features. For feature map fusion, a spatial attention mechanism is introduced to enhance the road areas in the map branches, considering the relatively low interference information in the map blocks and the more prominent road geometry features. Drawing on the spatial attention module of CBAM, the feature map $M_e^4$ is used to calculate the spatial attention weights. After multiplying the image feature map $B_e^4$ with the spatial attention matrix, the input channel attention module further aggregates the multichannel information to obtain the fused features $F_1$. The ASPP module is then used to capture multiscale information of the fused features. As a larger sampling rate does not yield better results, the dilation rates of the three cavity convolutions in the ASPP module are set to 6, 12, and 18, with reference to ReCo.

\textbf{Decoding stage}: Similar to DeepLabV3, the decoder used in this paper first upsampling the feature map, and the size of the sampled feature map is restored to $\frac{1}{4}$ that of the original map. The sampled feature map is then restored to its original size. The sampled feature map is restored to the size of the original map, and the low-level feature outputs from the first layer of the decoder are stitched together by the channel and passed through a $3\times3$ convolutional classifier and a residual refinement module. Finally, the features are restored to the original resolution after upsampling four times. Considering that directly stitching the low-level features $x_{low}$ with higher-level features may lead to data redundancy and yield negative results, the low-level features are processed in advance using a projection module with $1\times1$ convolution. Further, the representation module, which has a structure similar to that of the classifier, is used to generate feature maps with $\frac{1}{4}$ the original image size and channel = 256. The feature maps are then used to calculate the ReCo loss. The ReCo loss is discussed in detail in \secref{2.4}.

\subsubsection{Boundary Enhancement Module (BEM)}\label{2.3.2}
In road extraction tasks, the interior of the road elements is homogeneous and has an obvious geometric demarcation from the background. Therefore, low-level geometric feature information including nodes, lines, and edges provides important information for distinguishing roads from the background. Although a high-resolution remote sensing image is complex in texture and rich in low-level feature information, geometric features are easily submerged in complex features. Therefore, in this paper, BEM is proposed to extract and enhance the edge information in an image for this important geometric feature of the edge.

BEM is a lightweight feature extraction and encoding module located between the input image and the second layer of the backbone. BEM consists of two main components, a pretrained edge extraction network HED and a lightweight convolutional block. HED is an efficient multiscale edge detection network that continuously integrates and learns to obtain image-to-image edge-prediction maps of the input image. The image block is processed using the pretrained HED model to output a single-channel edge prediction map of the original image size. The values are between 0 and 1, with higher values indicating closeness to the edge. Next, the edge prediction map passes through an encoding layer composed of three convolution operations, and outputs a feature map of $\frac{1}{4}$ as the original image size. Subsequently, the edge feature map is concatenated with the backbone network features using a channel and input into the convolution module for fusion. To fully capture multiscale features, the convolution operation in the convolution module is replaced with MixConv, which combines convolution kernels of multiple sizes in a single unit, thus efficiently aggregating features of different resolutions. The fused multiscale feature map contains both the image texture and edge features.

\subsubsection{Residual Refinement Module (RRM)}\label{2.3.3}
As the convolution and pooling processes lose spatial detail information, the outputs of the decoder are coarse, and the edge regions are not sufficiently continuous and smooth. Therefore, an RRM is designed to refine the prediction results.

The RRM is a lightweight U-shaped module that can be easily integrated into any deep learning model. RRM is located after the classifier and takes the prediction results as input to outputs 2-channel prediction map of the same size. After passing through the classifier, the feature map passes through two convolution and pooling layers. After each pooling step, the size of the feature map is reduced by half. Instead of bilinear interpolation, RRM uses deconvolution to recover spatial resolution. The feature map sequentially undergoes two deconvolution processes consisting of deconvolution, BN, and activation layers in that order, to obtain a refined feature map. The refined feature map is regarded as the residual of the coarse prediction map $x_{coarse}$. The final prediction map is obtained by summing the coarse and refined predictions.

\subsection{Objective function}\label{2.4}
Three main types of objective functions are used to train  the model: supervised loss $Loss_{sup}$, unsupervised loss $Loss_{unsup}$, and consistency loss $Loss_{ctr}$. The overall training loss is a linear combination of these three loss functions, as shown in \equref{equ1}.
\begin{align}
L = Loss_{sup}+\alpha_{unsup}\times Loss_{unsup}+\alpha_{ctr}\times Loss_{ctr}  \label{equ1}
\end{align}
where $\alpha_{unsup}$ and $\alpha_{ctr}$ balance three loss functions within an order of magnitude ($\alpha_{unsup}$ was set to 0.7, and $\alpha_{ctr}$ was set to 0.2 ).

For supervised loss, BCEloss was used to measure the difference between the model prediction results and labels to improve the representation ability of the model. For unsupervised losses, the goal was to improve the generalization ability of unlabeled images. Following the existing semi-supervised segmentation method, the mean teacher model was used to generate pseudo-labels from unlabeled images, and the student model in the current epoch was used to generate prediction results on the augmented unlabeled images. The same loss function, BCELoss, was used to calculate $Loss_{unsup}$. However, in contrast to $Loss_{sup}$, $Loss_{unsup}$ operates only in regions where the confidence level of the prediction result is greater than $\delta _u$ ($\delta _u$ is set to 0.95). This reduces the interference of low-confidence regions in the training process.
\begin{align}
Loss_{sup}(P_l,Y_l)=&-\sum_{i=1}^{W}\sum_{j=1}^{H}[y_{ij}\cdot log(p_{ij}) \notag \\ &+(1-y_{ij})\cdot log(1-p_{ij})] \label{equ2} \\
Loss_{unsup}(P_u,Y_u)=&-\sum_{i=1}^{W}\sum_{j=1}^{H}[y_{ij}\cdot log(p_{ij}) \notag \\ &+(1-y_{ij})\cdot log(1-p_{ij})]\cdot w_{ij} \label{equ3}
\end{align}

We adopted a contrastive loss called ReCo loss to capture the invariant and specific information of the road category. Based on cluster assumption, the representations of positive pairs should be pulled closer, whereas those of negative pairs should be pulled apart. (The positive pairs indicate that the pixels belong to the same category). ReCo is a pixel-level contrast loss calculated across all available semantic categories in each minibatch. There are three elements in ReCo: queries $R_q^c$, positive keys $R_k^{(c,+)}$, and negative keys $R_k^{(c,-)}$. $R_q^c$ is sampled using the strategy described by \cite{Liu2021BootstrappingSS}. Given a query $r_q$ from $R_q^c$, taking the mean representation of the pixel category as the positive key $r_k^{(c,+)}$ and the mean representation of the other category as the negative key $r_k^{(c,-)}$, ReCo measures the distance between keys $r_k^{(c,+)}$ and $r_k^{(c,-)}$ and queries $r_q$ based on the NCE loss. This loss can be defined as:
\begin{align}
Loss_{ctr} &= \\ \notag
\sum_{c \in C}^{}\sum_{r_q \in R_q^c}^{}&-log\frac{(exp(r_q\cdot r_k^{(c,+)}/\tau ))}{(exp(r_q\cdot r_k^{(c,+)}/\tau )+\sum_{r_k^- \in R_k^c}exp(r_q\cdot r_k^{(c,-)}/\tau )')} \label{equ4}
\end{align}

\section{Experimental results and analysis}\label{3}
\subsection{Datasets}\label{3.1}
 \begin{figure*}[htb]
\small
\centering
		\begin{tabular}{cc}
 \includegraphics[width=0.42\textwidth]{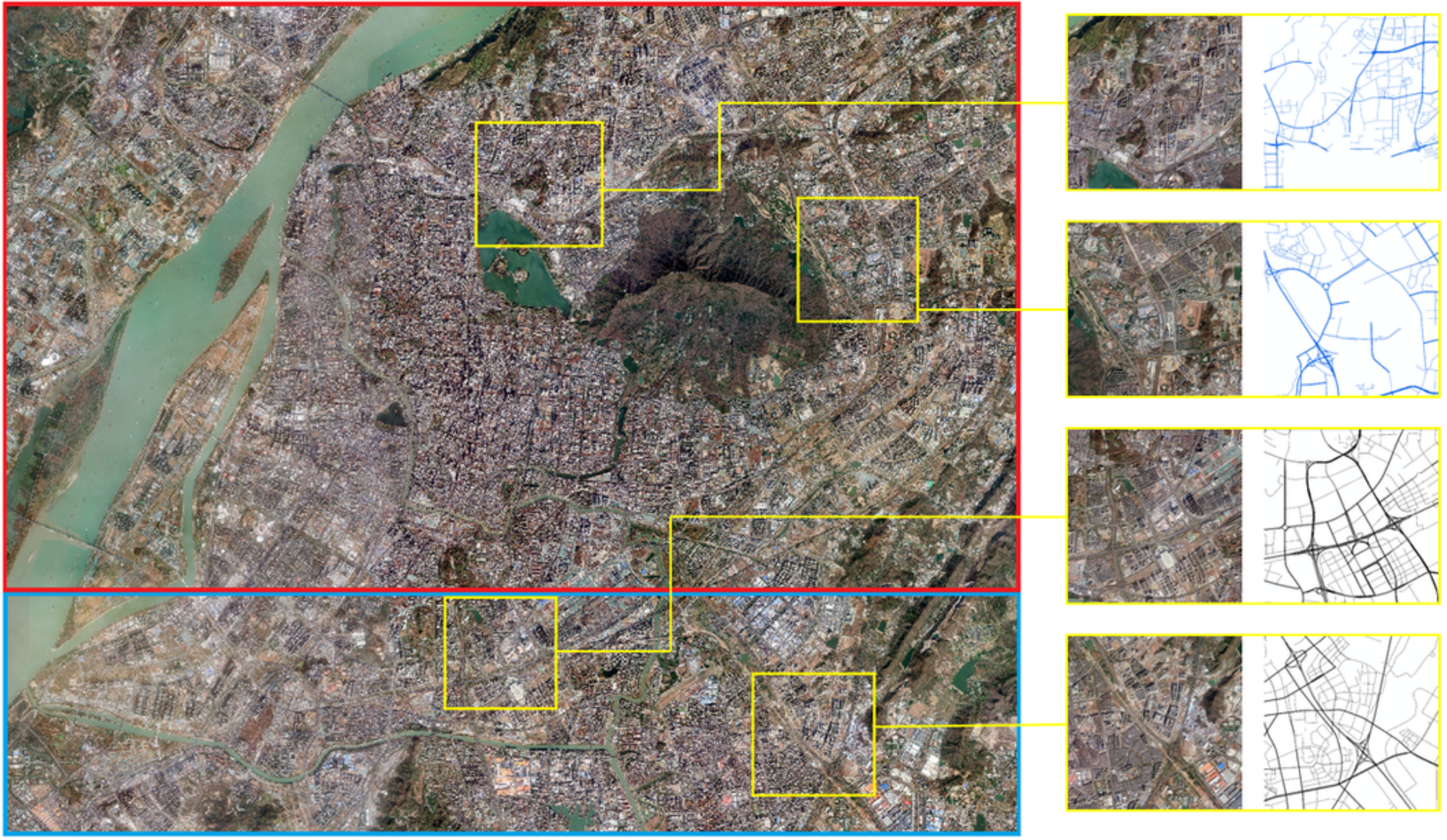}
			& 
\includegraphics[width=0.55\textwidth]{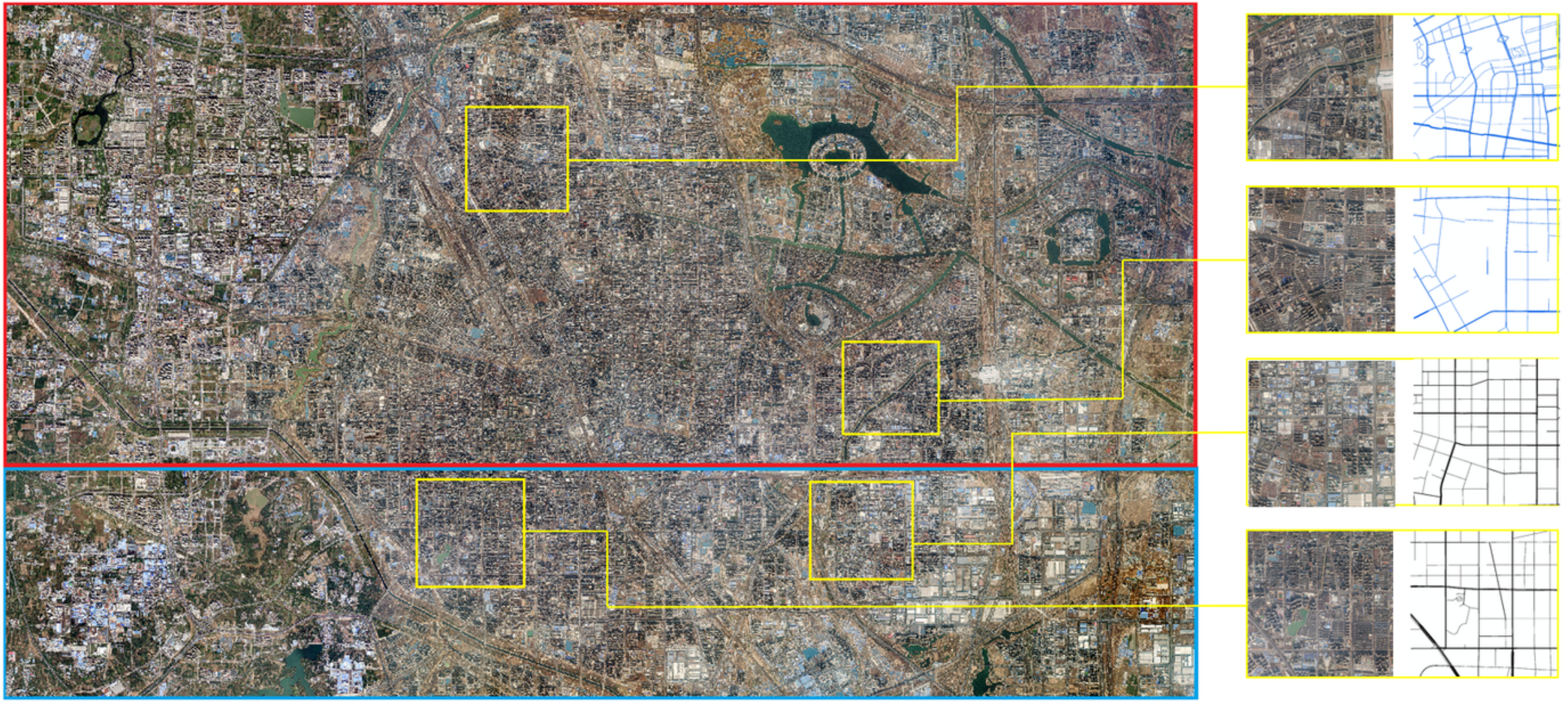}
		\\
118.647°E-118.937°E  , 31.935°N-32.137°N     &  113.480°E-113.835°E  , 34.680°N-34.850°N\\
area=853.3516$km^2$ &    area=910.1649$km^2$ \\
\textbf{(a)Nan Jing, Jiangsu, China} & \textbf{(b)Zheng Zhou, Henan, China}
\end{tabular}%
\caption{Overview of the self-constructed dataset. Labeled areas are marked in blue and unlabeled areas are marked in red. Detailed images are displayed for some areas, where labels in black indicate the corrected OSM roads and labels in blue indicate roads from the historical database. }
  \label{fig:Data} 
\end{figure*}

Two datasets were chosen for our experiments: the public DeepGlobe dataset and an experimental dataset constructed by us, covering 853.3516$km^2$ and 910.1649$km^2$ in Nanjing and Zhengzhou.

The DeepGlobe dataset was released in 2018. Images were sampled from the Digital Globe+Vivid Images dataset and have the GSD value of 50 cm/pixel. The DeepGlobe dataset covers Thailand, Indonesia, and India; the original geotiff images of the three regions have the same size, that is $19584\times19584$ pixels. In our experiments, 6226 images with a size of $512\times512$, spanning 1632 $km^2$, were randomly split into training and validation subsets in an 8:2 ratio. For the semi-supervised learning process, we randomly divided the training dataset into labeled and unlabeled subsets according to the parameter $\alpha_{lab-ratio}$, which represents the ratio of labeled data to all training data. For map patches, we generated a prior map based on pixel-level labeled data. The ground truth labels were randomly erased based on the parameter $\alpha_{masked-ratio}$, and incomplete data were regarded as part of the historical map. A higher $\alpha_{masked-ratio}$ value indicated that more road labels were erased and that the simulated historical map was less accurate. The simulated historical map was then transferred to the GaoDe map style, which consisted of three channels.

The second dataset was a self-constructed dataset covering two areas in Nanjing and Zhengzhou (as shown in \figref{fig:Data}). The urban morphology of the two areas is diverse. The road network of Zhengzhou is a typical square type, whereas Nanjing is more complex and comprises various types. This makes the dataset diverse and rich, meeting the needs for testing the robustness of our model. 
For labeled data, we manually edited the road labels based on data downloaded from OSM, which contained 6.90\% positive and 93.10\% negative pixels. 
For historical data, we downloaded the SHP file from OSM (https://www.openstreetmap.org/) and manually edited the data to exaggerate the changed areas.
A 1:250,000 scale SHP file (produced in 2017) from the National Geomatics Center of China contains 6.08\% positive and 93.92\% negative pixels. We divided the images in a ratio of 7:3 into labeled training data and unlabeled prediction data. We then optimized our road updating model on the training subset, calculated the metric values on the unlabeled subset, and reported the predicted results on the unlabeled subset.

Given the spatial resolution of the images and the general width of roads, roads were divided into primary and secondary classes, where the masks of the primary road had a buffer of seven pixels and those of the secondary road had a buffer of four pixels.

\subsection{Evaluation Metrics}
Four metrics widely used in segmentation tasks were used to evaluate the performances of the proposed and comparison methods. These included the 
intersection over union (IoU), precision (P), recall (R), and F1 scores These metrics are defined as follows:
\begin{align}
IoU &= \frac{TP}{TP+FN+FP}\times 100\%\\
P &= \frac{TP}{TP+FP}\times 100\%\\
R &= \frac{TP}{TP+FN}\times 100\%\\
F1 &= 2\times \frac{P\times R}{P+R}\times 100\%
\end{align}
where true positive (TP), true negative (TN), false true (FT), and false negative (FN) are the numbers of pixels being classified correctly or not. For example, TP refers to the number of correctly classified object pixels. TN is the number of pixels correctly classified as background pixels. FT refers to pixels that are incorrectly classified as object pixels. FN refers to the number of pixels incorrectly classified as background pixels.

\subsection{Implementation details}
Our experiments were implemented using Pytorch 1.11.0, on two Nvidia A100 (40GB) graphics cards. The proposed network was trained in a semi-supervised manner. with a learning rate of 0.001, momentum of 0.9, and weight decay of 0.0005. Inspired by the work using ReCo, a polynomial annealing policy was adopted to schedule the learning rate, which helped the network better converge to the optimal solution. Both the DeepGlobe and self-constructed datasets were trained for 40 epochs, and all experiments were conducted under the same experimental conditions. The input images were $512\times512$ pixels and were augmented using Gaussian blur, color jittering, and random horizontal and vertical flips. For the ReCo loss, there were three important parameters: number of query samples, number of key samples, and parameter temperature $\tau $. They were set to 0128, 256, and 0.5, respectively, for each image batch. The output representation feature map was half the size of the original, with the channel set to 256.

\subsection{Comparison experiments} \label{4.1}
To verify the effectiveness of our road-updating methods, both supervision-based road extraction and semi-supervised segmentation methods were selected for comparison. For supervision-based road extraction methods, experiments were conducted in full-label and partial-label modes, which were used to verify the performance of the supervised learning method. For semi-supervised segmentation methods, both backbone architecture and semi-supervised training strategies had a significant influence on the results. We fixed the prediction network architecture to DeepLabV3 and conducted experiments using six semi-supervised training strategies: CutOut \citep{devries2017cutout}, CutMix \citep{french2019cutmix}, ReCo \citep{Liu2021BootstrappingSS}, Adv \citep{Hung2018Adv}, S4l \citep{zhai2019s4l}, and Gct \citep{ke2020Gct}. In comparative experiments, CutOut, CutMix, ReCo, and our model were implemented using the official code of ReCo \citep{Liu2021BootstrappingSS}, whereas others were implemented using PixelSeg \citep{ke2020Gct}.

\subsubsection{Comparative results on DeepGlobe dataset}
\begin{table}[!bt]
\centering
  \caption{Comparative results on DeepGlobe dataset (Higher is better, and the best results are shown in bold). All the experiments are trained with 12.5\% labeled data. The proposed method achieves the best IoU score among the listed methods.}\label{tab_deep_com}
\resizebox{\linewidth}{!}{
\begin{tabular}{cccccc}
\toprule
{\textbf{Methods}} & \textbf{IoU(\%)}   & \textbf{mIoU(\%)}  & \textbf{P(\%)} & \textbf{R(\%)} & \textbf{F1(\%)} \\ 
\midrule
Adv		&30.75	&63.75	& 82.14	&32.95   &47.04  \\
Gct      &29.42	&63.70	&83.78	&31.19    &45.46	\\
S4l	    &44.89   &70.96    &70.85  &55.06   &61.97    \\
Cutout  &59.70	&78.75	&76.21	&73.37 &74.76   \\
Cutmix	&60.22	&79.02	&75.88	&74.49  &75.17	\\
Classmix    &59.53  & 78.68  & 77.44  & 72.02   & 74.63  \\
Reco    & 60.52  &79.19    &77.36  &73.54   &75.40    \\
\midrule
\makecell[c]{SRUNet  \\ (without map) }  &62.49	&80.26	&80.87	&73.33  &76.92    \\
\makecell[c]{SRUNet  \\ (Ours) }  & \textbf{74.89}    &\textbf{86.81}   &\textbf{86.70}   &\textbf{84.60}  &\textbf{85.64} \\
\bottomrule
\end{tabular}
}
\end{table}

\begin{figure*}[tb]
\small
   \centering
		\newcommand{\tabincell}[2]{\begin{tabular}{@{}#1@{}}#2\end{tabular}}
		\begin{tabular}{cccccccc}
\includegraphics[width=0.1\textwidth]{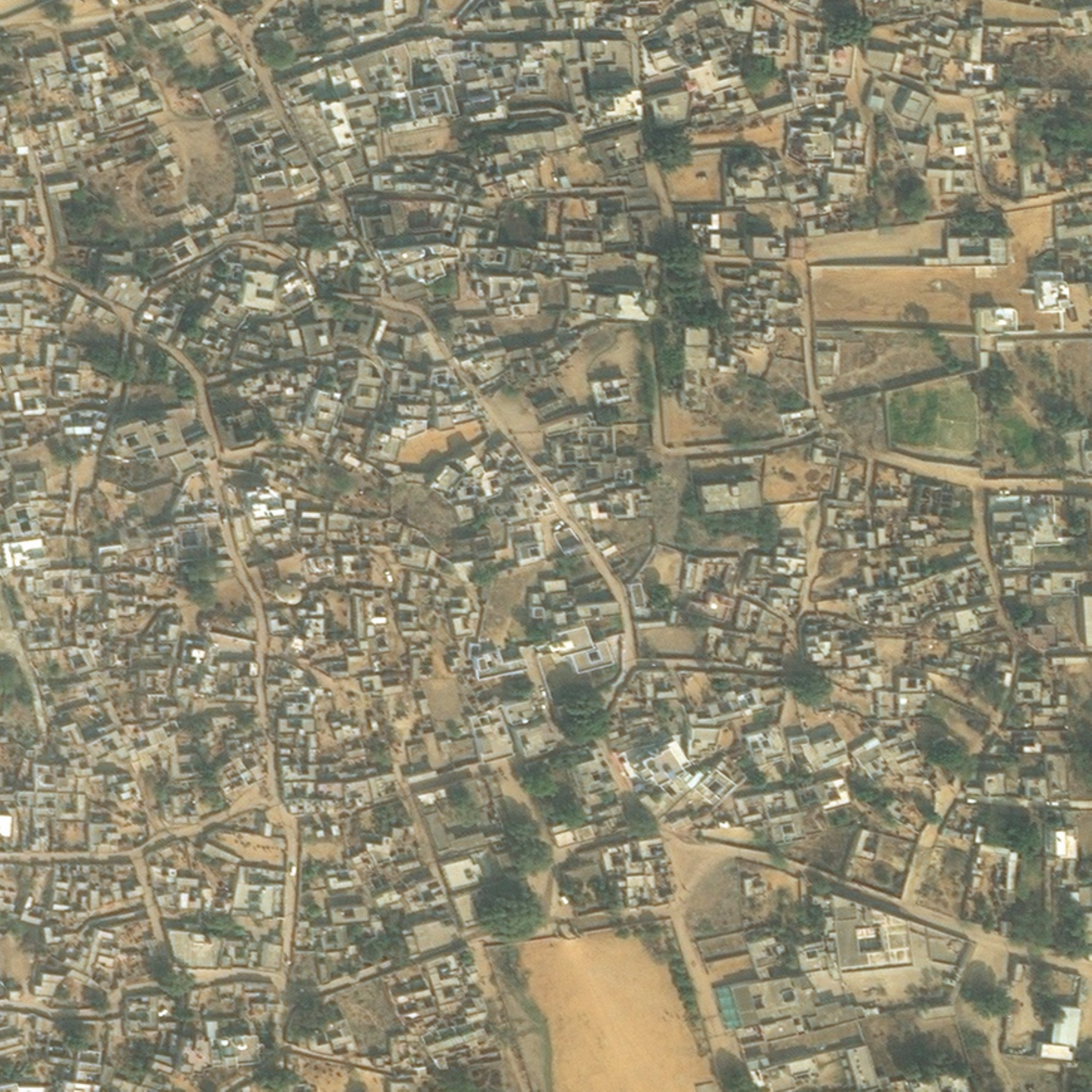}
			& 
\includegraphics[width=0.1\textwidth]{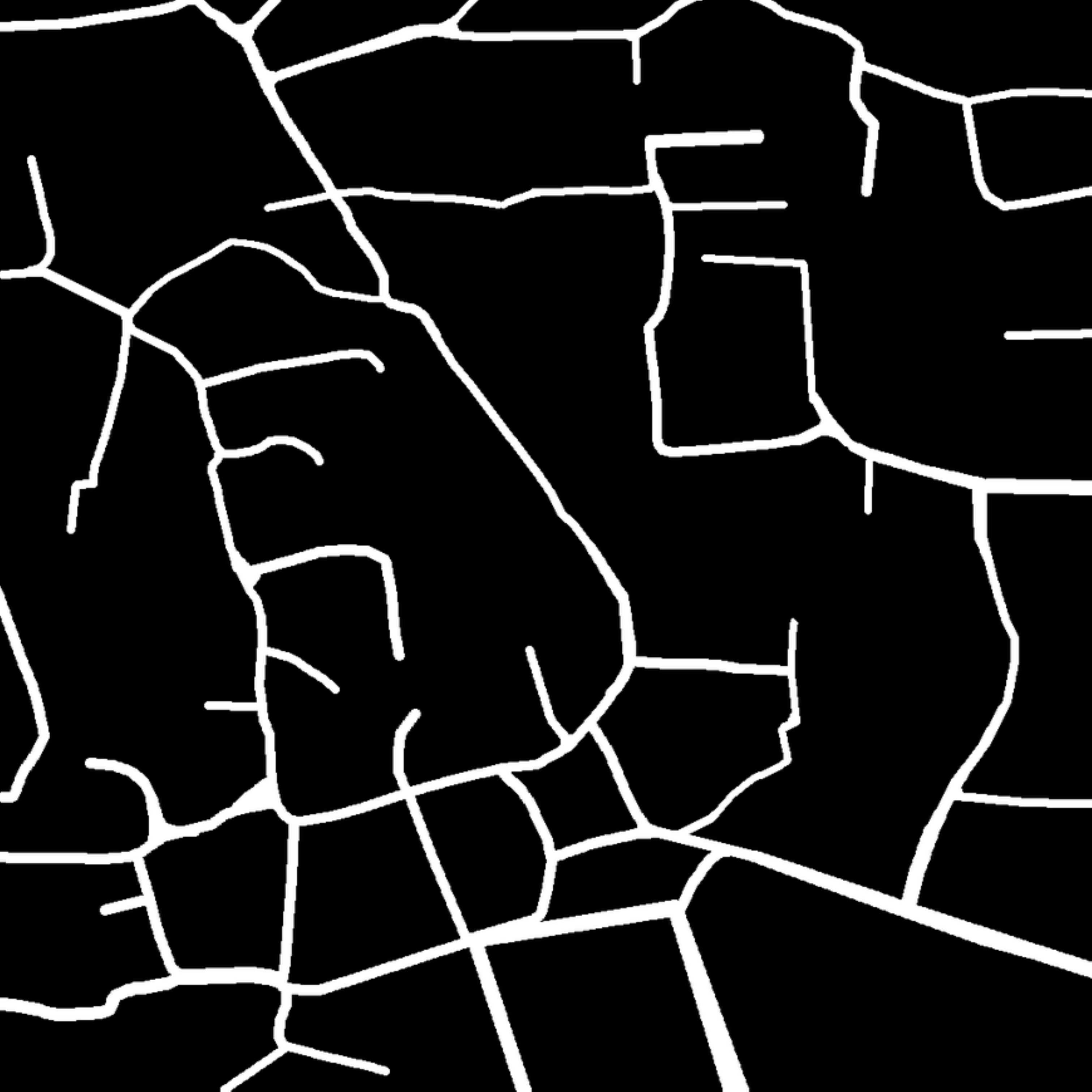}
			& 
\includegraphics[width=0.1\textwidth]{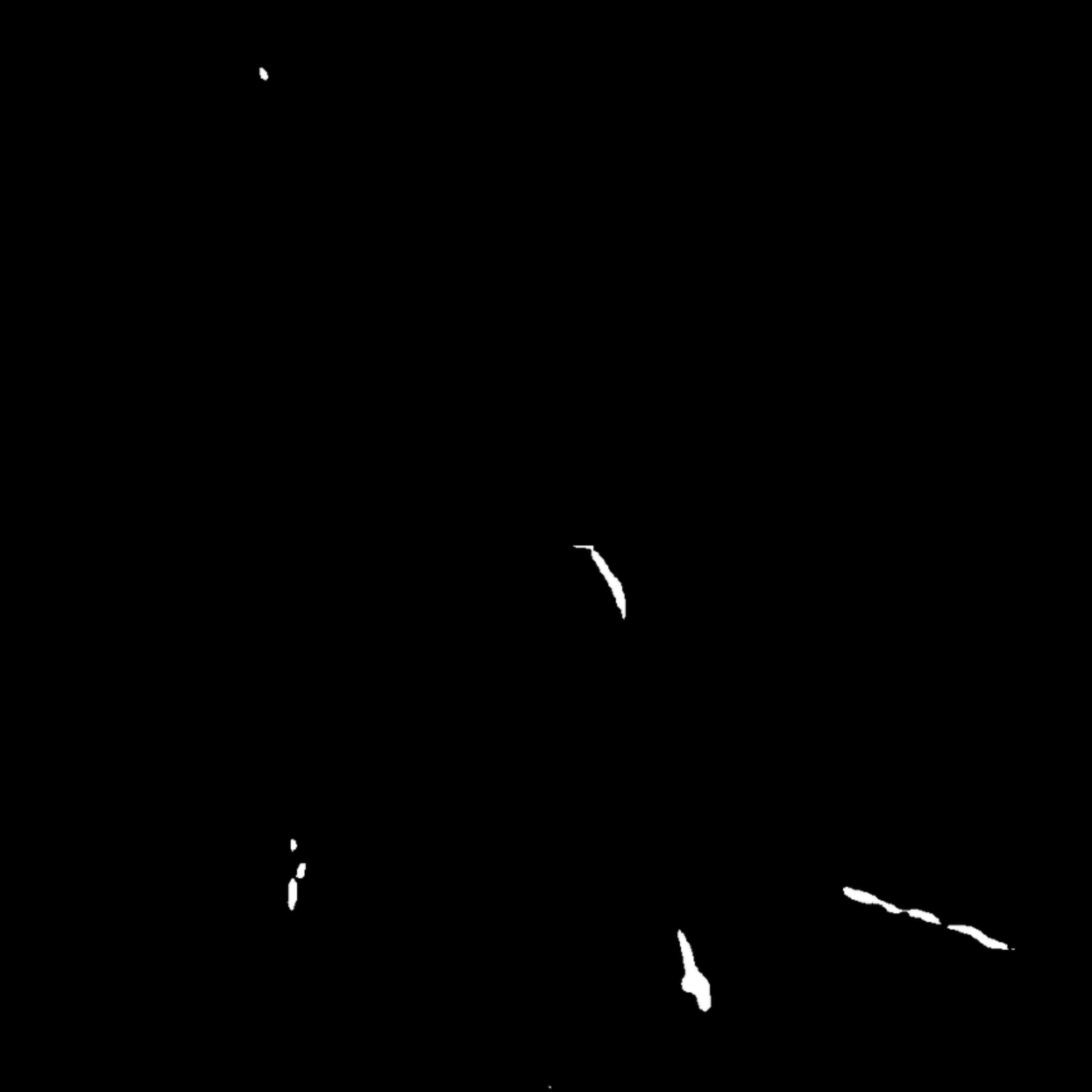}
			& 
\includegraphics[width=0.1\textwidth]{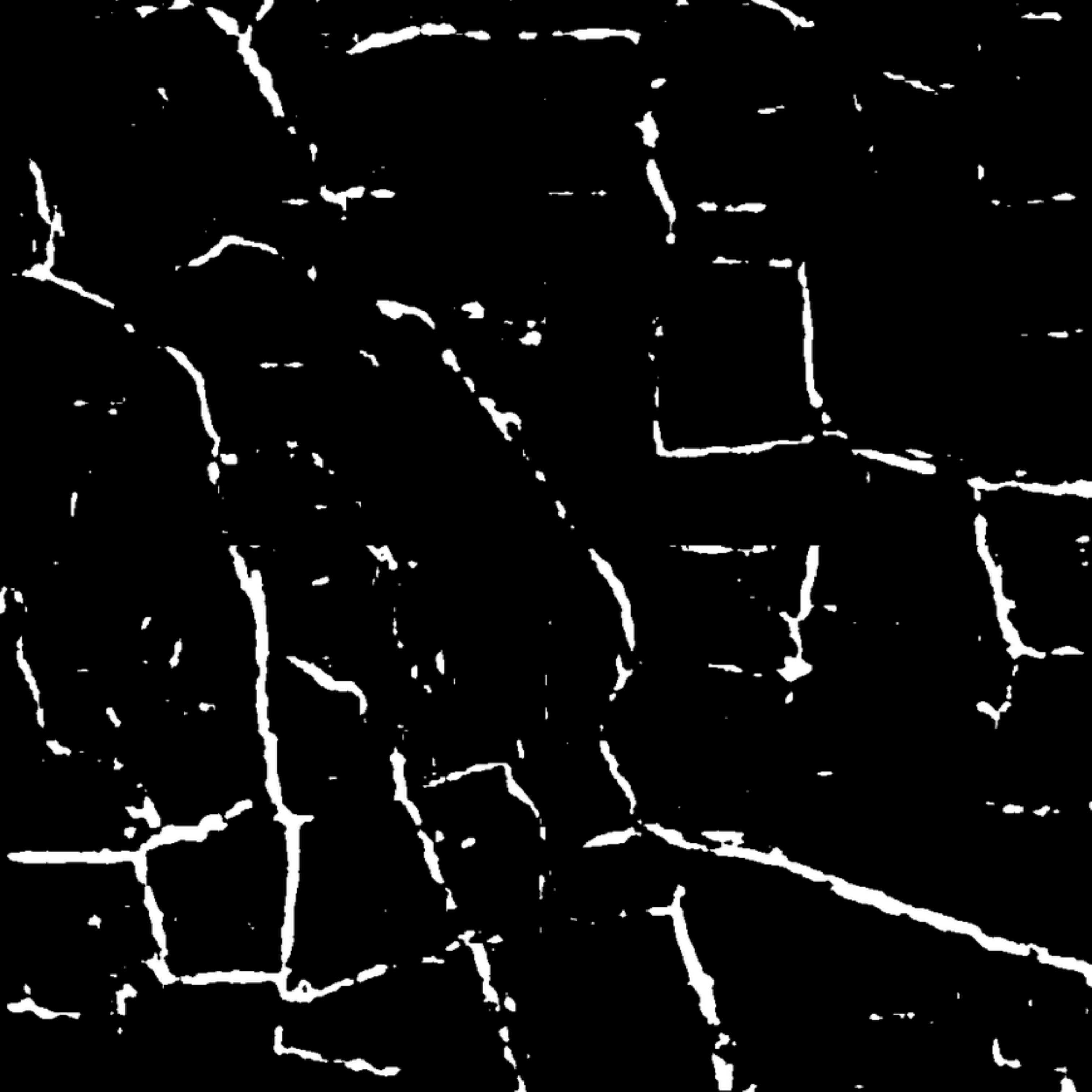}
			& 
\includegraphics[width=0.1\textwidth]{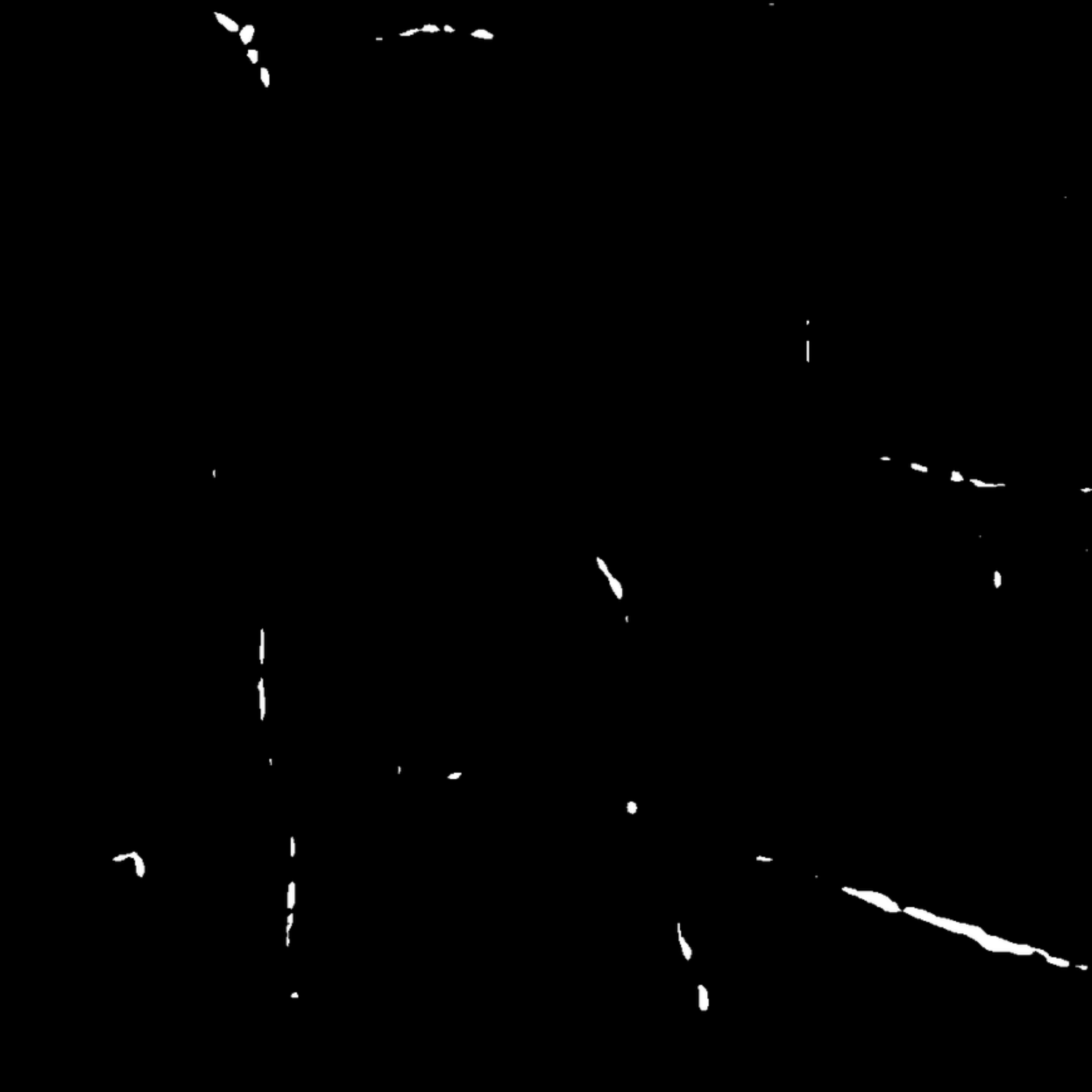}
			& 
\includegraphics[width=0.1\textwidth]{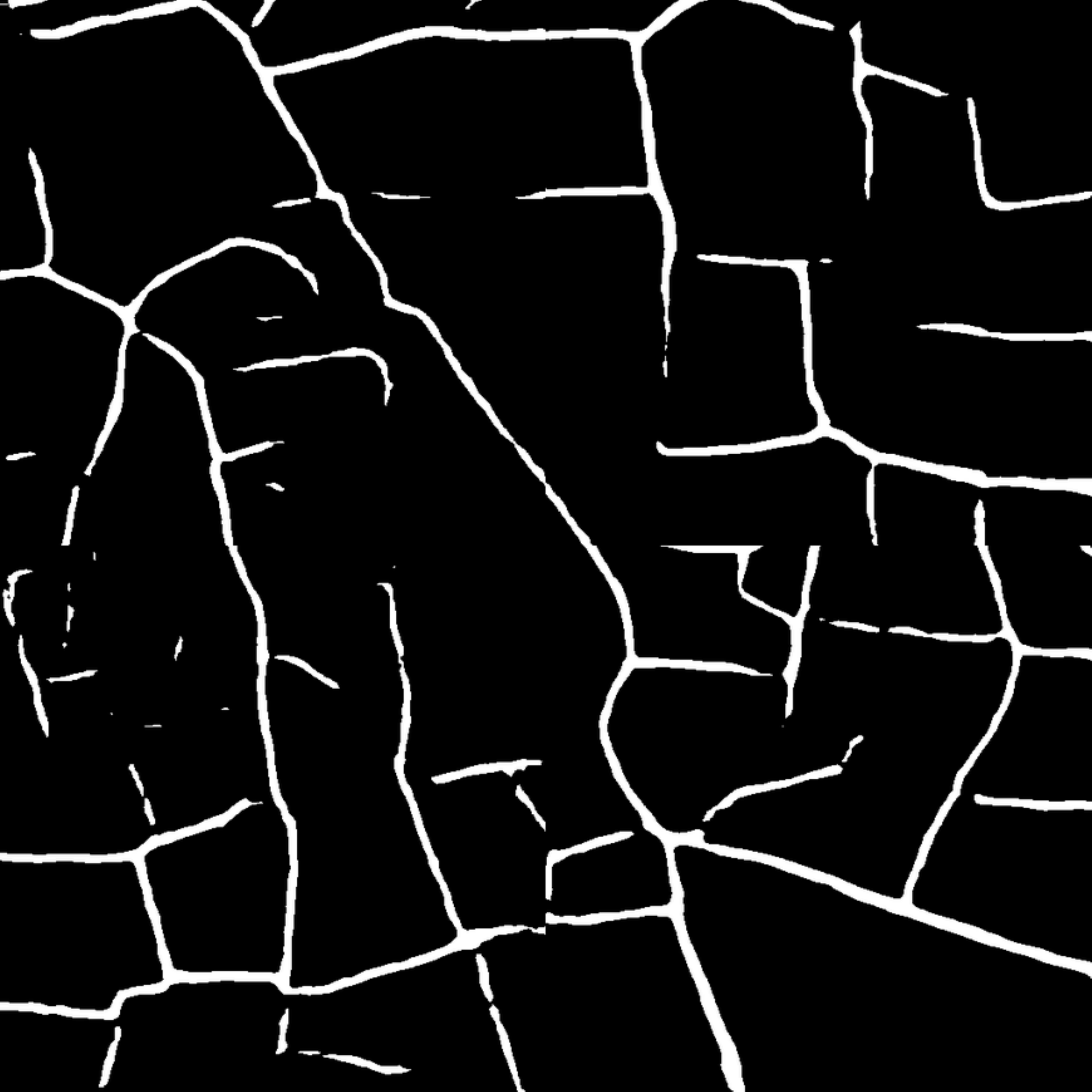}
			& 
\includegraphics[width=0.1\textwidth]{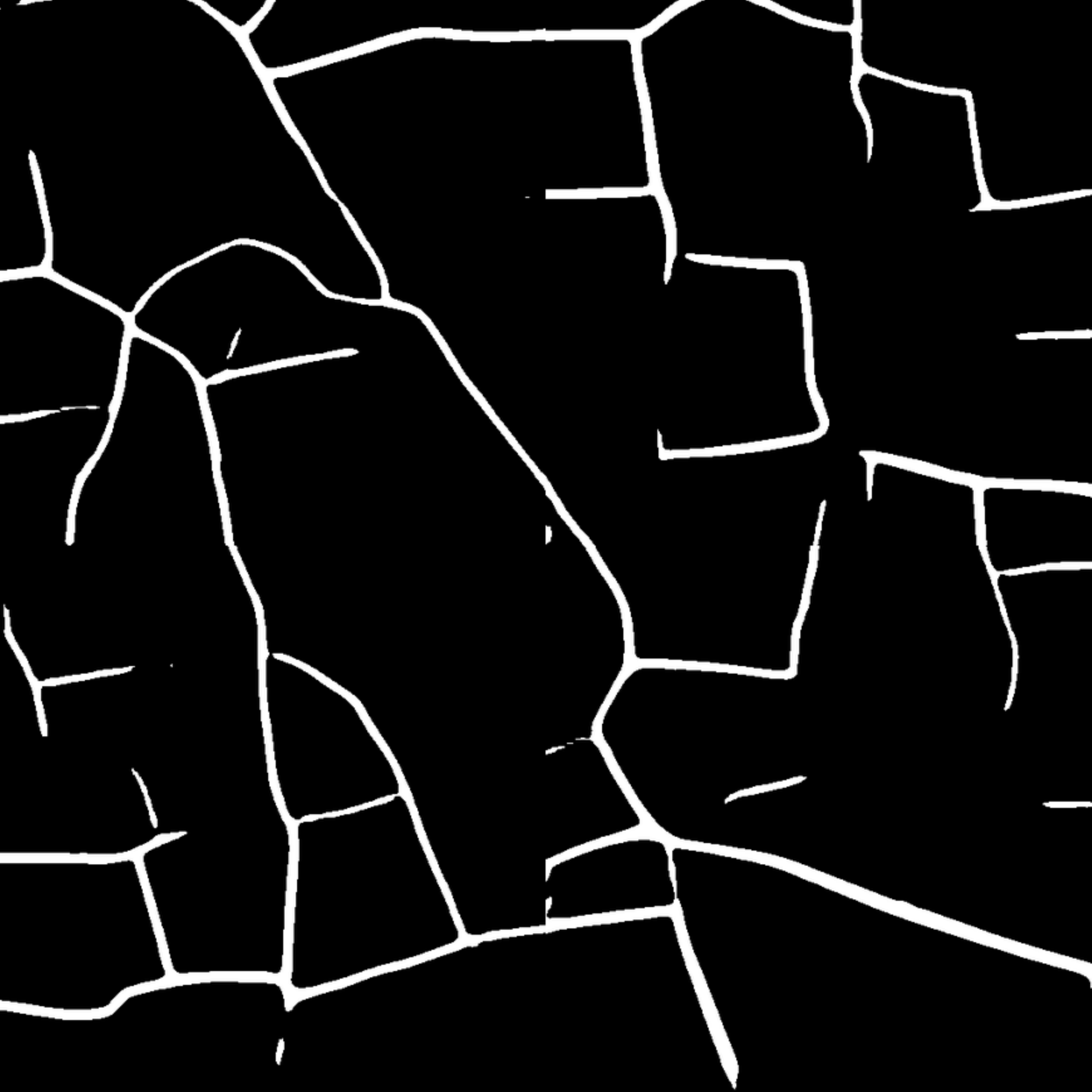}
            &
\includegraphics[width=0.1\textwidth]{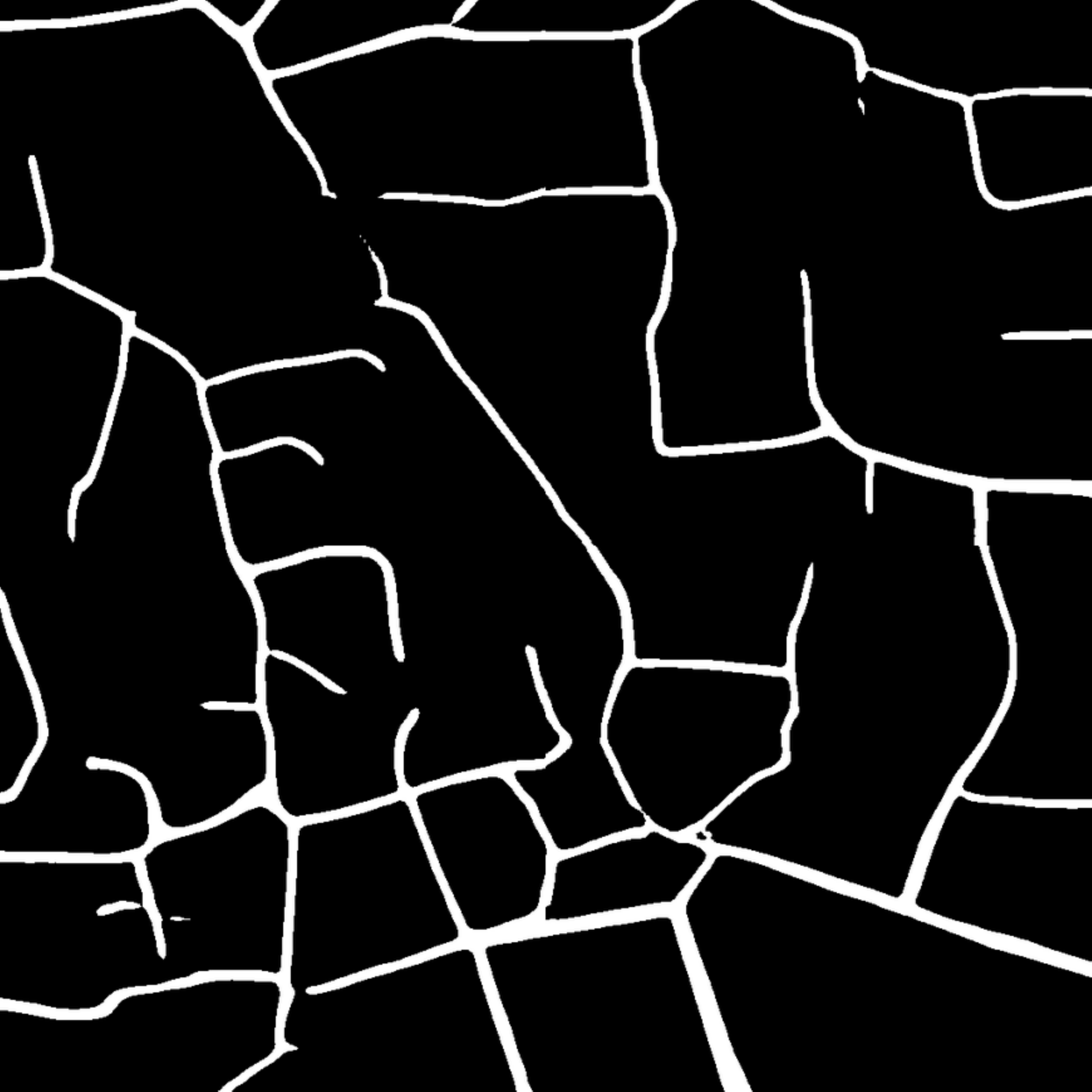}
		\\
\includegraphics[width=0.1\textwidth]{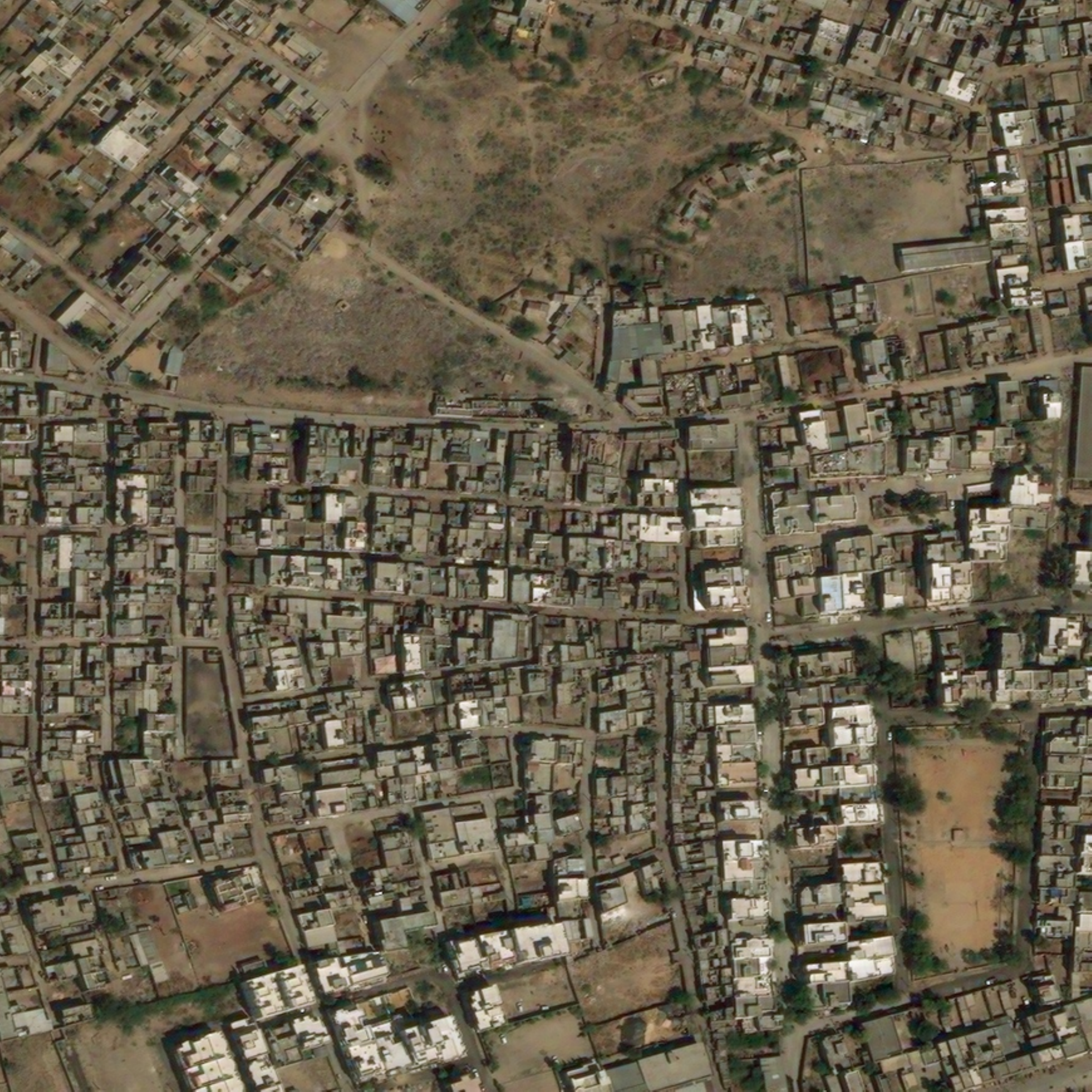}
			& 
\includegraphics[width=0.1\textwidth]{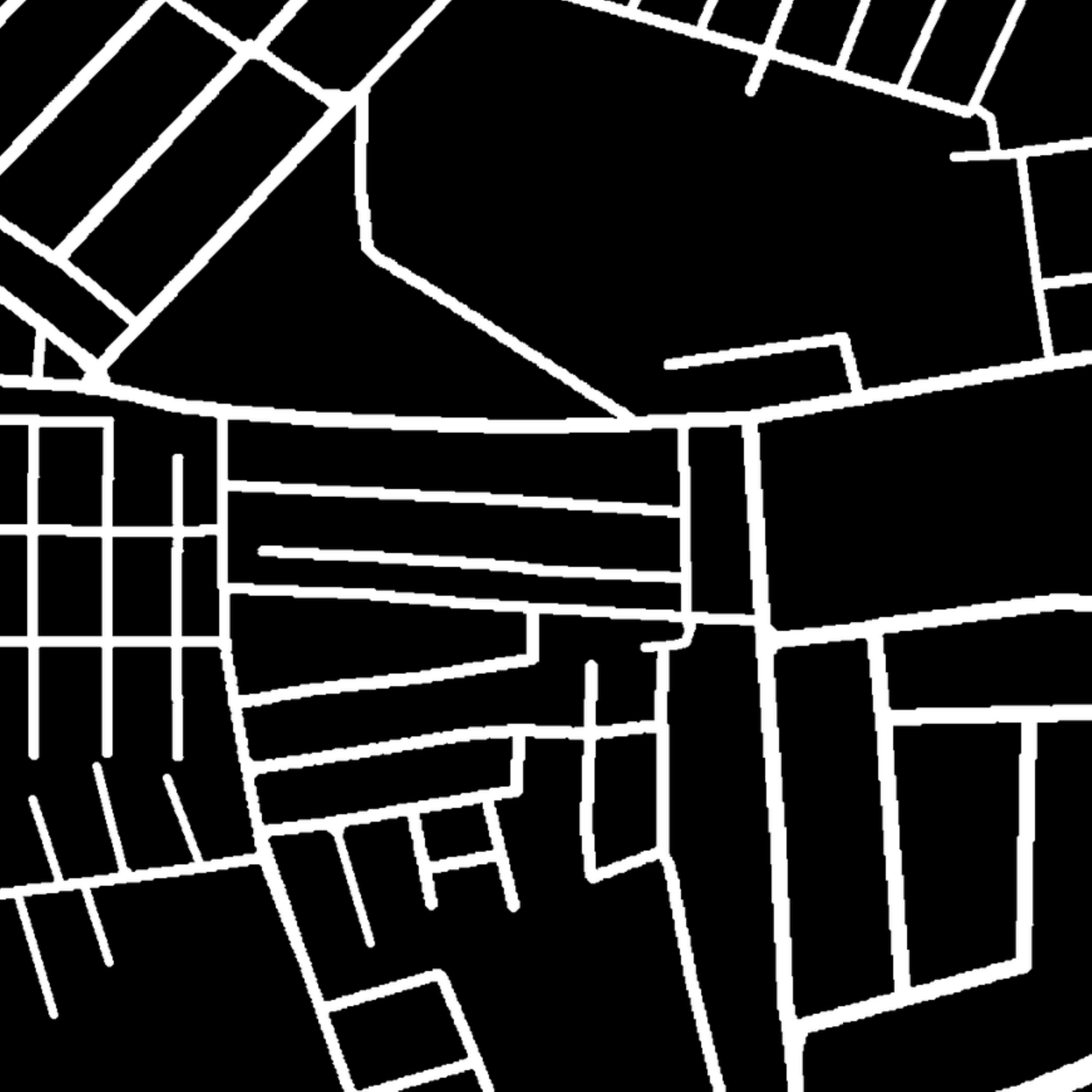}
			& 
\includegraphics[width=0.1\textwidth]{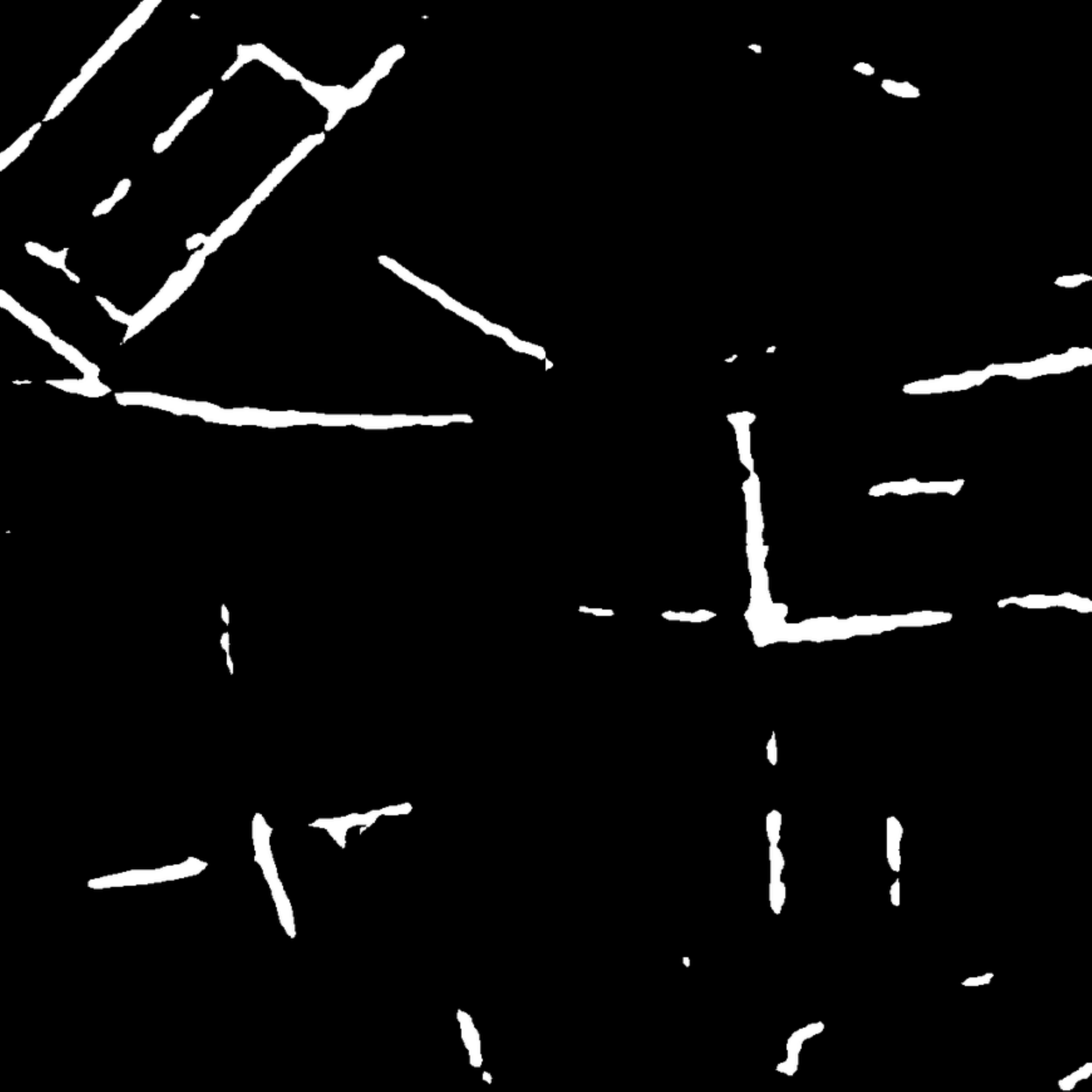}
			& 
\includegraphics[width=0.1\textwidth]{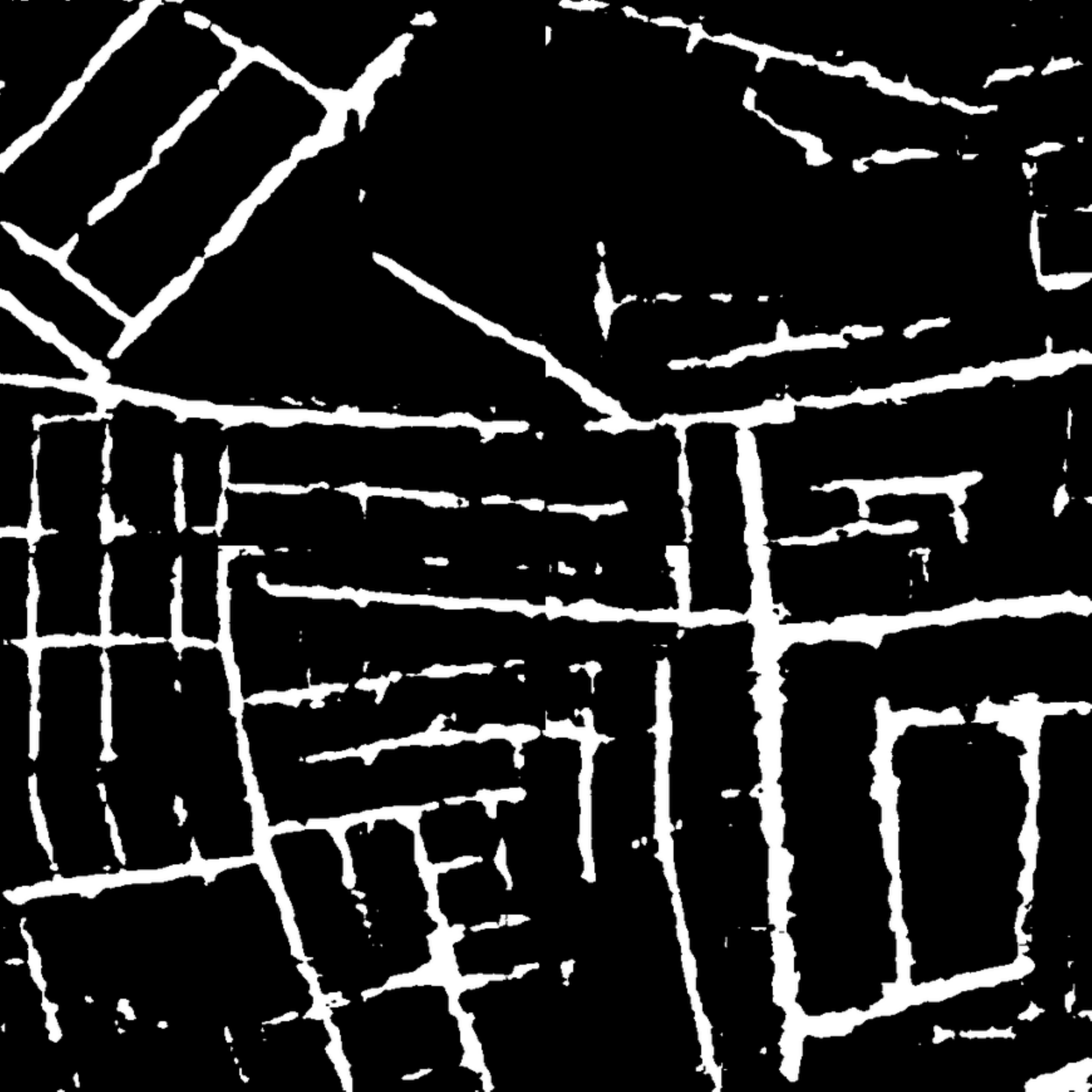}
			& 
\includegraphics[width=0.1\textwidth]{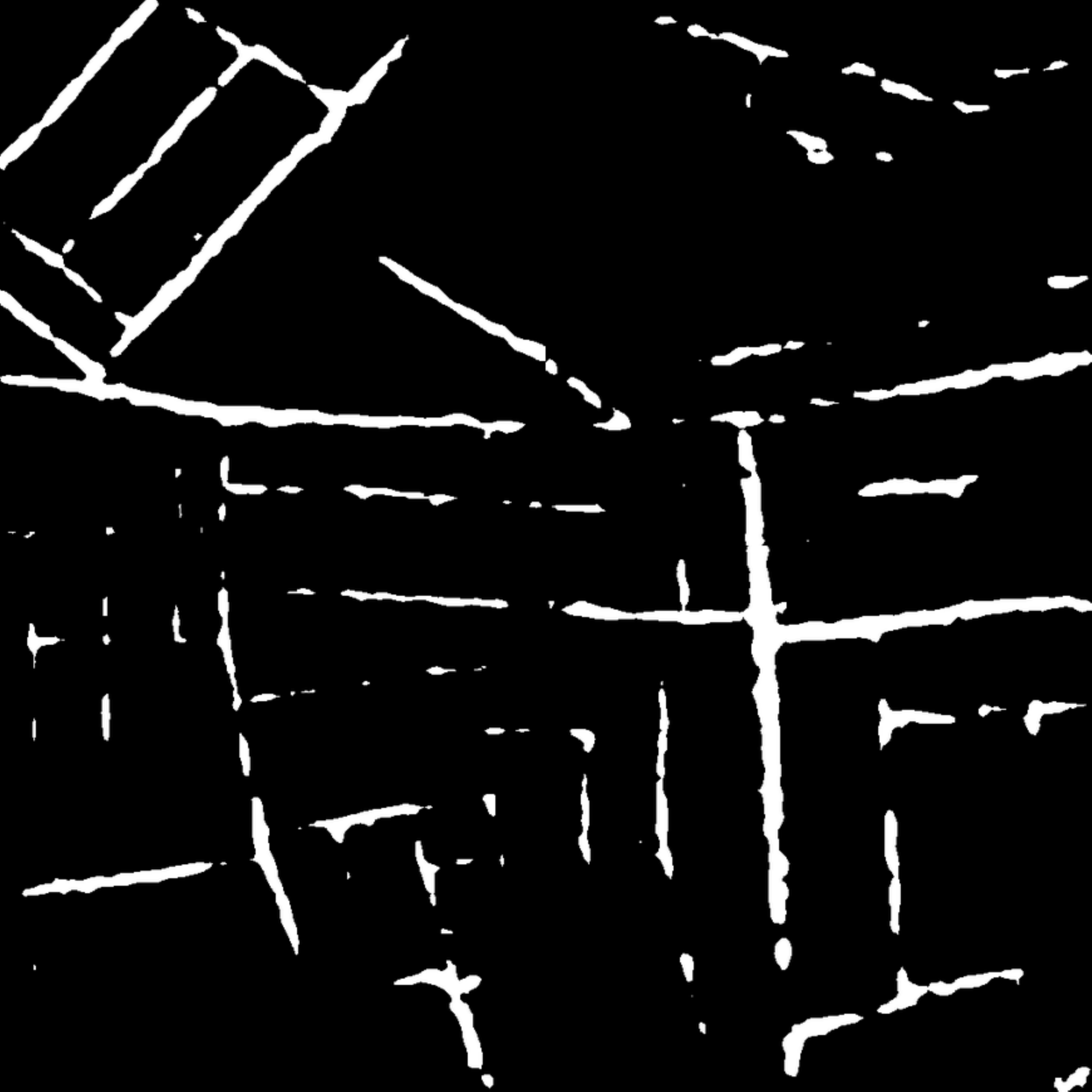}
			& 
\includegraphics[width=0.1\textwidth]{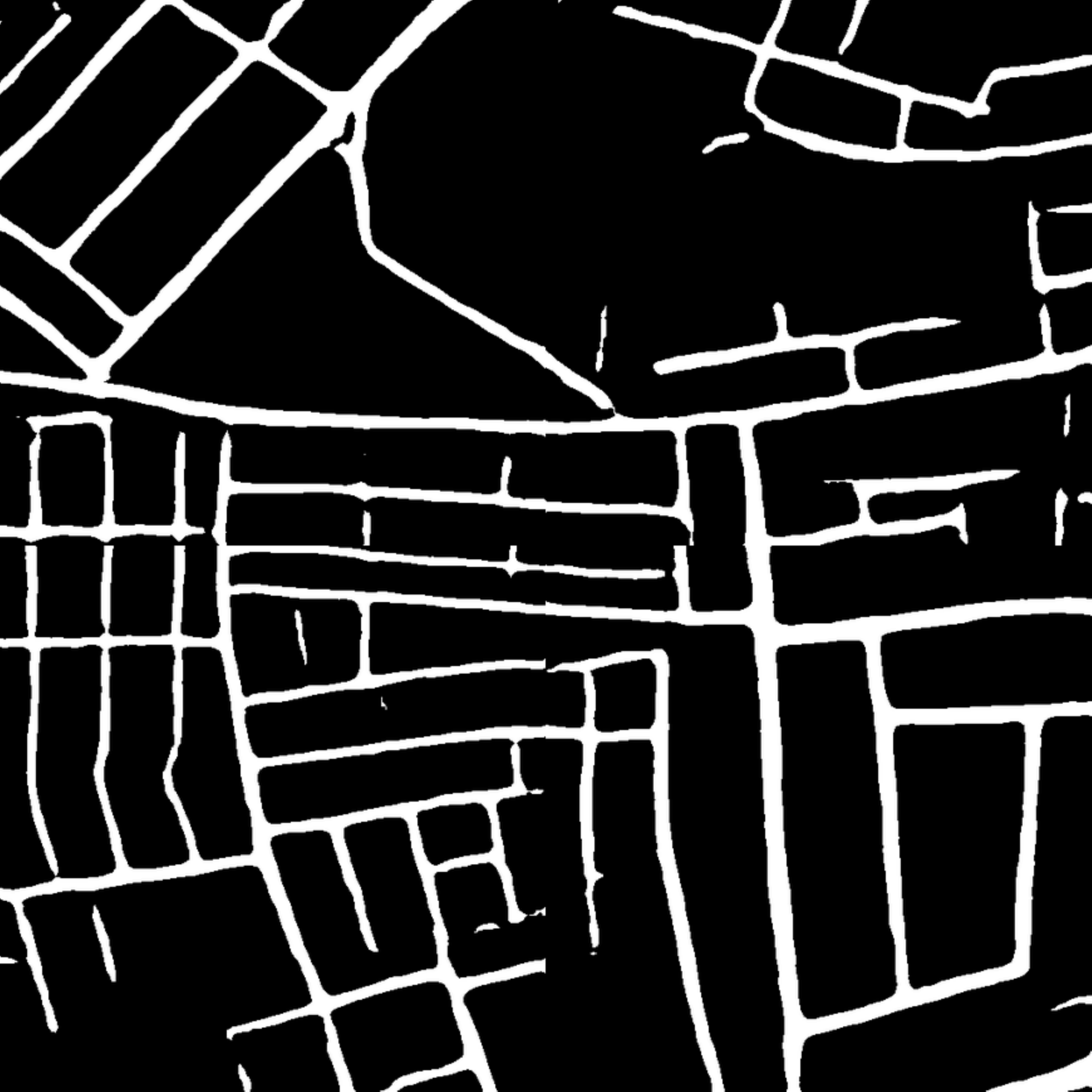}
			& 
\includegraphics[width=0.1\textwidth]{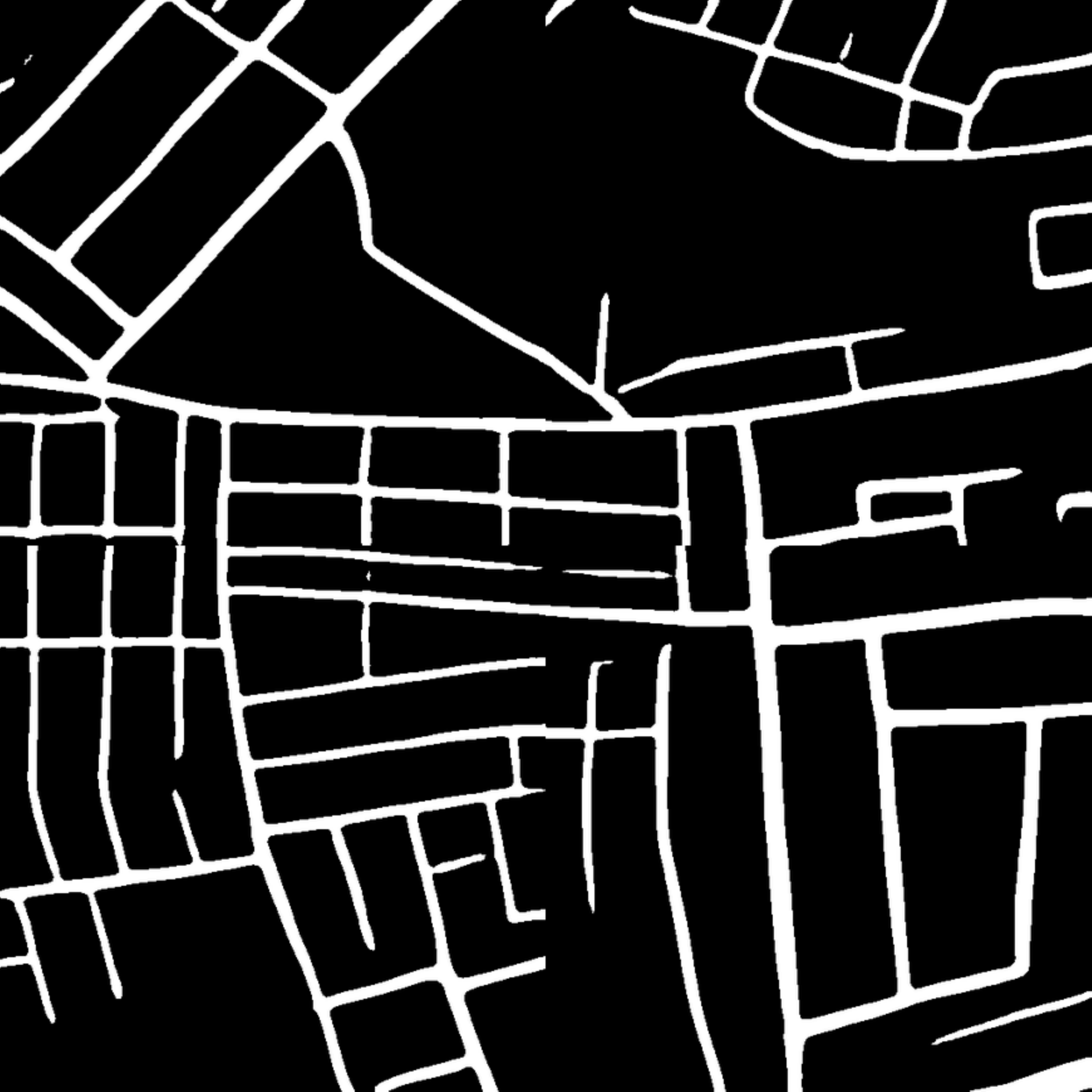}
            &
\includegraphics[width=0.1\textwidth]{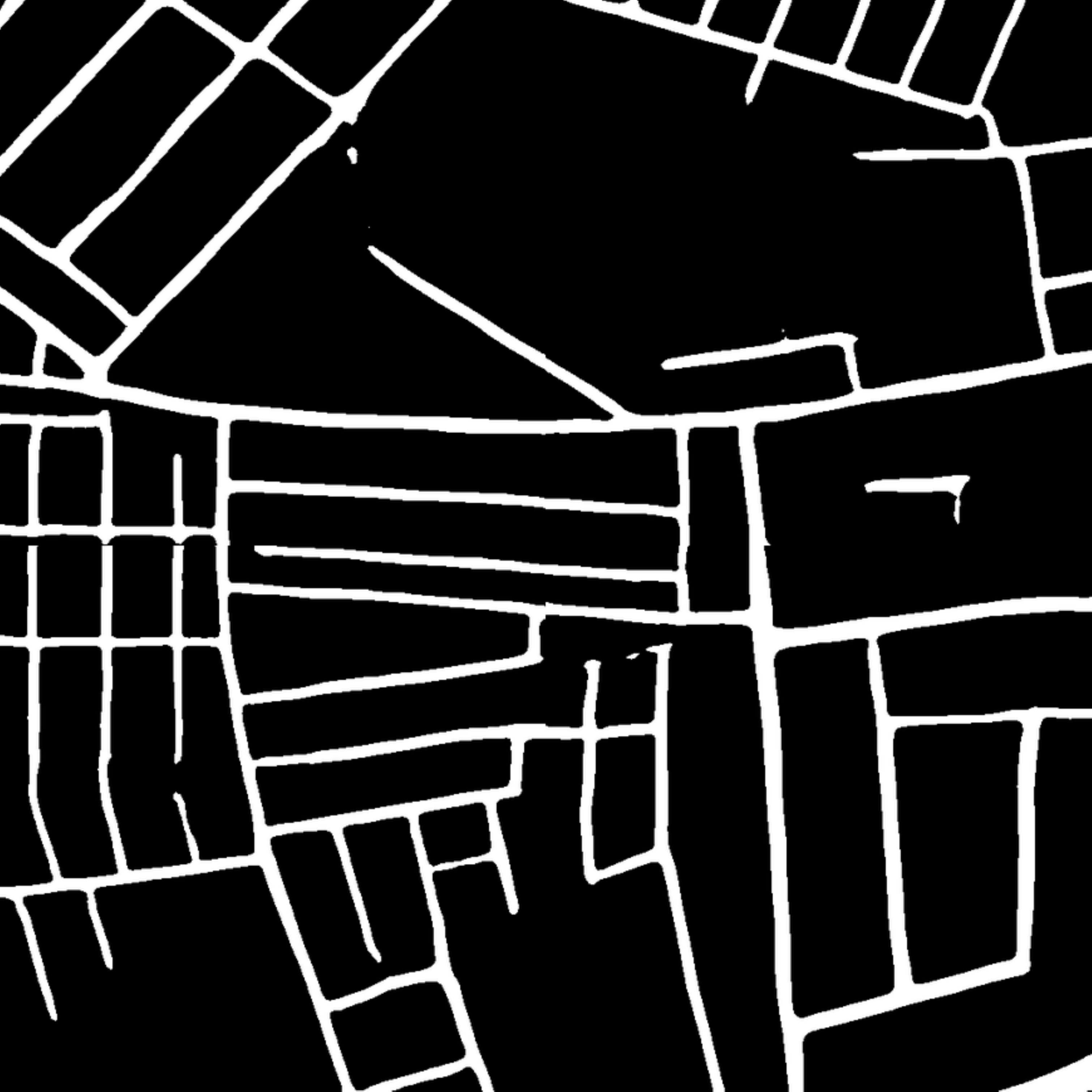}
		\\
\includegraphics[width=0.1\textwidth]{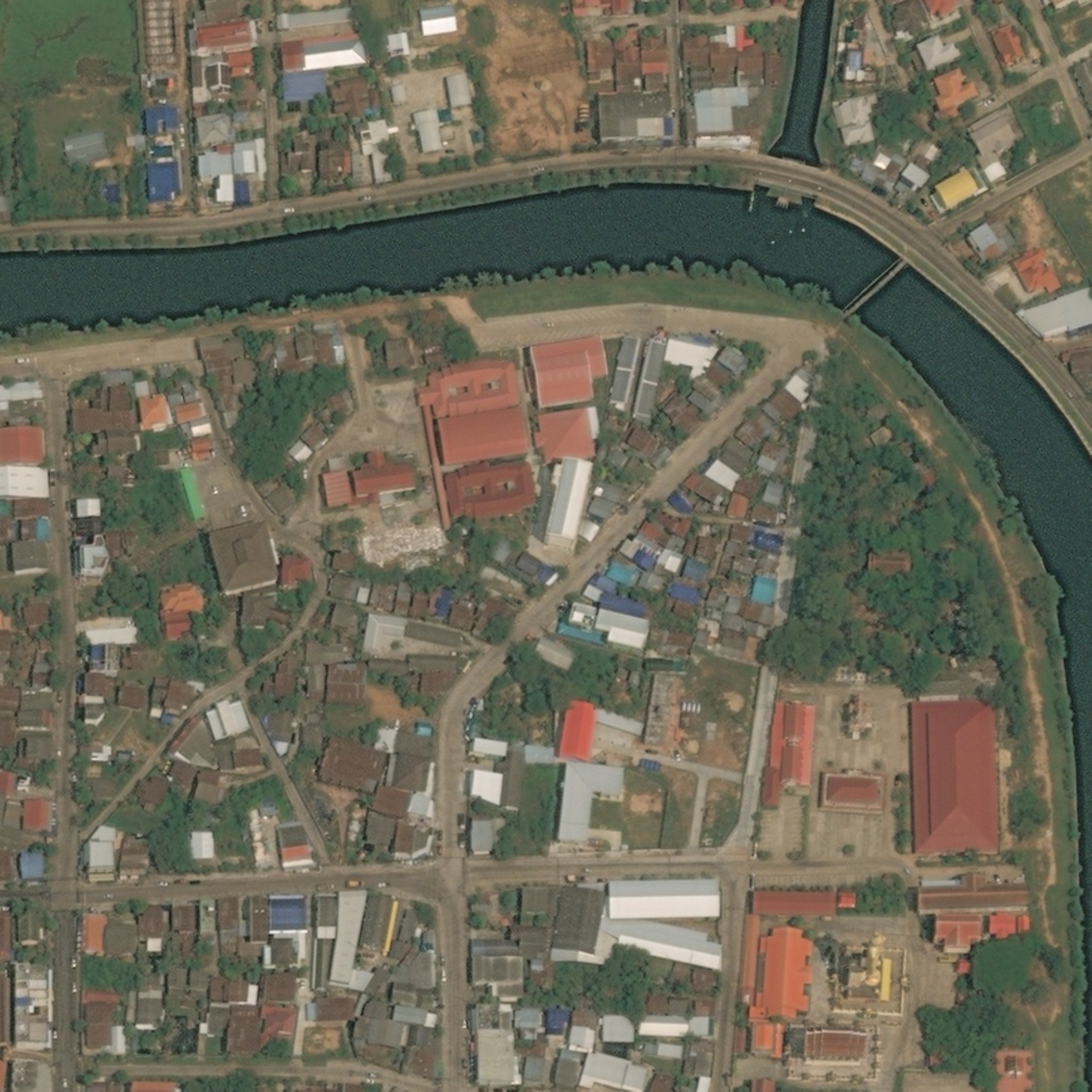}
			& 
\includegraphics[width=0.1\textwidth]{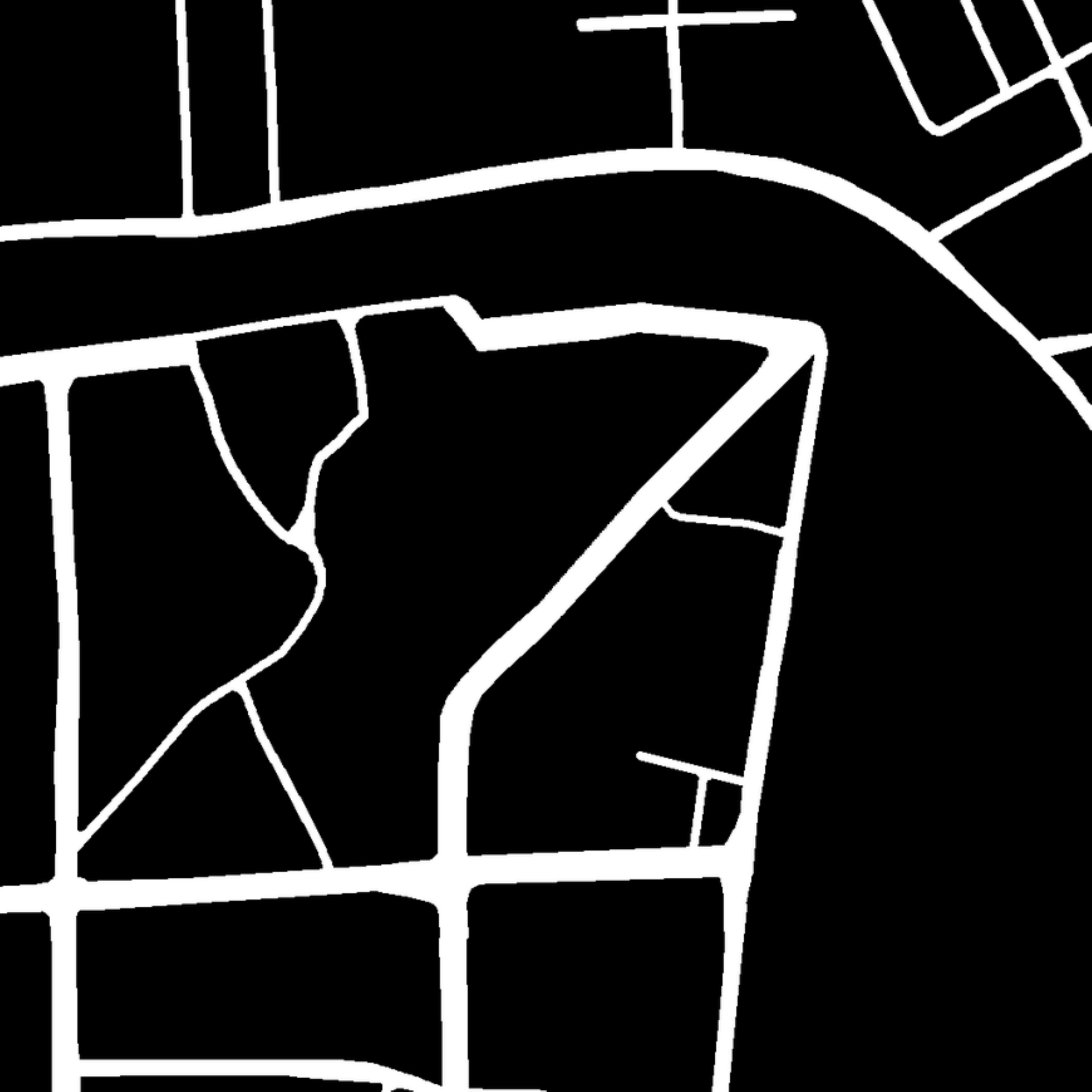}
			& 
\includegraphics[width=0.1\textwidth]{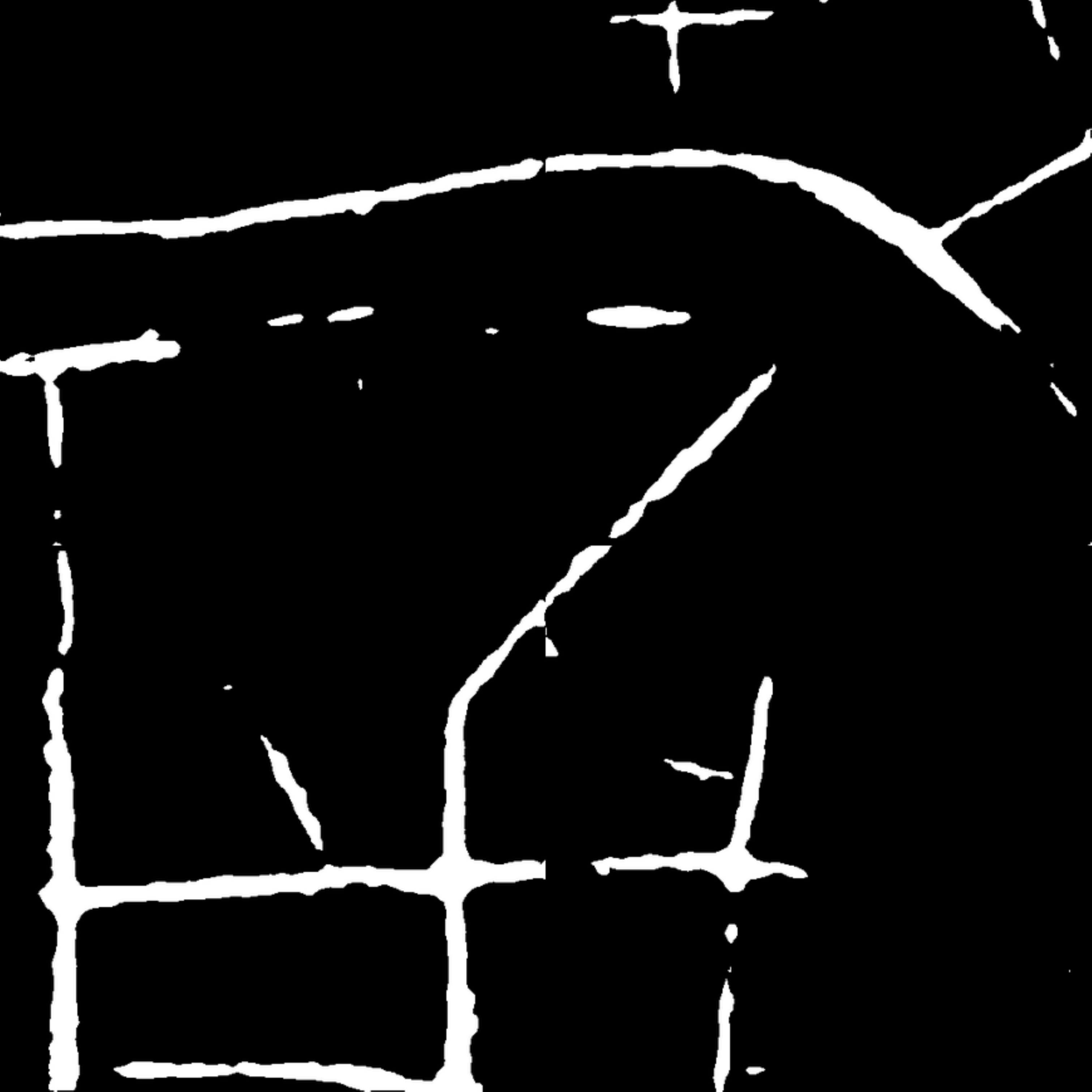}
			& 
\includegraphics[width=0.1\textwidth]{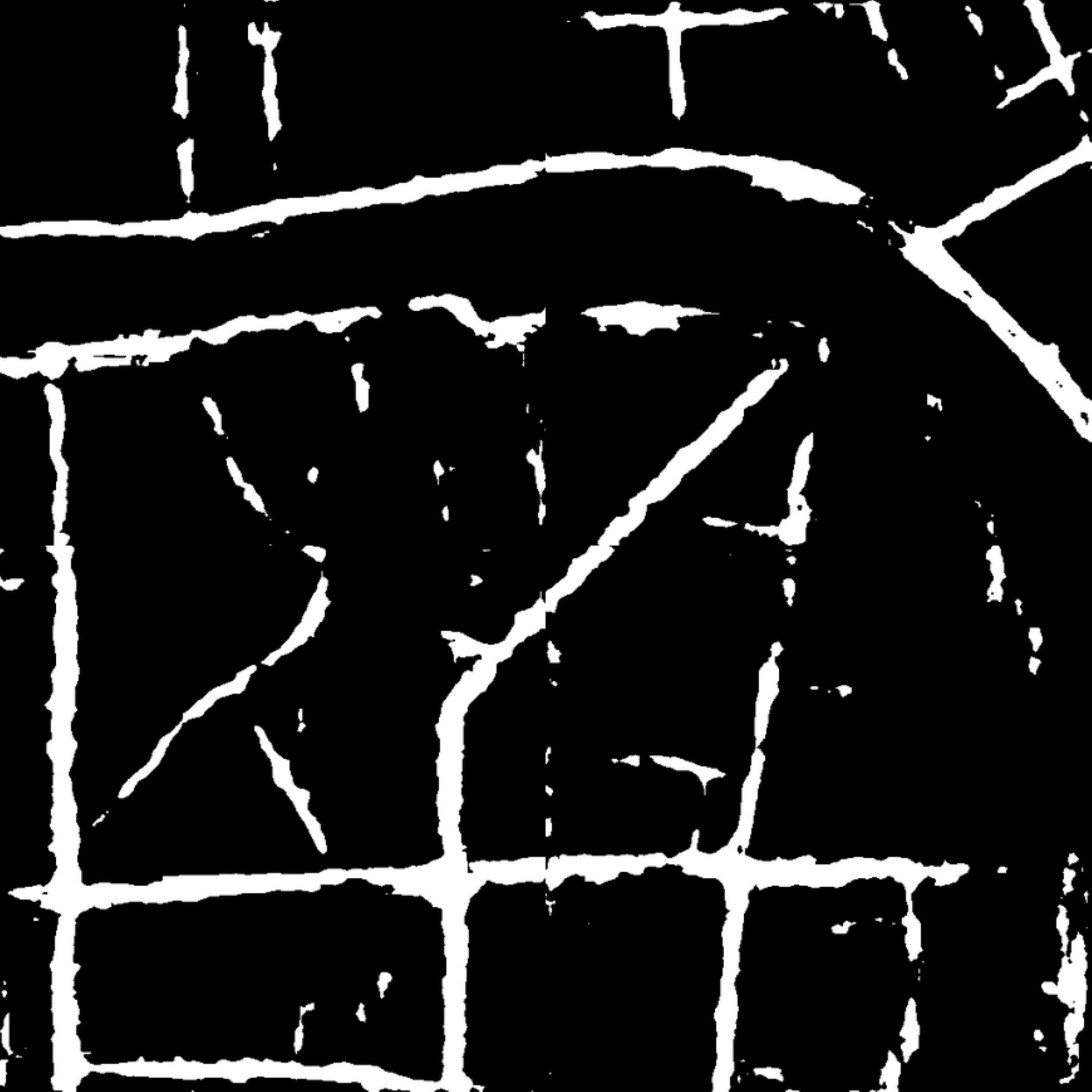}
			& 
\includegraphics[width=0.1\textwidth]{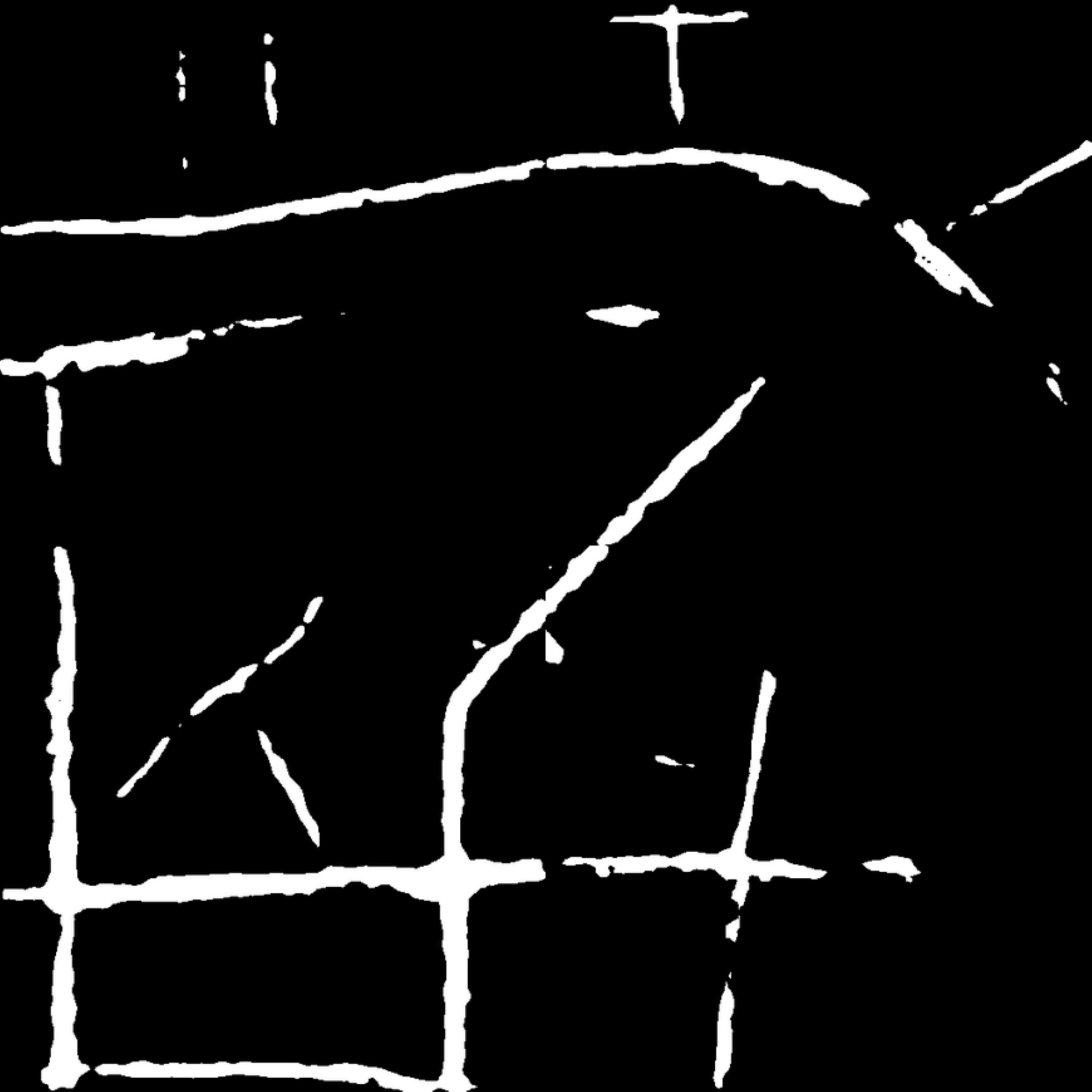}
			& 
\includegraphics[width=0.1\textwidth]{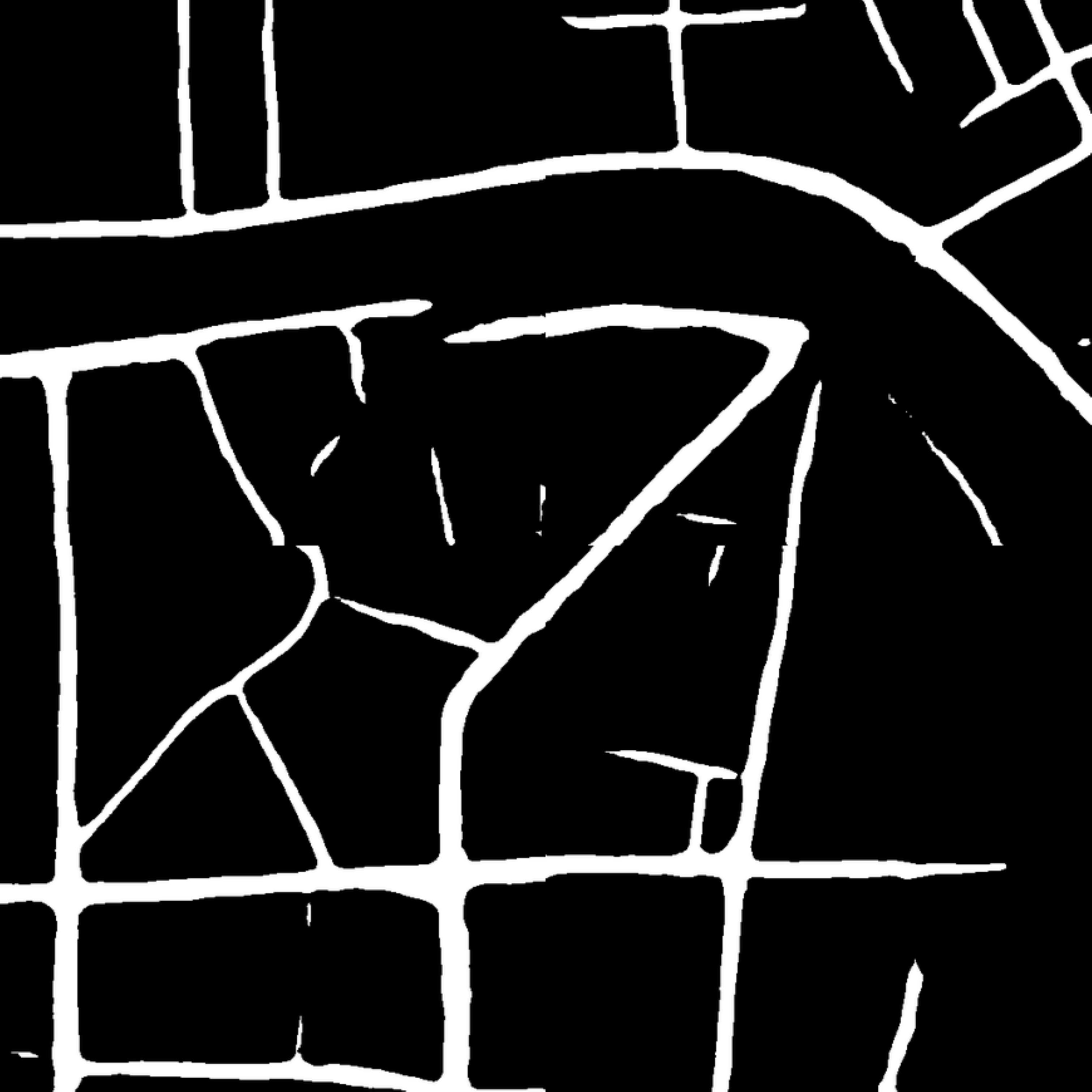}
			& 
\includegraphics[width=0.1\textwidth]{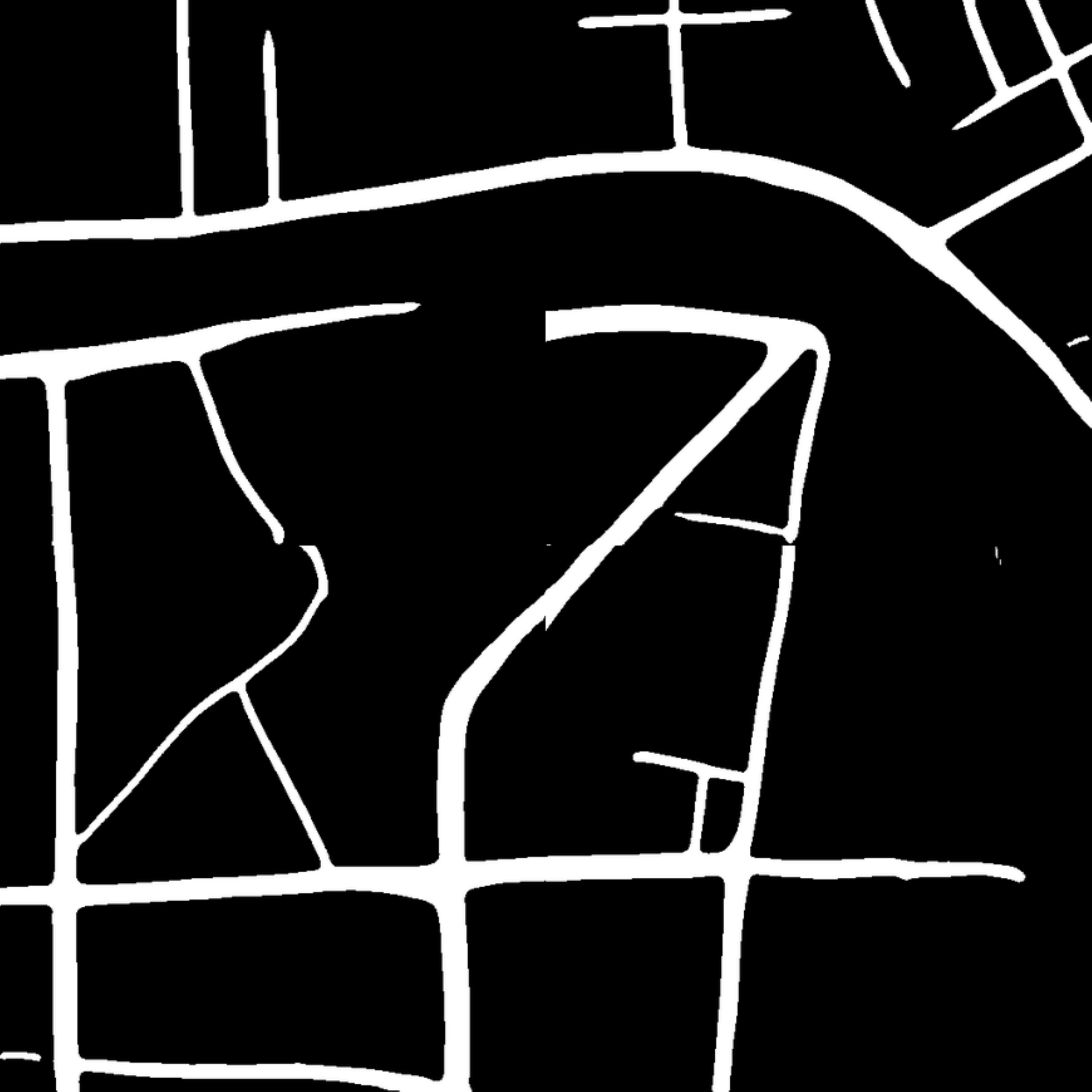}
            &
\includegraphics[width=0.1\textwidth]{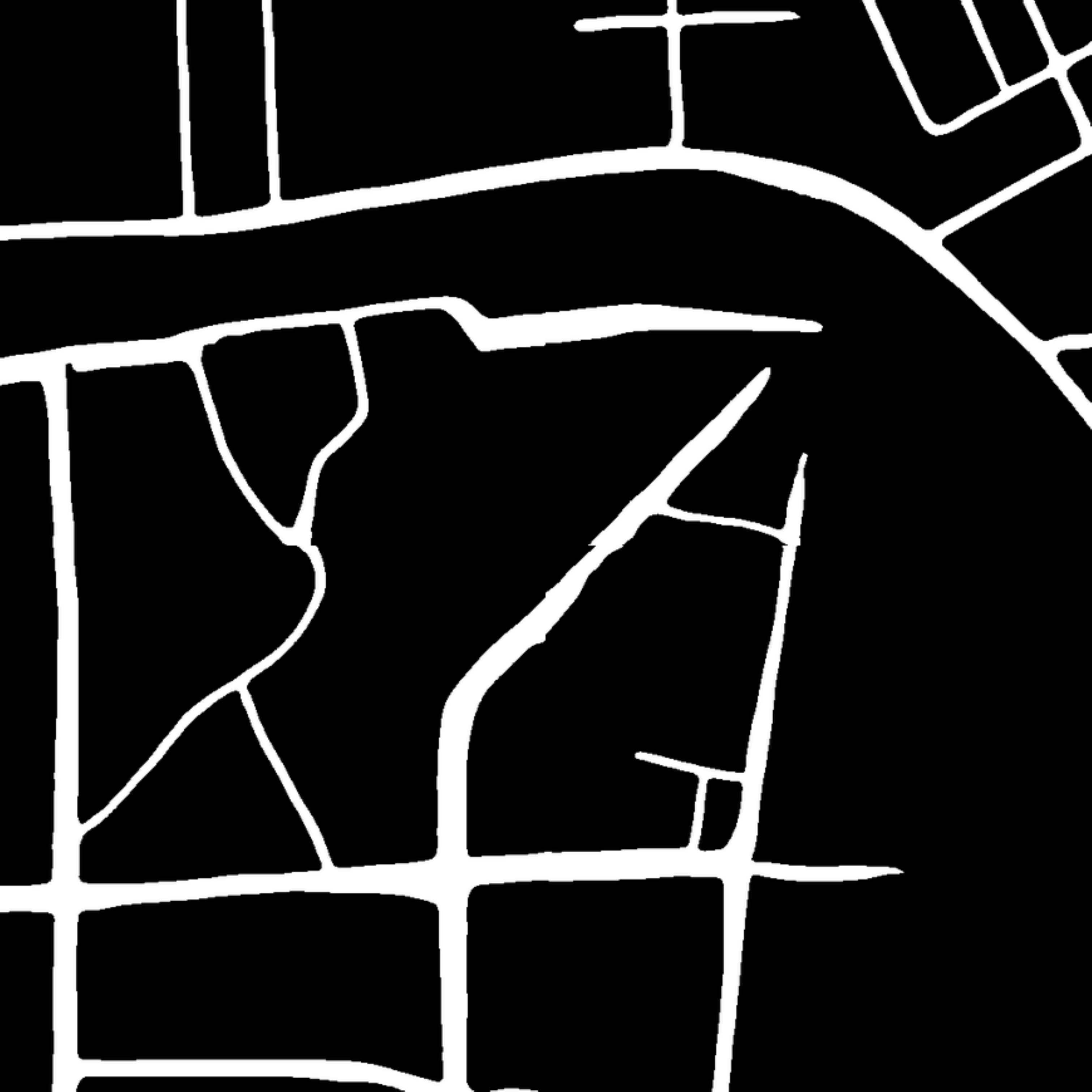}
		\\
\includegraphics[width=0.1\textwidth]{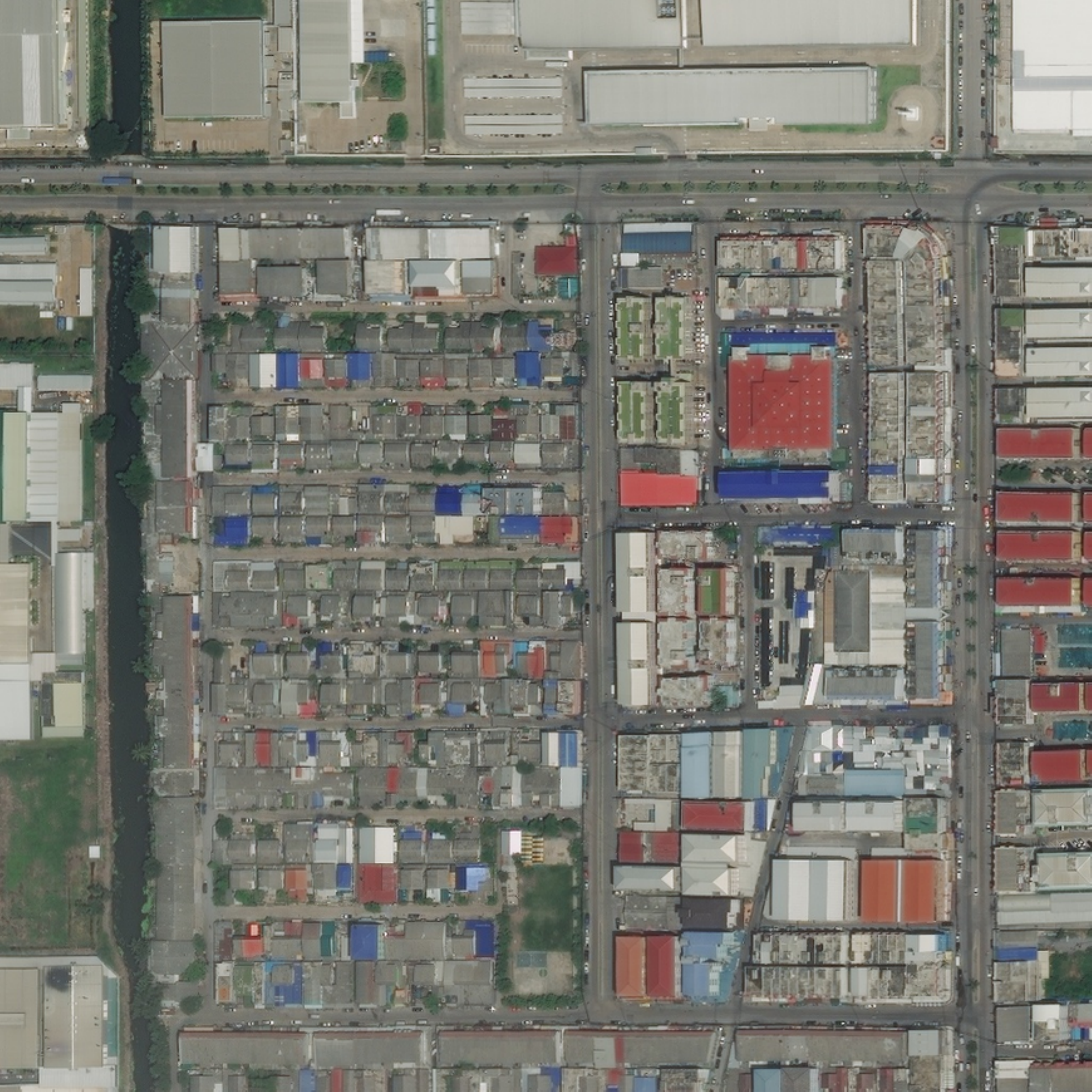}
			& 
\includegraphics[width=0.1\textwidth]{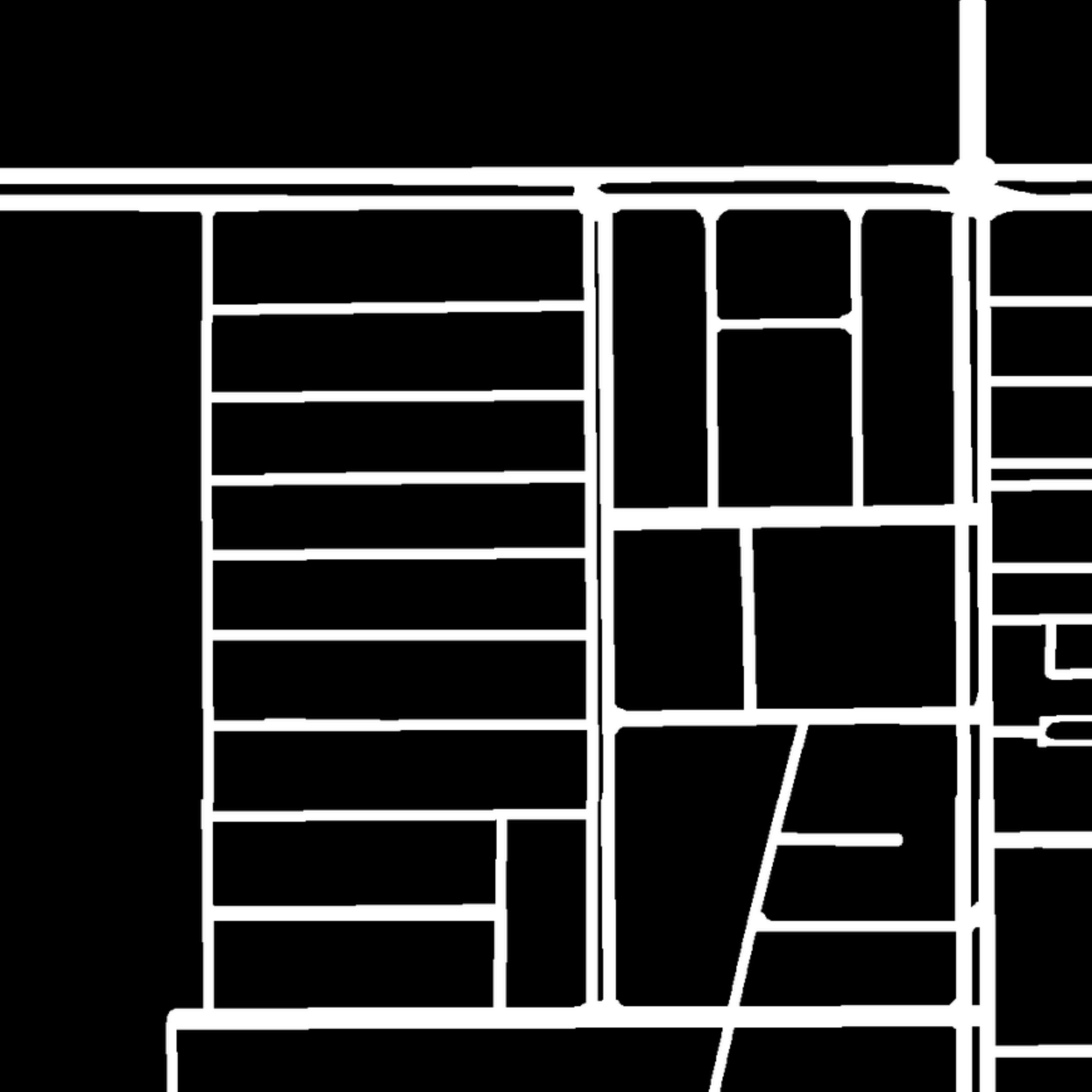}
			& 
\includegraphics[width=0.1\textwidth]{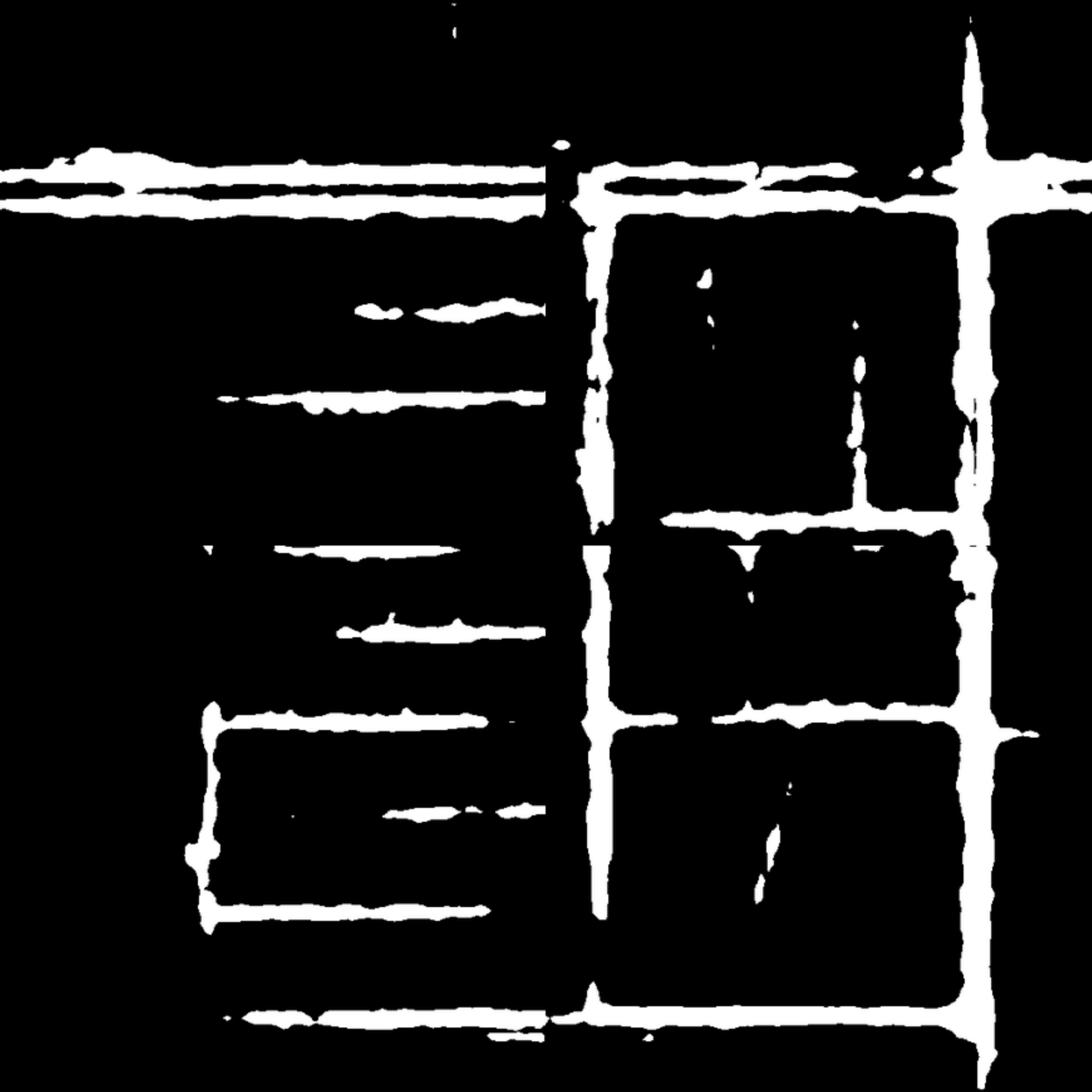}
			& 
\includegraphics[width=0.1\textwidth]{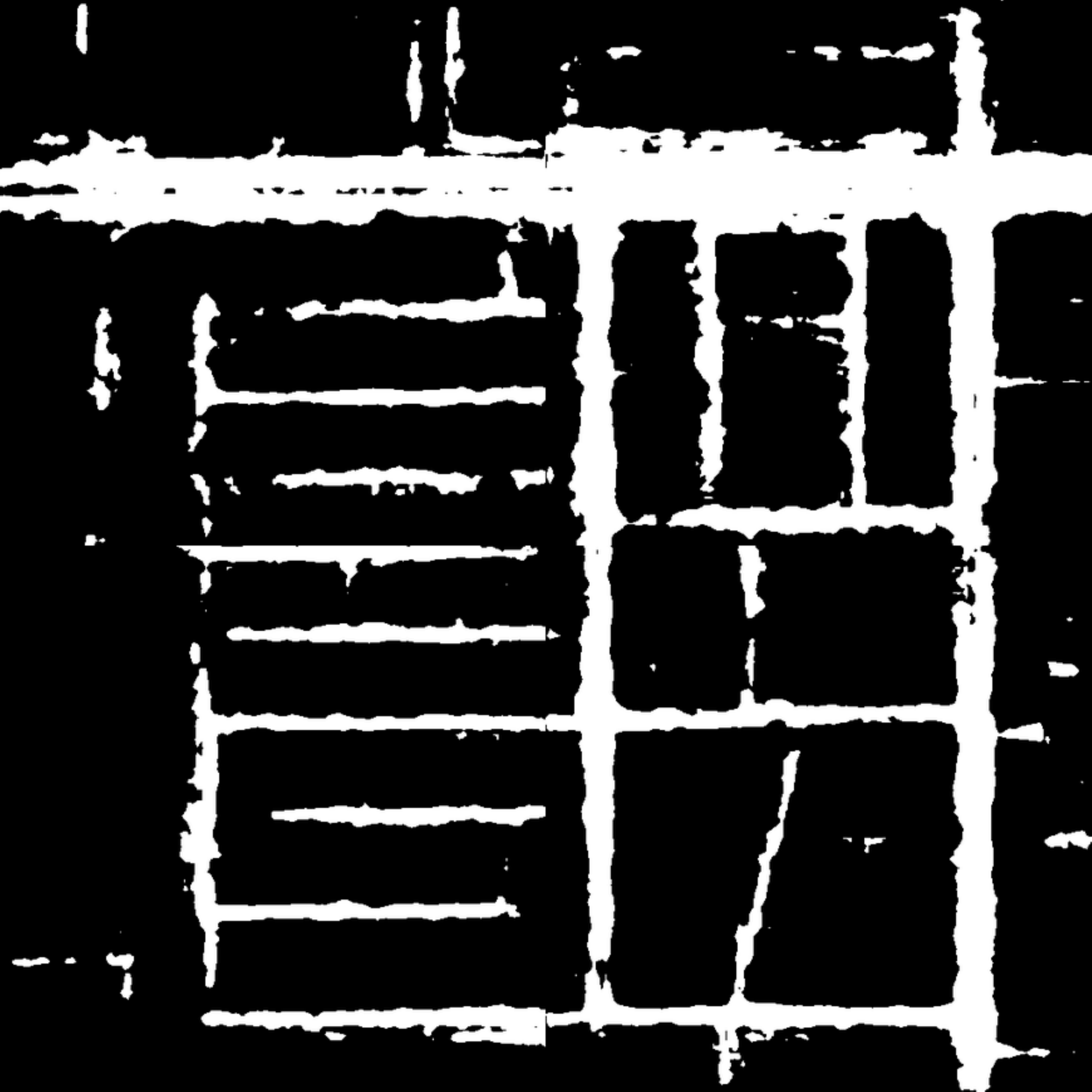}
			& 
\includegraphics[width=0.1\textwidth]{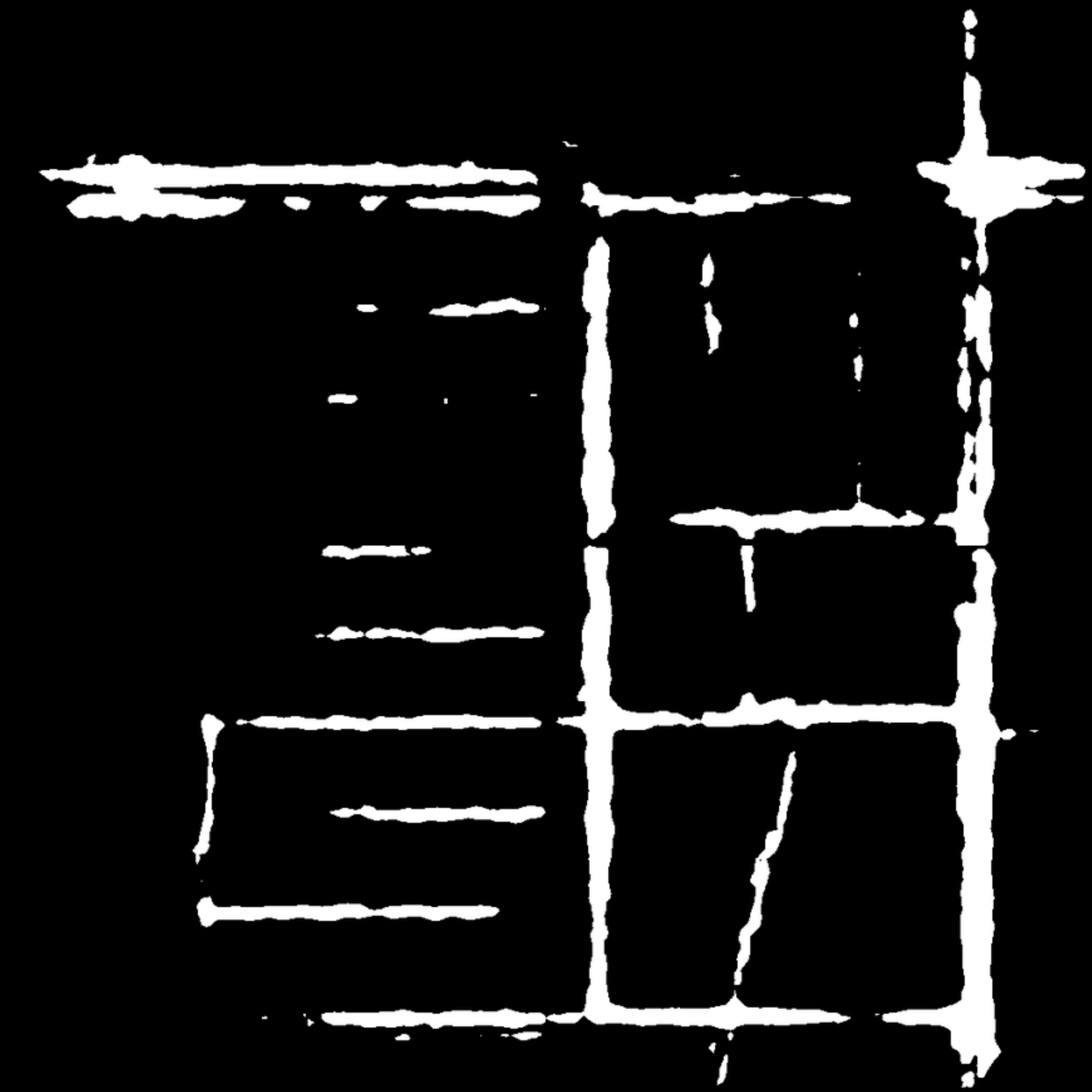}
			& 
\includegraphics[width=0.1\textwidth]{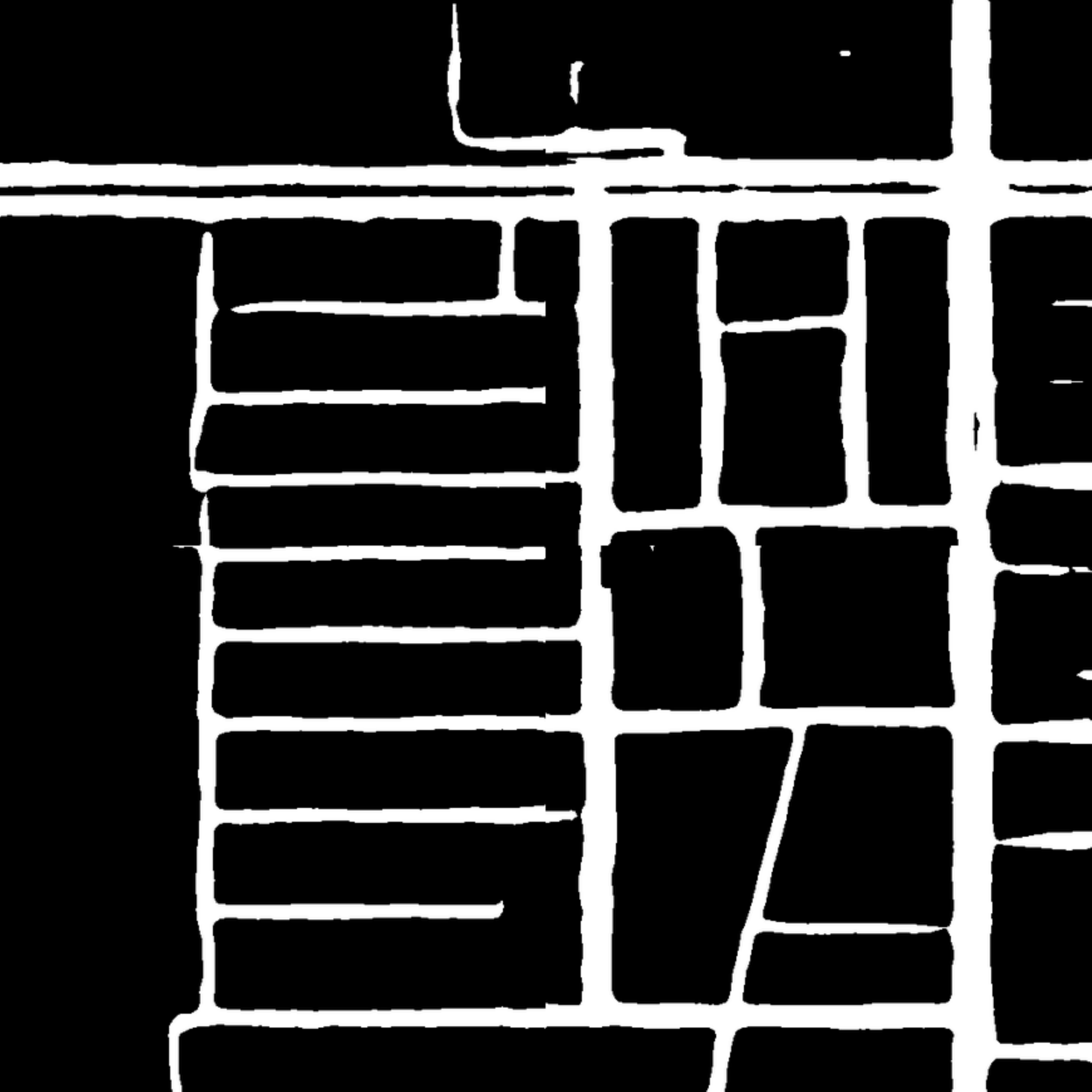}
			& 
\includegraphics[width=0.1\textwidth]{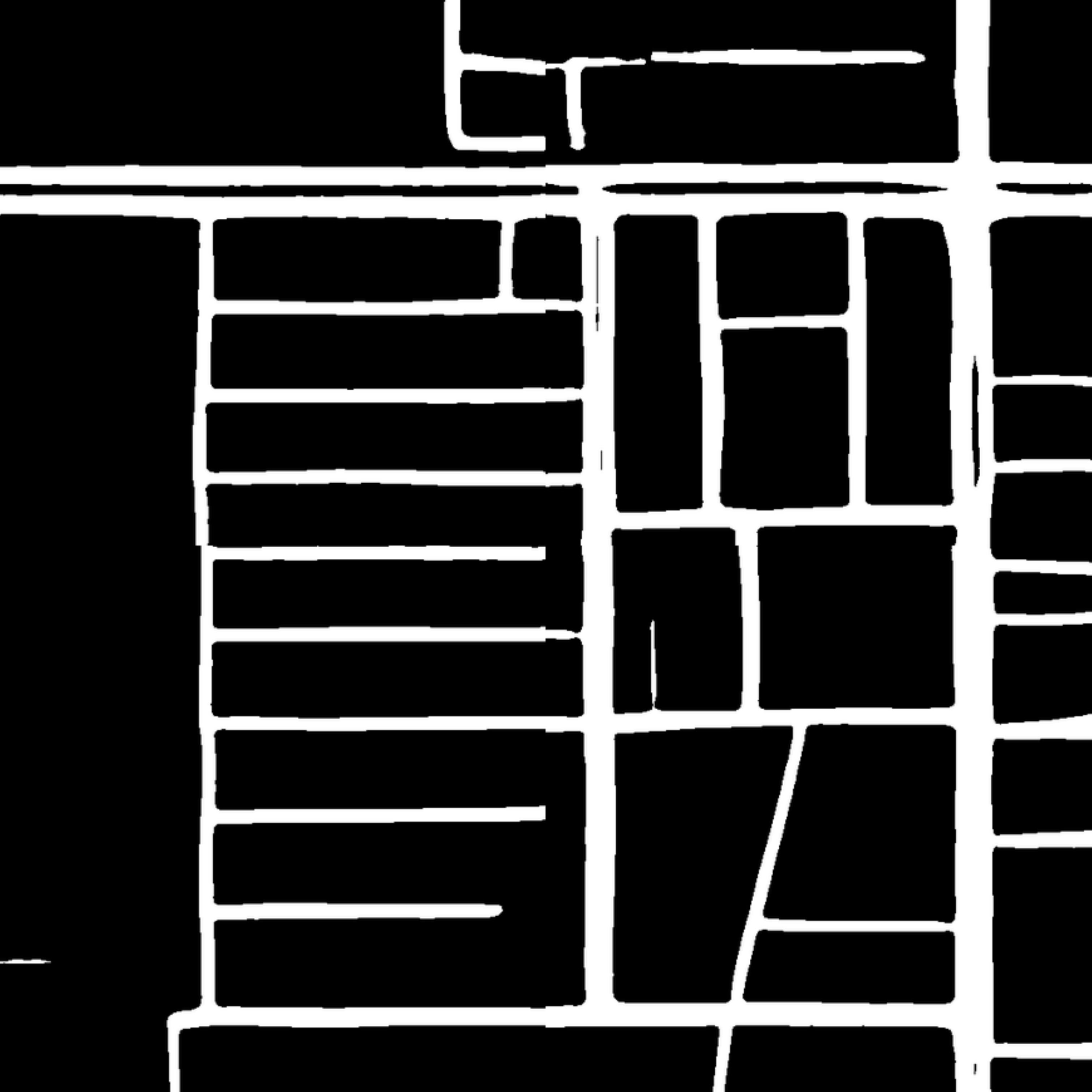}
            &
\includegraphics[width=0.1\textwidth]{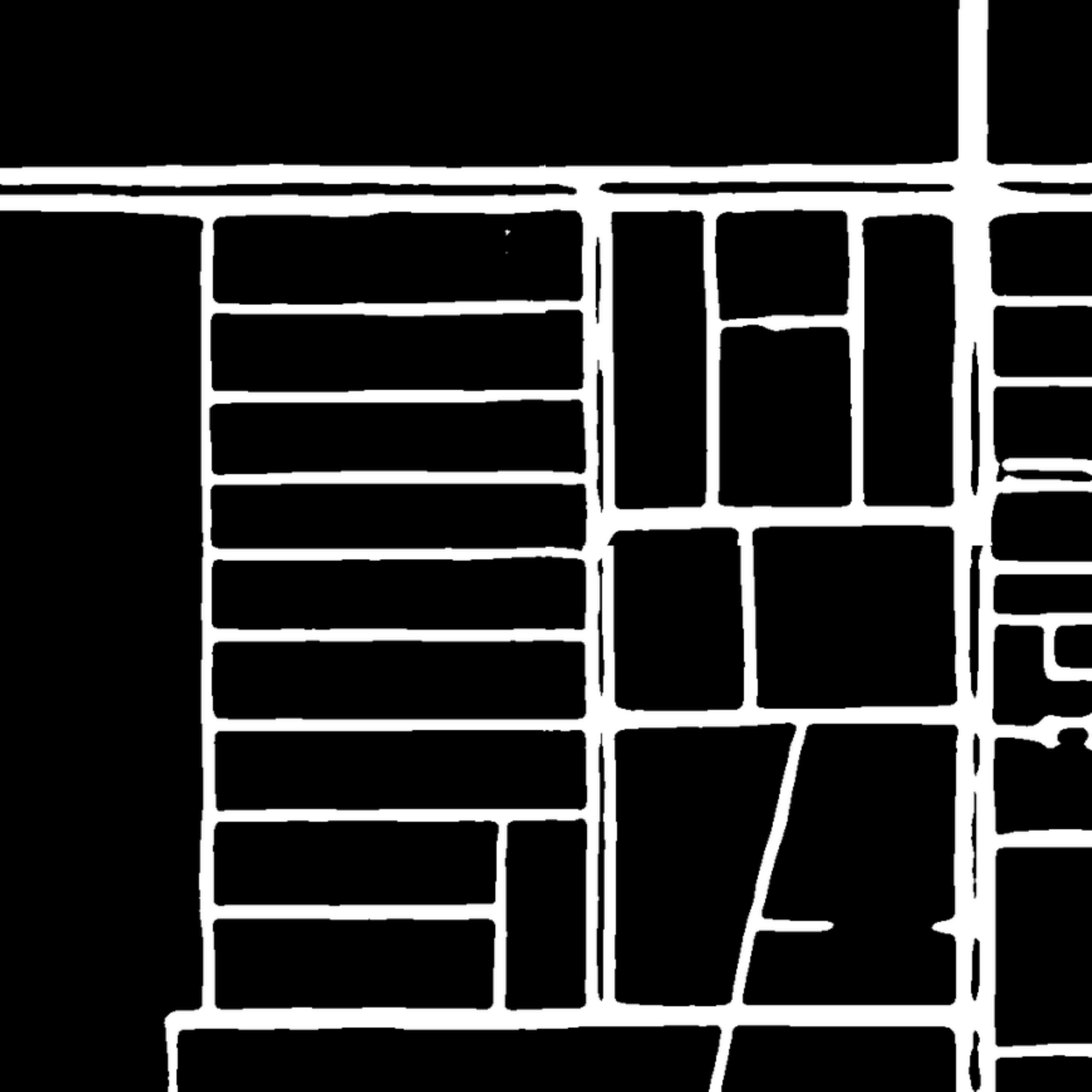}
		\\
\includegraphics[width=0.1\textwidth]{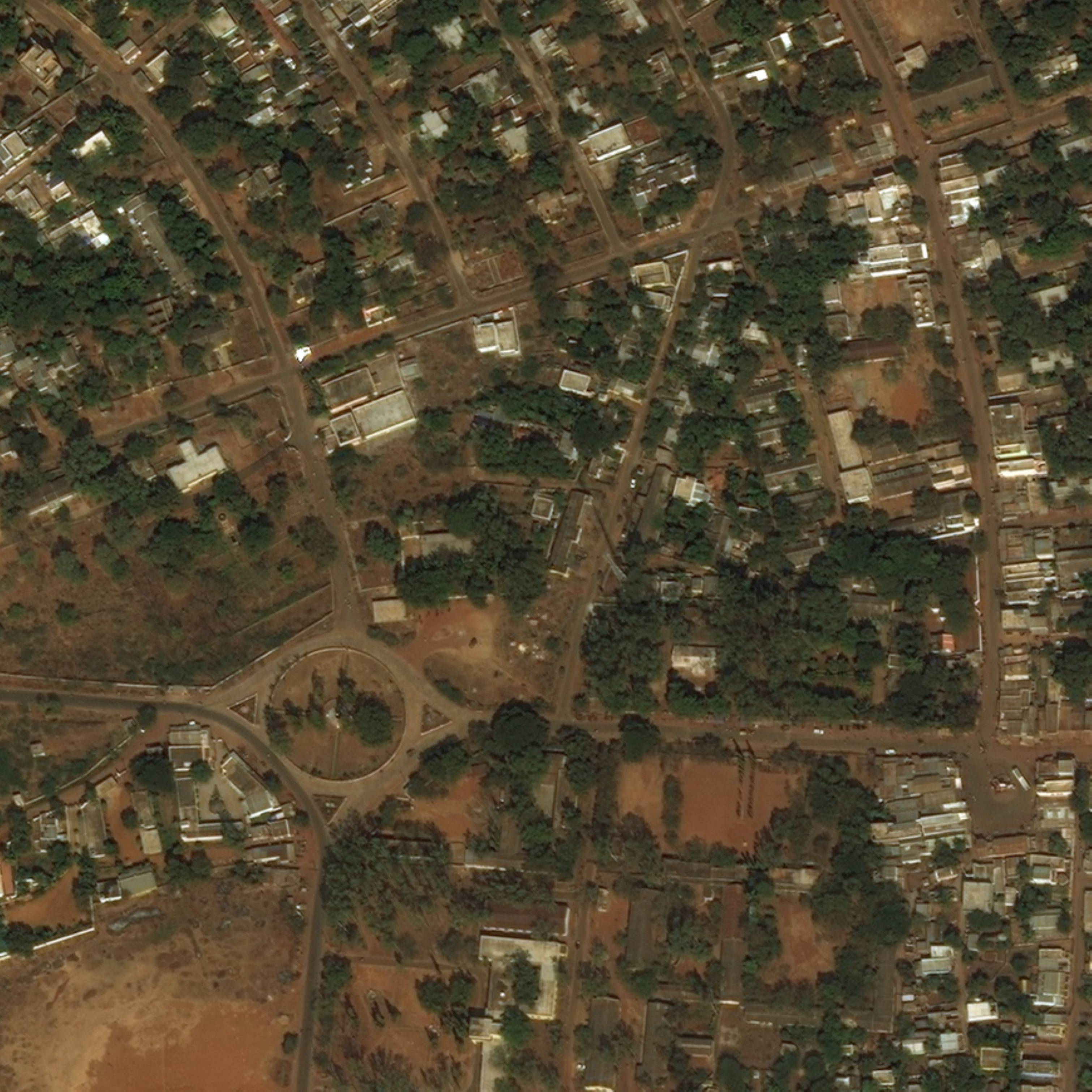}
			& 
\includegraphics[width=0.1\textwidth]{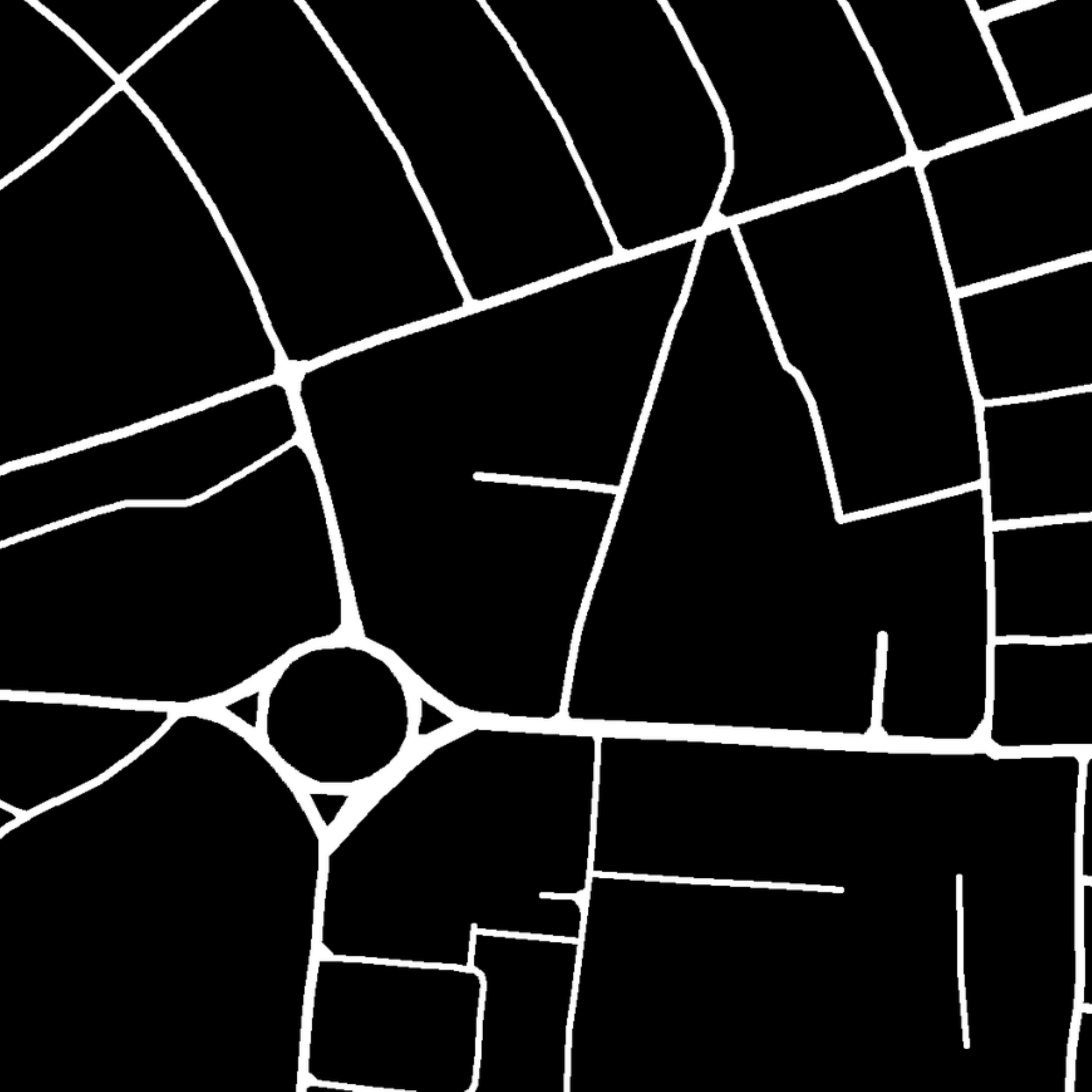}
			& 
\includegraphics[width=0.1\textwidth]{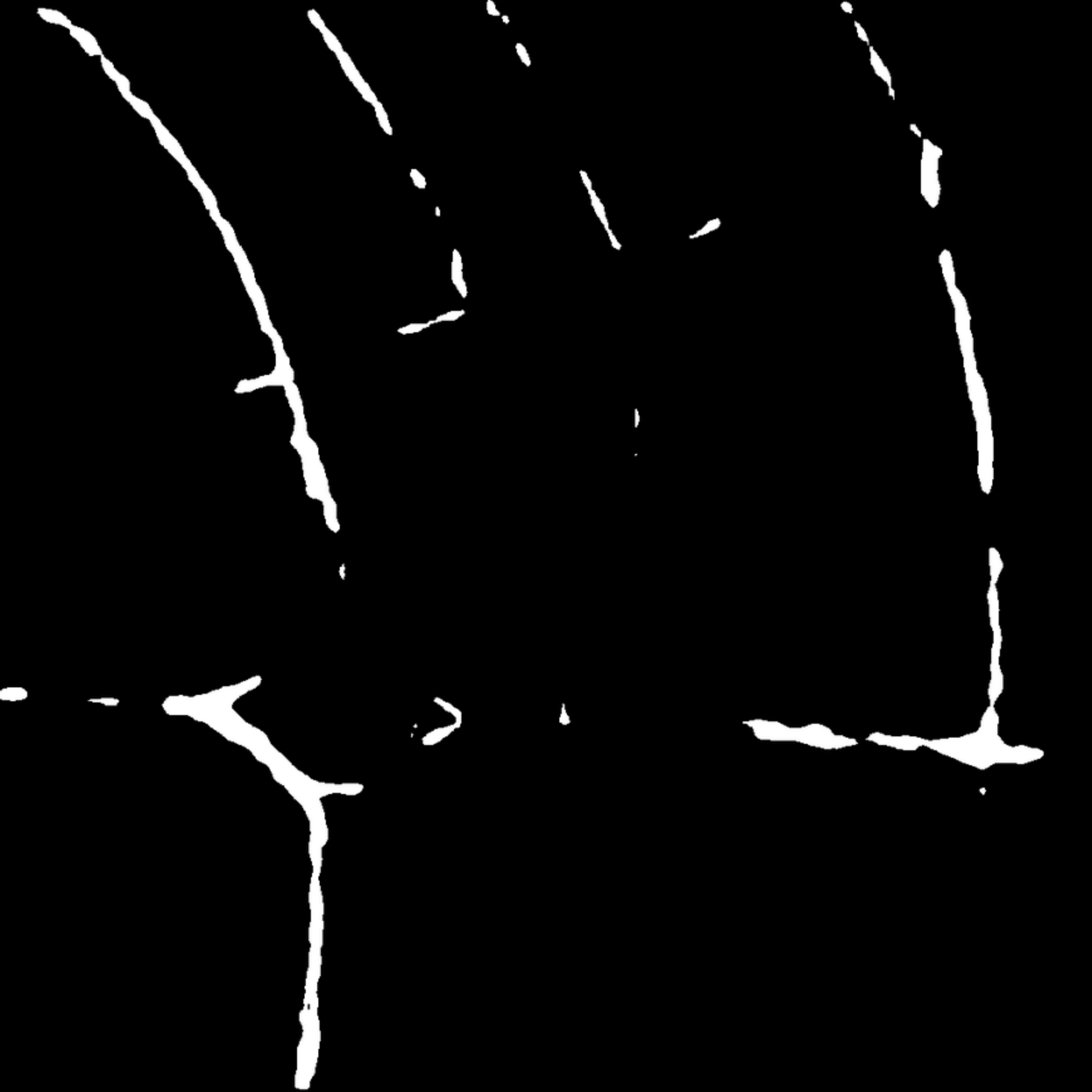}
			& 
\includegraphics[width=0.1\textwidth]{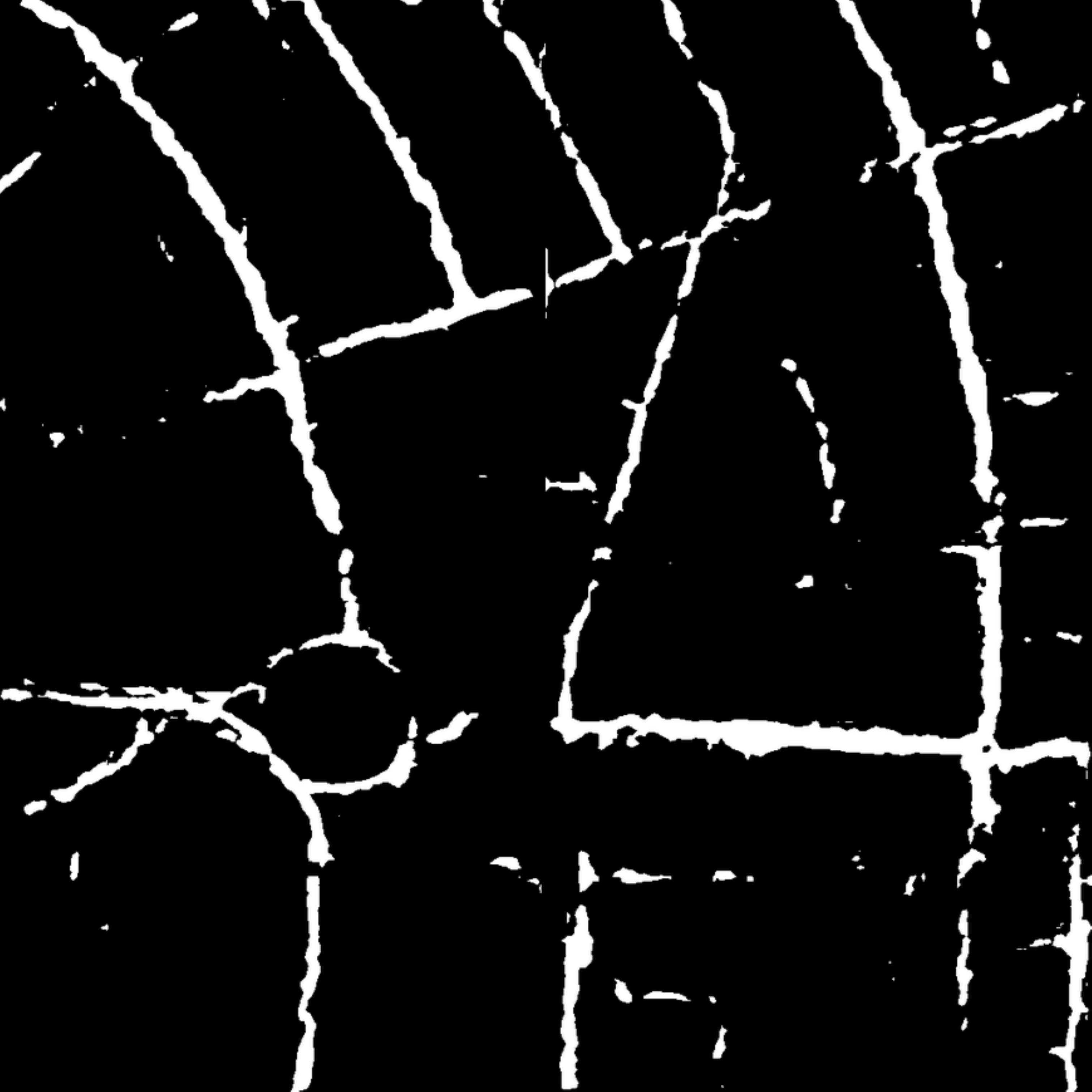}
			& 
\includegraphics[width=0.1\textwidth]{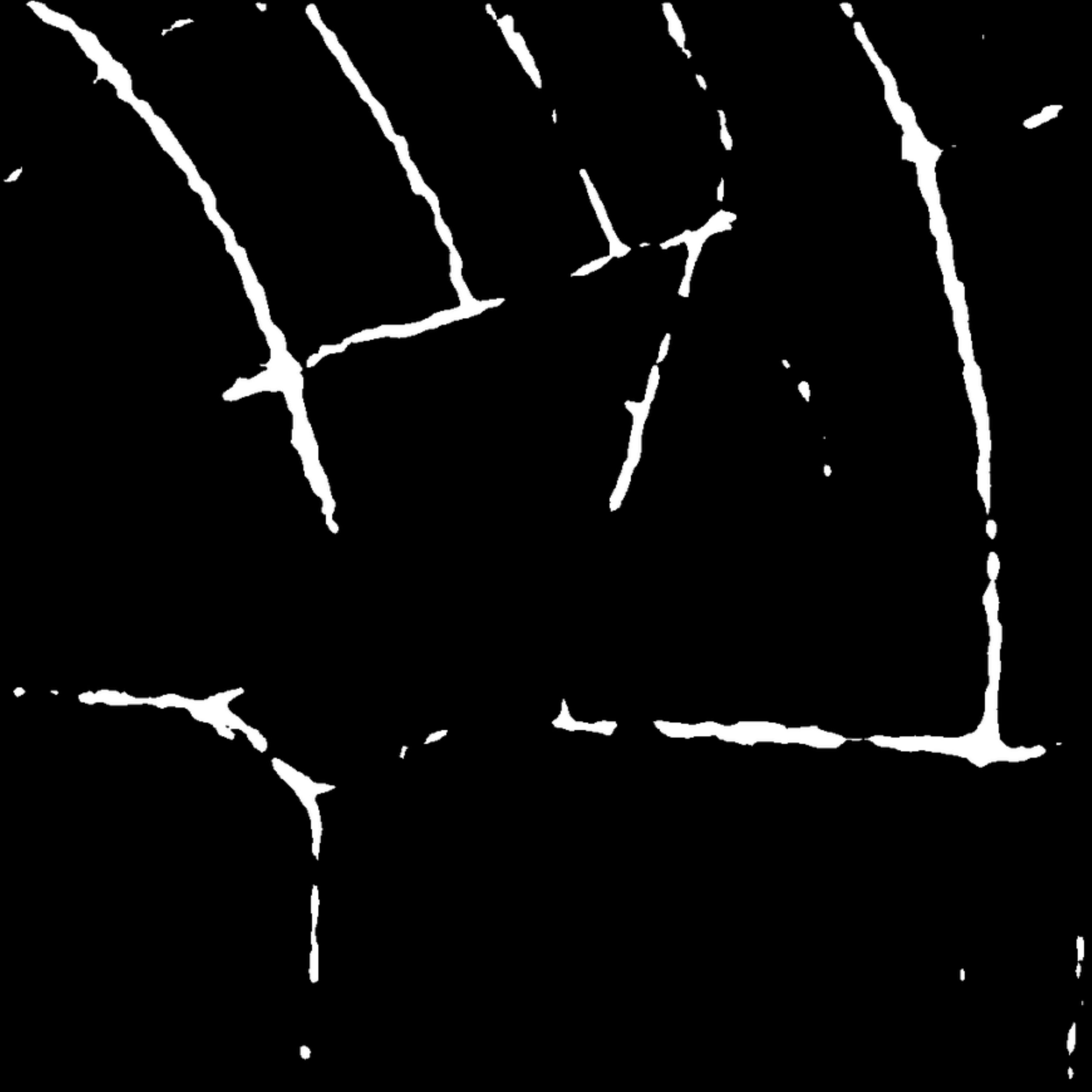}
			& 
\includegraphics[width=0.1\textwidth]{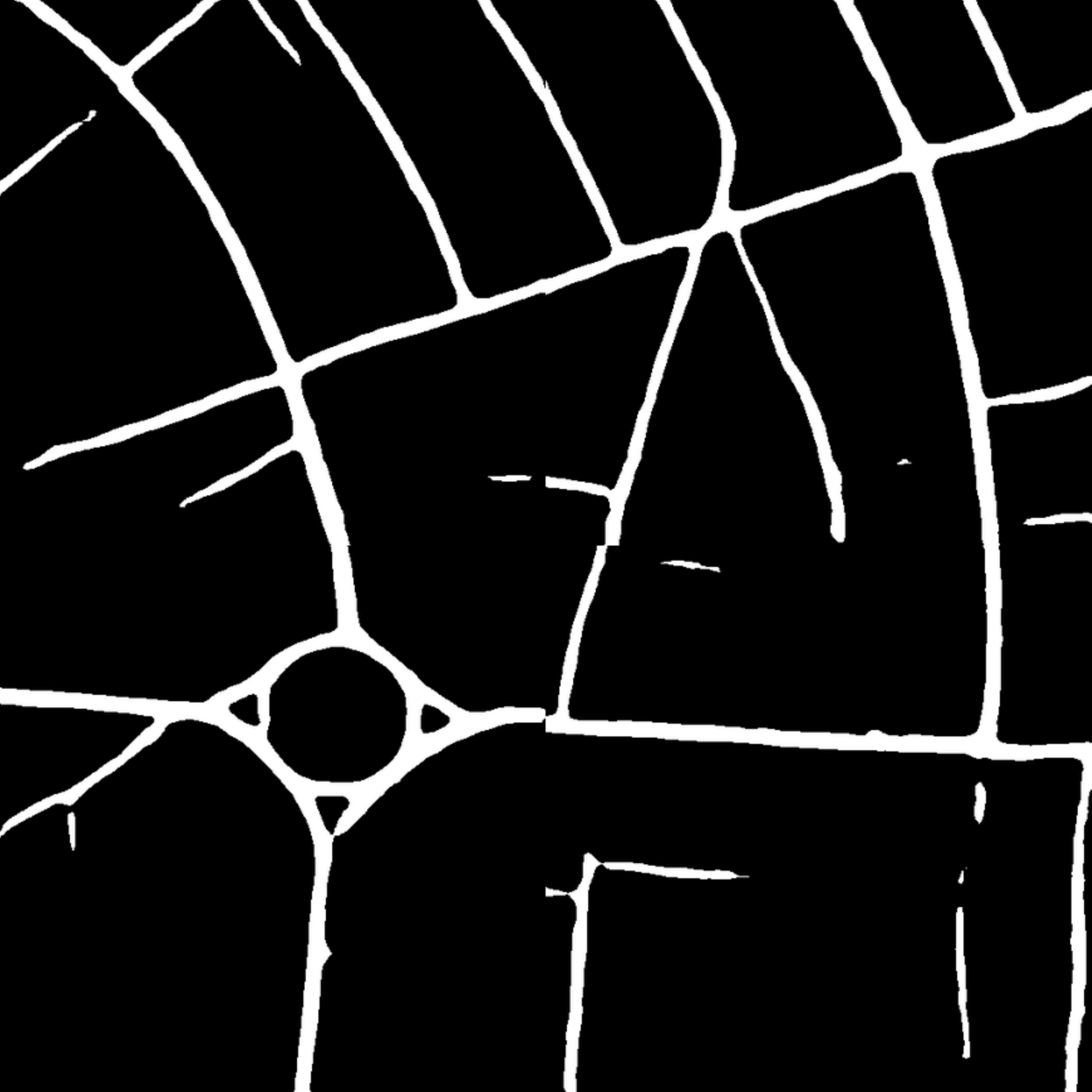}
			& 
\includegraphics[width=0.1\textwidth]{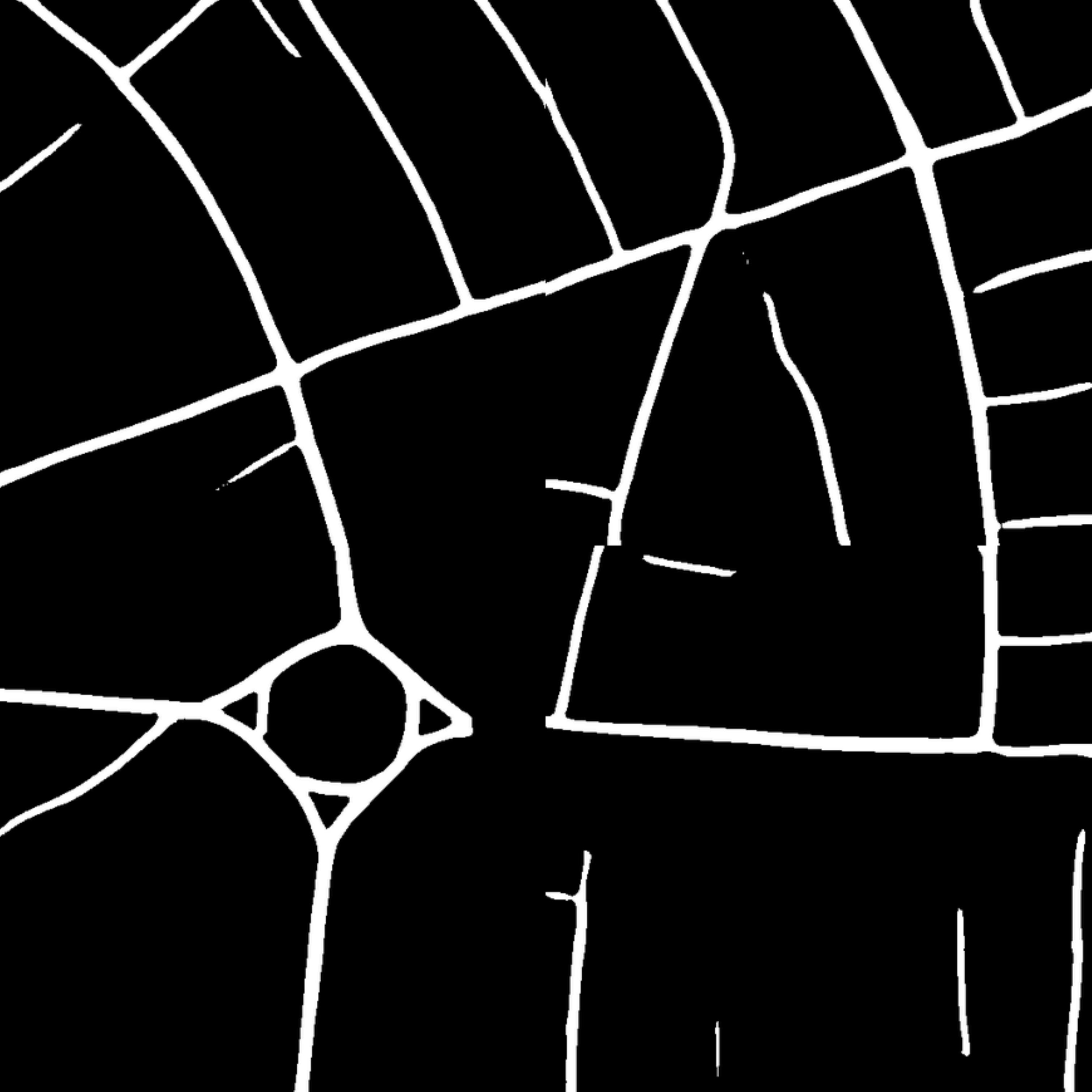}
            &
\includegraphics[width=0.1\textwidth]{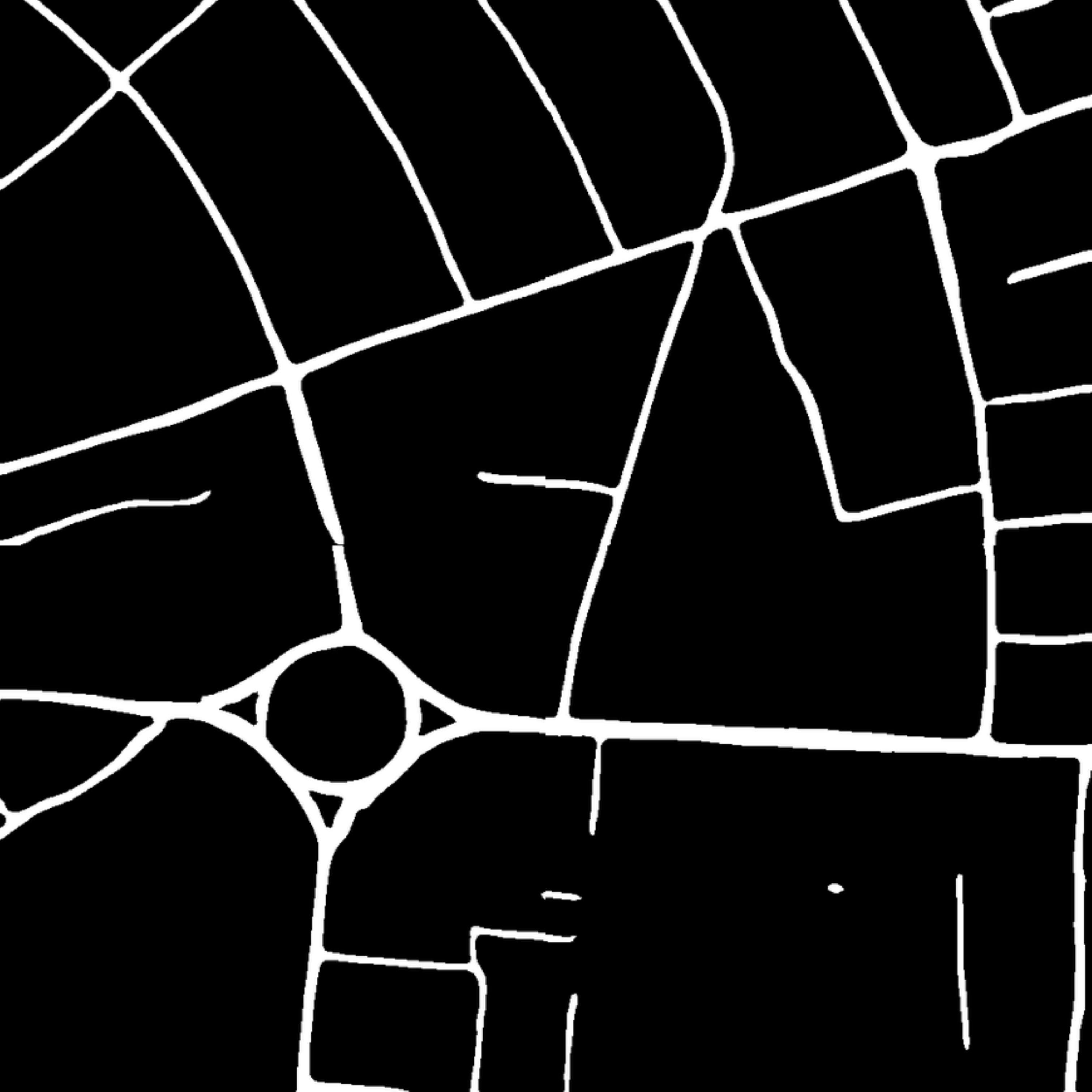}
		\\
(a)Image & (b)GT & (c) Adv & (d) s4l & (e) gct & (f) reco & (g) \makecell[c]{SRUNet  \\ (without map) }  & (h)\makecell[c]{SRUNet  \\ (Ours) } 
\end{tabular}%
\caption{Visual comparison on the DeepGlobe Road dataset. (a) Original images. (b) Ground truth. (c) Adv. (d) s4l. (e) gct. (f) reco. (g) SRUNet(without map prior). (h) SRUNet.} 
\label{vis_deep_com} 
\end{figure*}

\tabref{tab_deep_com} shows the comparative results on DeepGlobe dataset. The IoU scores of our model were 20\% and 14\% higher than those of the DeeplabV3 supervised method and ReCo semi-supervised method, respectively. In particular, our model achieved the highest recall value, meaning that it better recognized missing roads and achieved more complete results. 

The visualization results of the models are shown in \figref{vis_deep_com}, where the semi-supervised methods are generally superior to the supervised methods for the same amount of training data. Across the results of the semi-supervised methods, the models were found to perform better in dense road areas than in sparse road areas, and our model could better detect the spatial details in the boundary regions. As for the completeness of the prediction results, common problems with the proposed methods include false and missing detections. As shown in \figref{vis_deep_com}, roads covered by buildings and trees could hardly be detected using limited training data. Our proposed model thus shows better classification ability, indicating its stronger feature representation ability.

This demonstrates that, A) semi-supervised learning methods are superior to supervised learning methods. As supervised learning methods do not consider the additional auxiliary information of the unlabeled data, they have poor generalization ability. B) Our model considers the geometry information and uses a contrastive learning loss to strengthen classification ability. Experiments show that the steps mentioned above have a positive effect on the model performance.

\subsubsection{Comparative results on the self-constructed dataset}
\begin{figure*}[tb]
\small
   \centering
		\newcommand{\tabincell}[2]{\begin{tabular}{@{}#1@{}}#2\end{tabular}}
		\begin{tabular}{cccccccc}
\multirow{3}*{\rotatebox[origin=c]{90}{\scriptsize Images from Nanjing, Jiangsu Province}} 
            &
\includegraphics[width=0.12\textwidth]{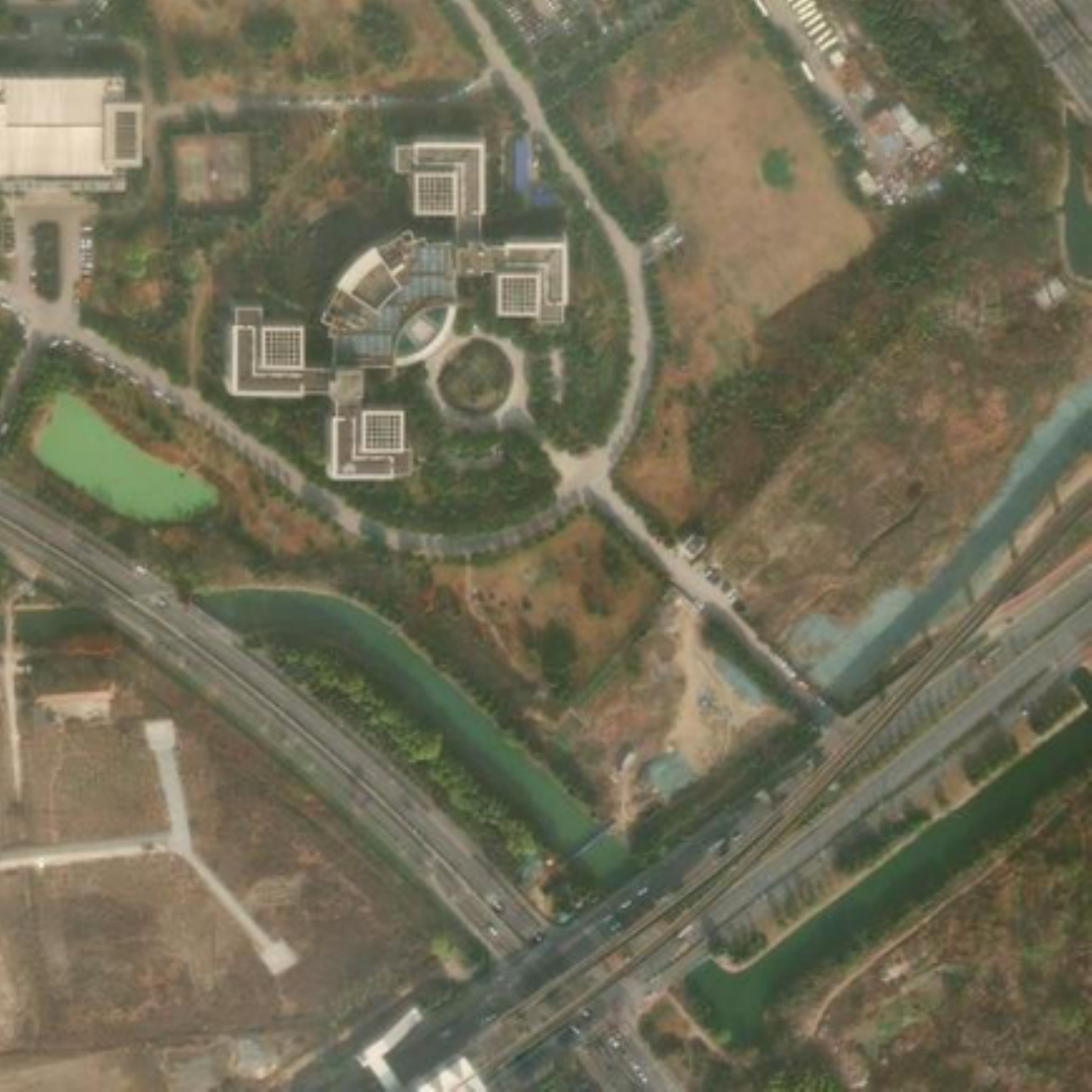}
			& 
\includegraphics[width=0.12\textwidth]{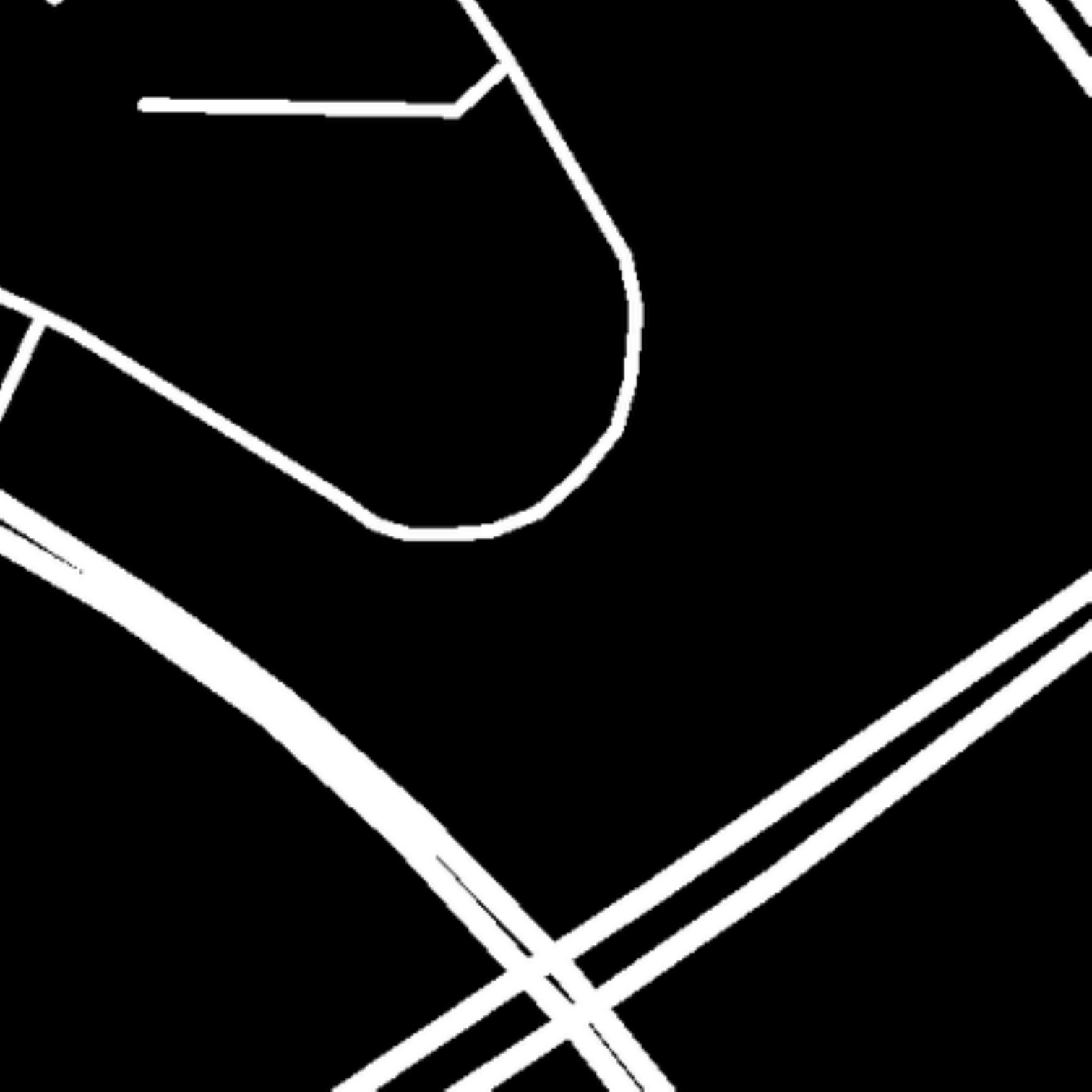}
			& 
\includegraphics[width=0.12\textwidth]{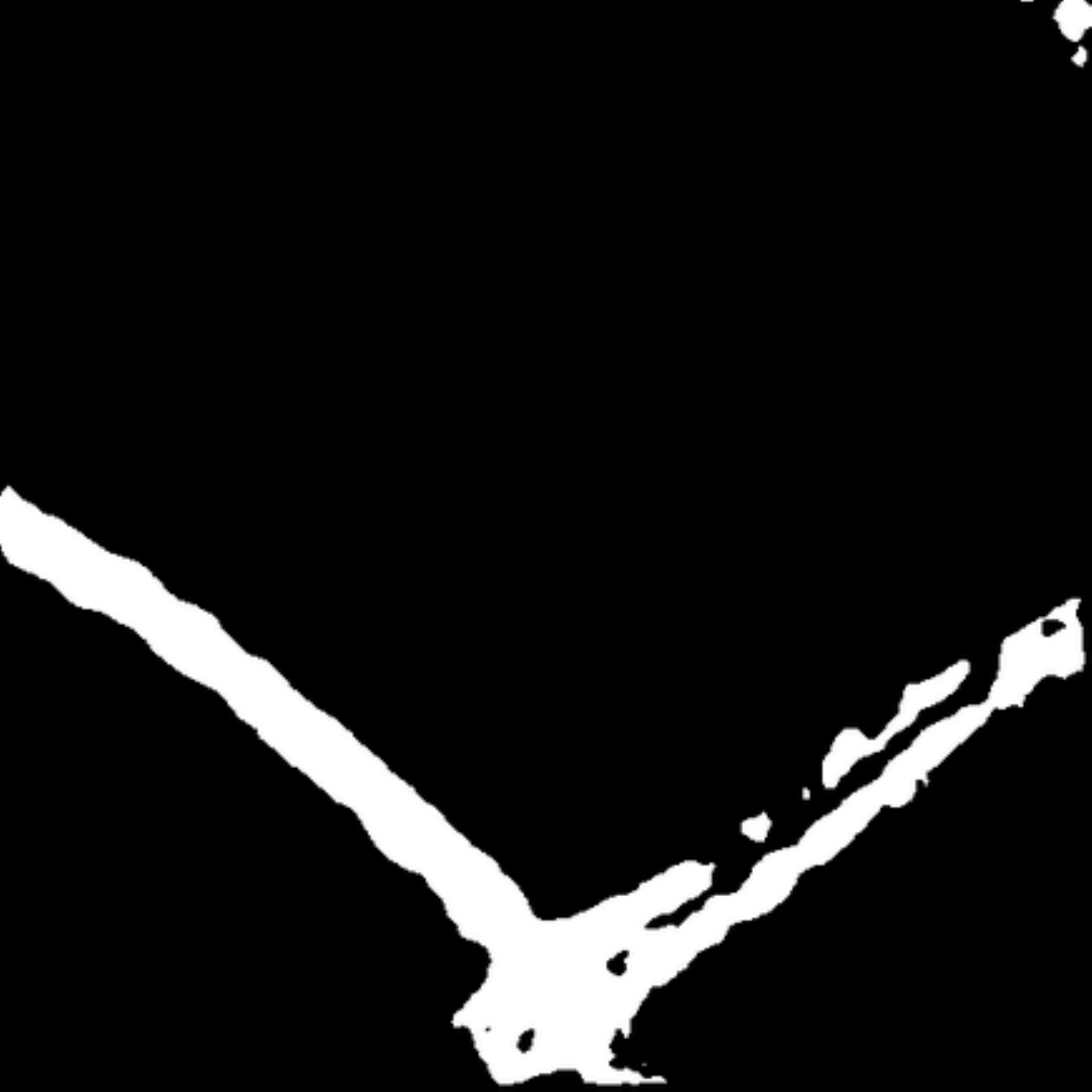}
			& 
\includegraphics[width=0.12\textwidth]{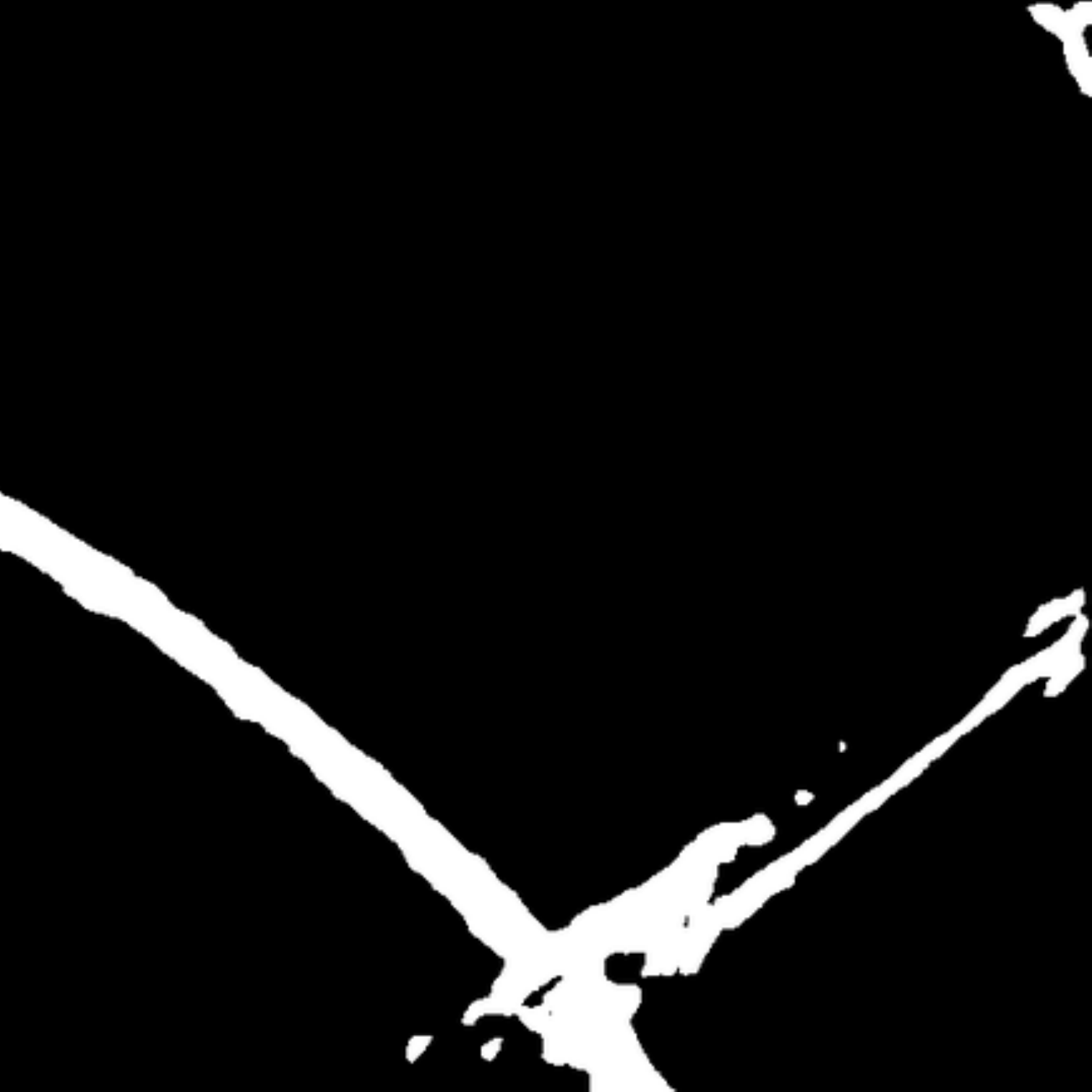}
			& 
\includegraphics[width=0.12\textwidth]{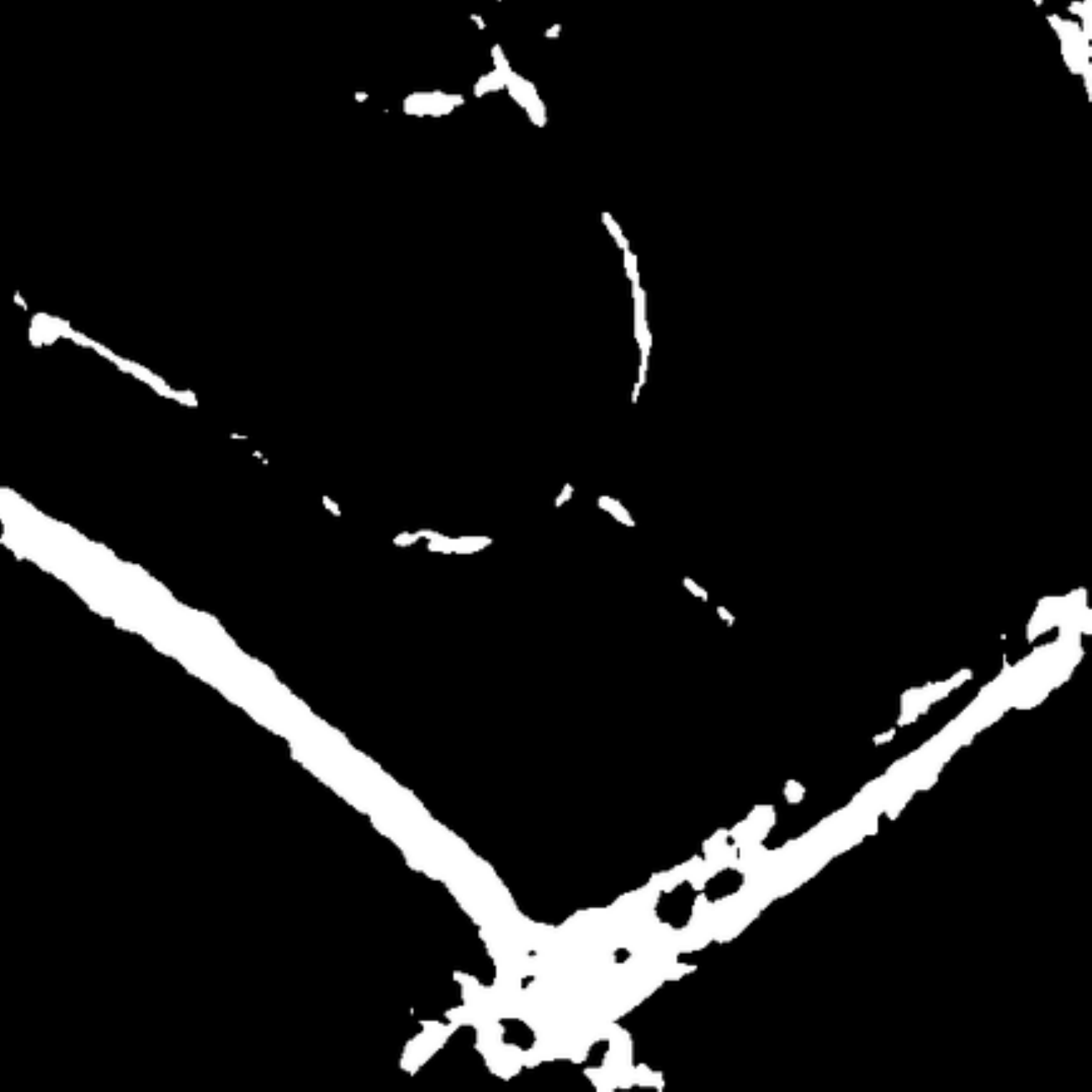}
			& 
\includegraphics[width=0.12\textwidth]{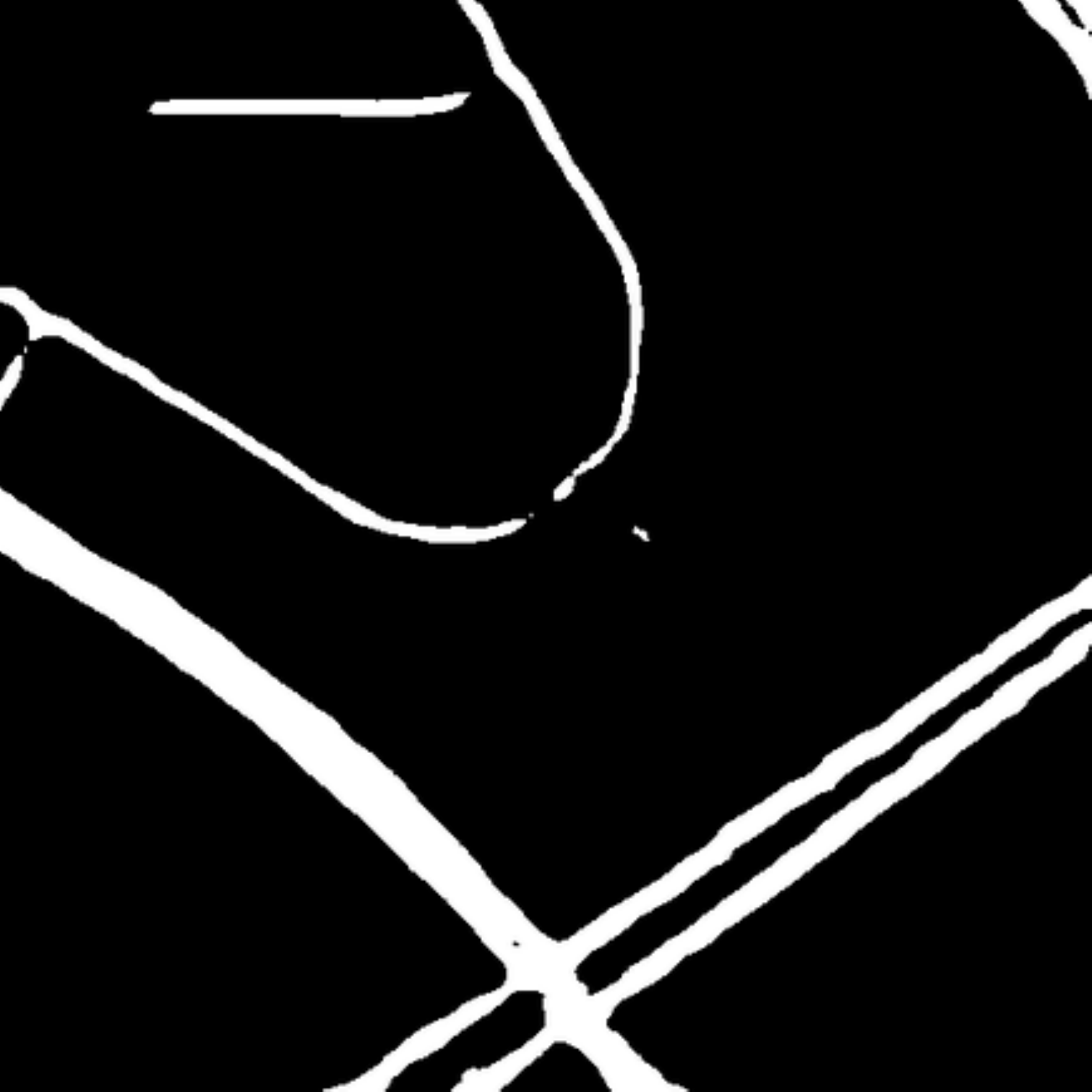}
			& 
\includegraphics[width=0.12\textwidth]{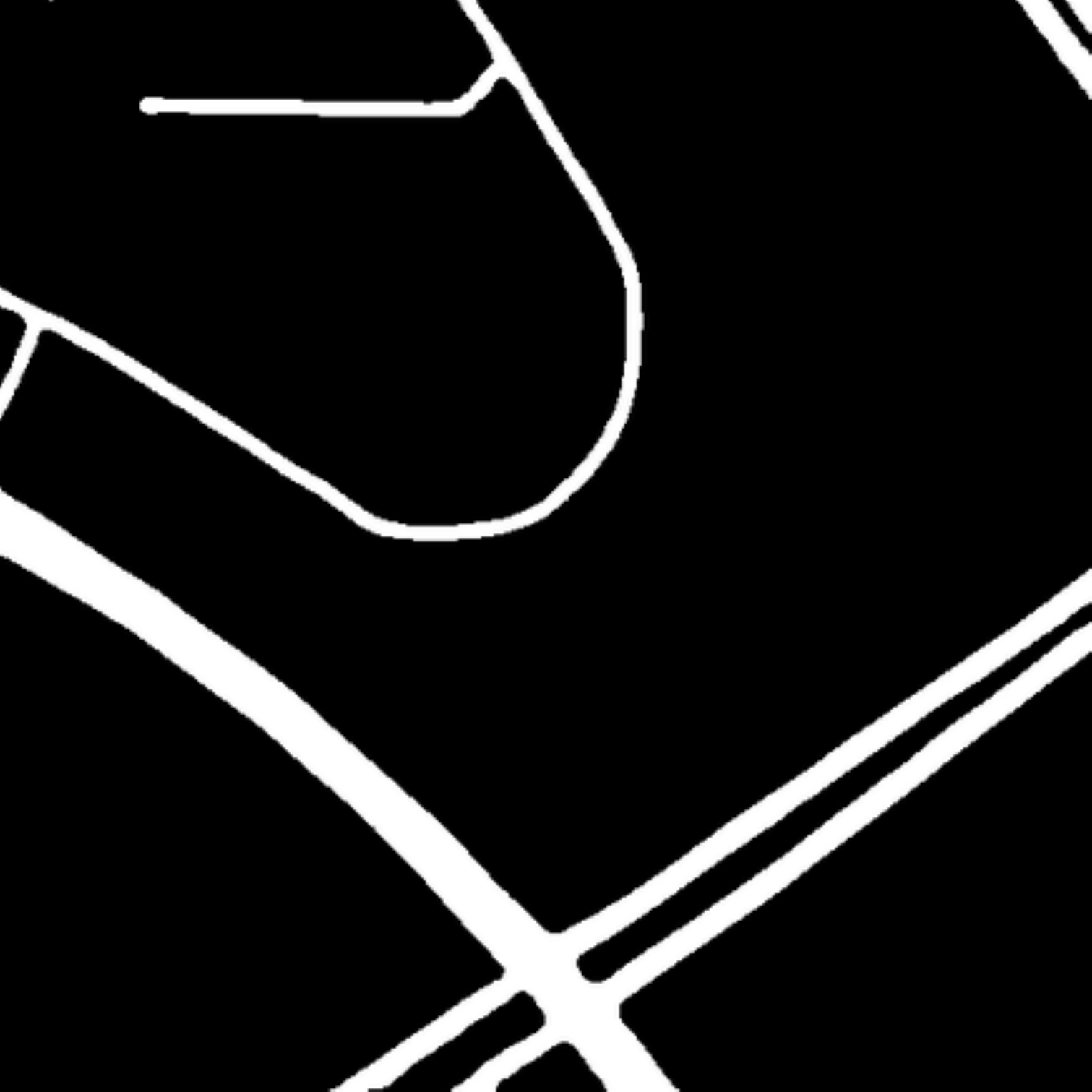}
		\\
            &
\includegraphics[width=0.12\textwidth]{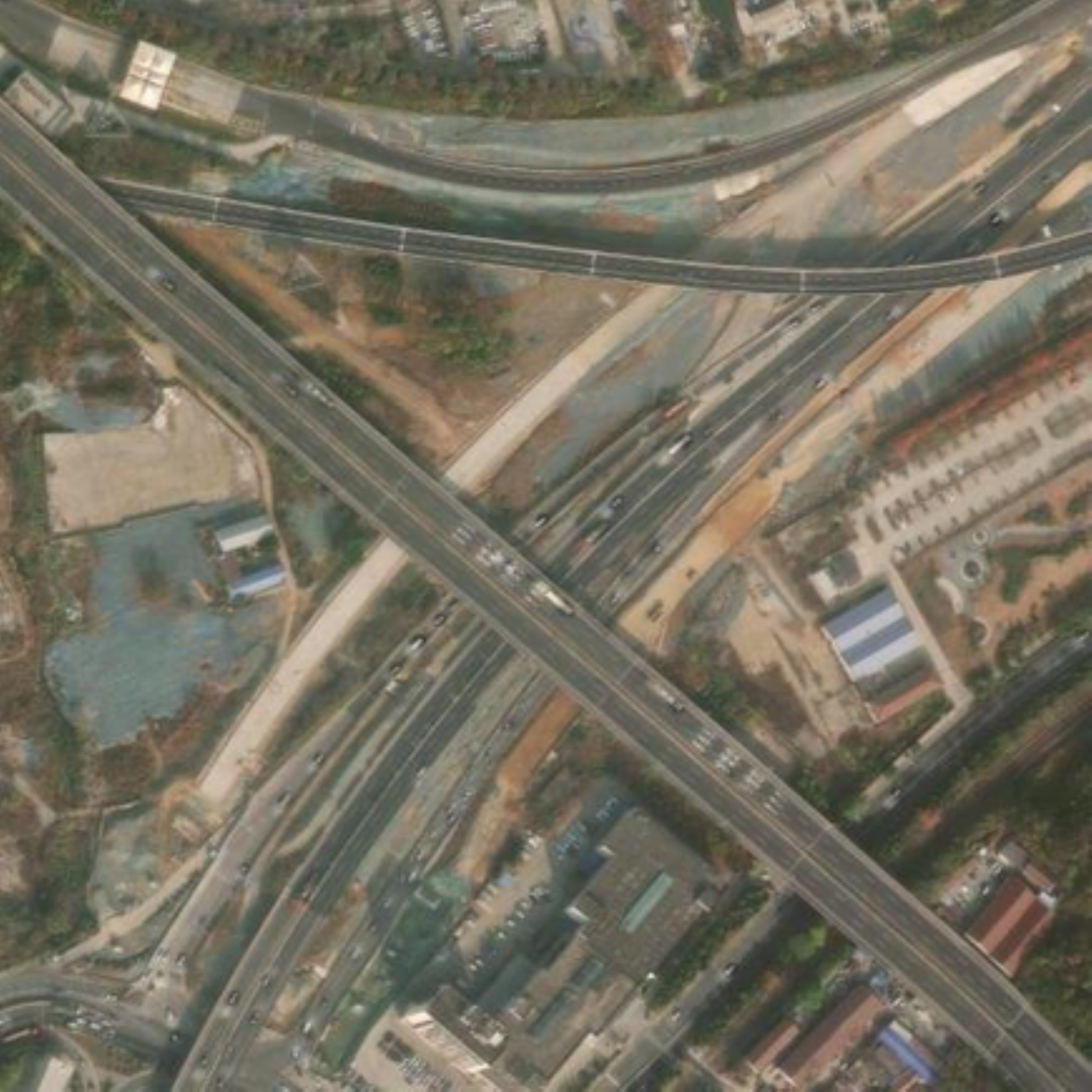}
			& 
\includegraphics[width=0.12\textwidth]{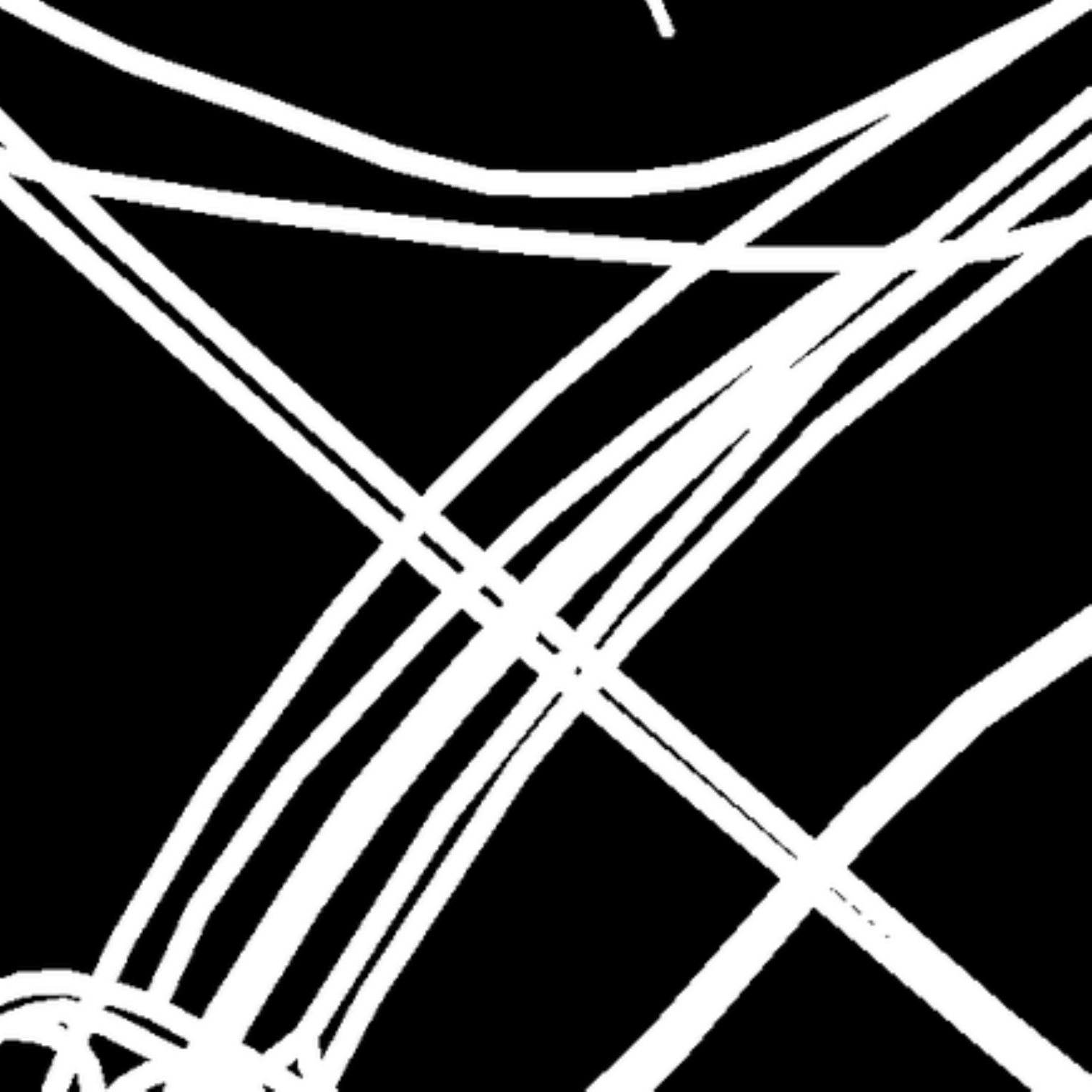}
			& 
\includegraphics[width=0.12\textwidth]{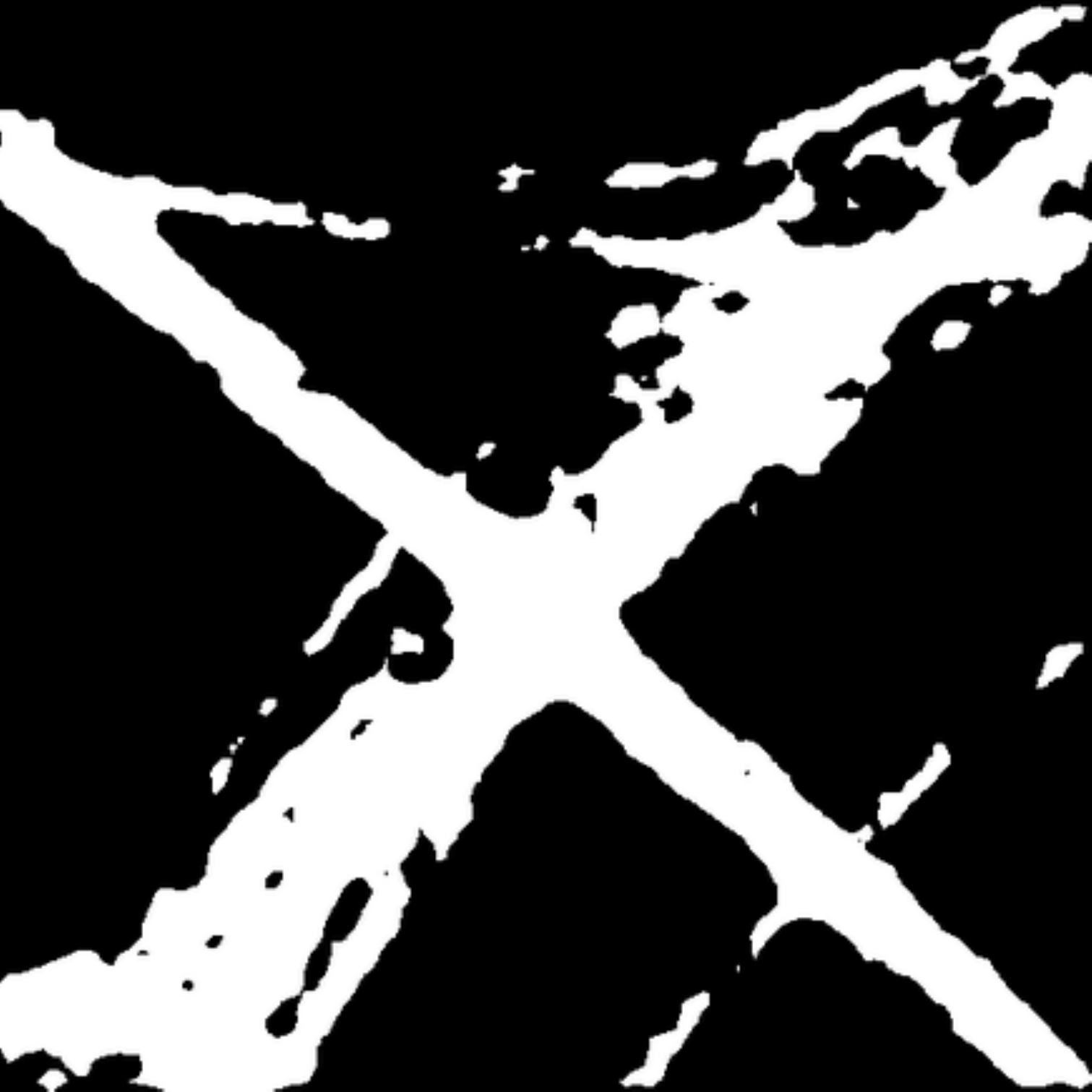}
			& 
\includegraphics[width=0.12\textwidth]{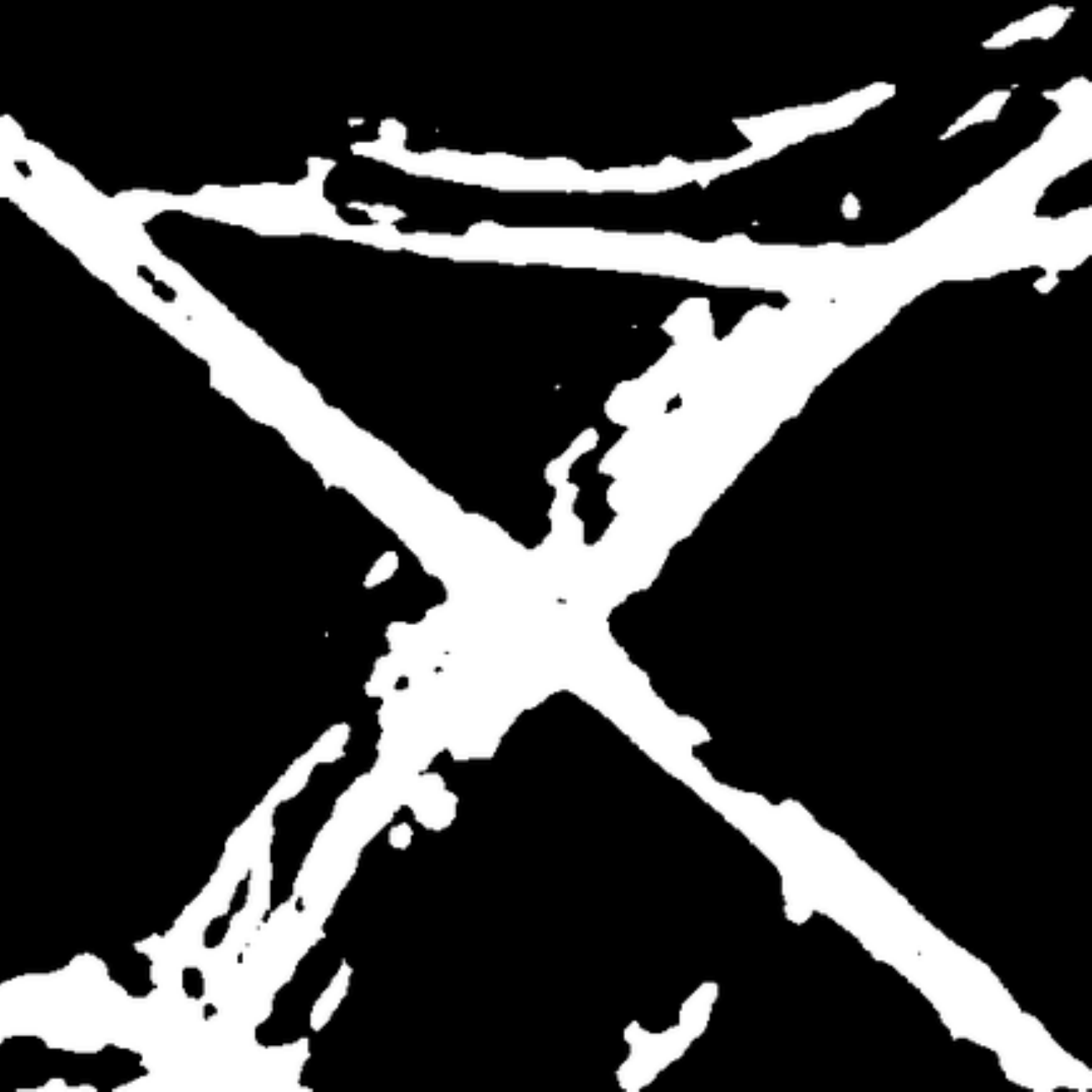}
			& 
\includegraphics[width=0.12\textwidth]{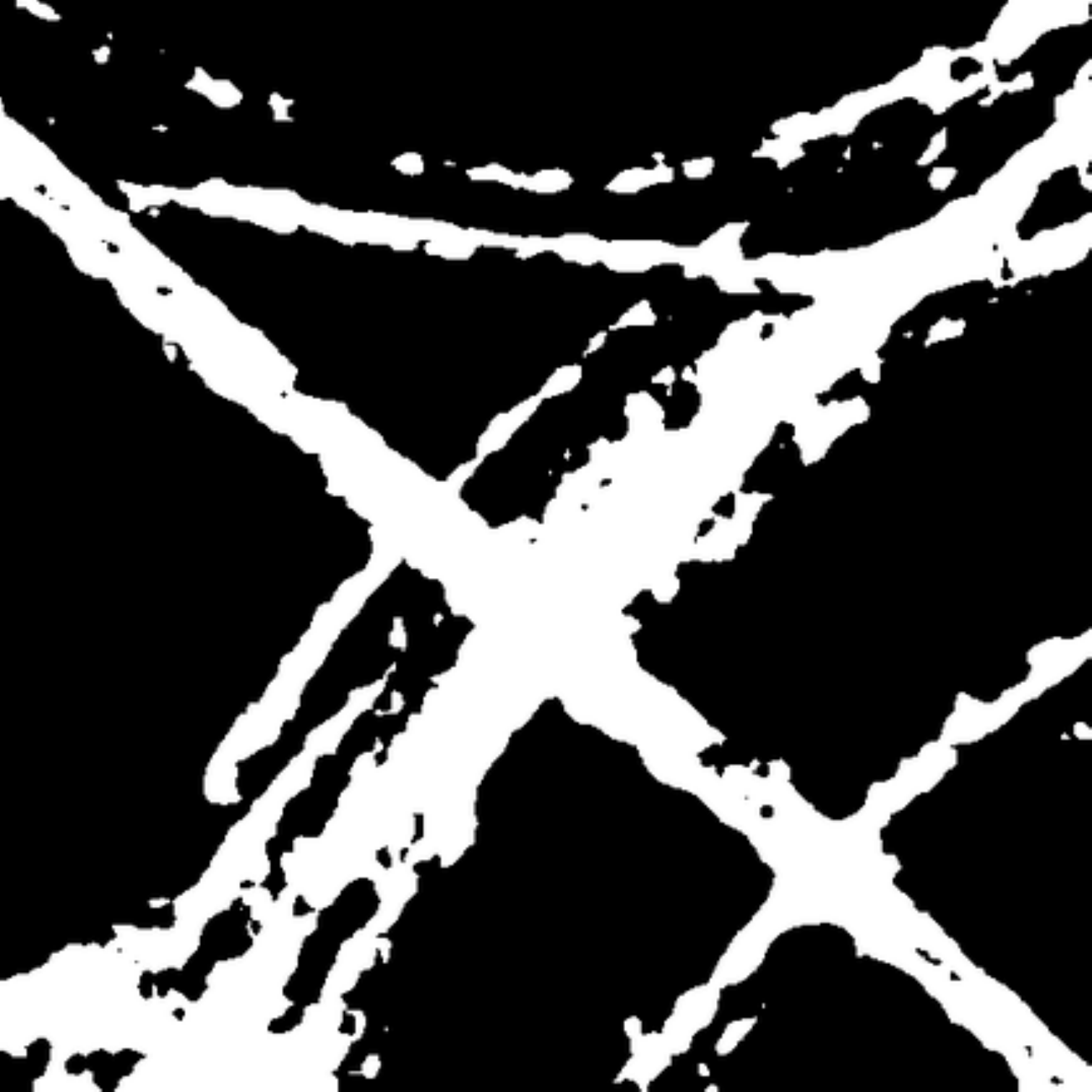}
			& 
\includegraphics[width=0.12\textwidth]{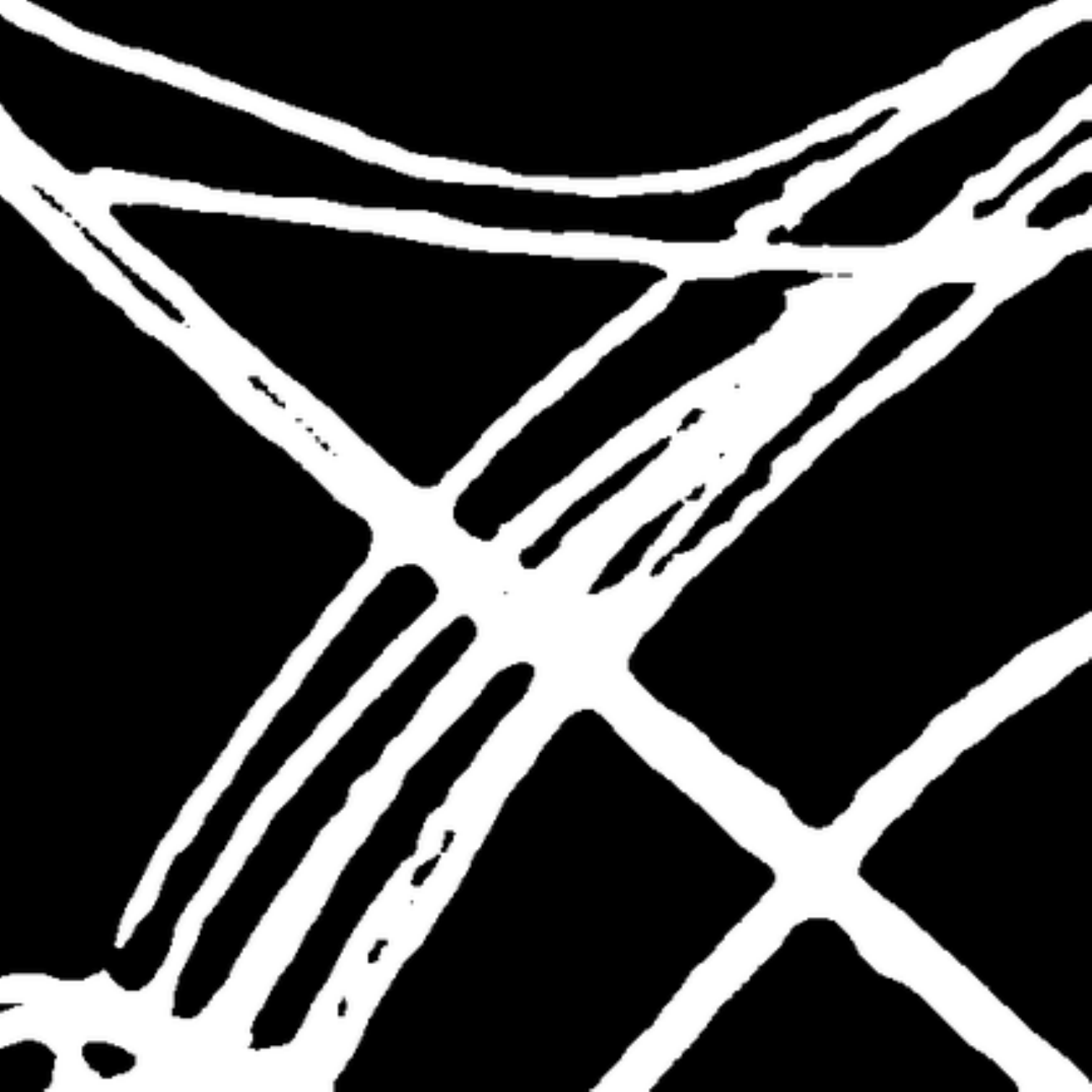}
			& 
\includegraphics[width=0.12\textwidth]{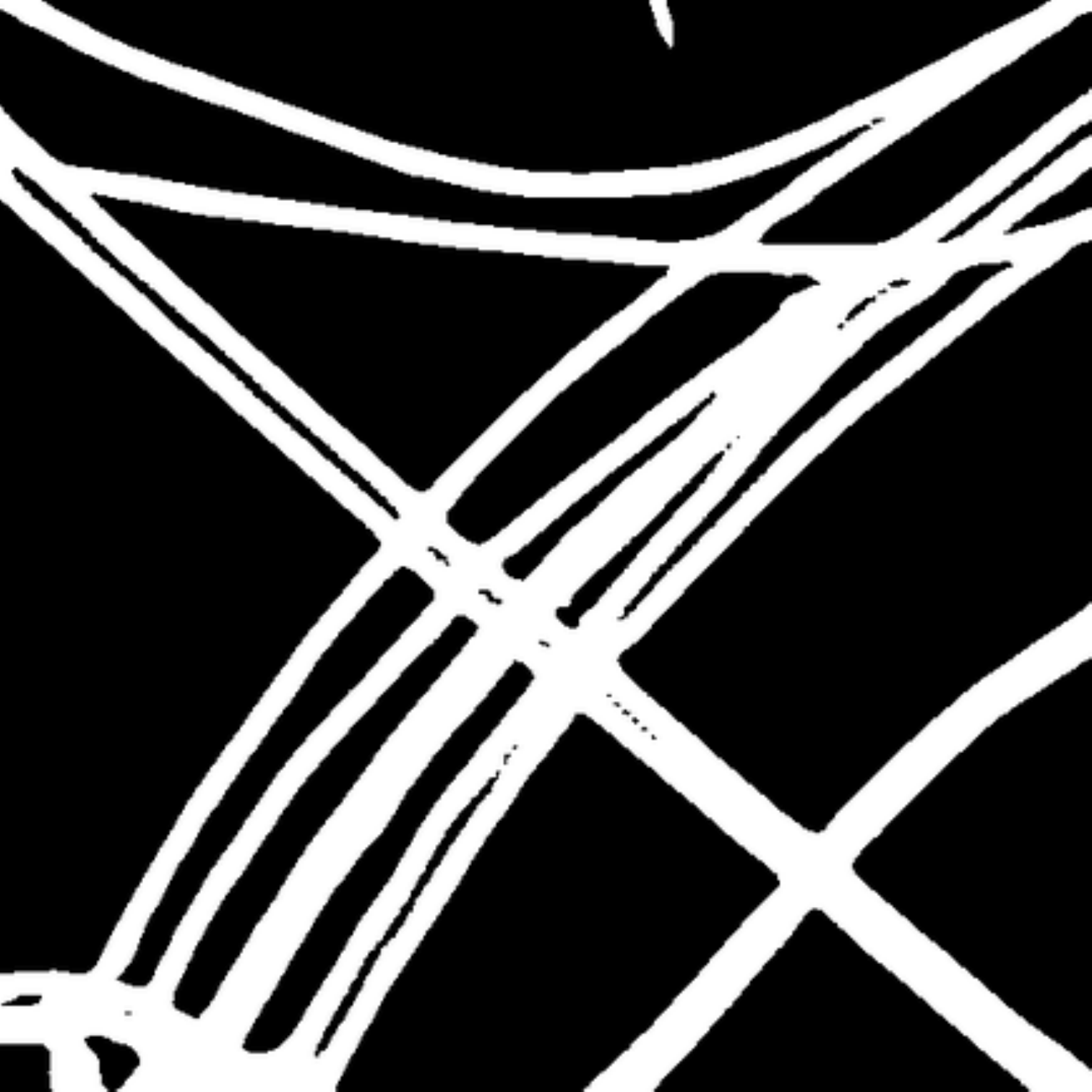}
		\\
            &
\includegraphics[width=0.12\textwidth]{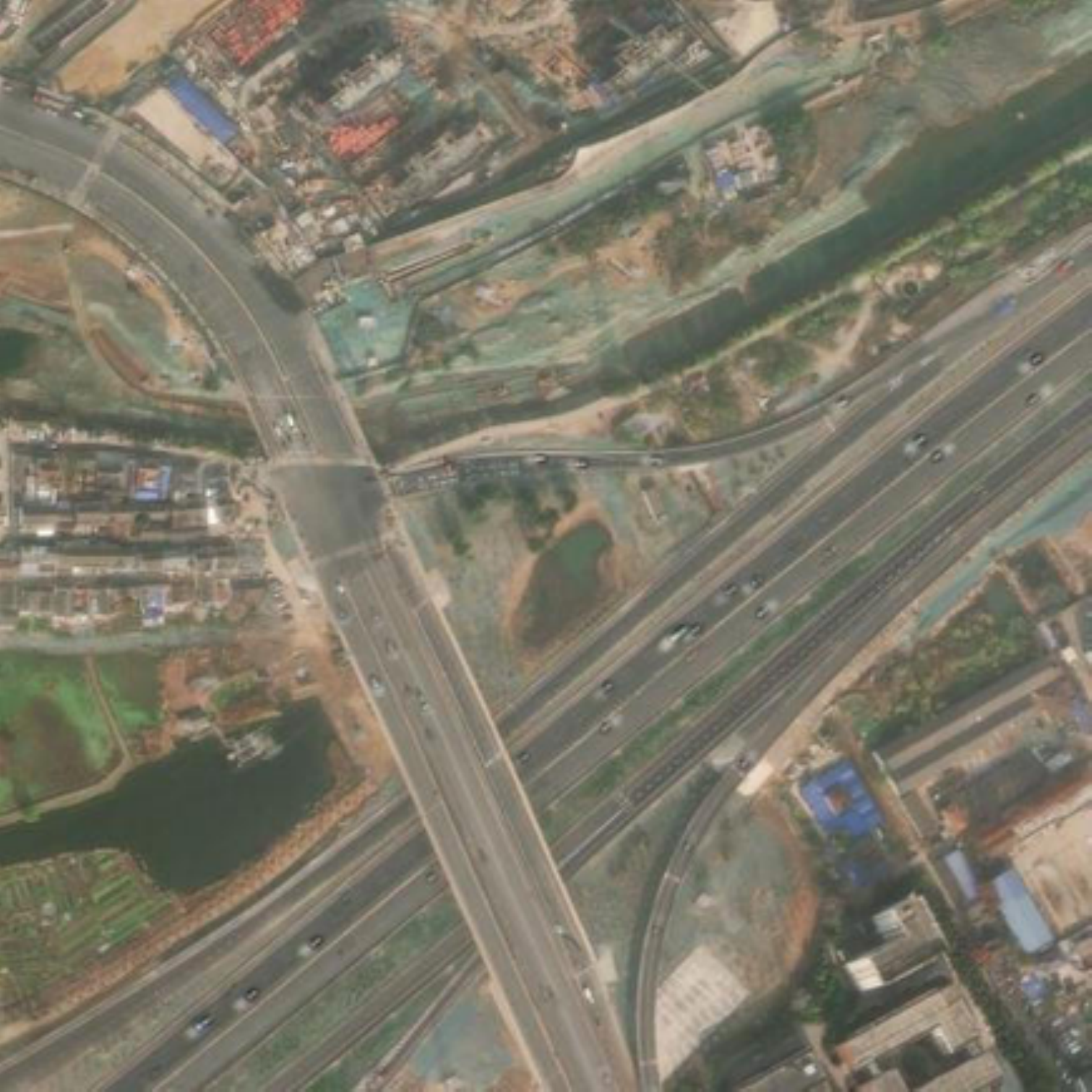}
			& 
\includegraphics[width=0.12\textwidth]{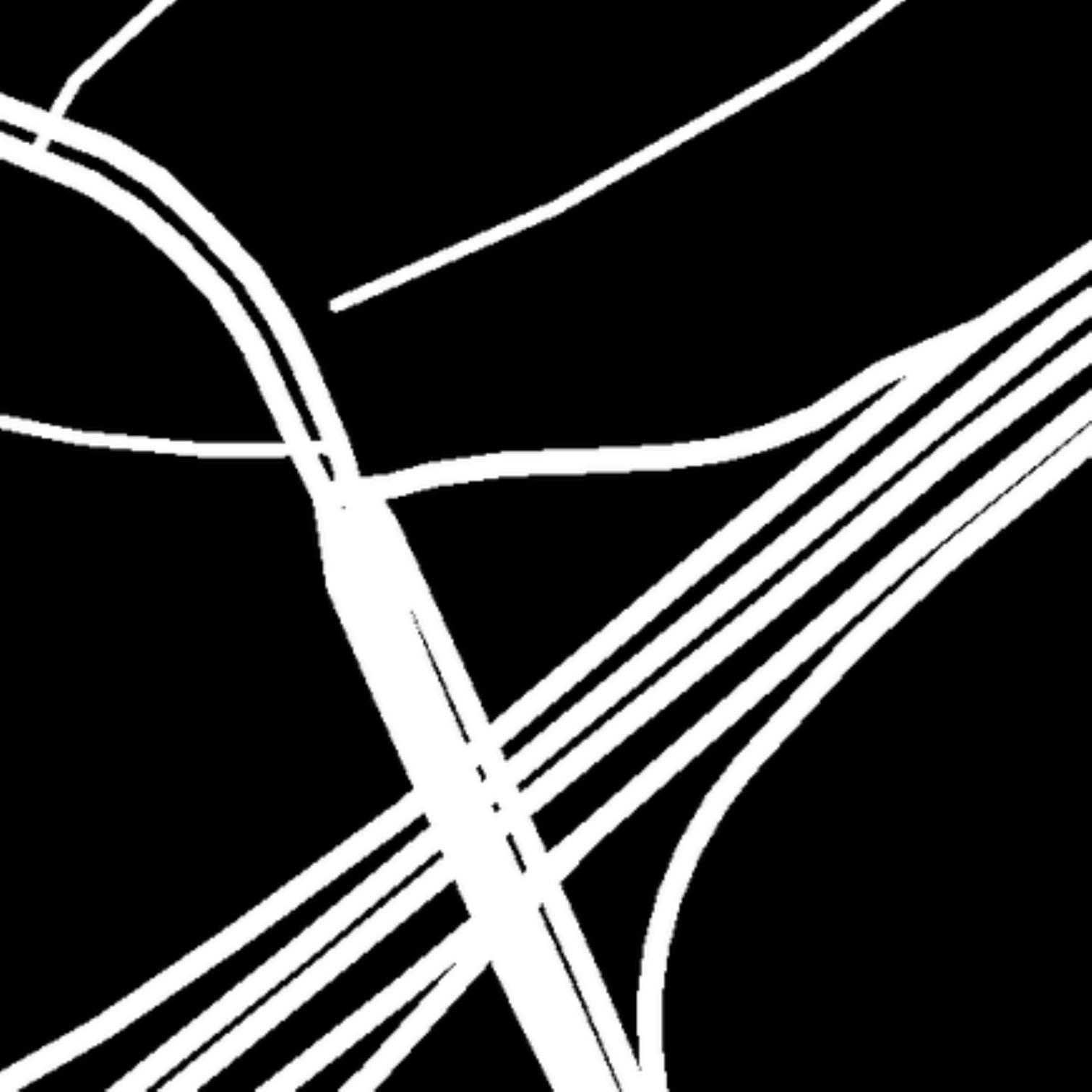}
			& 
\includegraphics[width=0.12\textwidth]{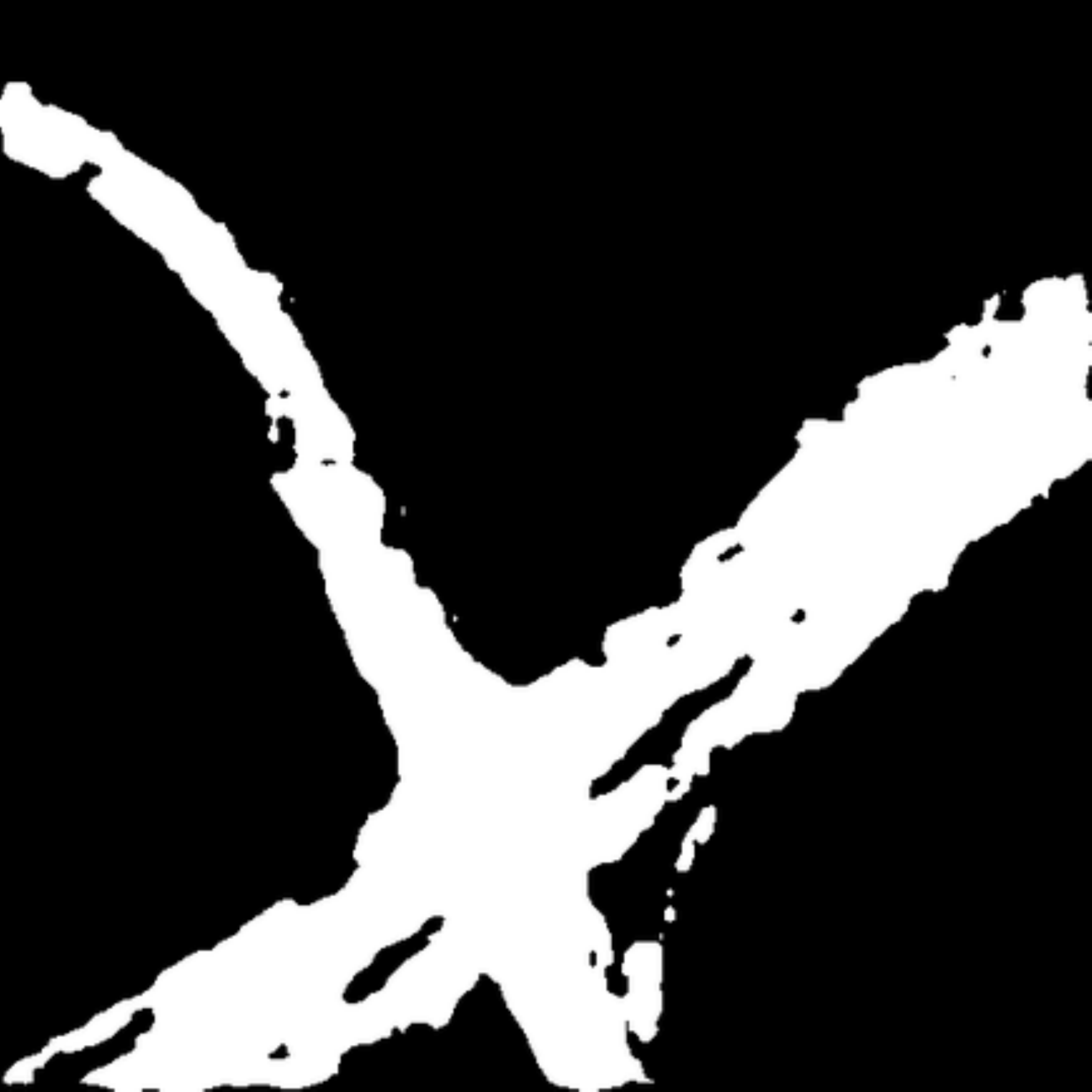}
			& 
\includegraphics[width=0.12\textwidth]{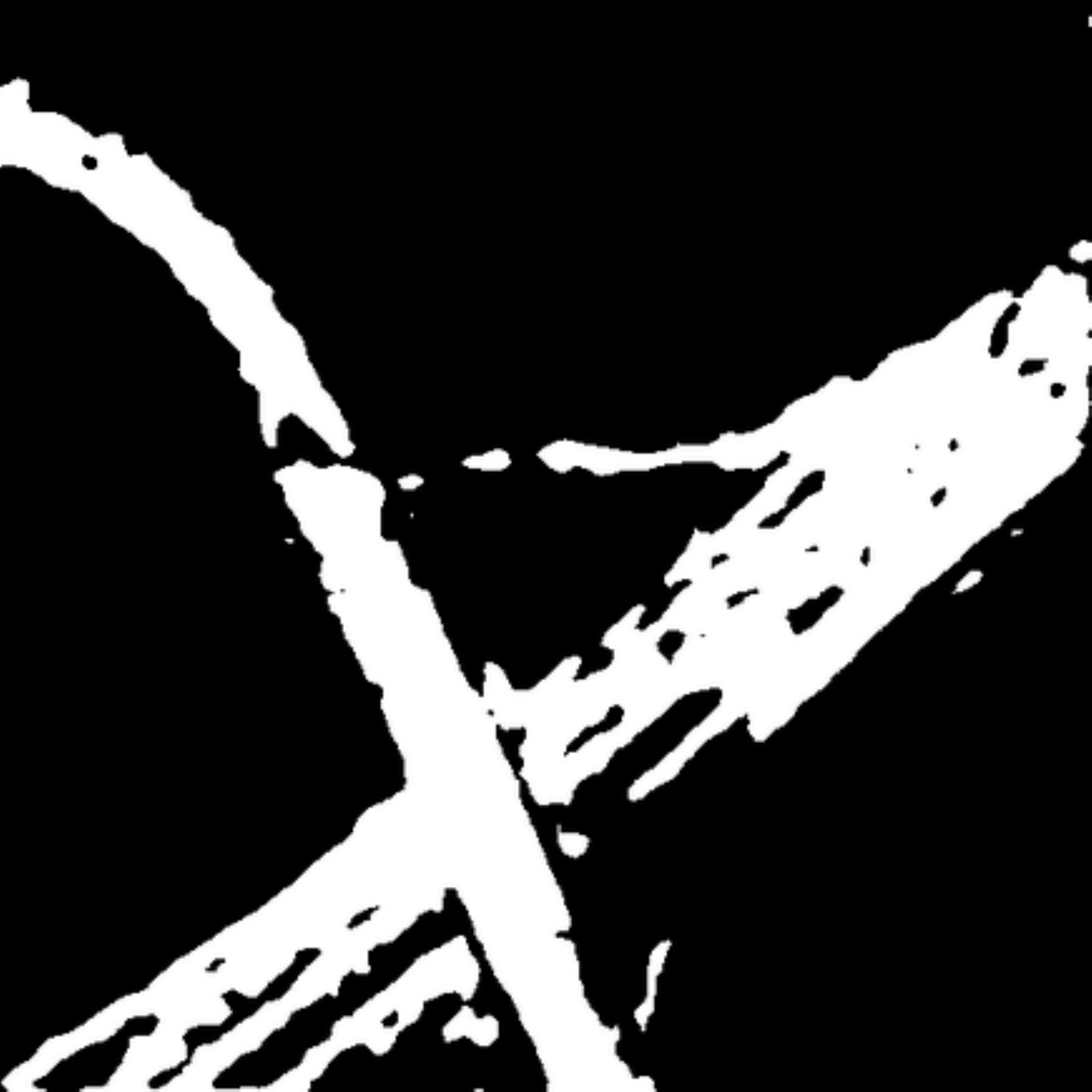}
			& 
\includegraphics[width=0.12\textwidth]{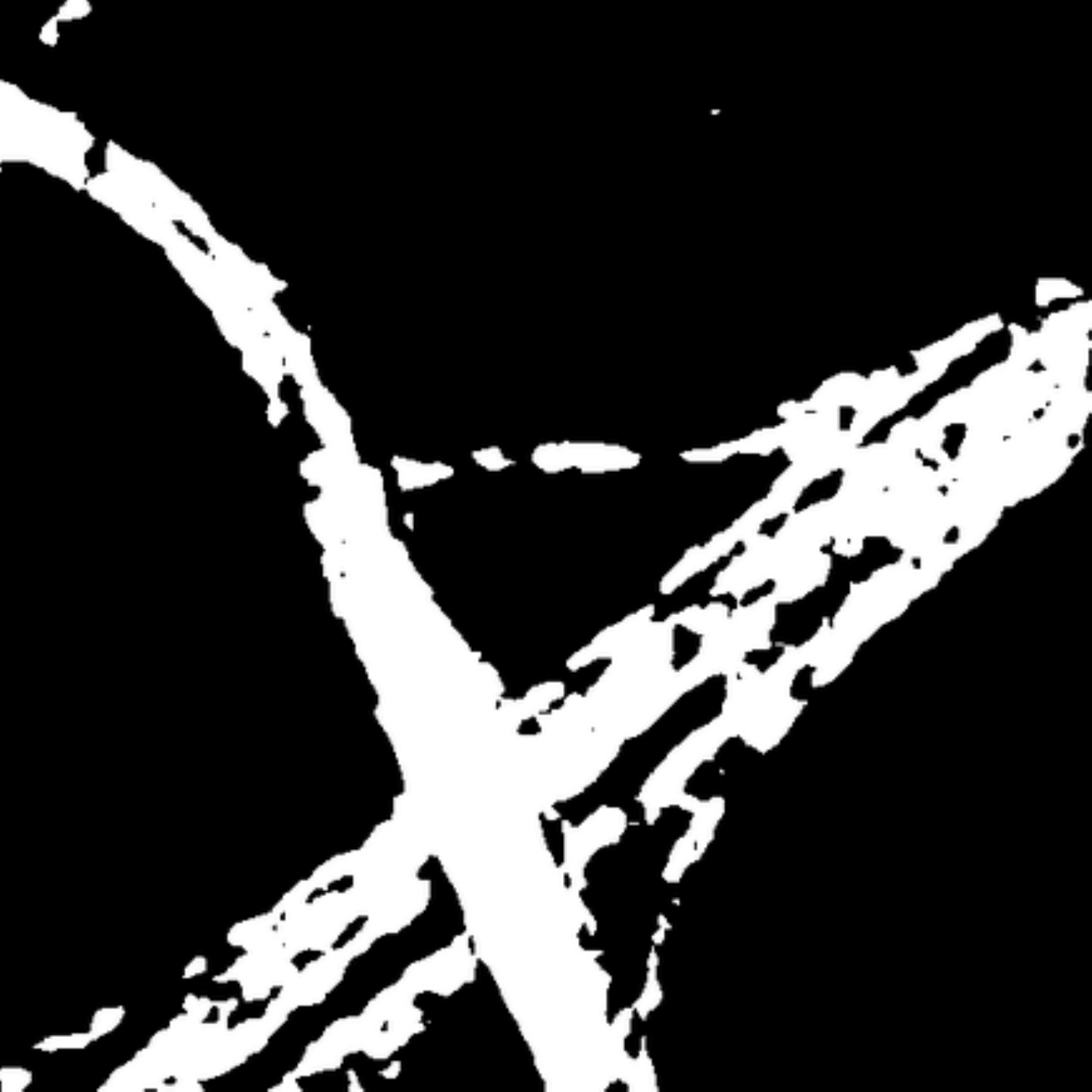}
			& 
\includegraphics[width=0.12\textwidth]{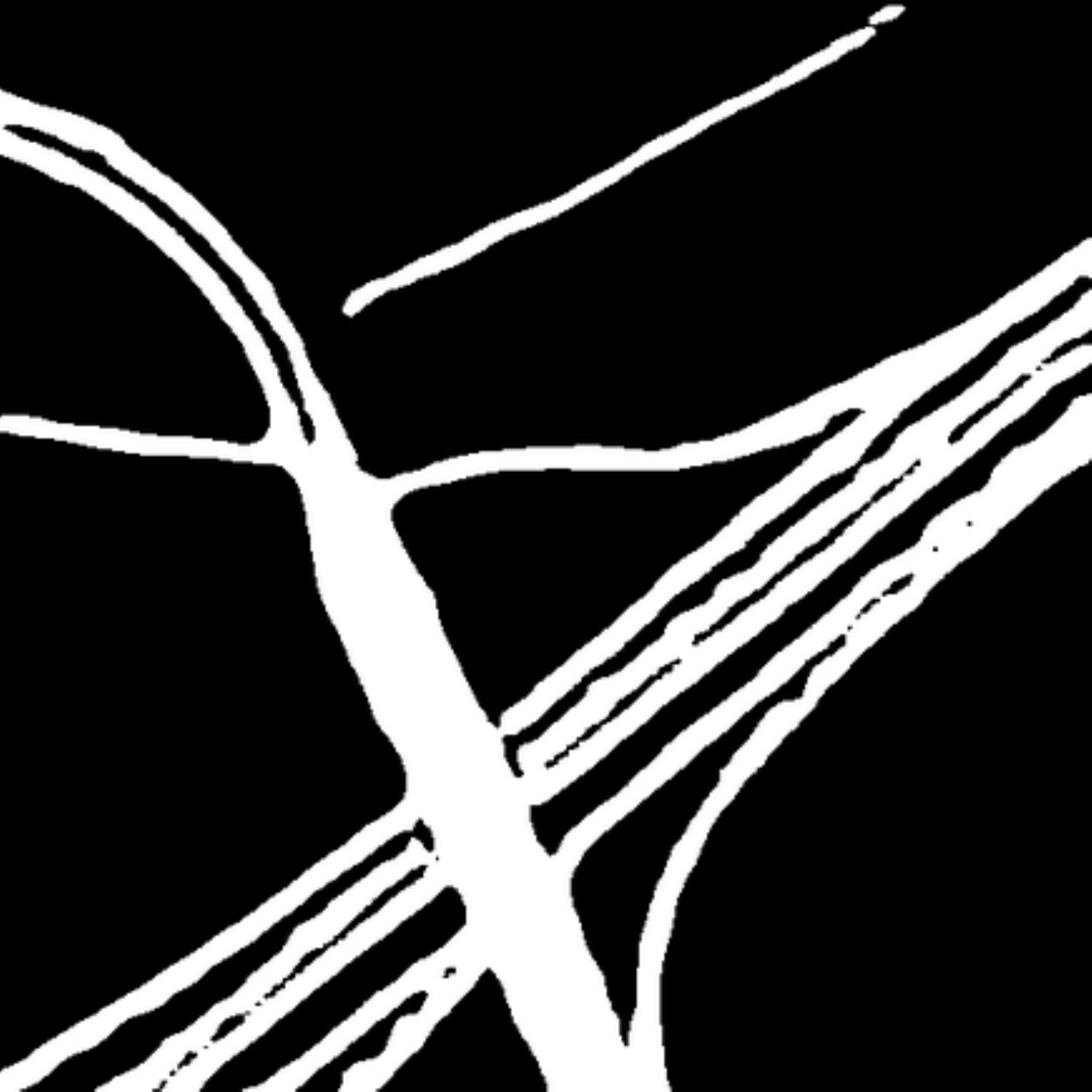}
			& 
\includegraphics[width=0.12\textwidth]{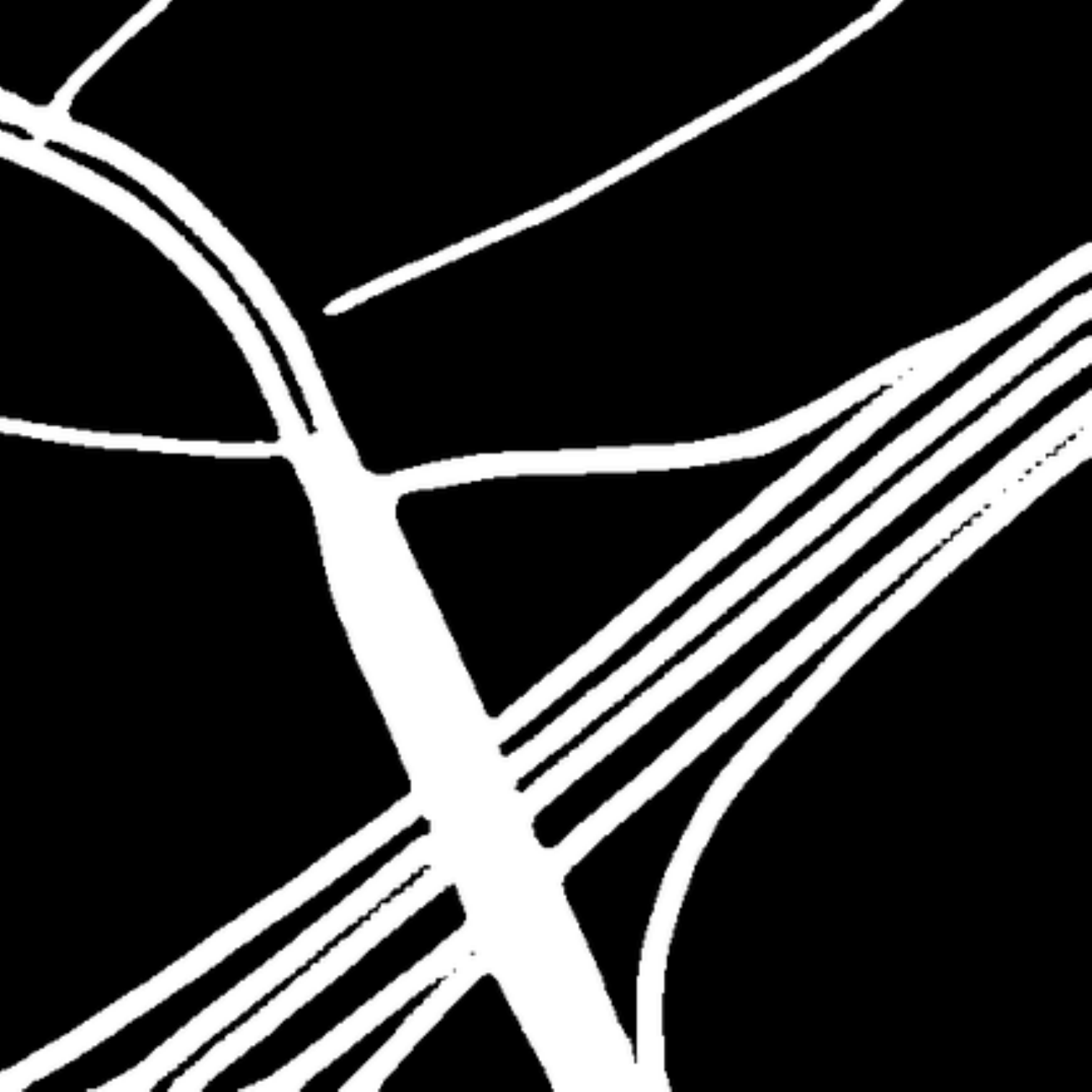}
		\\
\multirow{3}*{\rotatebox[origin=c]{90}{\scriptsize Images from Zhengzhou, Henan Province}} 
            &
\includegraphics[width=0.12\textwidth]{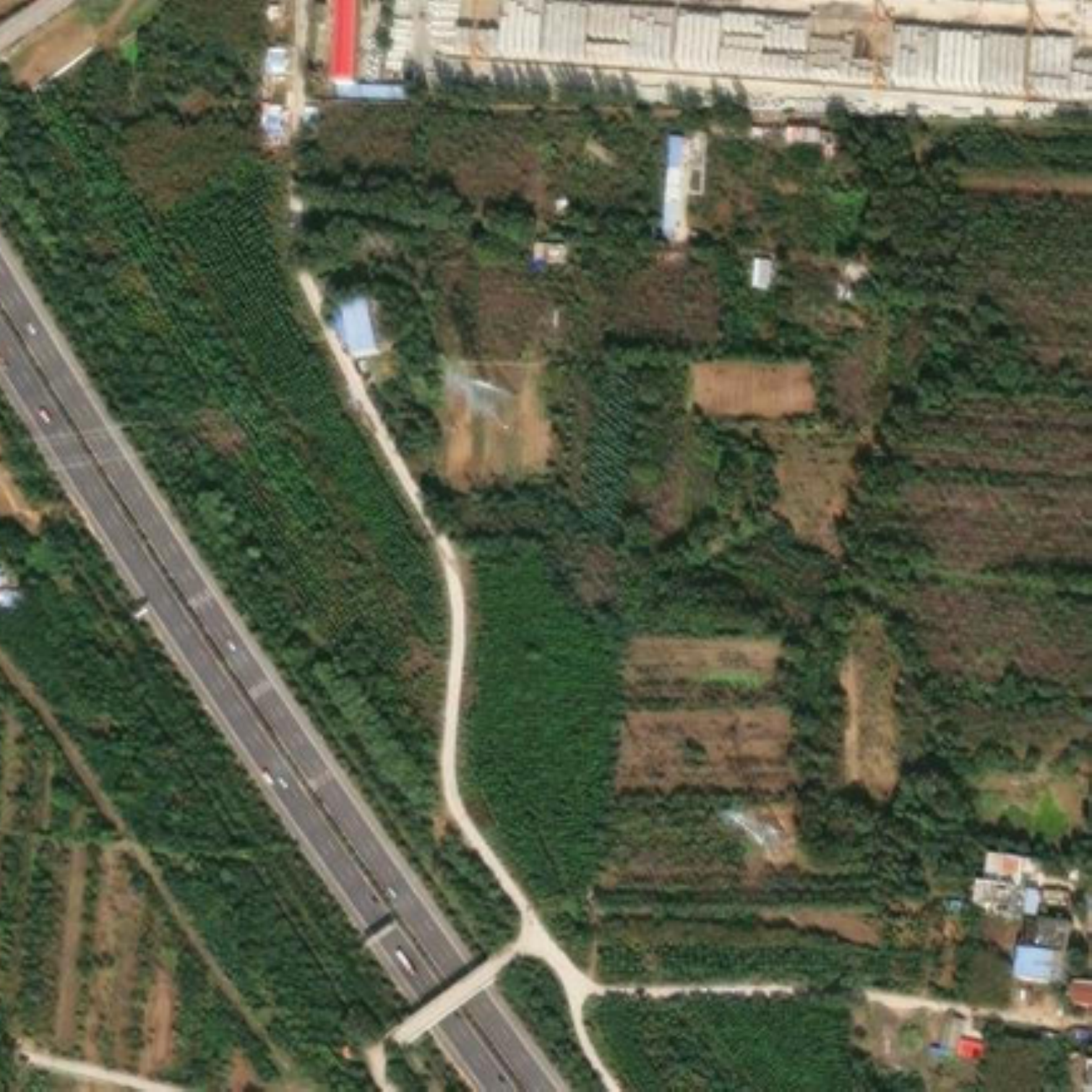}
			& 
\includegraphics[width=0.12\textwidth]{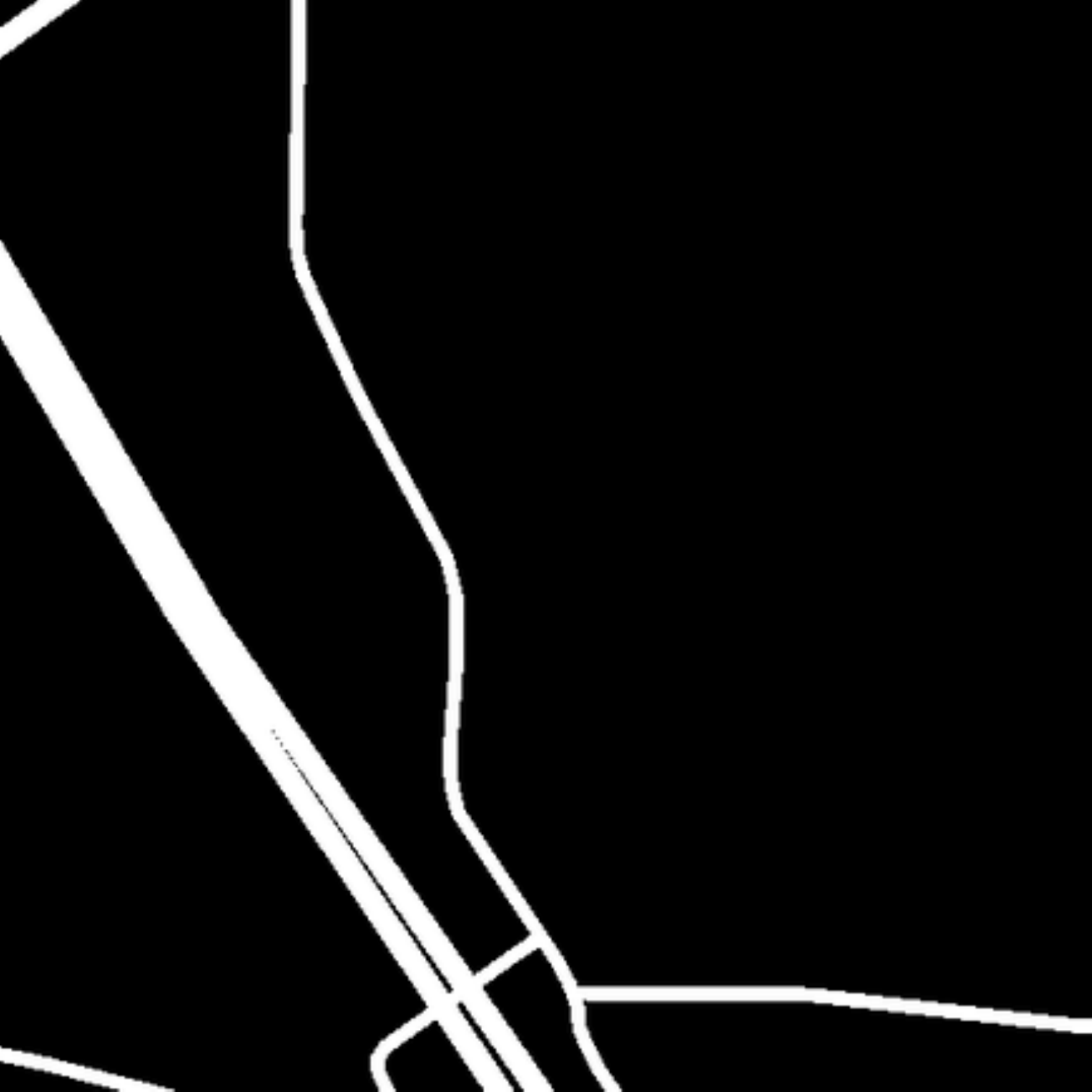}
			& 
\includegraphics[width=0.12\textwidth]{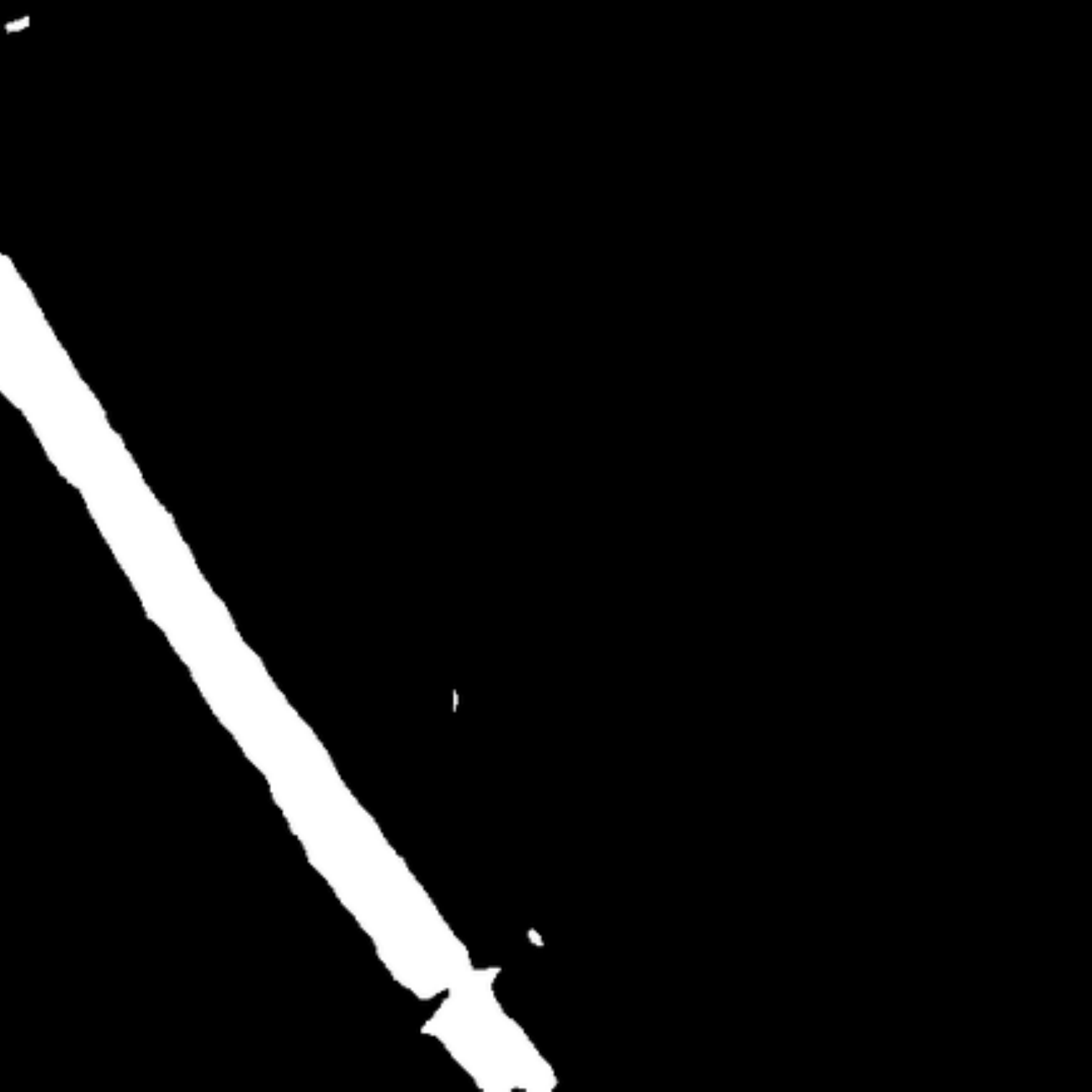}
			& 
\includegraphics[width=0.12\textwidth]{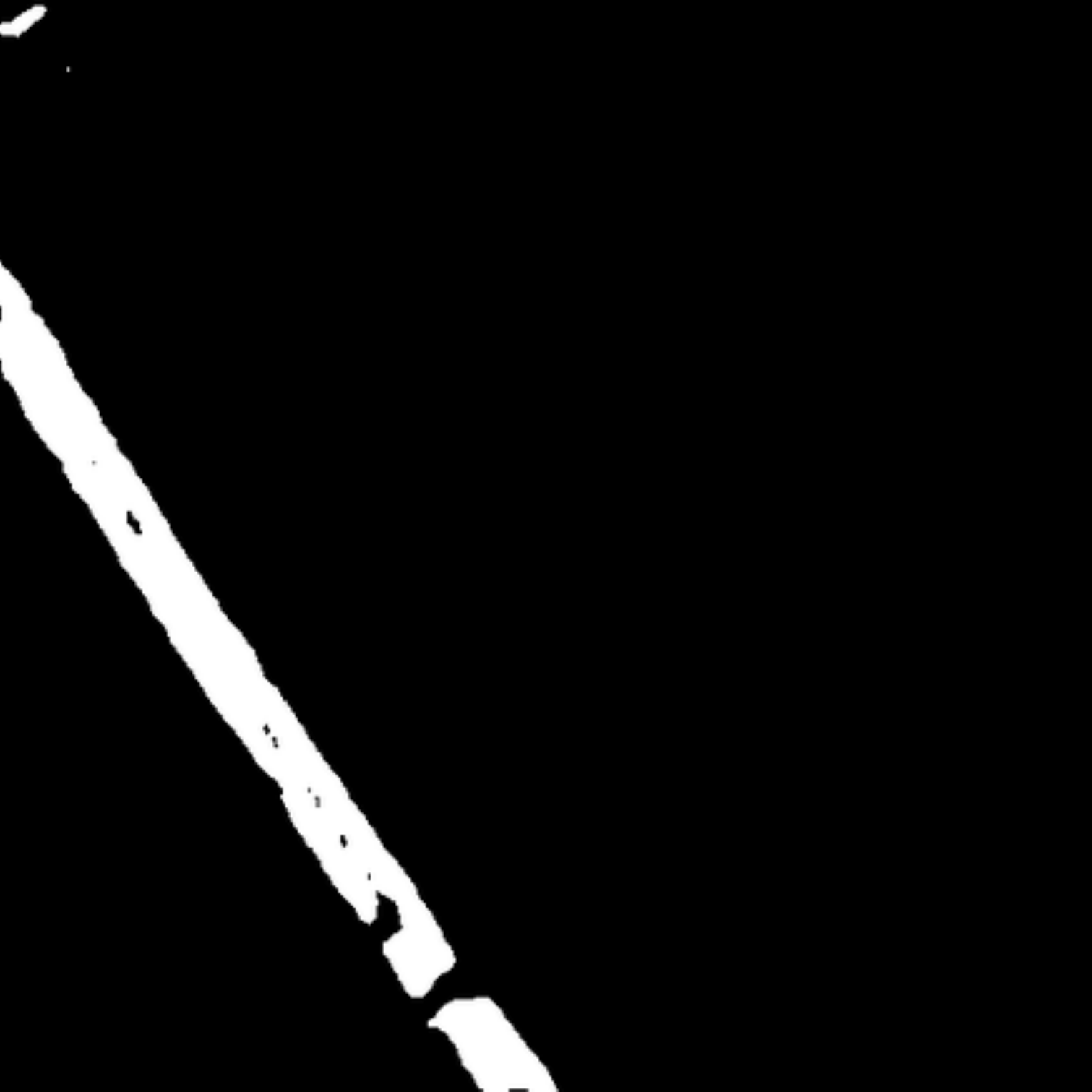}
			& 
\includegraphics[width=0.12\textwidth]{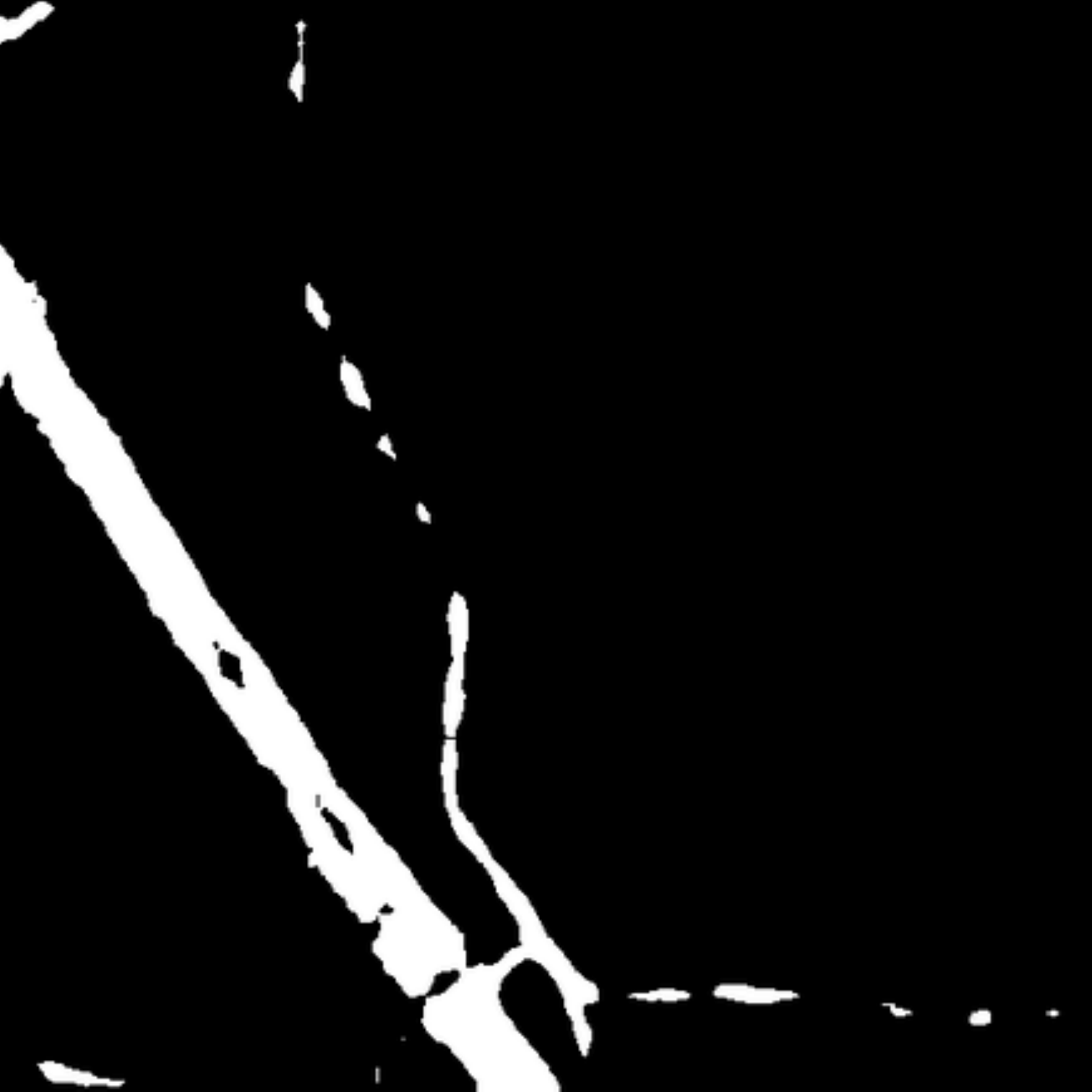}
			& 
\includegraphics[width=0.12\textwidth]{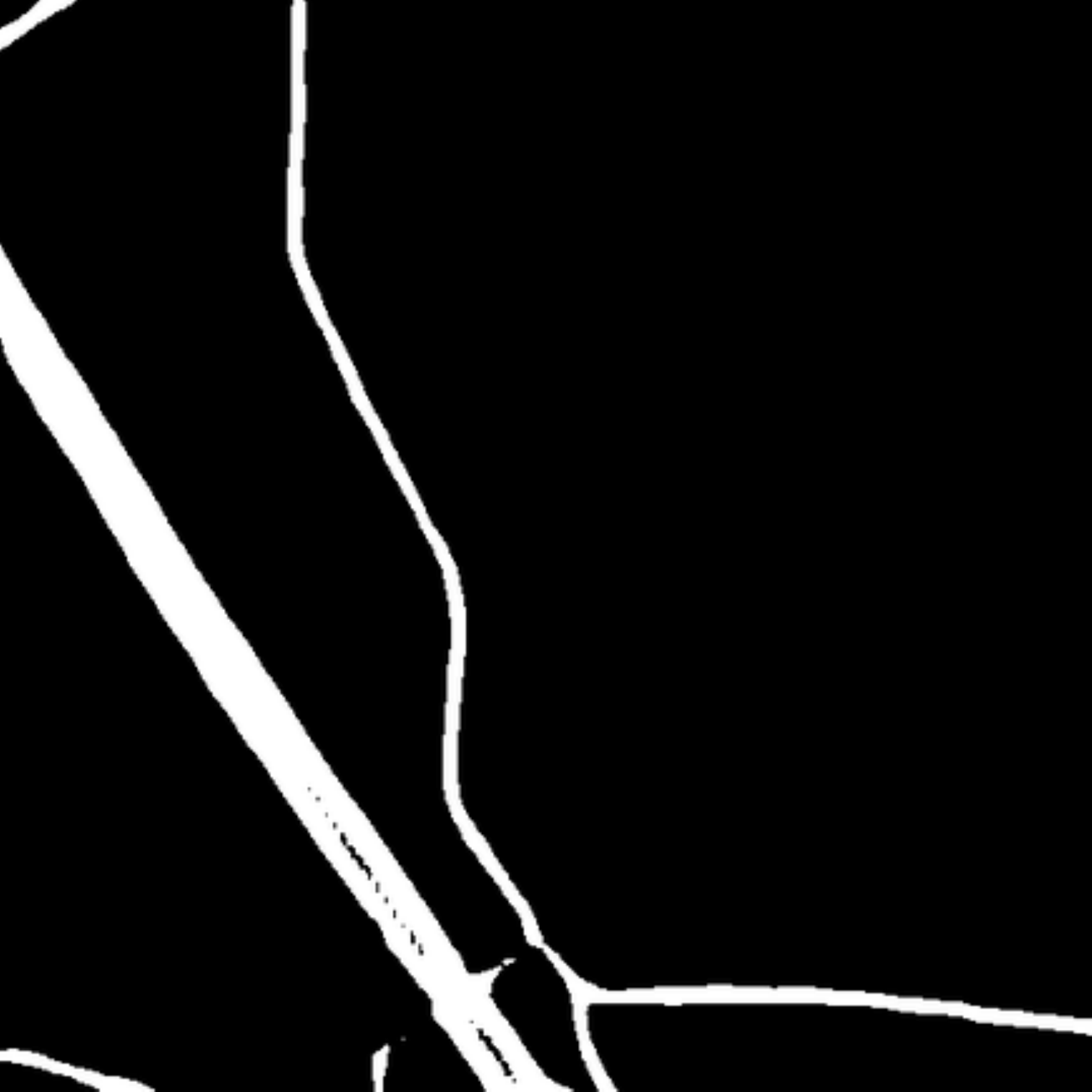}
			& 
\includegraphics[width=0.12\textwidth]{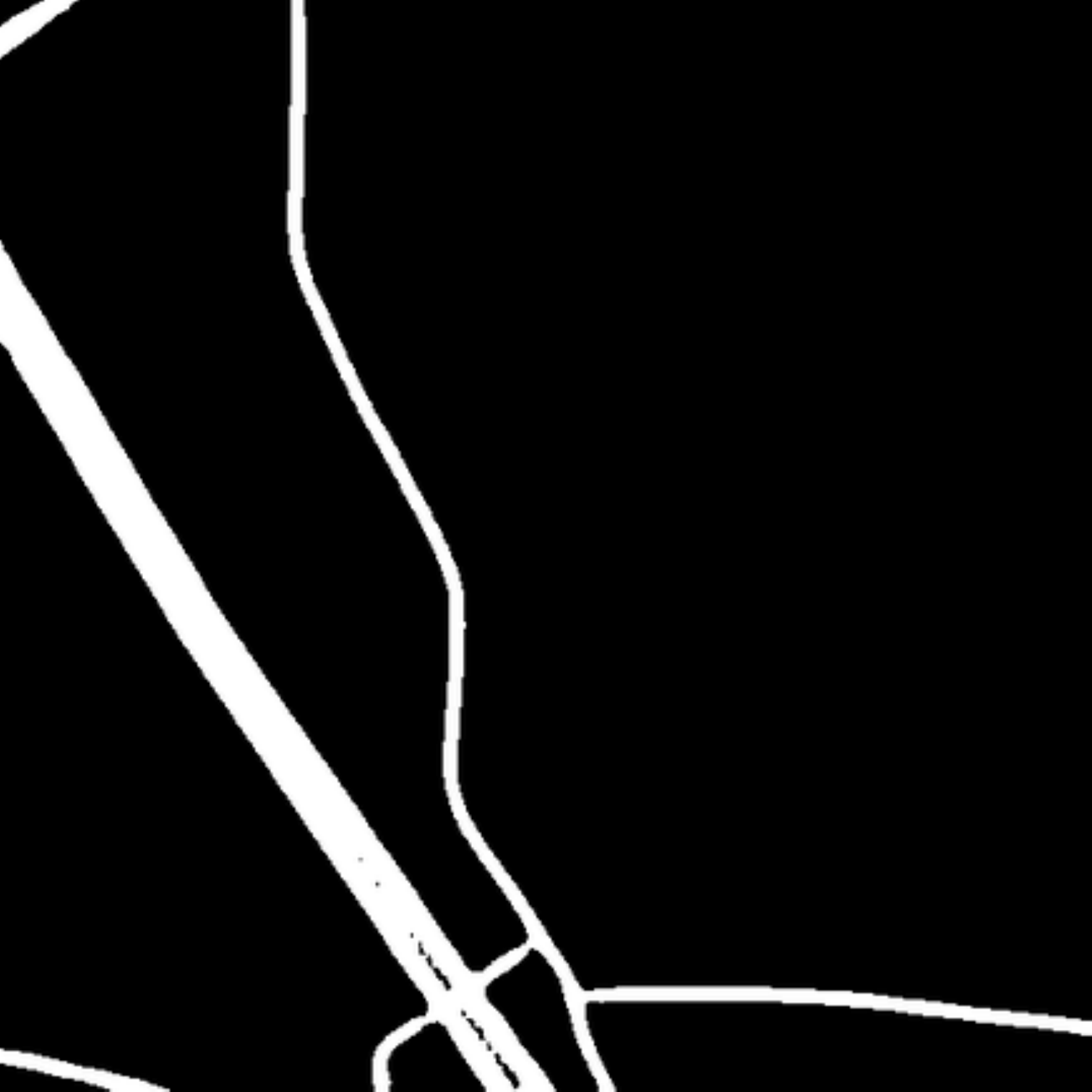}
		\\
            &
\includegraphics[width=0.12\textwidth]{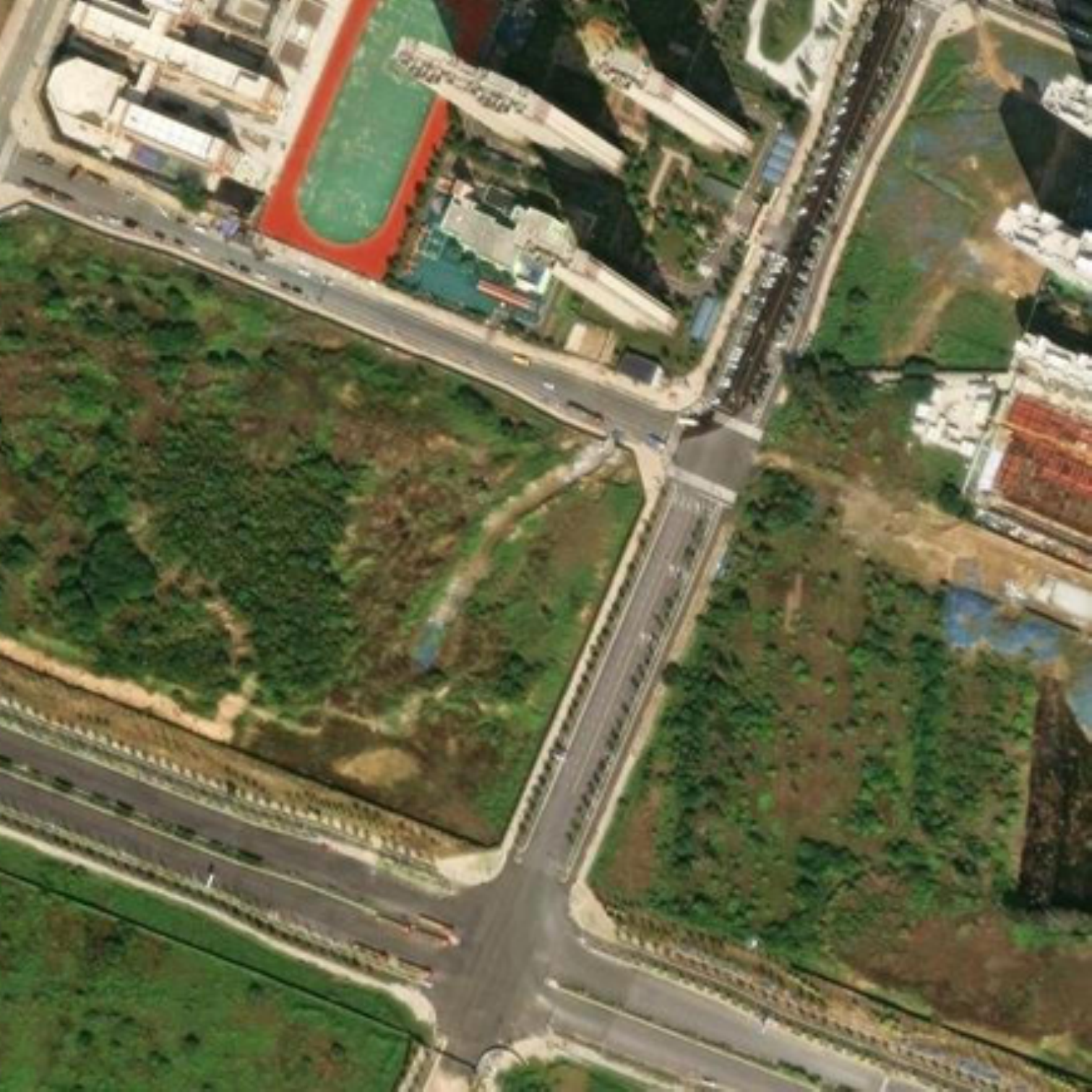}
			& 
\includegraphics[width=0.12\textwidth]{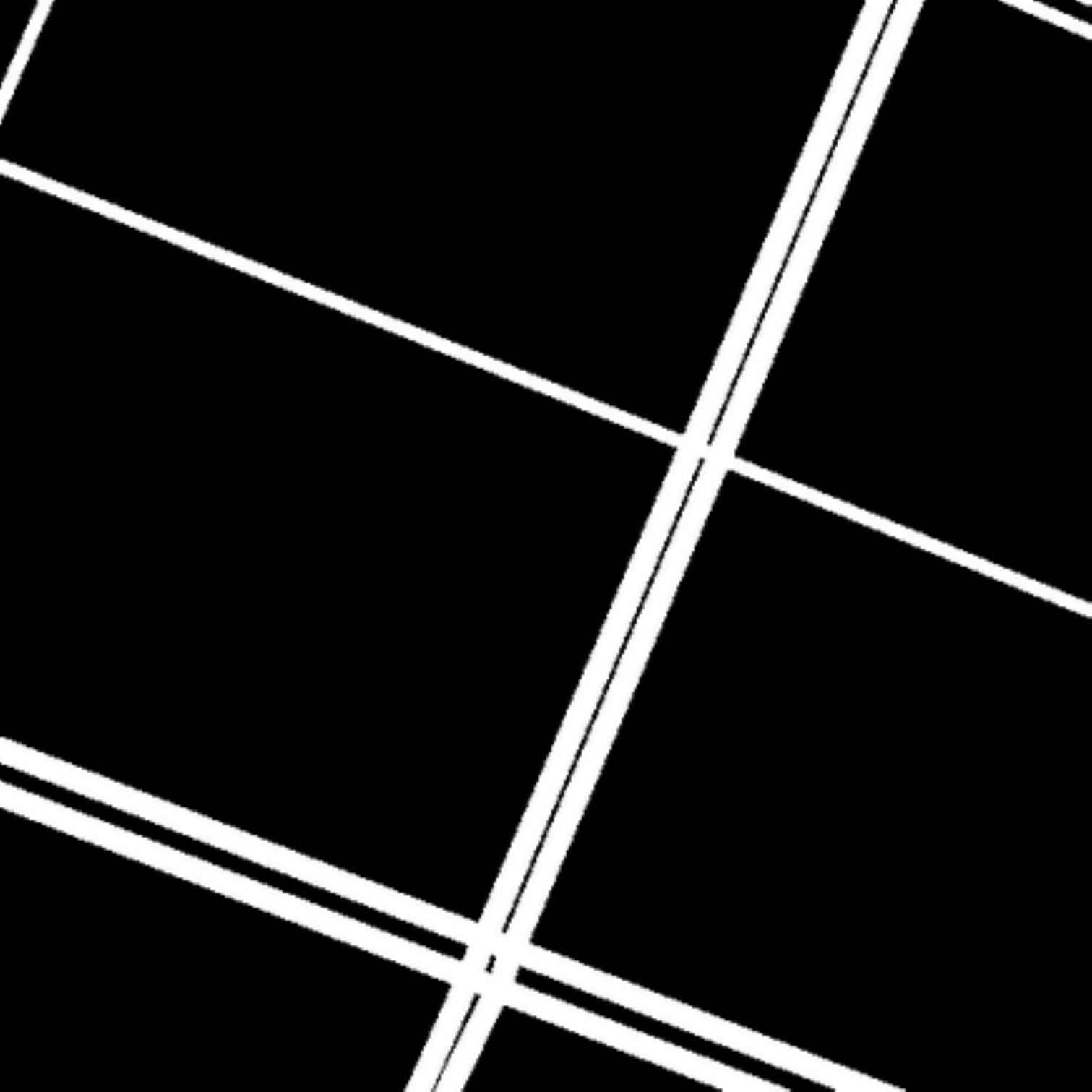}
			& 
\includegraphics[width=0.12\textwidth]{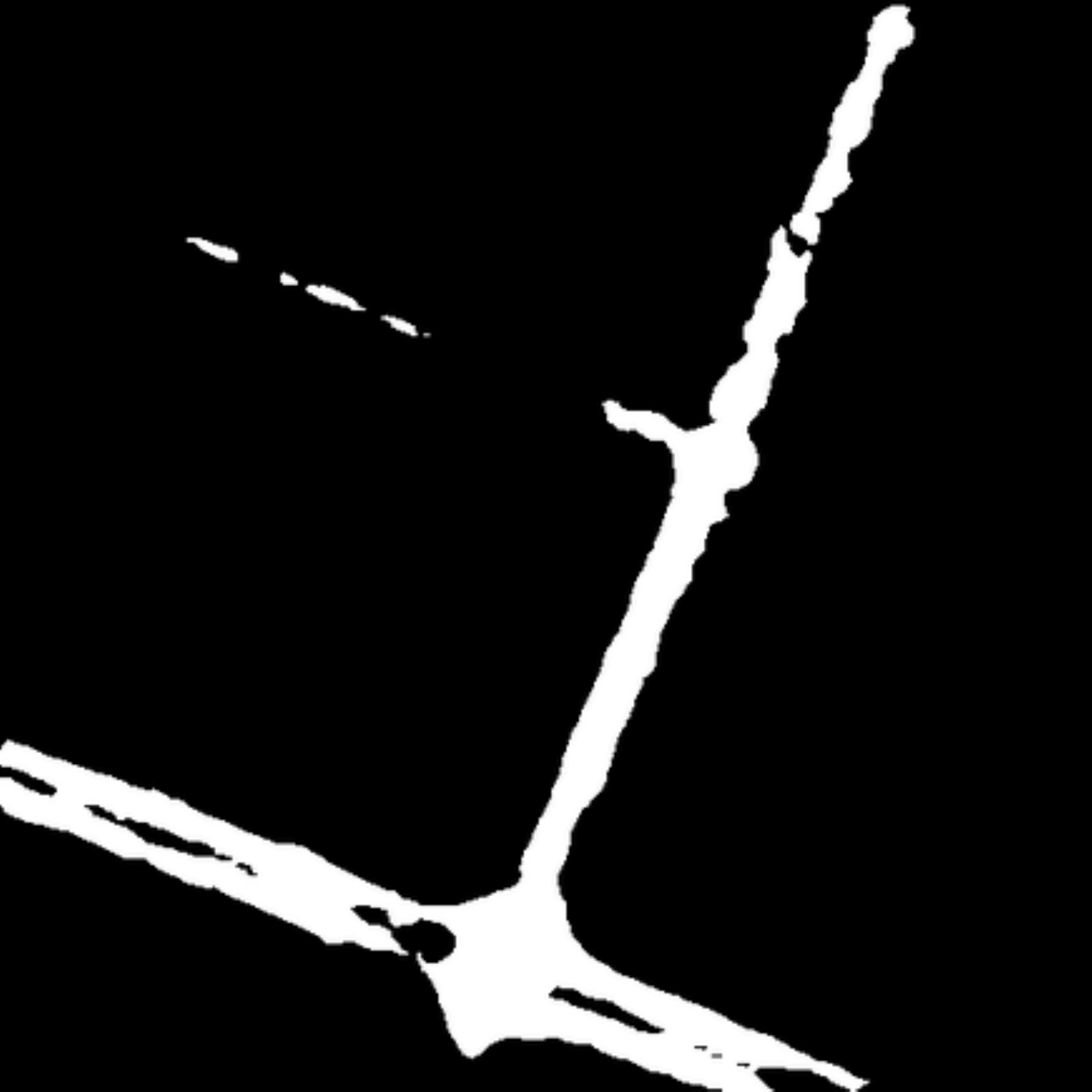}
			& 
\includegraphics[width=0.12\textwidth]{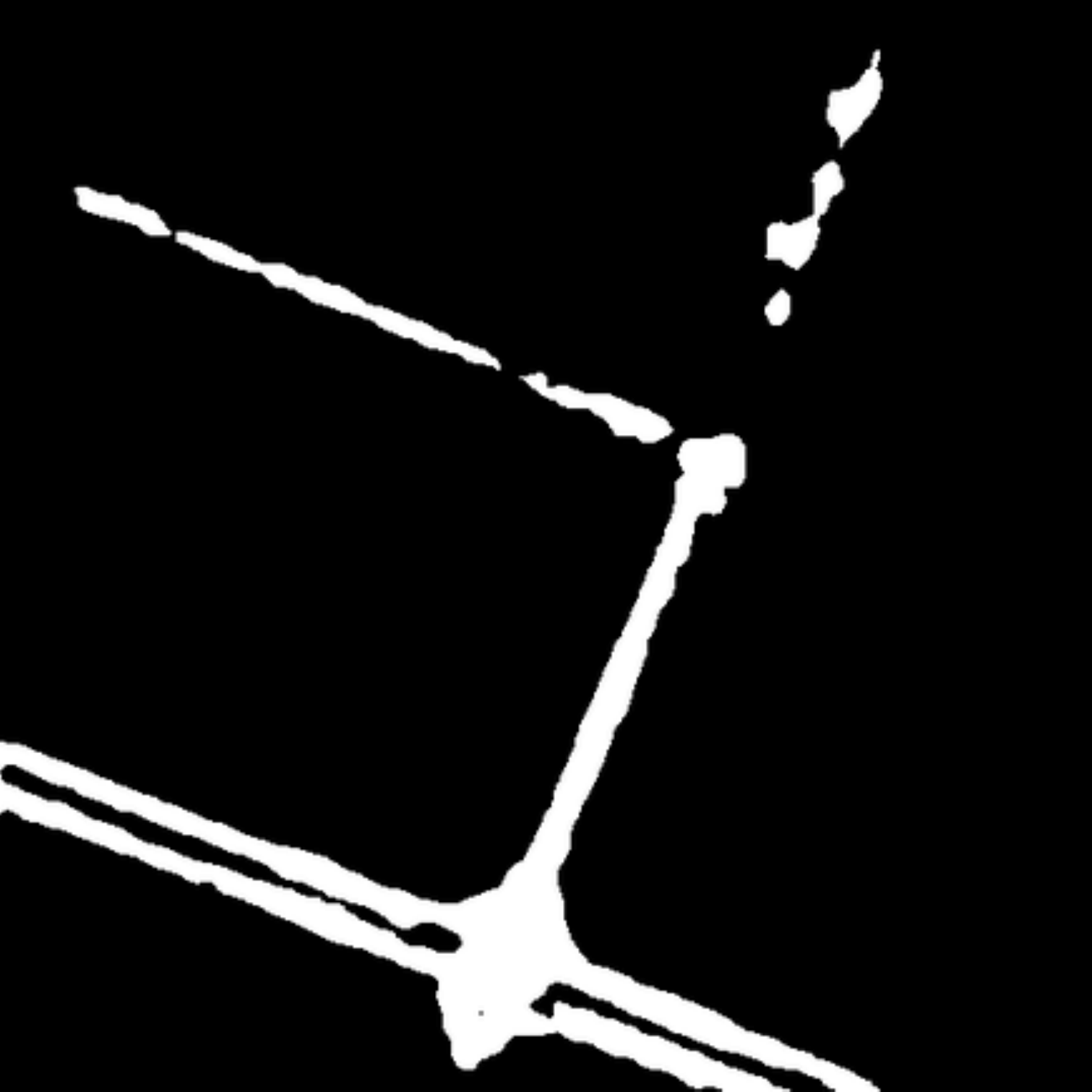}
			& 
\includegraphics[width=0.12\textwidth]{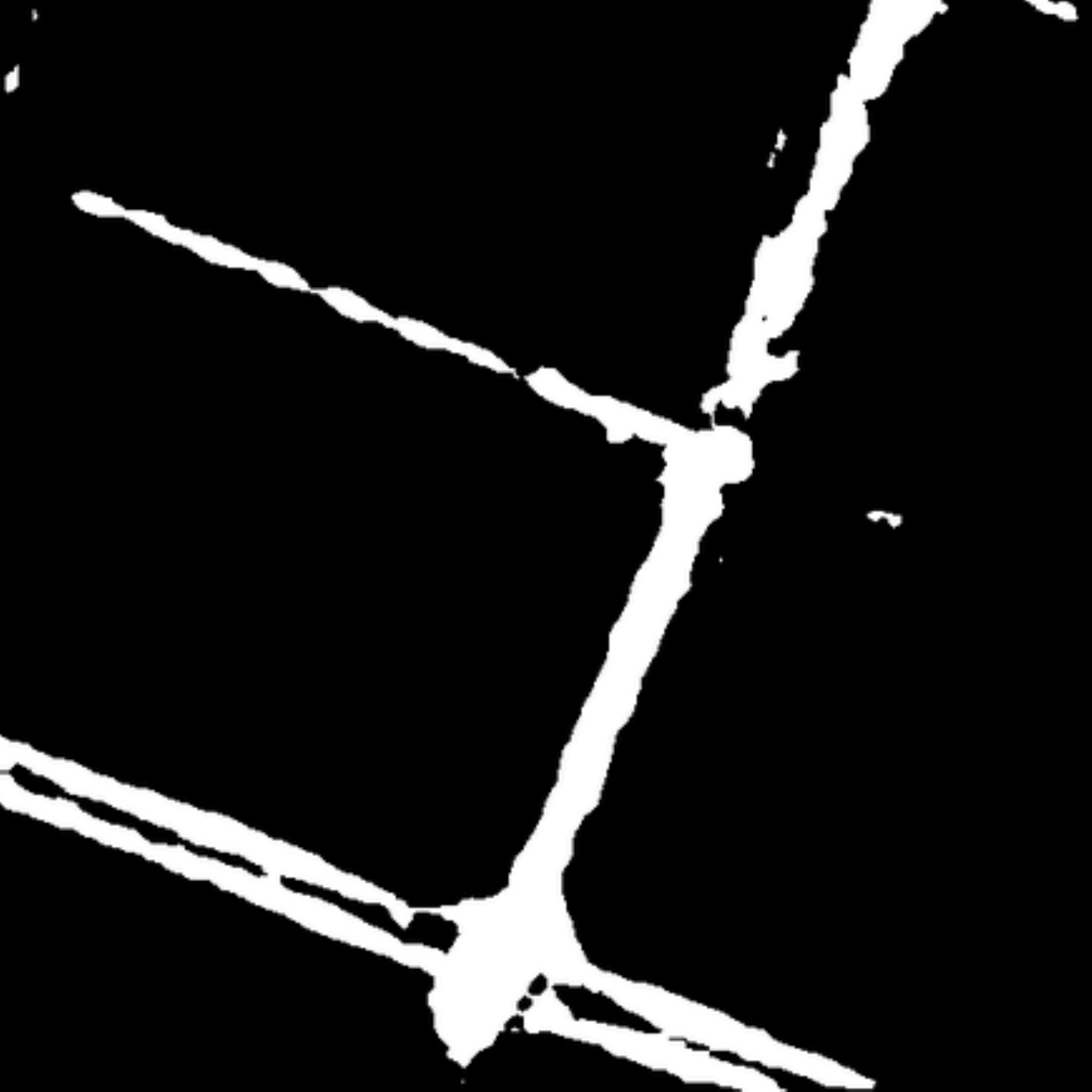}
			& 
\includegraphics[width=0.12\textwidth]{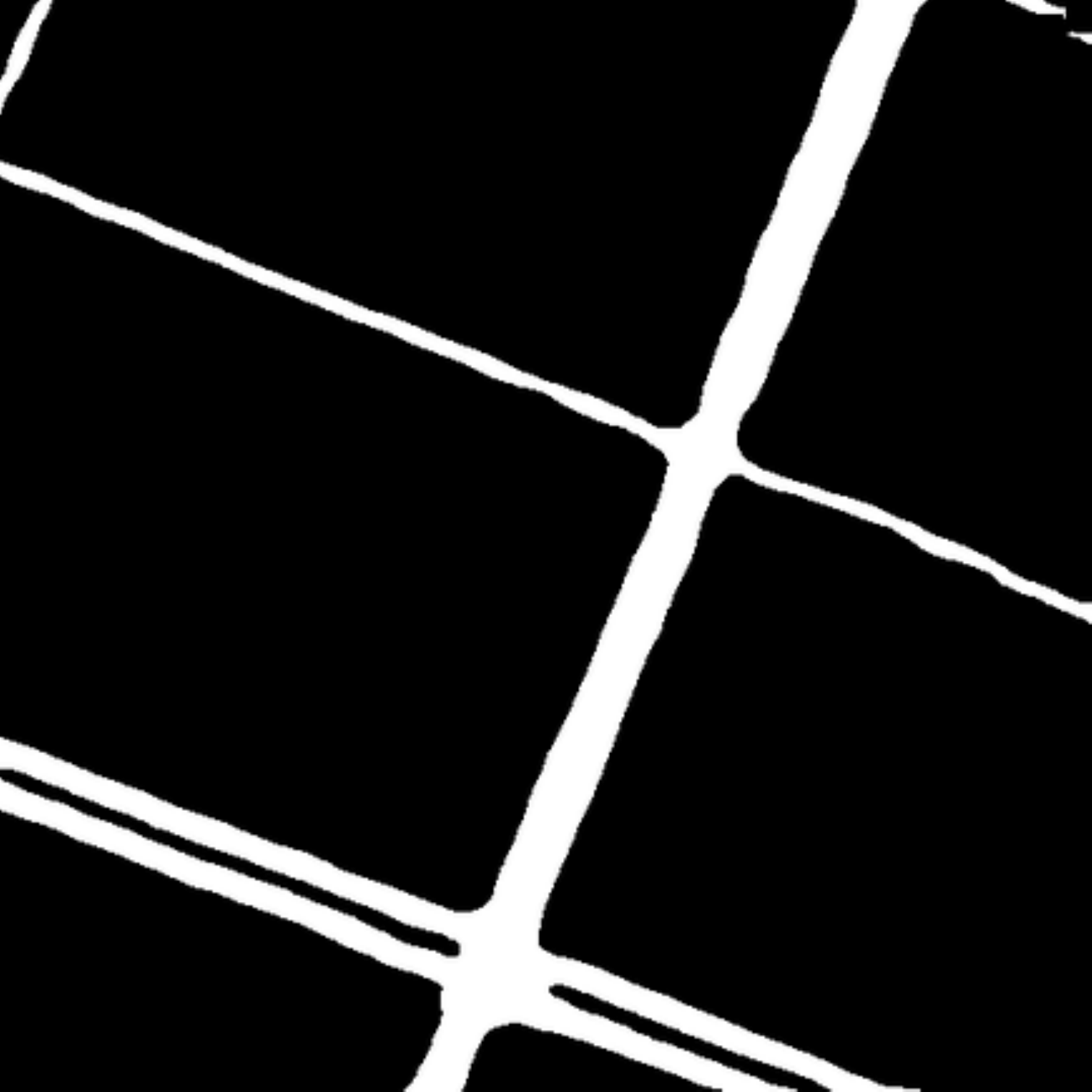}
			& 
\includegraphics[width=0.12\textwidth]{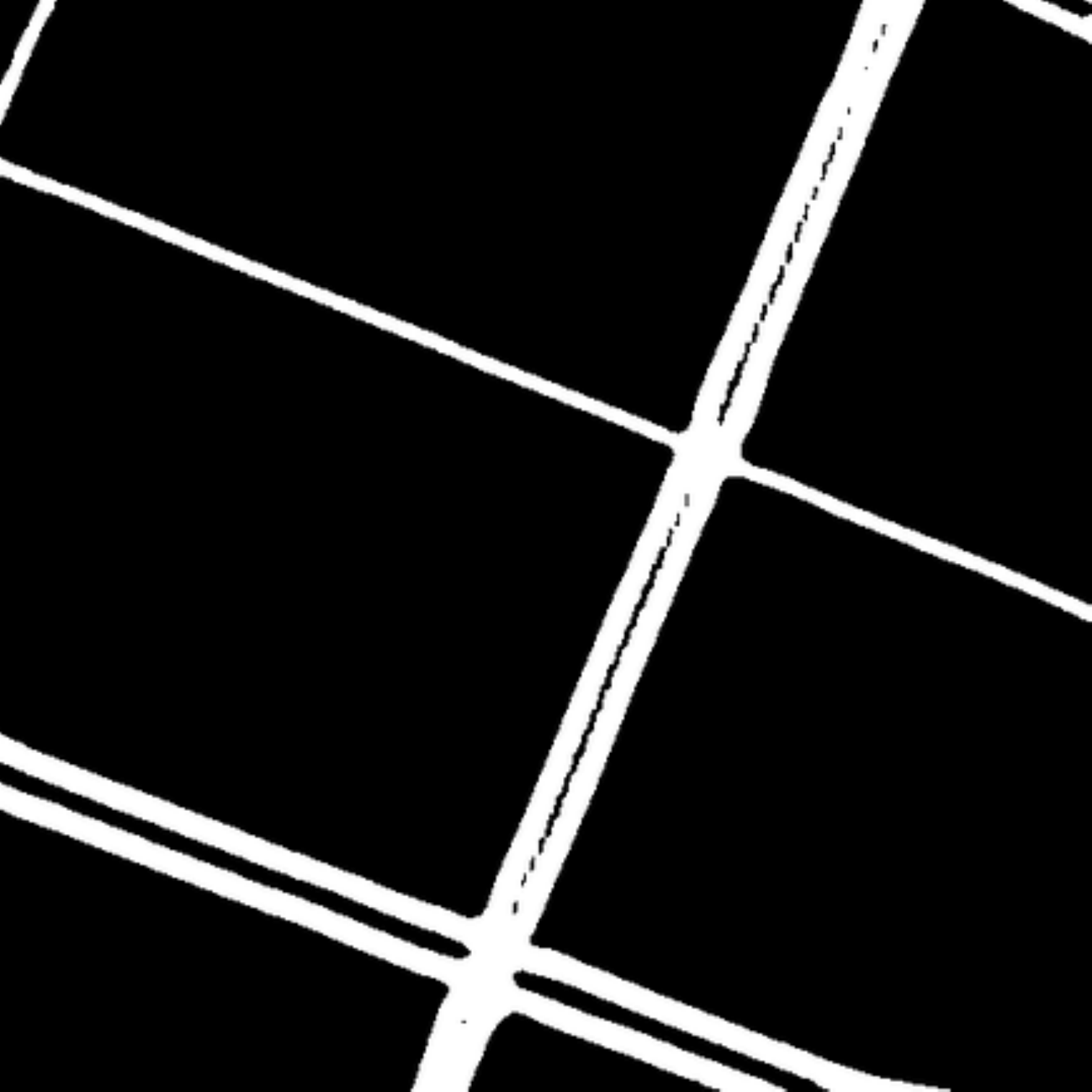}
		\\
            &
\includegraphics[width=0.12\textwidth]{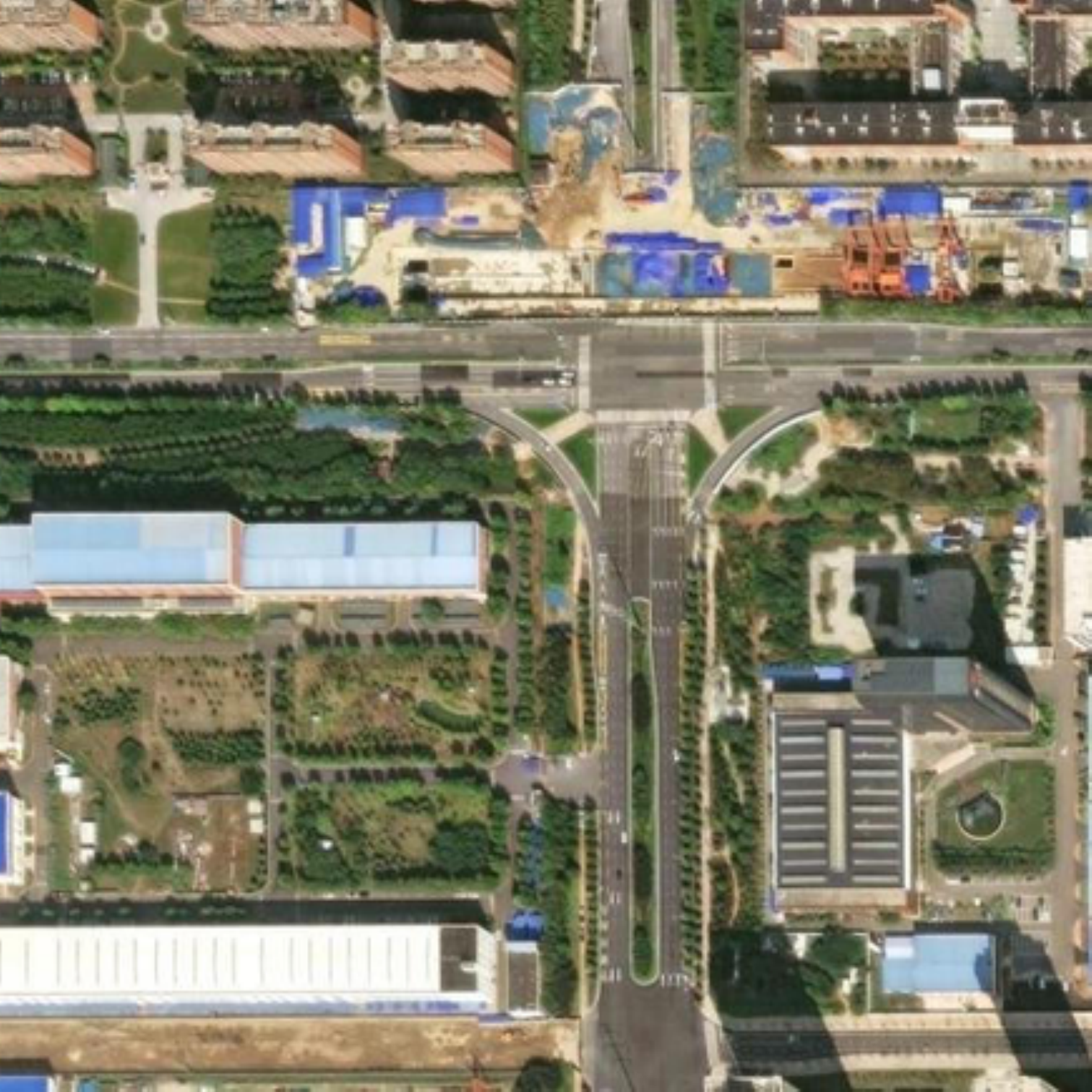}
			& 
\includegraphics[width=0.12\textwidth]{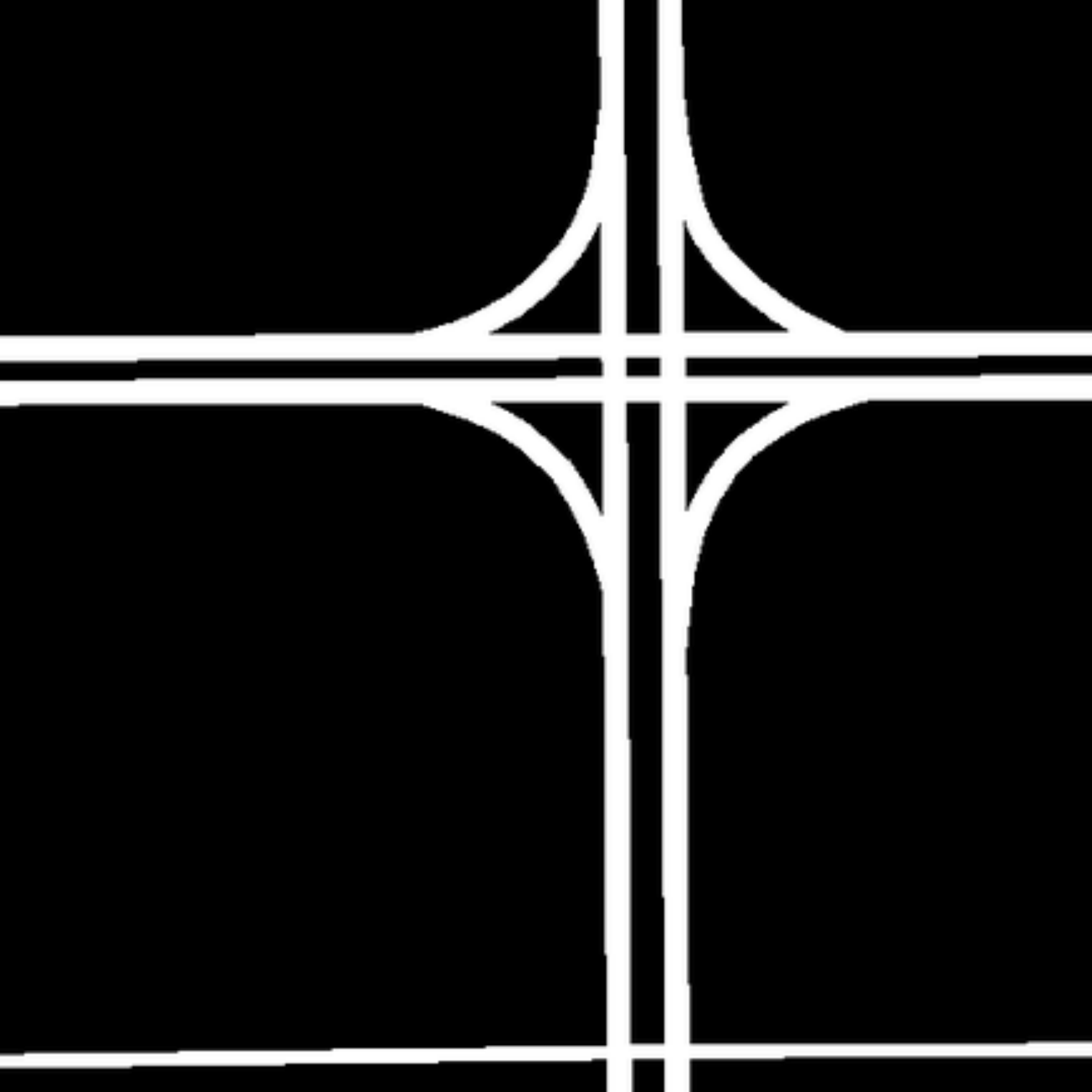}
			& 
\includegraphics[width=0.12\textwidth]{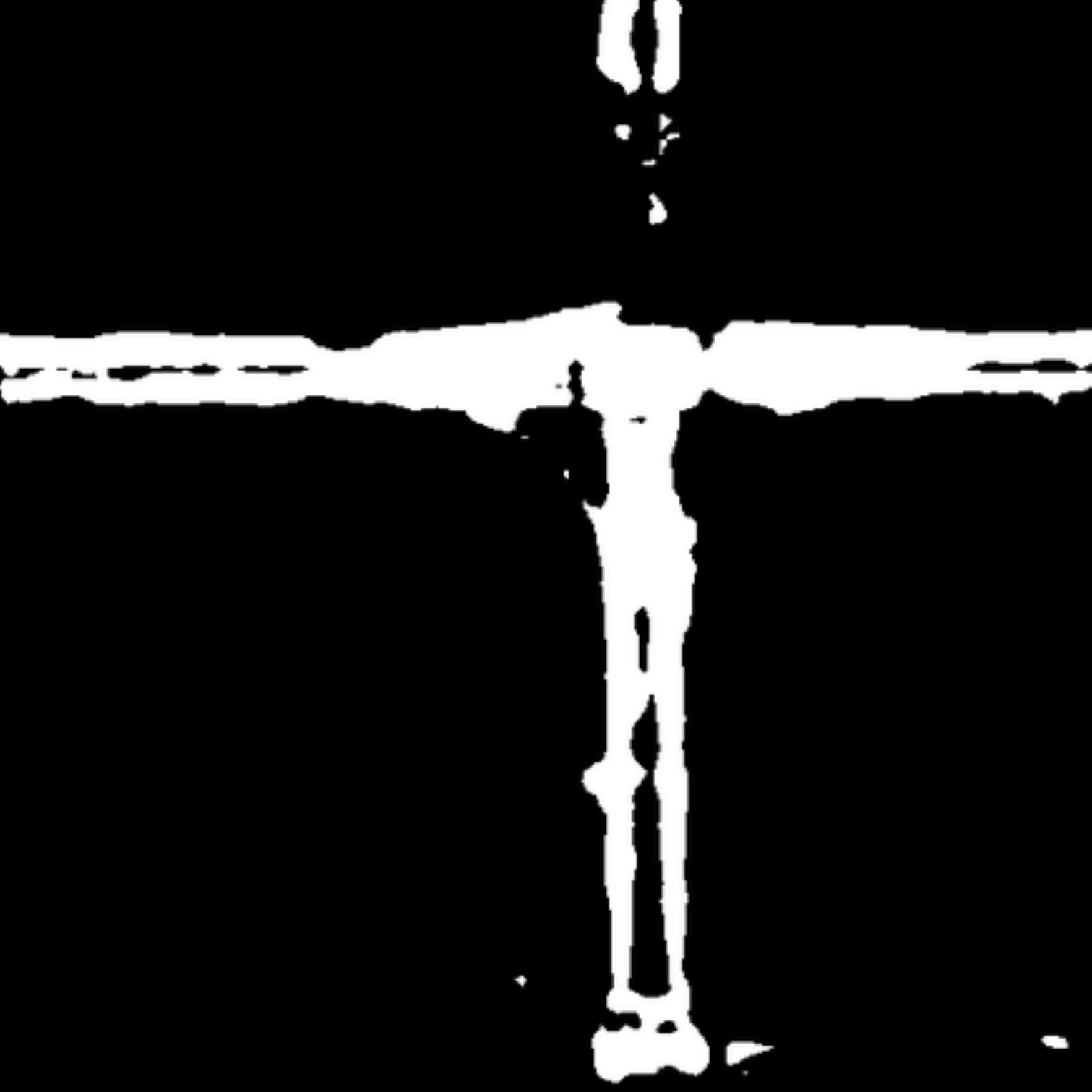}
			& 
\includegraphics[width=0.12\textwidth]{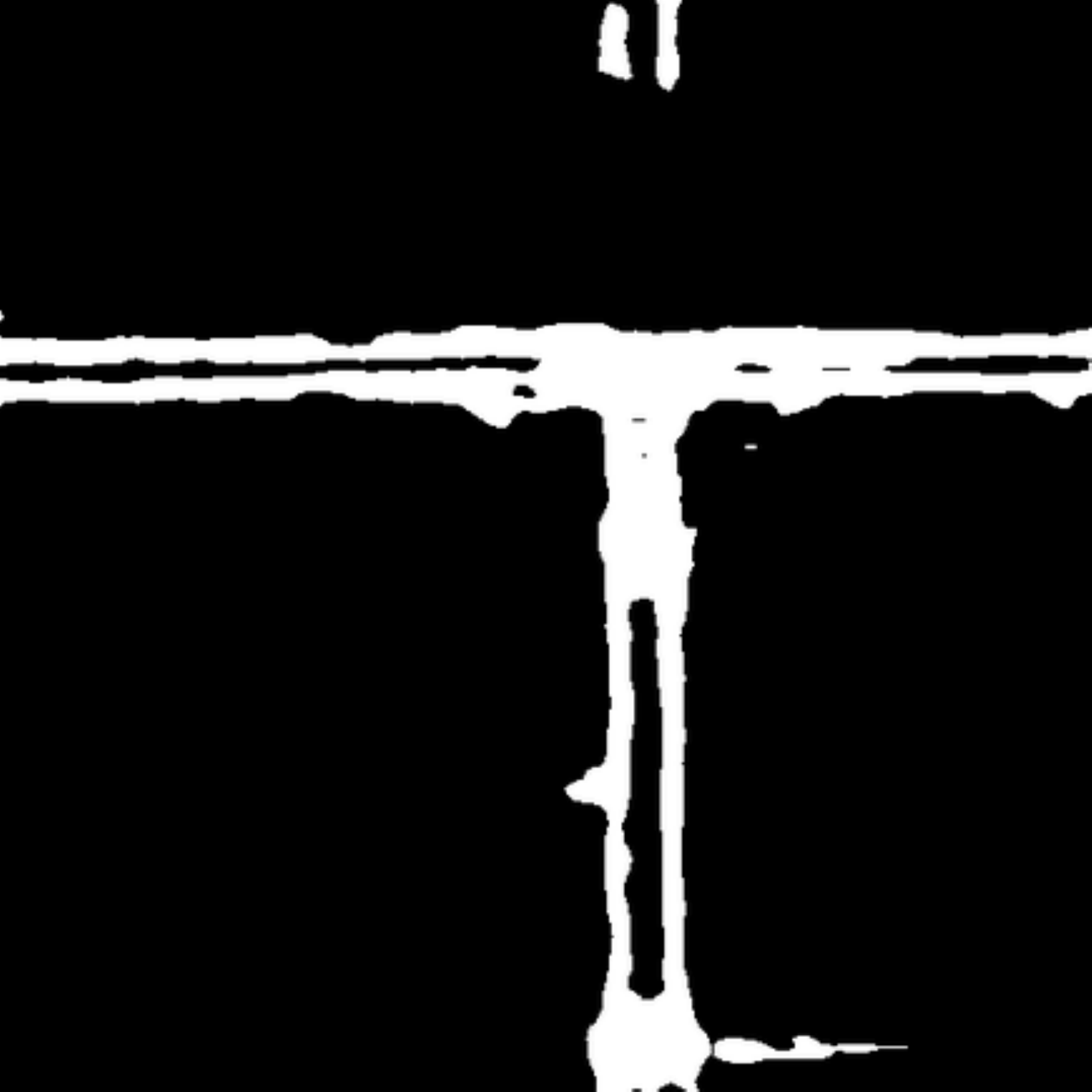}
			& 
\includegraphics[width=0.12\textwidth]{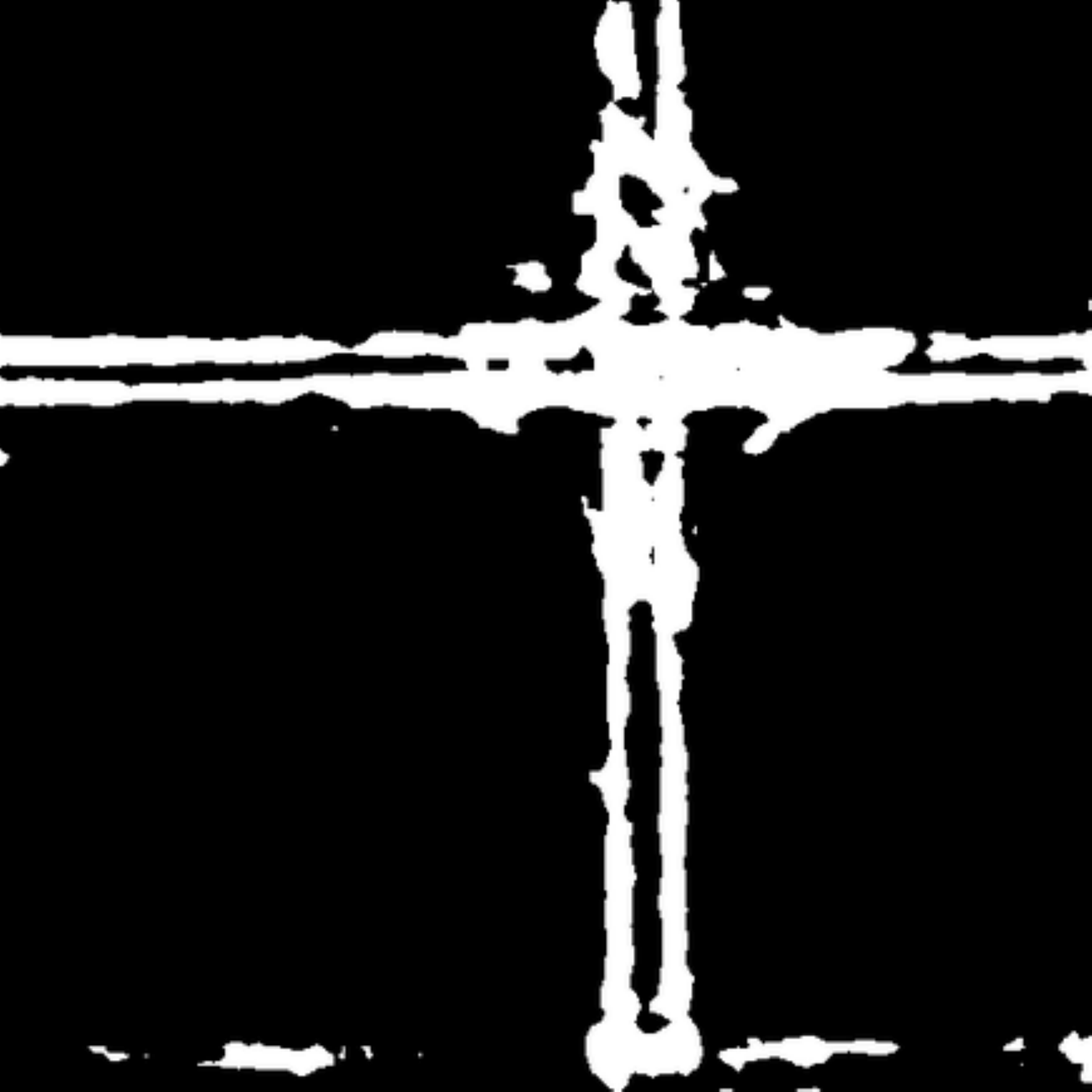}
			& 
\includegraphics[width=0.12\textwidth]{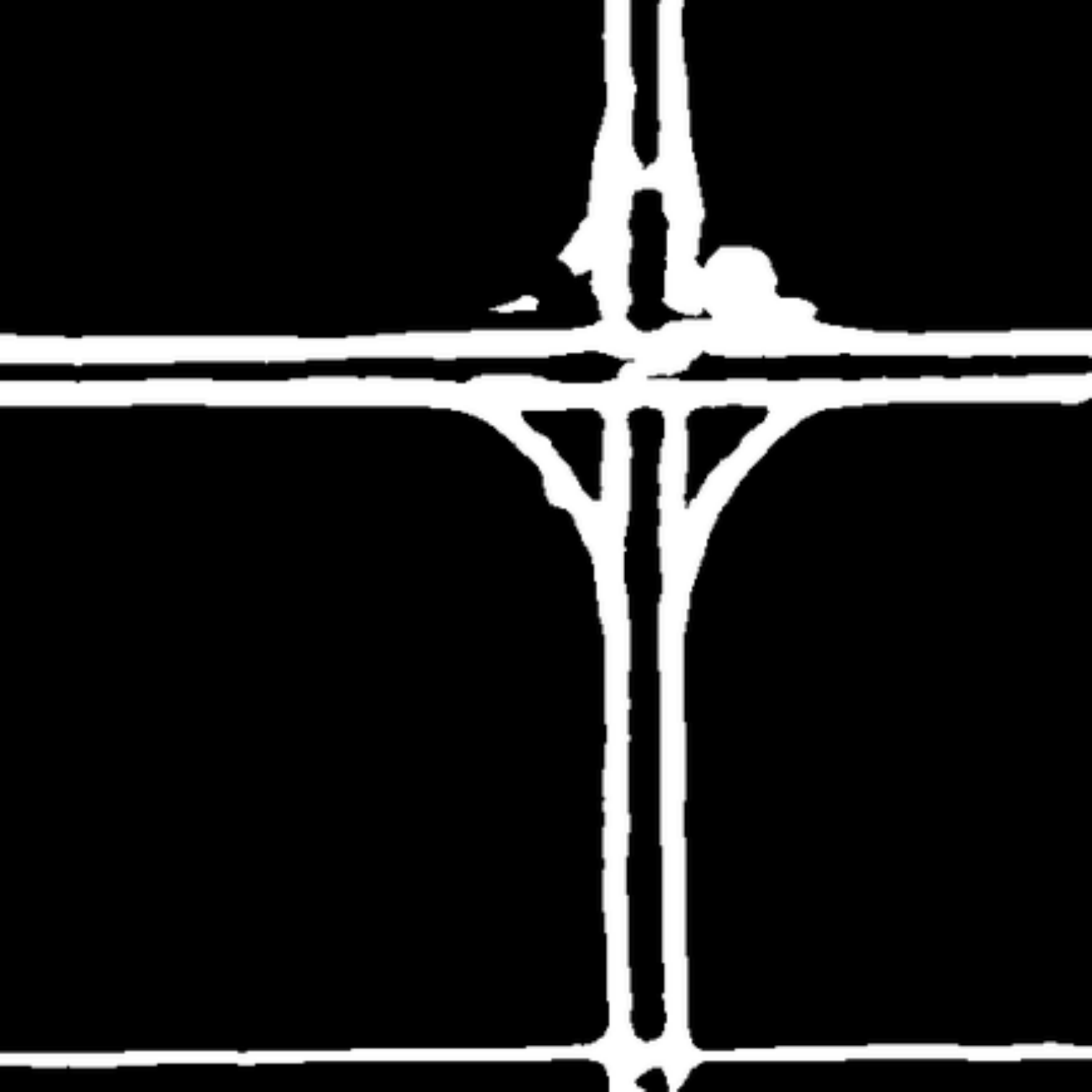}
			& 
\includegraphics[width=0.12\textwidth]{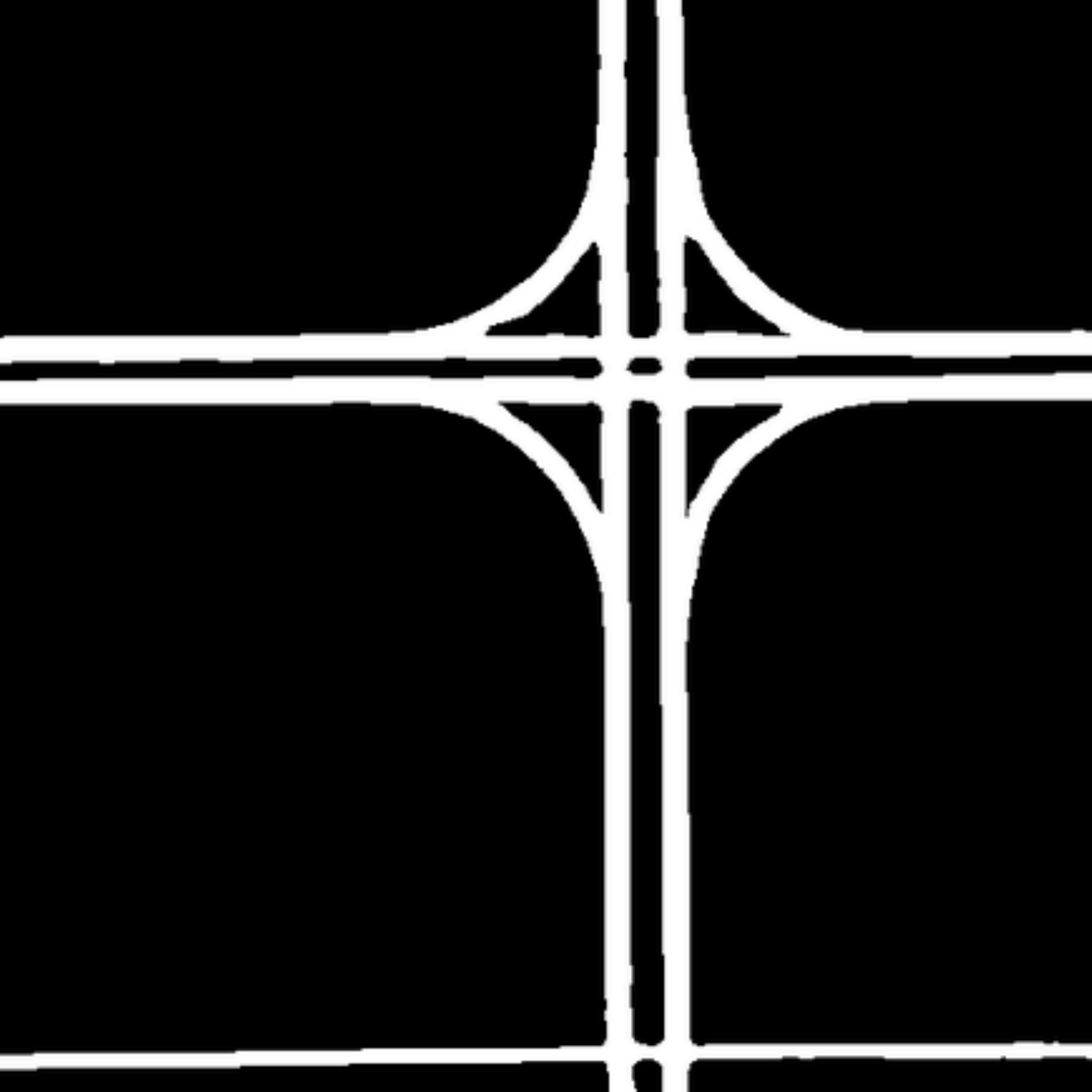}
		\\
& (a)Image & (b)GT & (c) gct & (d) s4l & (e) adv & (f) reco & (g) \makecell[c]{SRUNet  \\ (Ours) } 
\end{tabular}%
\caption{Visual comparison on the self-constructed dataset. (a) Original images. (b) Ground truth. (c) gct. (d) s4l. (e) adv. (f) reco. (g) SRUNet }
\label{vis_suyu_com} 
\end{figure*}

\begin{table}[bt]
\centering
  \caption{Comparative results on self-constructed dataset (Higher is better, and the best results are shown in bold). The proposed method achieves the best IoU score among the listed methods.}\label{tab_suyu_com}
\resizebox{\linewidth}{!}{
\begin{tabular}{cccccc}
\toprule
{\textbf{Methods}} & \textbf{IoU(\%)}   & \textbf{mIoU(\%)}  & \textbf{P(\%)} & \textbf{R(\%)} & \textbf{F1(\%)} \\ 
\midrule
Adv		&22.18	&58.46	&76.72	&23.78   &36.30   \\
Gct      &10.92	&52.58	&76.84	&11.28    &19.69	\\
S4l	    &36.08   &65.53  &65.72  &44.44   &53.03    \\
Classmix  &52.12	&73.96	&66.39	&70.83  &68.52   \\
Cutmix	&51.32	&73.62	&69.36	&66.37  &67.83	\\
Cutout   &51.67  &73.79   &68.95  &67.33   &68.14  \\
Reco    & 52.76  &74.38    &69.22  &68.93   &69.08    \\
\midrule
\makecell[c]{SRUNet  \\ (Ours) }  & 74.52  &86.30  &85.24 &85.57   &85.40 \\
\makecell[c]{SRUNet  \\ (Nan jing) }  & 72.96  &85.57  &\textbf{86.67}  &82.18   &84.37 \\
\makecell[c]{SRUNet  \\ (Zheng zhou) }  & \textbf{75.74}  &\textbf{86.86}  &84.19  &\textbf{88.28}   &\textbf{86.19} \\
\bottomrule
\end{tabular}
}
\end{table}                             
To verify effectiveness in practical application, we also conducted comparative experiments on the self-constructed dataset. Considering that the ground truth was generated from centerline data with different widths, the prediction precision was generally lower than that of the DeepGlobe dataset. \tabref{tab_suyu_com} shows our results; we also report the individual results from the Nanjing and Zhengzhou test sets.

The results are presented in \tabref{tab_suyu_com}. First, our model achieved optimal scores on all five metrics and showed an improvement of approximately 20\% in the IoU metric compared with that obtained using ReCo. This was partly because of the improvements in the model itself, as our model concentrated more on the edge details, and the ReCo-based semi-supervised approach could improve the model ability to classify different categories. However, the main changes in road elements occurred in urban and rural areas, and the available data provided rich prior knowledge for urban areas, improving the stability of the prediction process in road detection. 
Second, we found that the stability of the semi-supervised methods varied on the same dataset. This may be attributed to the different objective functions for various tasks and methods, and the significantly varying learning ability of the models, particularly when extracting different category features and distinguishing between different categories. For the remote-sensing image task, the contrastive-based learning method used in this study achieved better results than those obtained with other methods. 
Third, the prediction results for the Zhengzhou area were better than those for the Nanjing area, and the IoU index was superior by 2.78 \%. The images show that the roads in Zhengzhou are relatively regular and clear, whereas those in Nanjing are more complex. This shows that the ability of the current methods to deal with complex road conditions still has  potential for further improvement.

As shown in \figref{vis_suyu_com}, we selected six typical images ( $512\times 512$ ), of which the first three rows were from Nanjing, the last three rows were from Zhengzhou, and the first two columns were the images and ground truth. As seen from the visualization results, for road areas with obvious features, all experimental methods could detect road areas to a certain extent; however, for areas with rural paths, only ReCo and SRUNet proposed in this study could effectively restore spatial details. Moreover, the detection results of the proposed method were more continuous and complete, particularly for parallel roads and complex intersection areas. Thus, our proposed method can obtain clearer extraction results, demonstrating the effectiveness of our method in practical use.

\subsection{Road Updating Results and Analysis}
\begin{figure*}[!h]
        \small      
           \centering
        		\newcommand{\tabincell}[2]{\begin{tabular}{@{}#1@{}}#2\end{tabular}}
        		\begin{tabular}{cc}
        \multirow{8}{*}[11.5ex]{\includegraphics[width=0.7\textwidth]{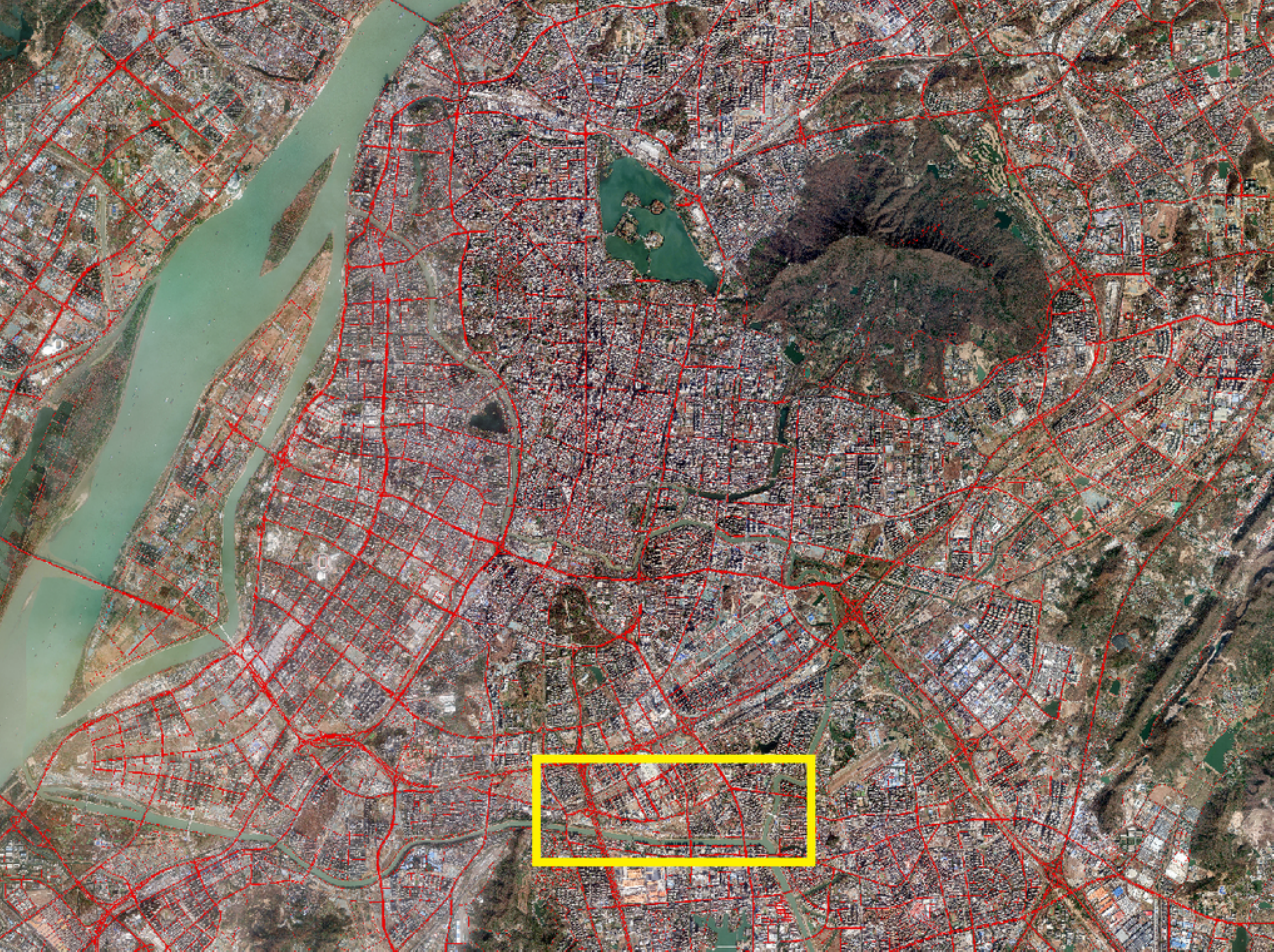}}  
        			& 
        \includegraphics[width=0.3\textwidth]{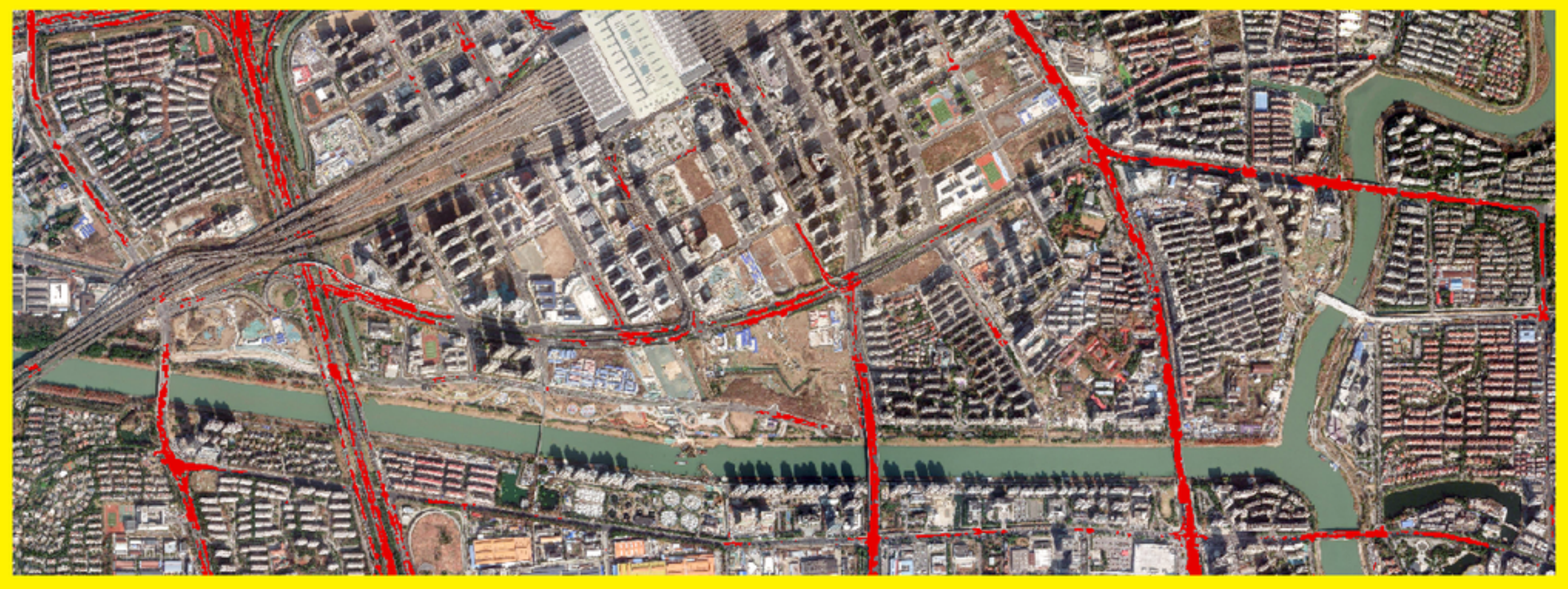}
                    \\
        
        			&        (a) Adv
                    \\
        
        			& 
        \includegraphics[width=0.3\textwidth]{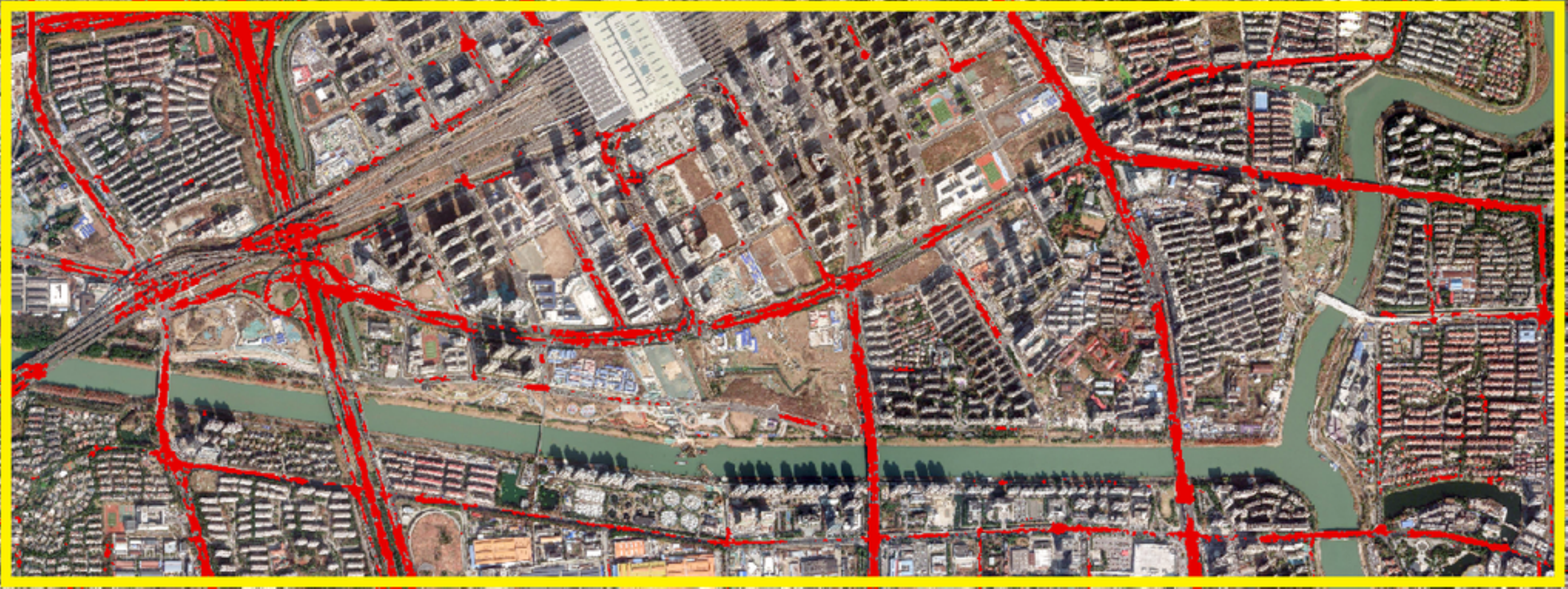}
                    \\
        
        			&         (b) S4l
                    \\
        
        			& 
        \includegraphics[width=0.3\textwidth]{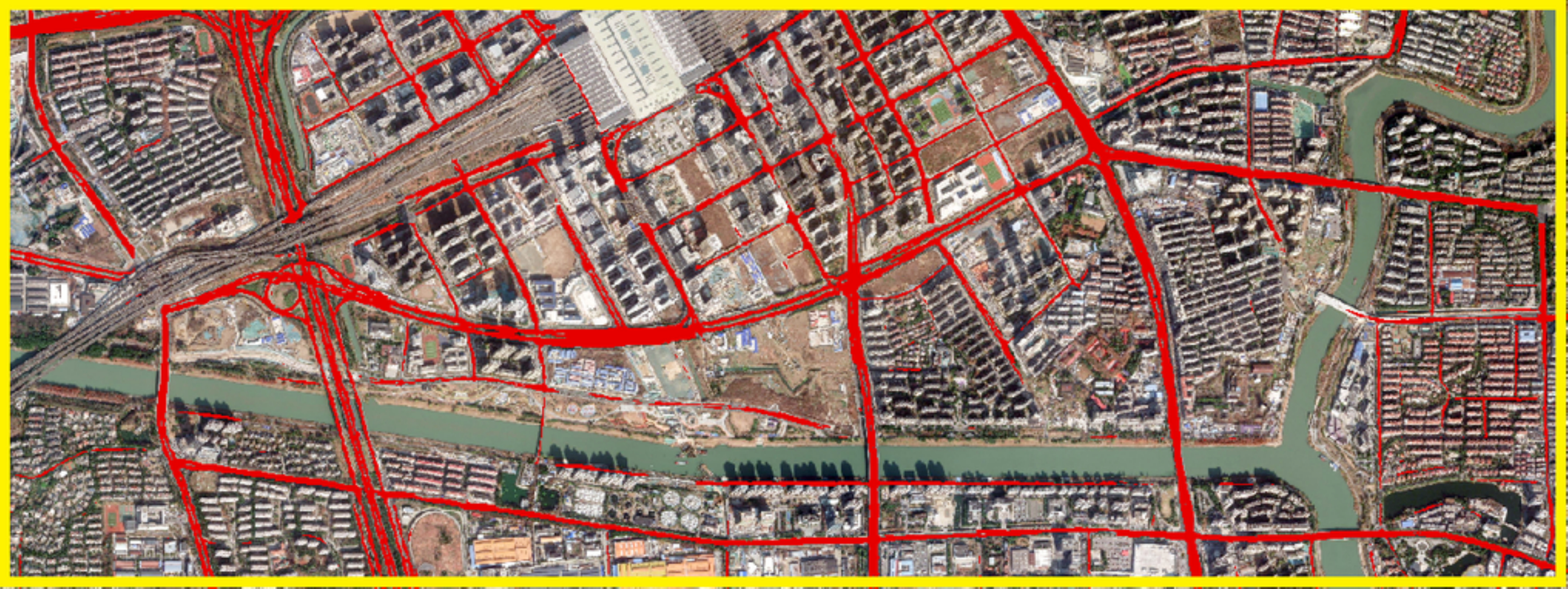}
                    \\
        
        			&         (c) ReCo
                    \\
        
        			& 
        \includegraphics[width=0.3\textwidth]{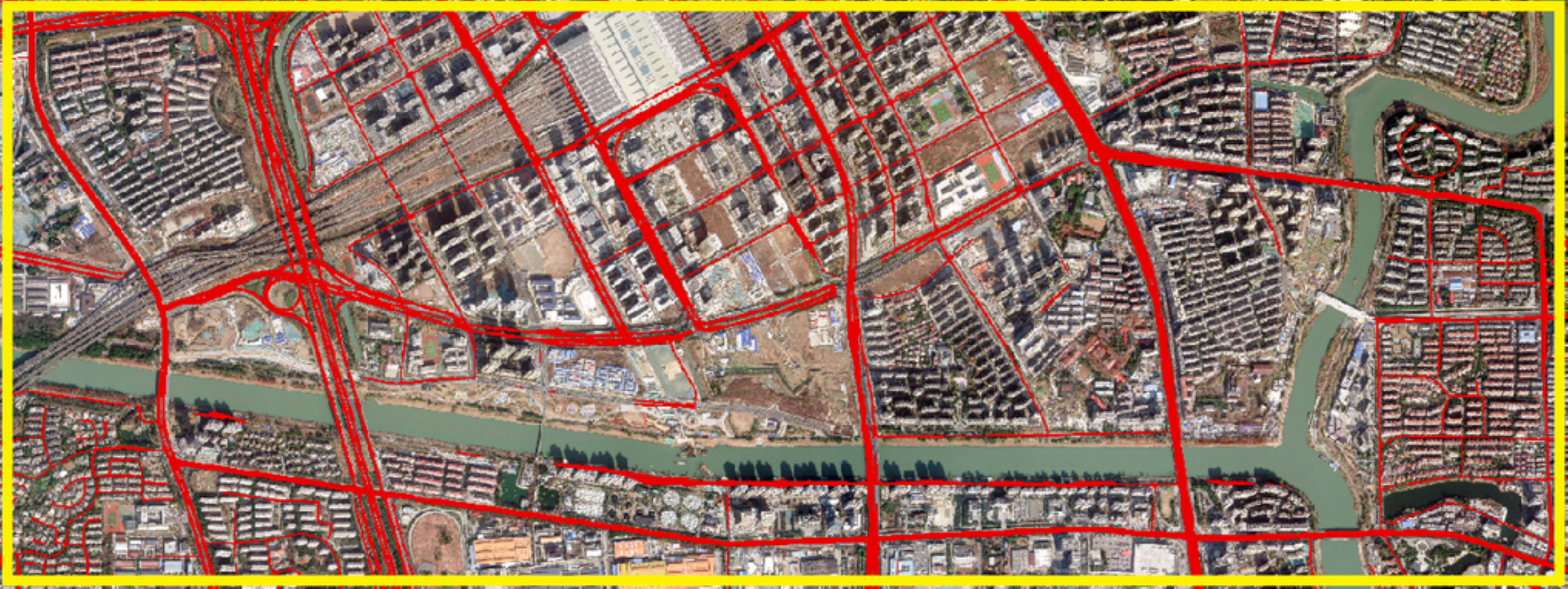}
                    \\
        
        			&         (d) SRUNet
        		  \\
        \multicolumn{2}{c}{(A) Nan Jing, Jiangsu Province, China}
                    \\
        \multirow{4}{*}[11.5ex]{\includegraphics[width=0.7\textwidth]{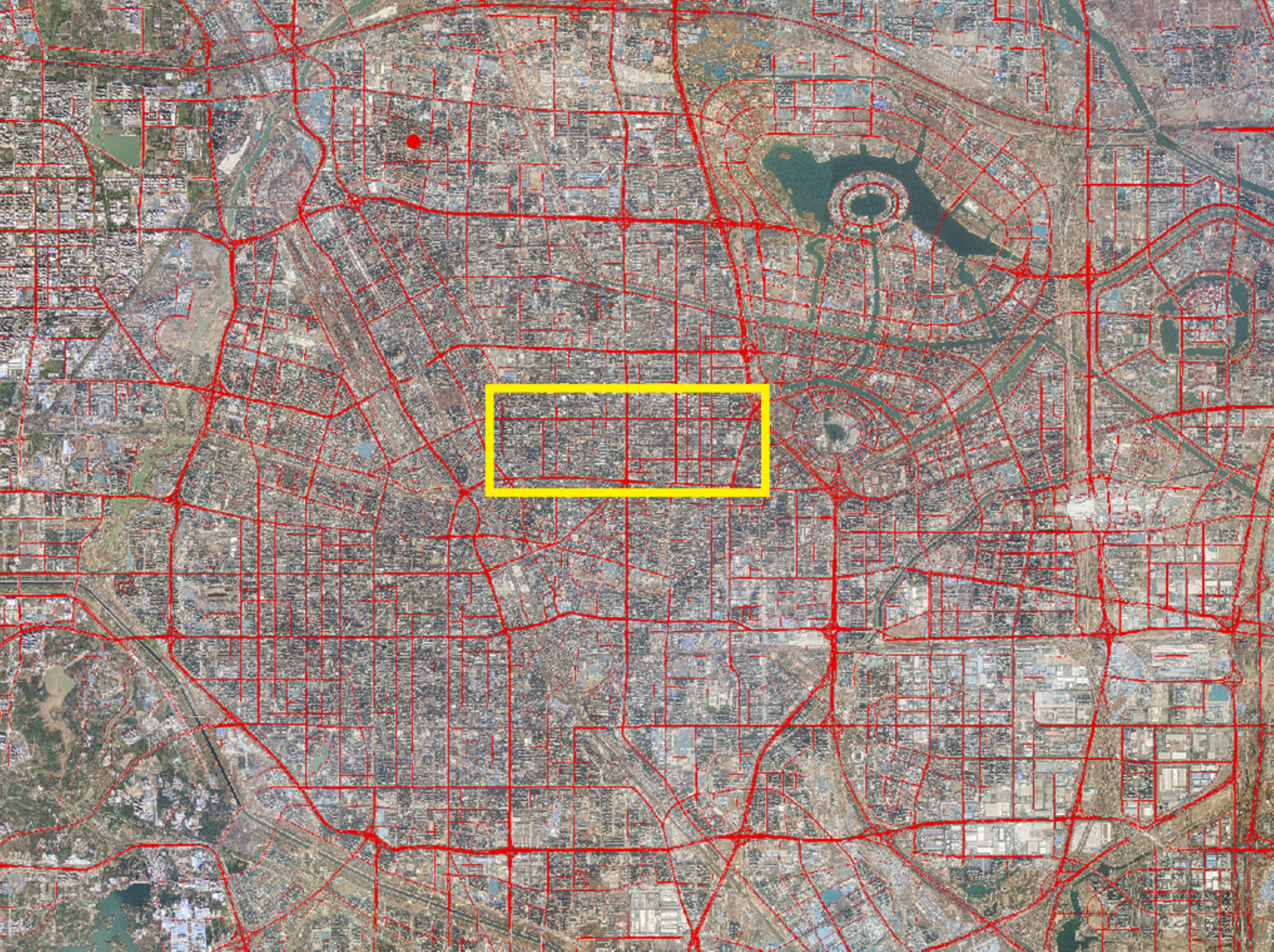}}  
        			& 
        \includegraphics[width=0.3\textwidth]{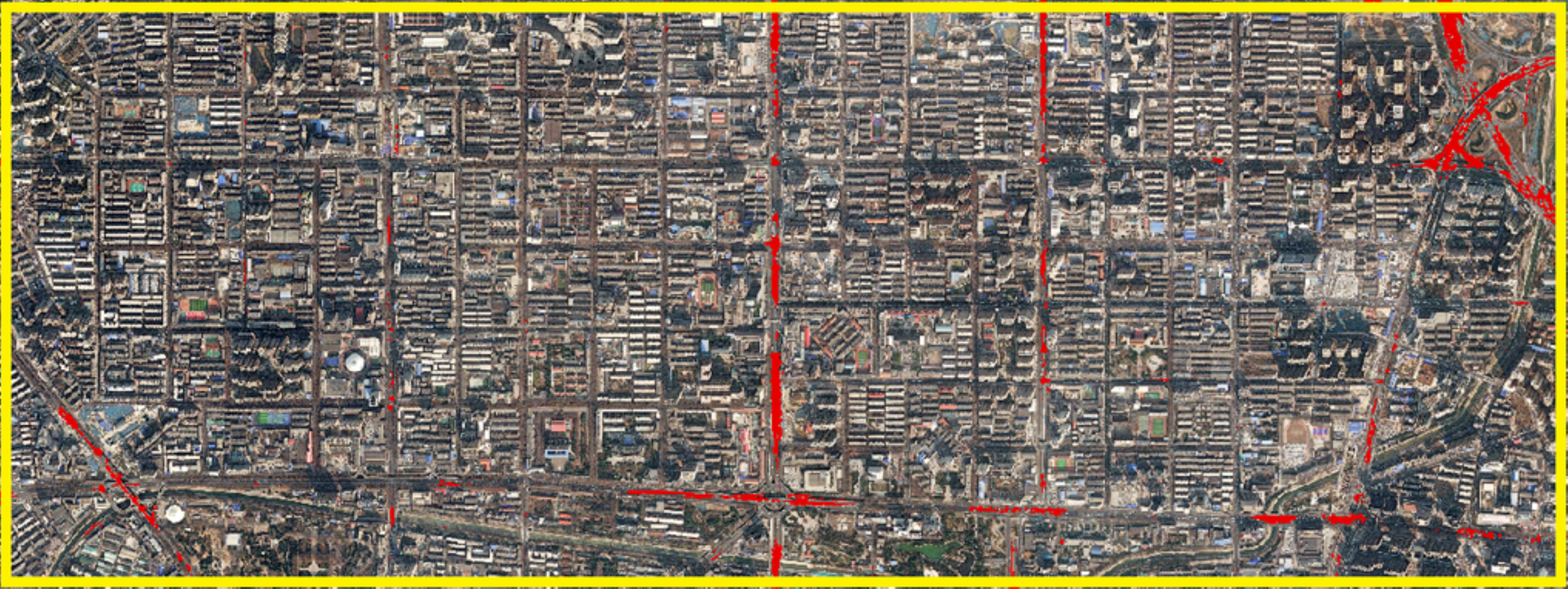}
                    \\
        
        			&        (a) Adv
                    \\
        
        			& 
        \includegraphics[width=0.3\textwidth]{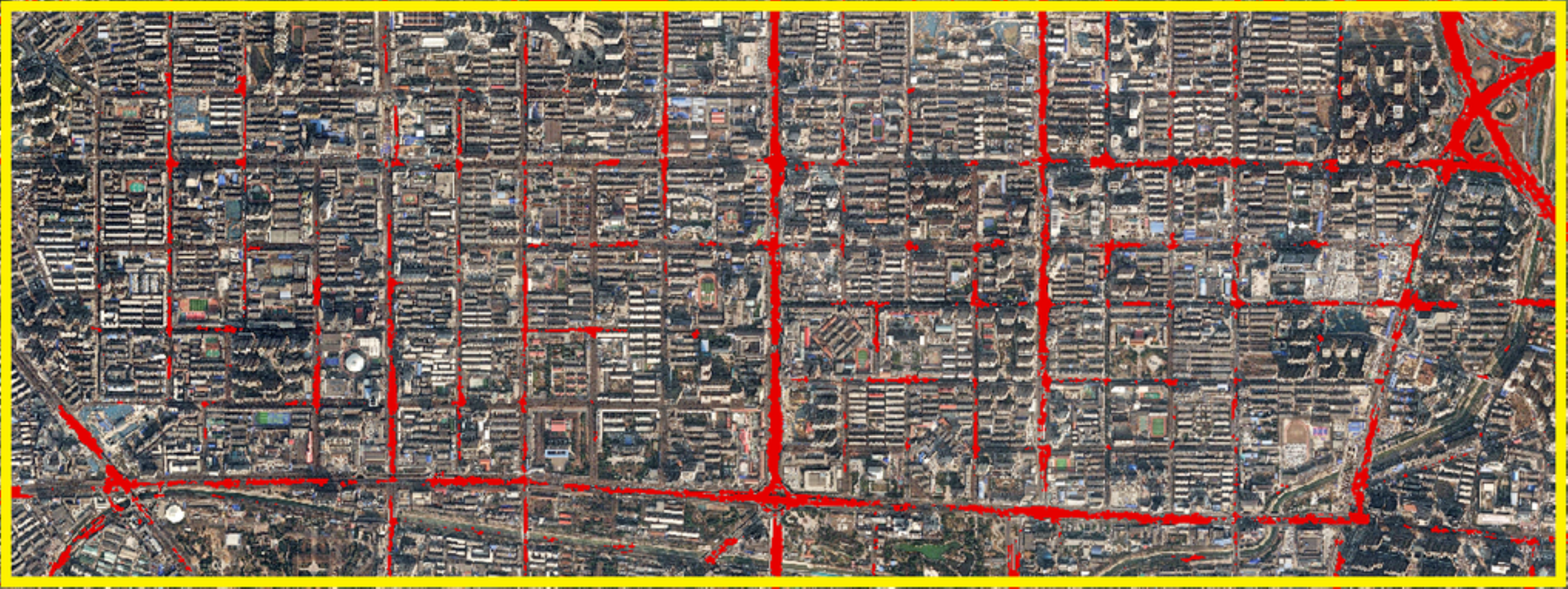}
                    \\
        
        			&         (b) S4l
                    \\
        
        			& 
        \includegraphics[width=0.3\textwidth]{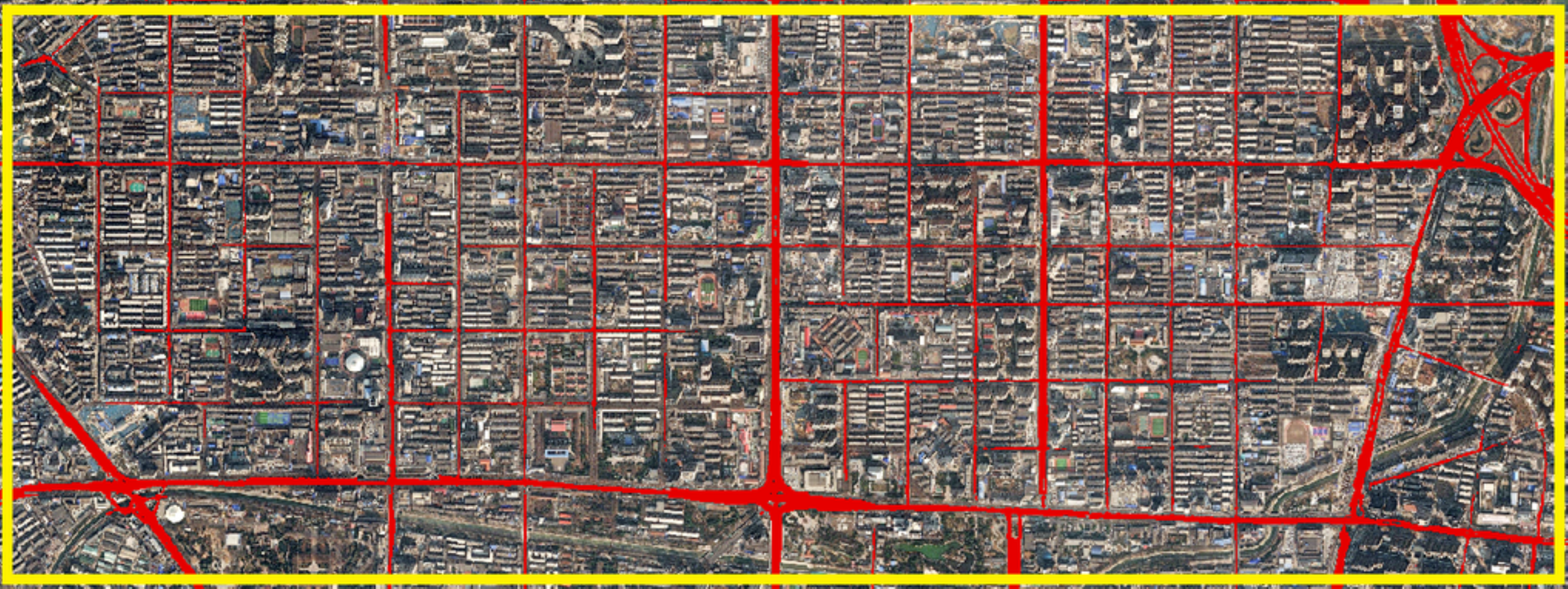}
                    \\
        
        			&         (c) ReCo
                    \\
        
        			& 
        \includegraphics[width=0.3\textwidth]{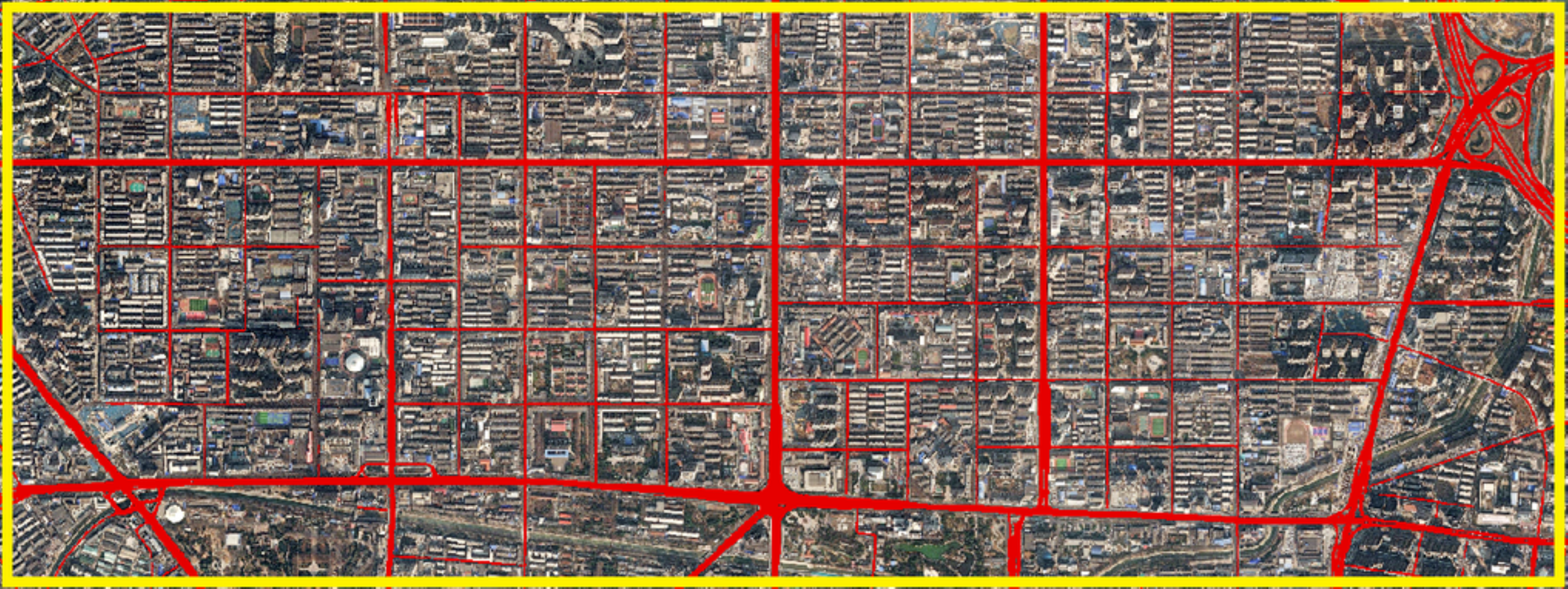}
                    \\
        
        			&         (d) SRUNet
        		  \\
        \multicolumn{2}{c}{(B) Zheng Zhou, Henan Province, China}
        \end{tabular}%
        \caption{Visualization results of comparison methods on the large-scale self-constructed Road Dataset. In a large image area, the proposed method shows better continuity.} 
        \label{update_results_large} 
    \end{figure*}  

\begin{figure*}[htb]
\small
   \centering
		\newcommand{\tabincell}[2]{\begin{tabular}{@{}#1@{}}#2\end{tabular}}
		\begin{tabular}{c c c c}
\includegraphics[width=0.15\textwidth]{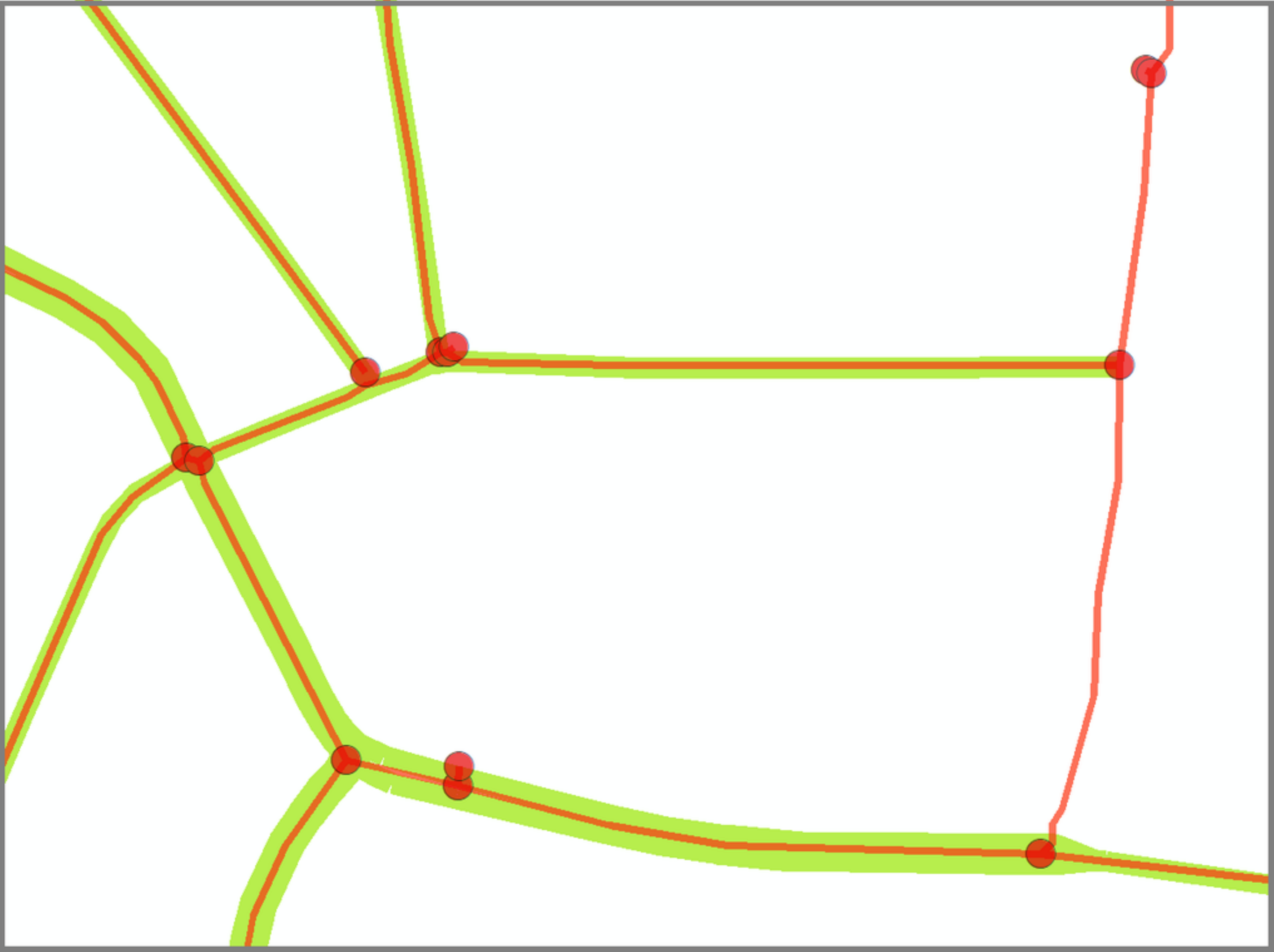}
			& 
\includegraphics[width=0.15\textwidth]{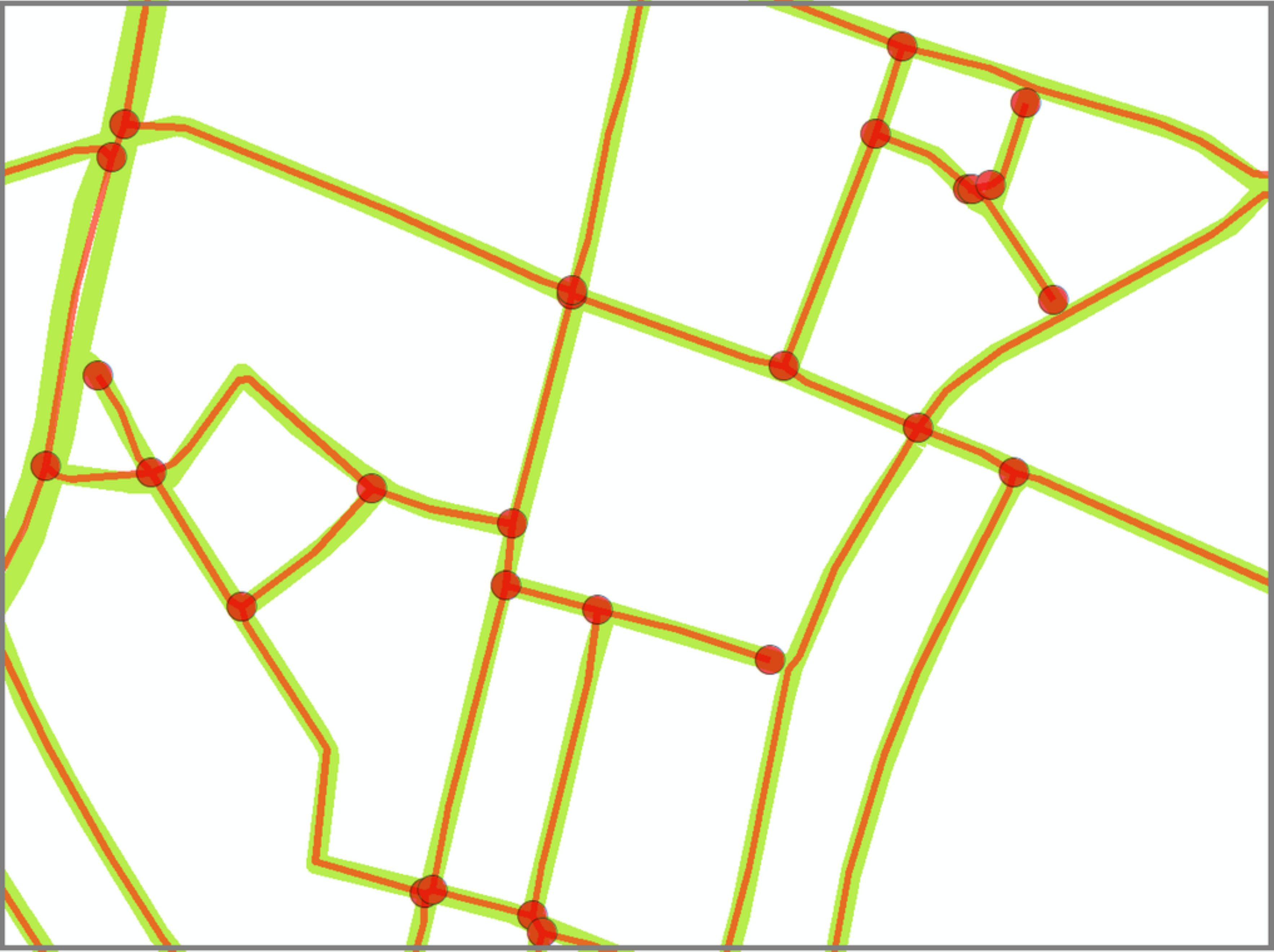}
			& 
\includegraphics[width=0.15\textwidth]{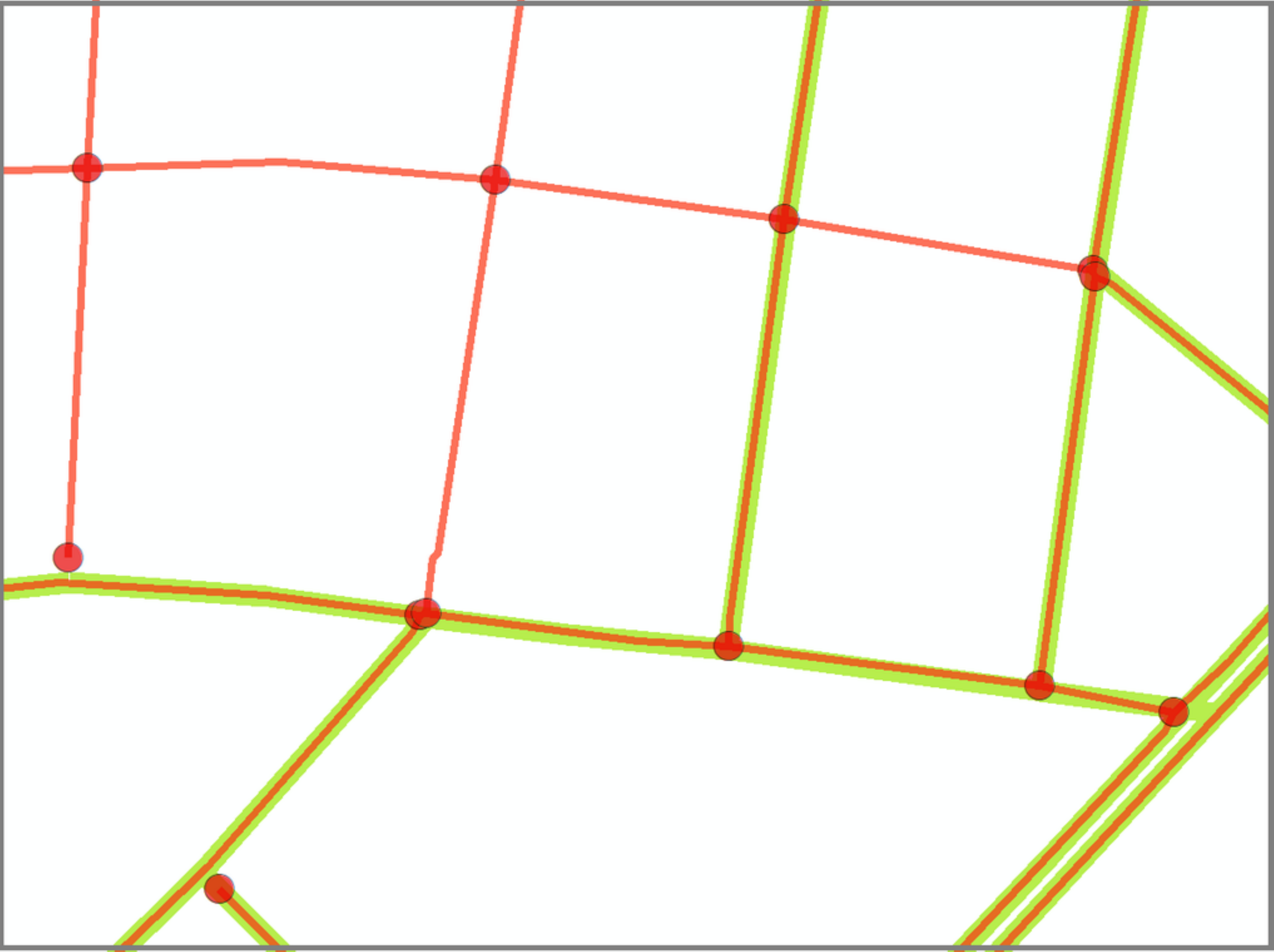}
			& 
\includegraphics[width=0.15\textwidth]{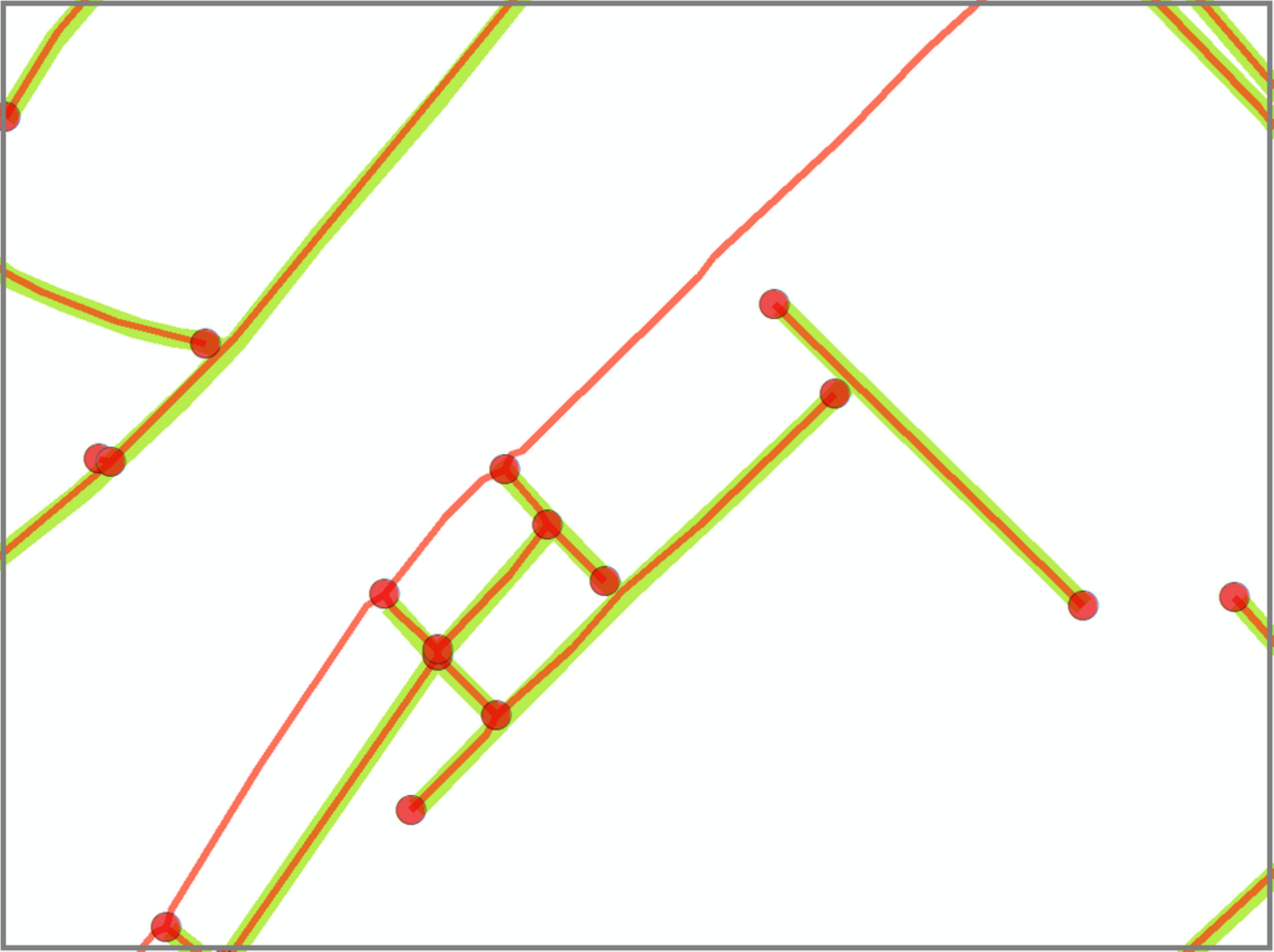}
			\\ 
\includegraphics[width=0.15\textwidth]{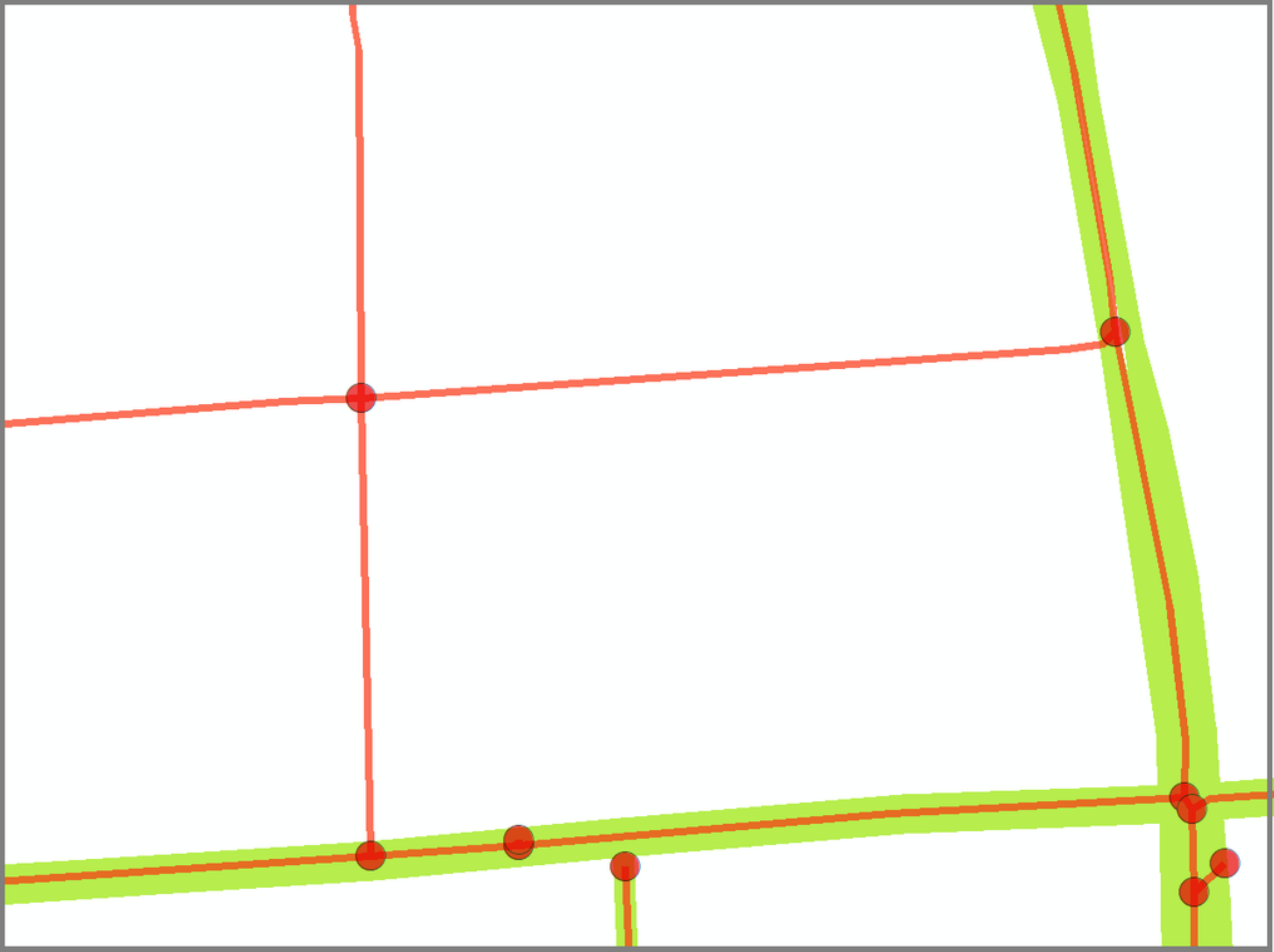}
			& 
\includegraphics[width=0.15\textwidth]{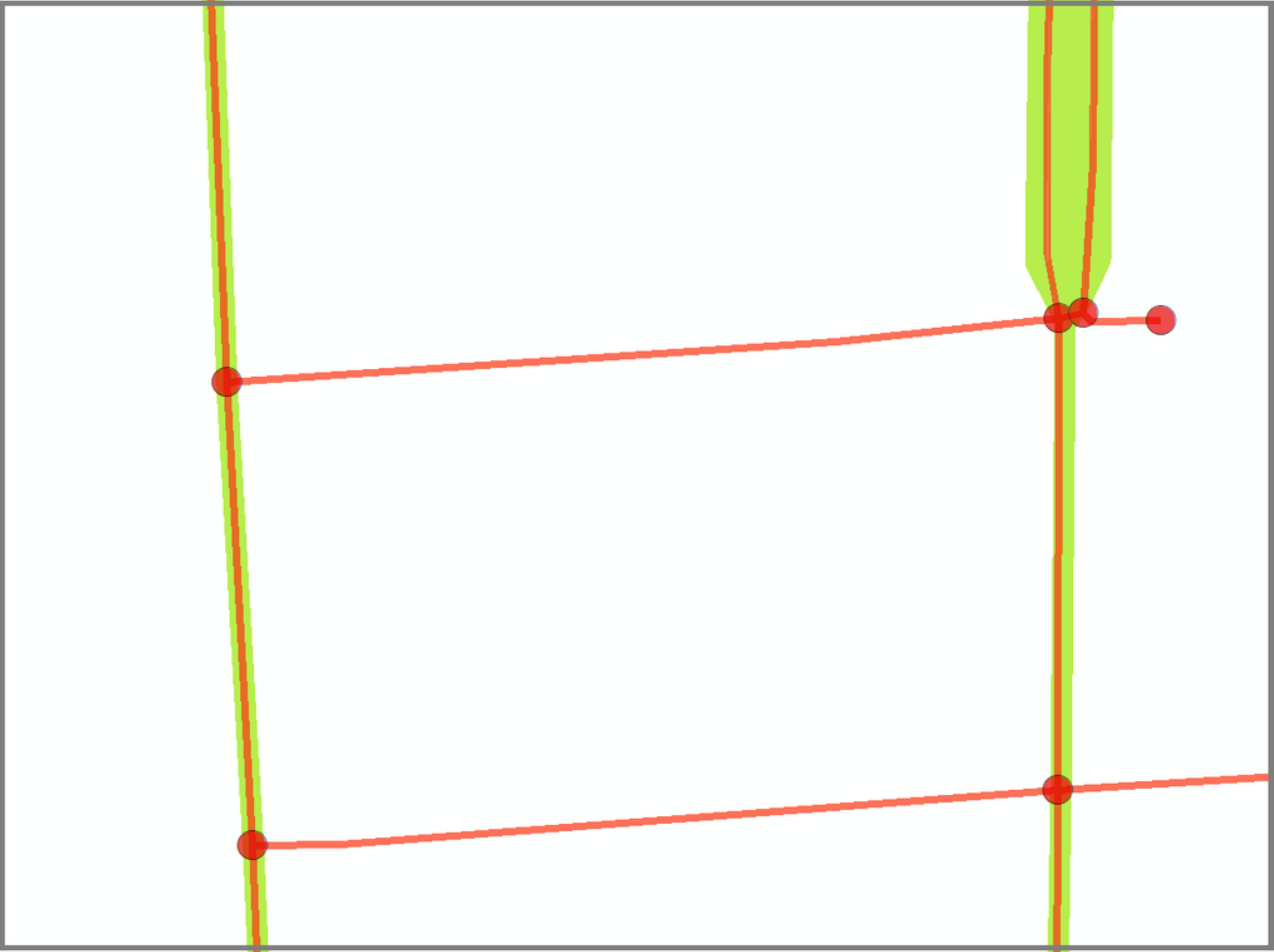}
			& 
\includegraphics[width=0.15\textwidth]{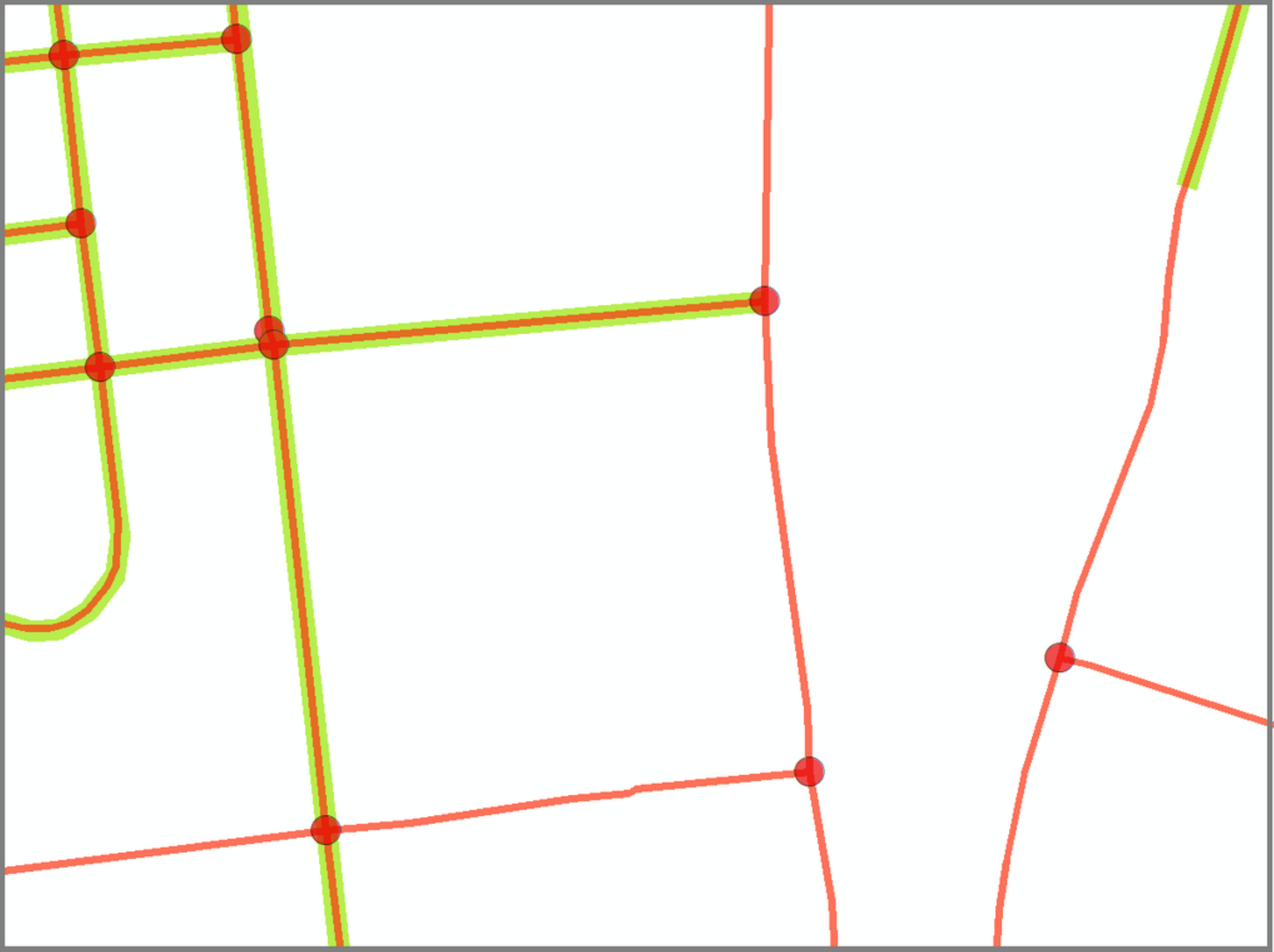}
			& 
\includegraphics[width=0.15\textwidth]{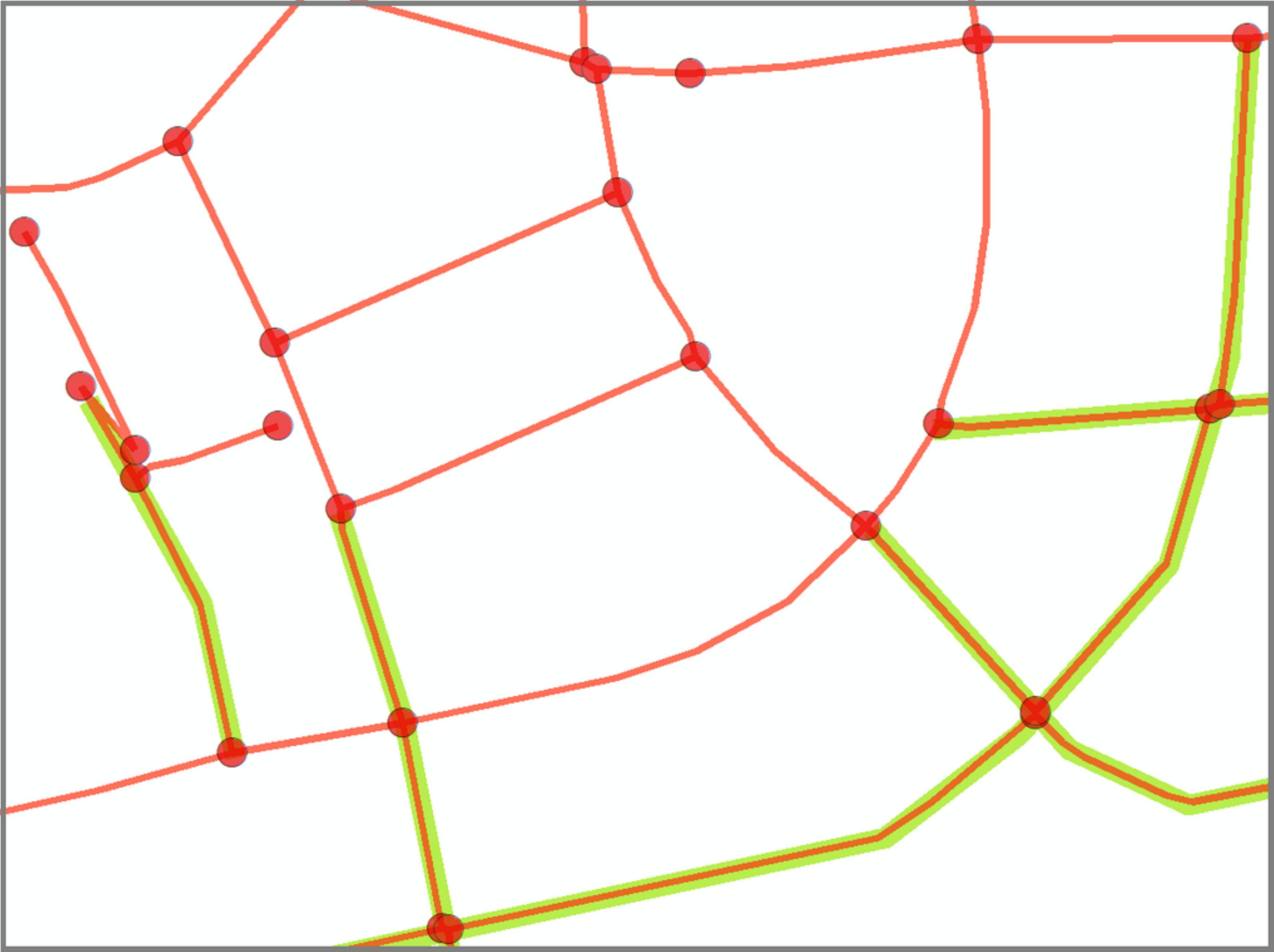}
\end{tabular}%
\caption{Results of vectorized results for changed areas in Zhengzhou and Nanjing. Red represents the up-to-date road vectors. Green represents the historical road vectors. Results show that the updated road network can maintain good consistency with the historical data and has good topological consistency.} 
\label{update_results_vec} 
\end{figure*}

\begin{figure*}[h]
\small
   \centering
		\newcommand{\tabincell}[2]{\begin{tabular}{@{}#1@{}}#2\end{tabular}}
		\begin{tabular}{m{1.7cm}<{\centering} m{1.9cm}<{\centering} m{1.9cm}<{\centering} m{1.9cm}<{\centering} m{1.9cm}<{\centering} m{1.9cm}<{\centering}m{1.9cm}<{\centering}m{1.9cm}<{\centering}}
Image
            &
\includegraphics[width=0.12\textwidth]{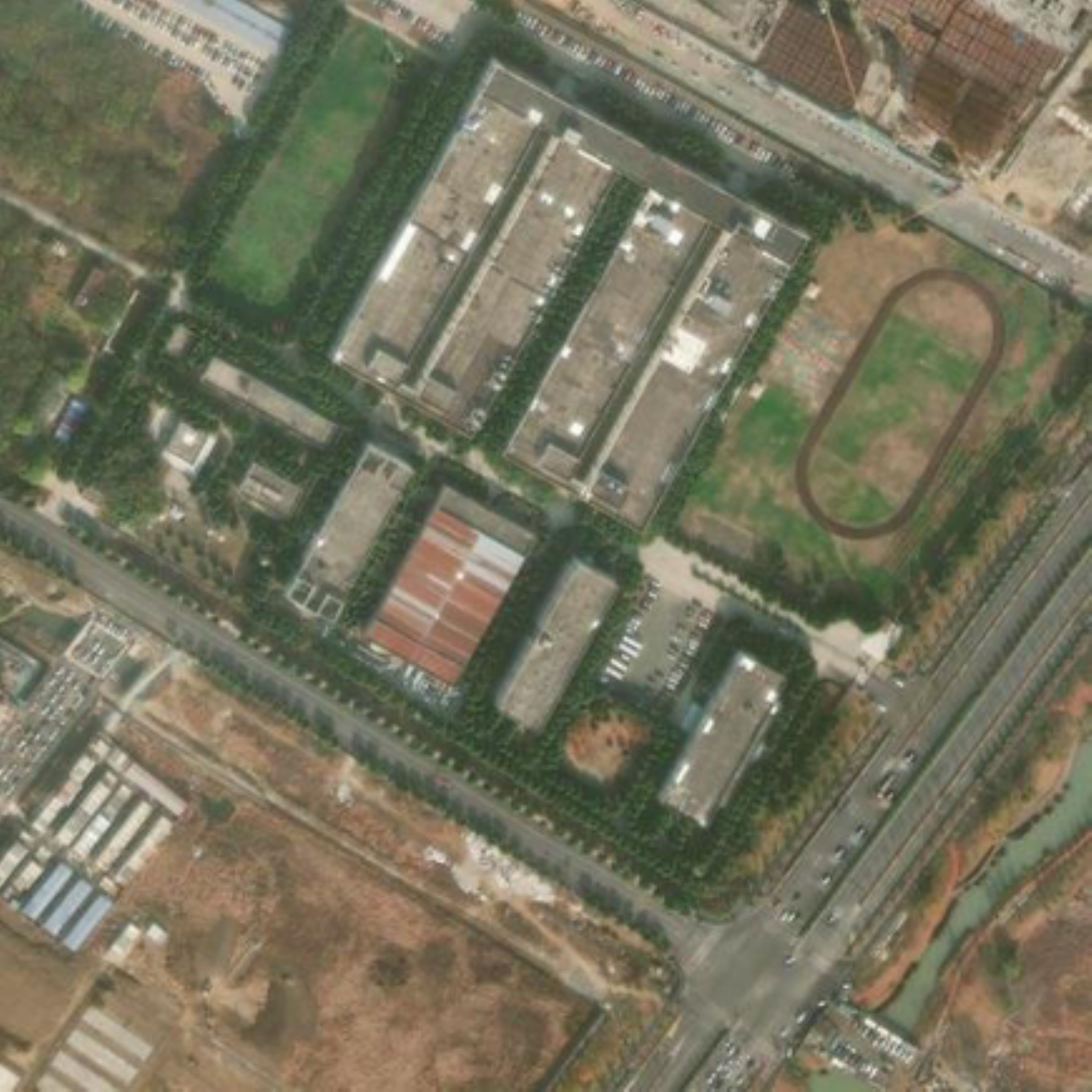}
			& 
\includegraphics[width=0.12\textwidth]{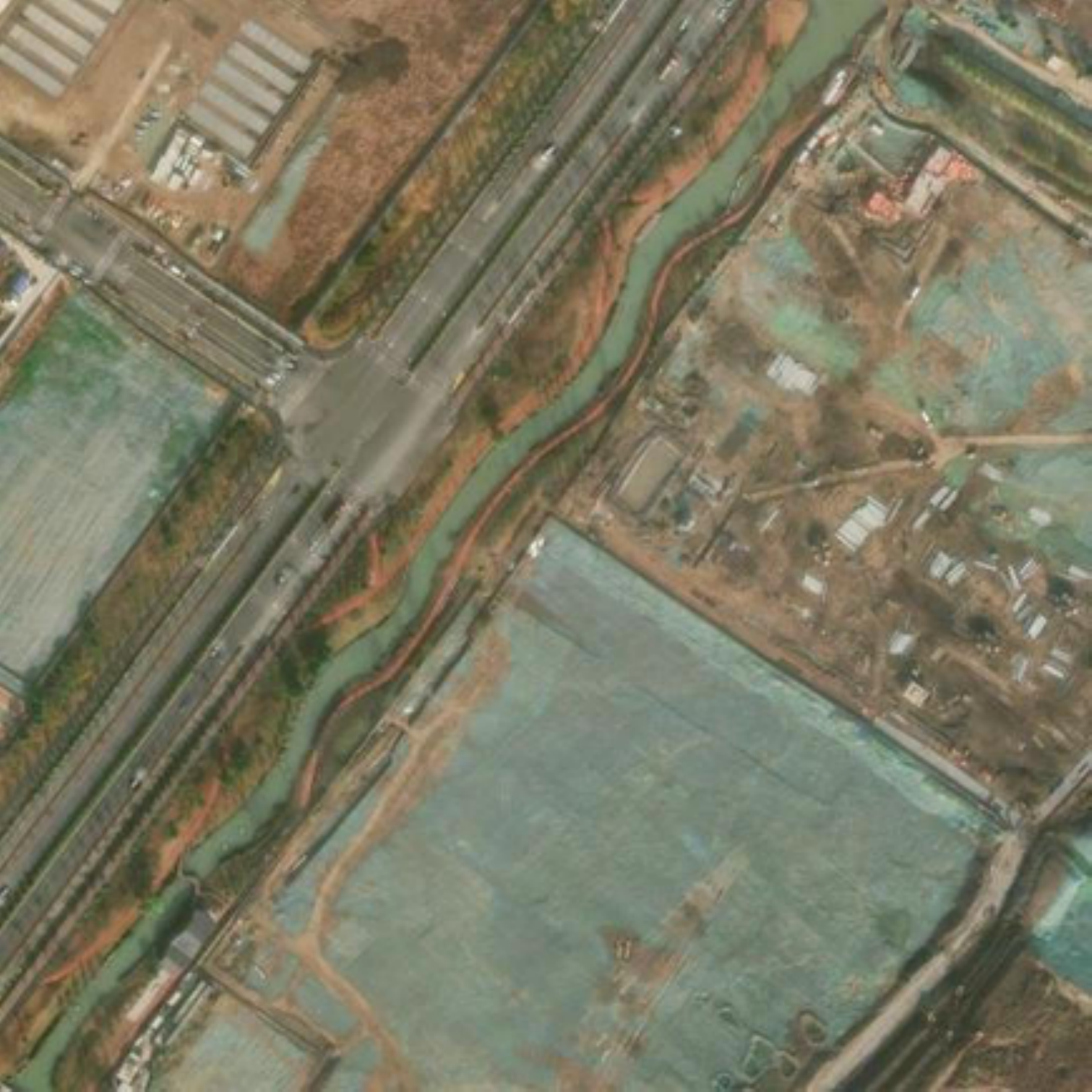}
			& 
\includegraphics[width=0.12\textwidth]{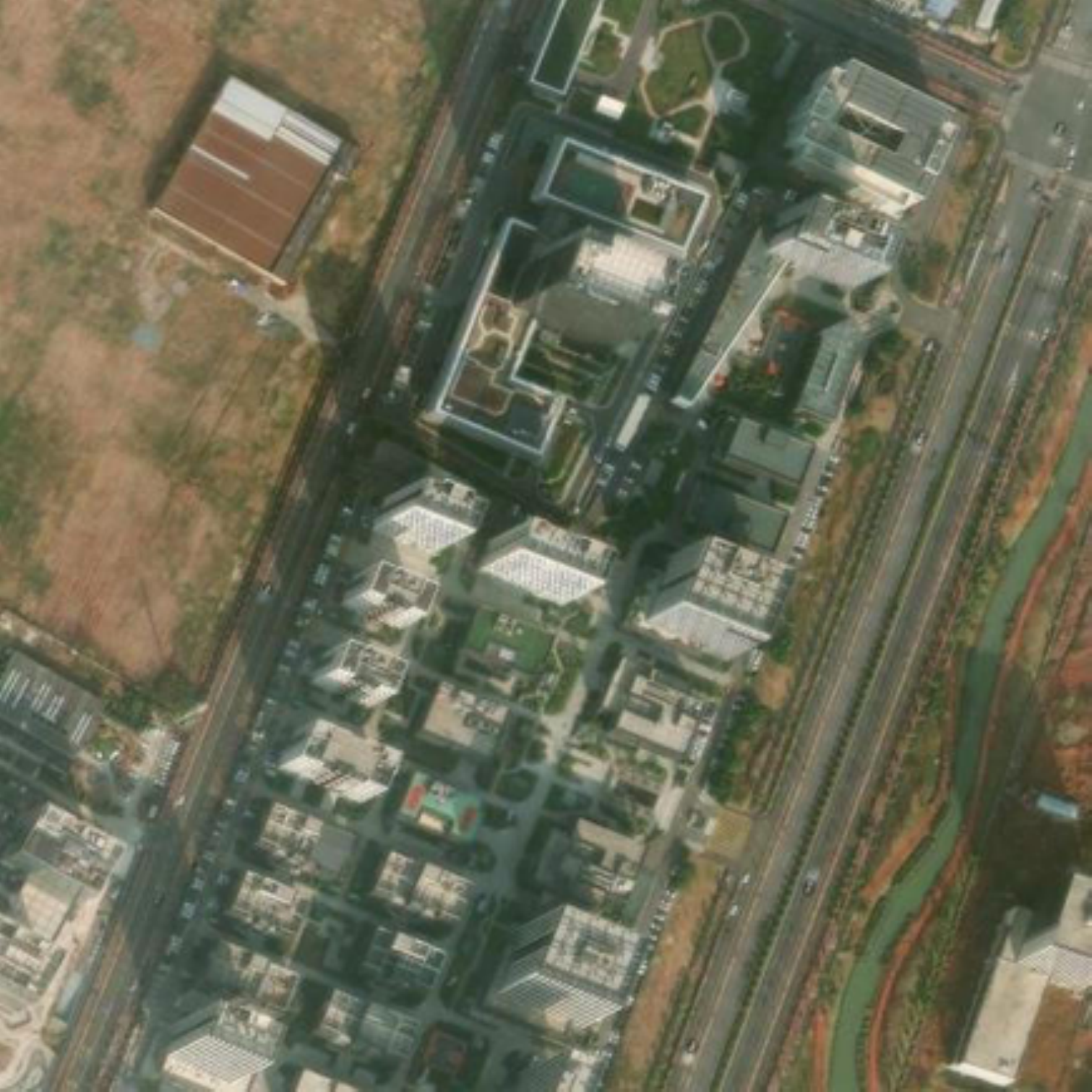}
			& 
\includegraphics[width=0.12\textwidth]{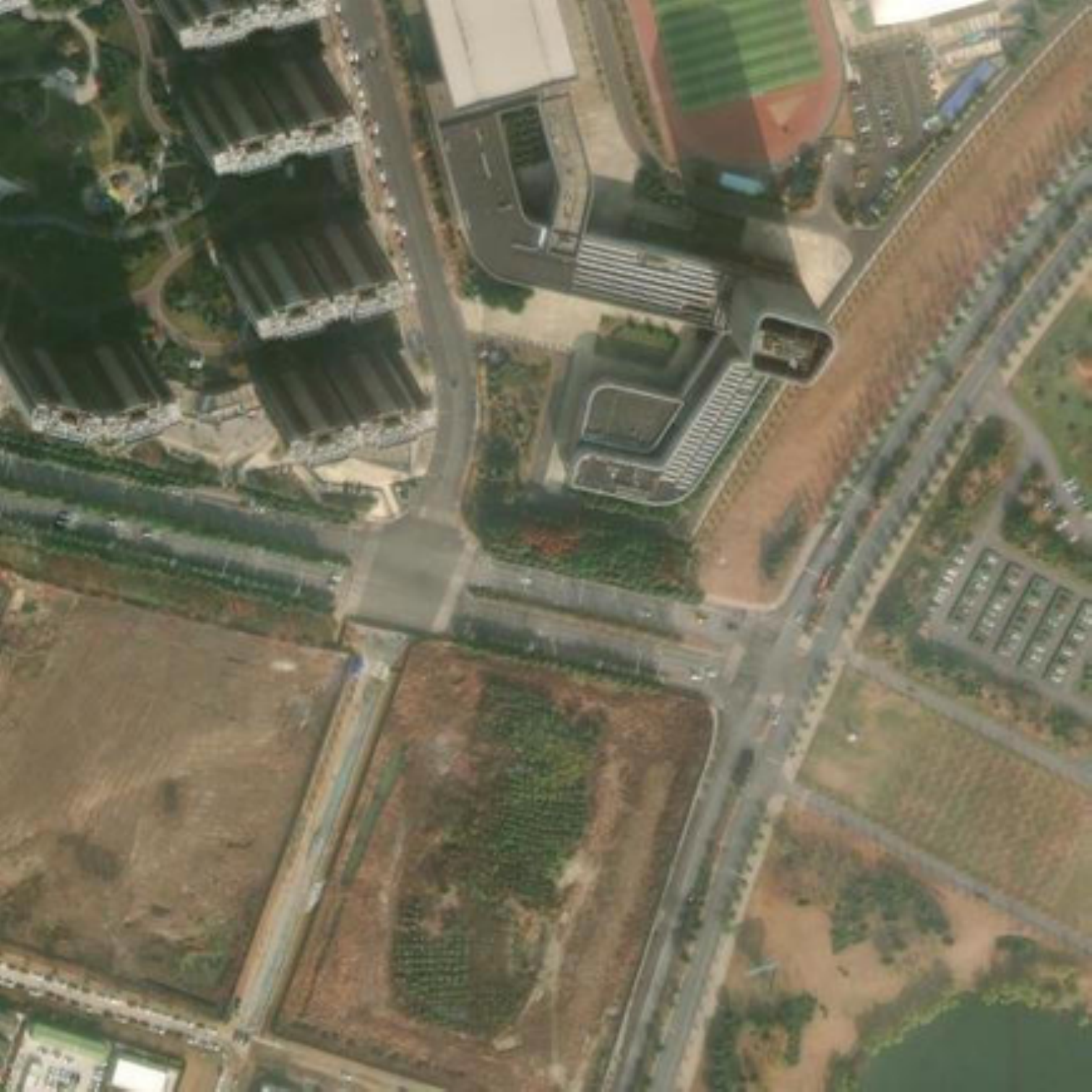}
			& 
\includegraphics[width=0.12\textwidth]{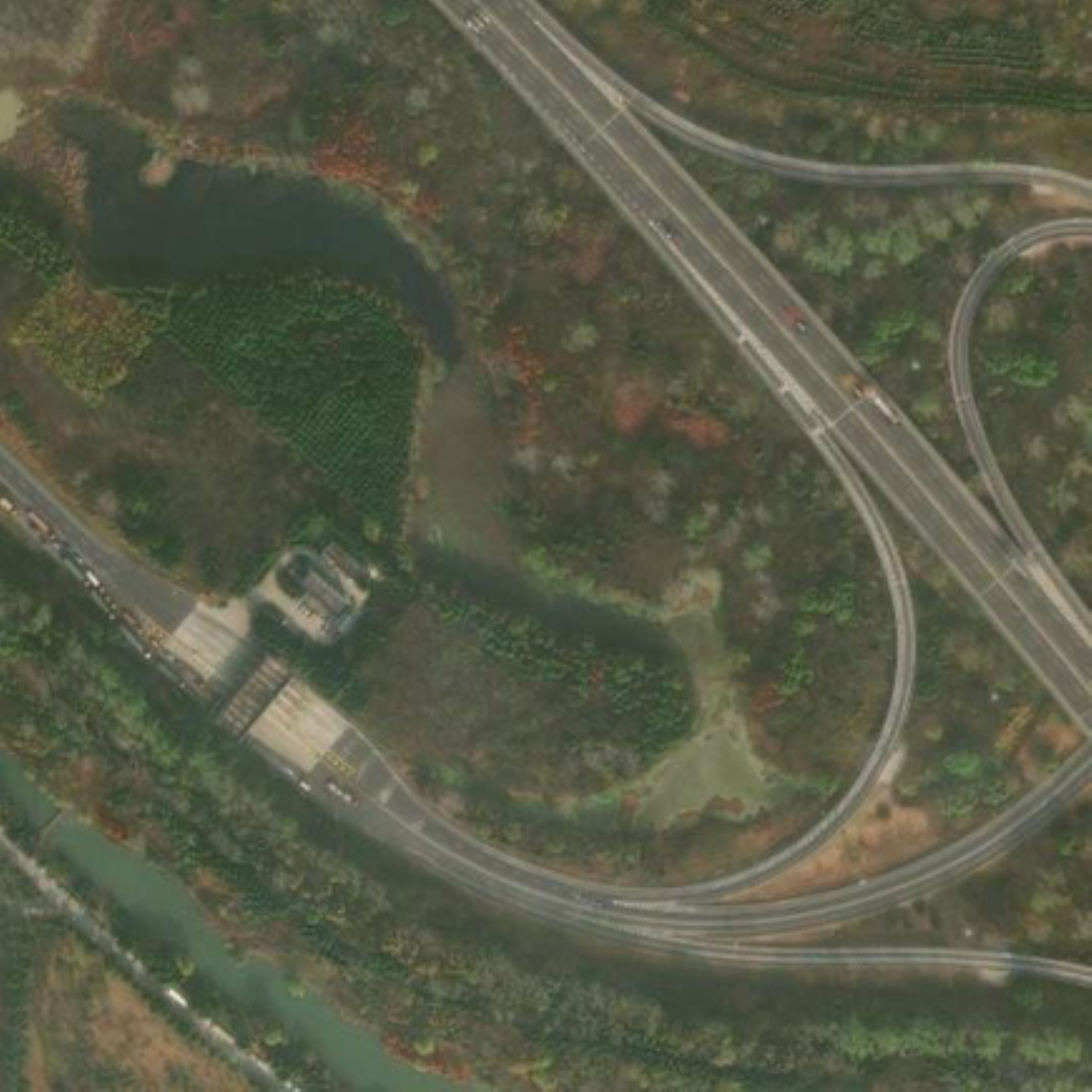}
			& 
\includegraphics[width=0.12\textwidth]{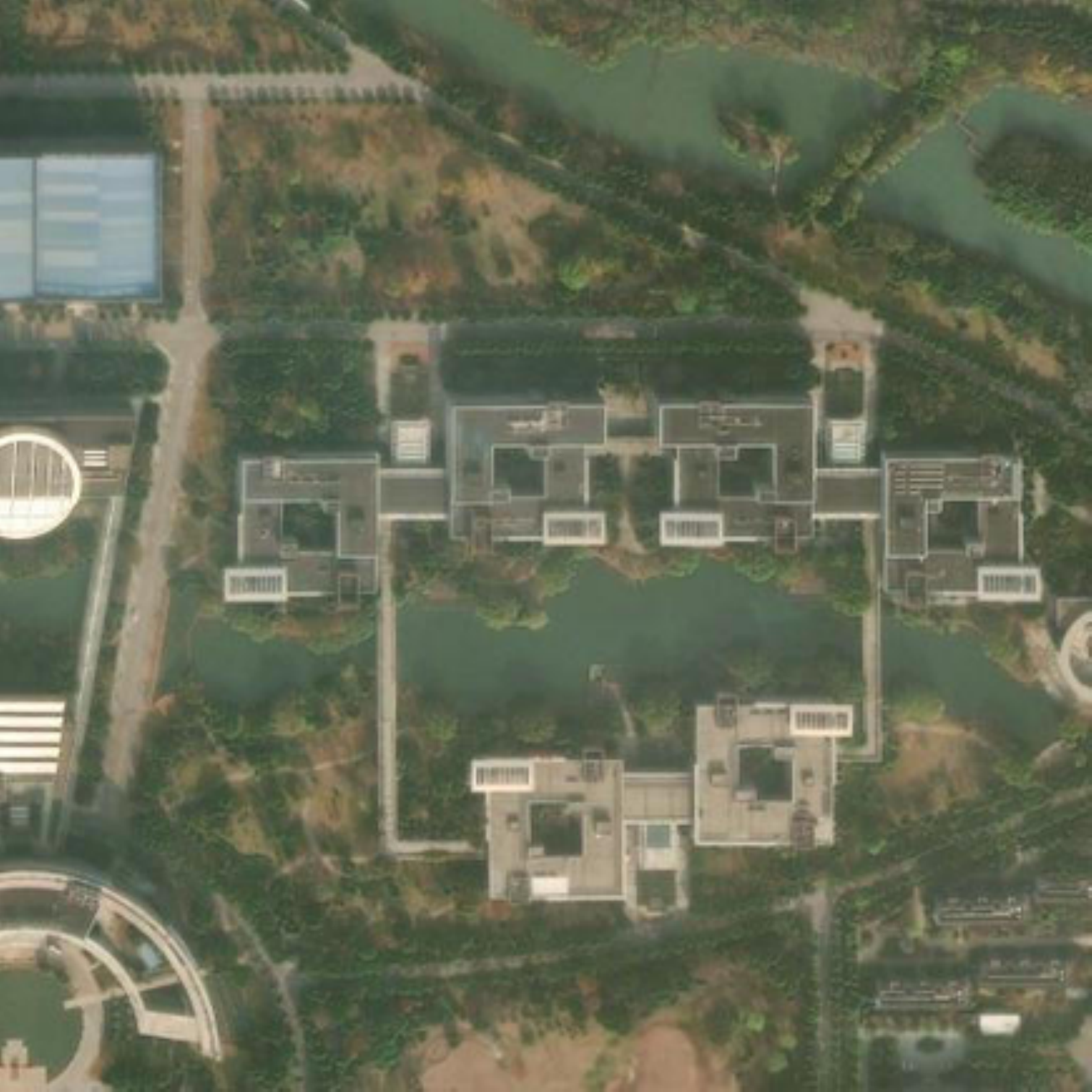}
			& 
\includegraphics[width=0.12\textwidth]{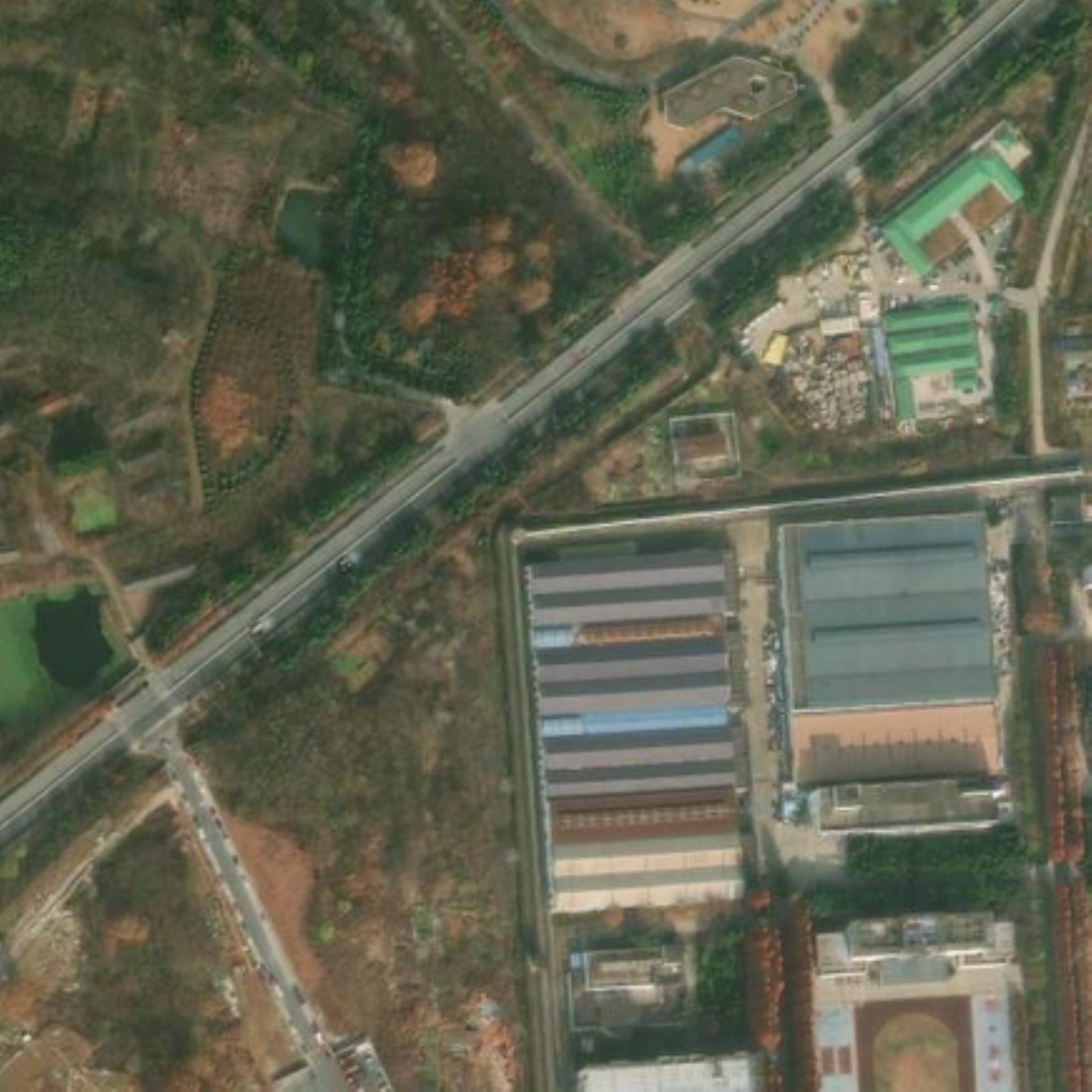}
		\\
\makecell*[c]{Historical \\ Map}
            &
\includegraphics[width=0.12\textwidth]{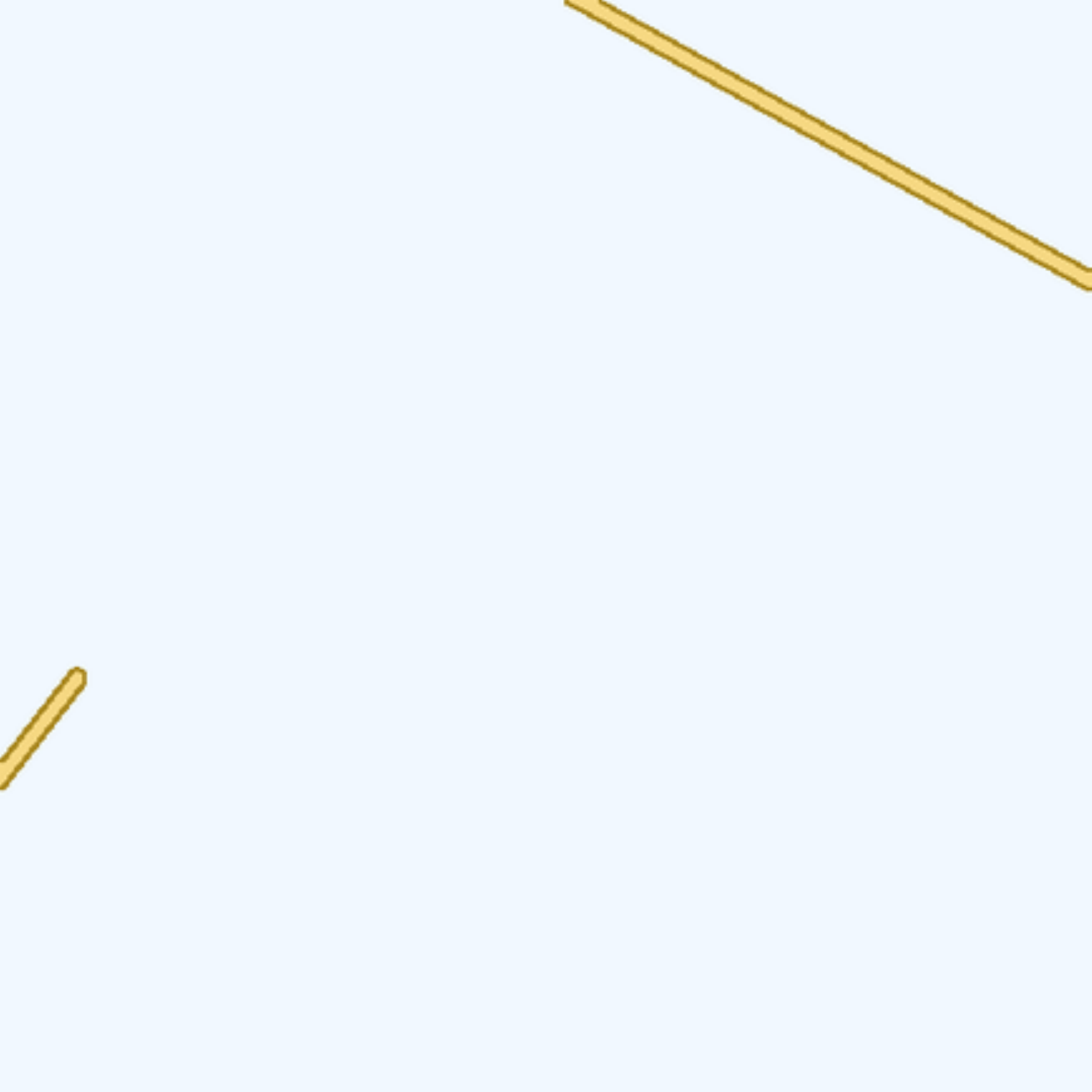}
			& 
\includegraphics[width=0.12\textwidth]{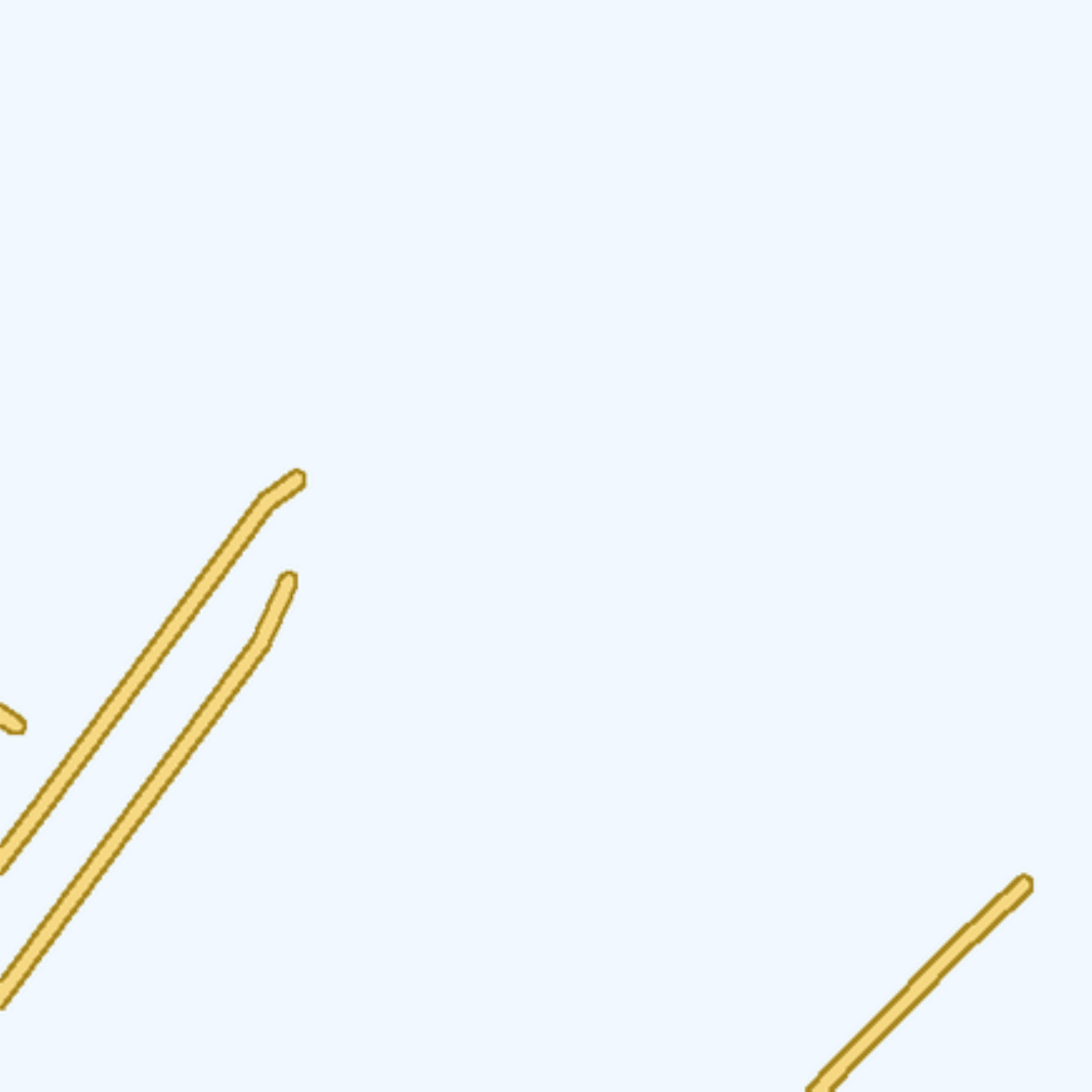}
			& 
\includegraphics[width=0.12\textwidth]{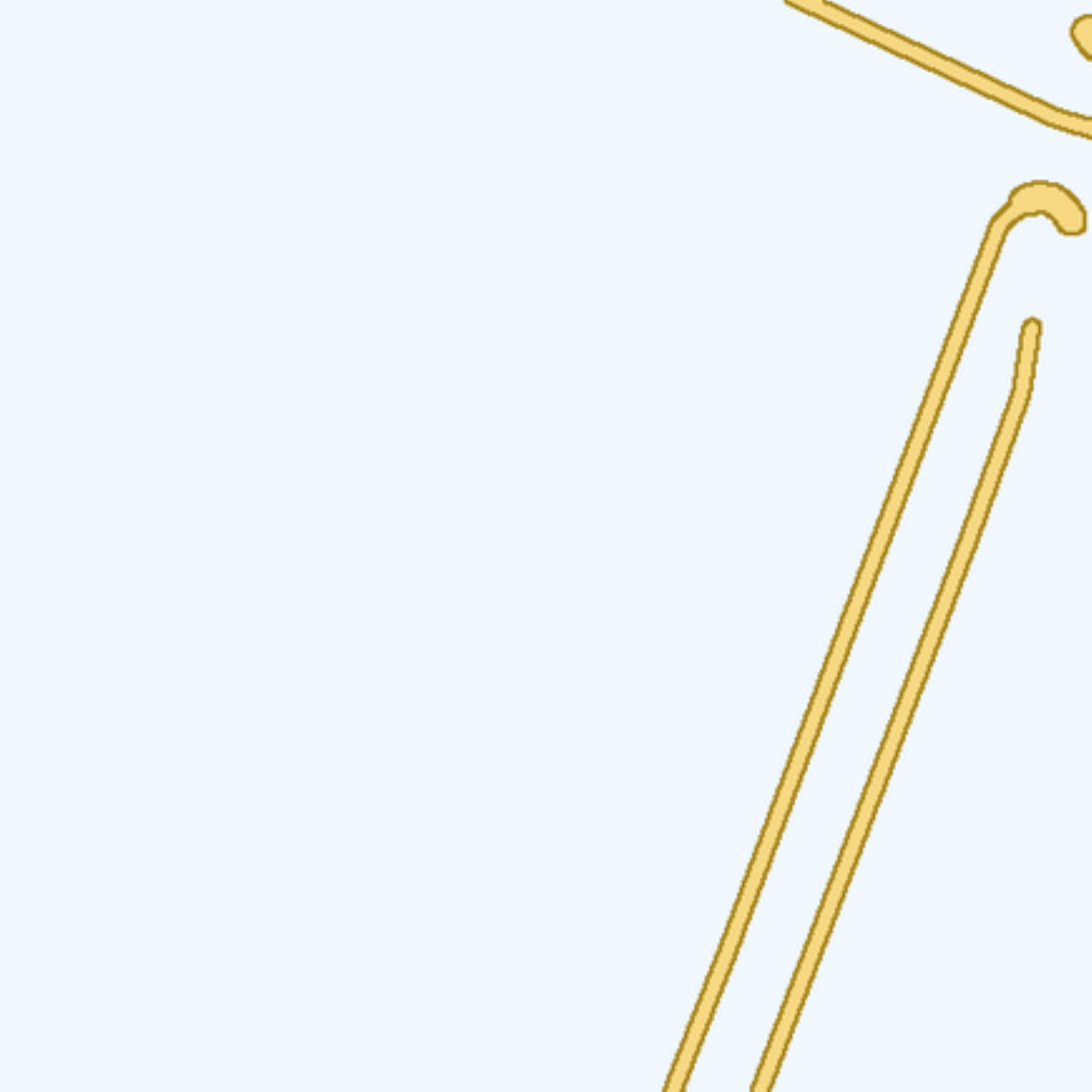}
			& 
\includegraphics[width=0.12\textwidth]{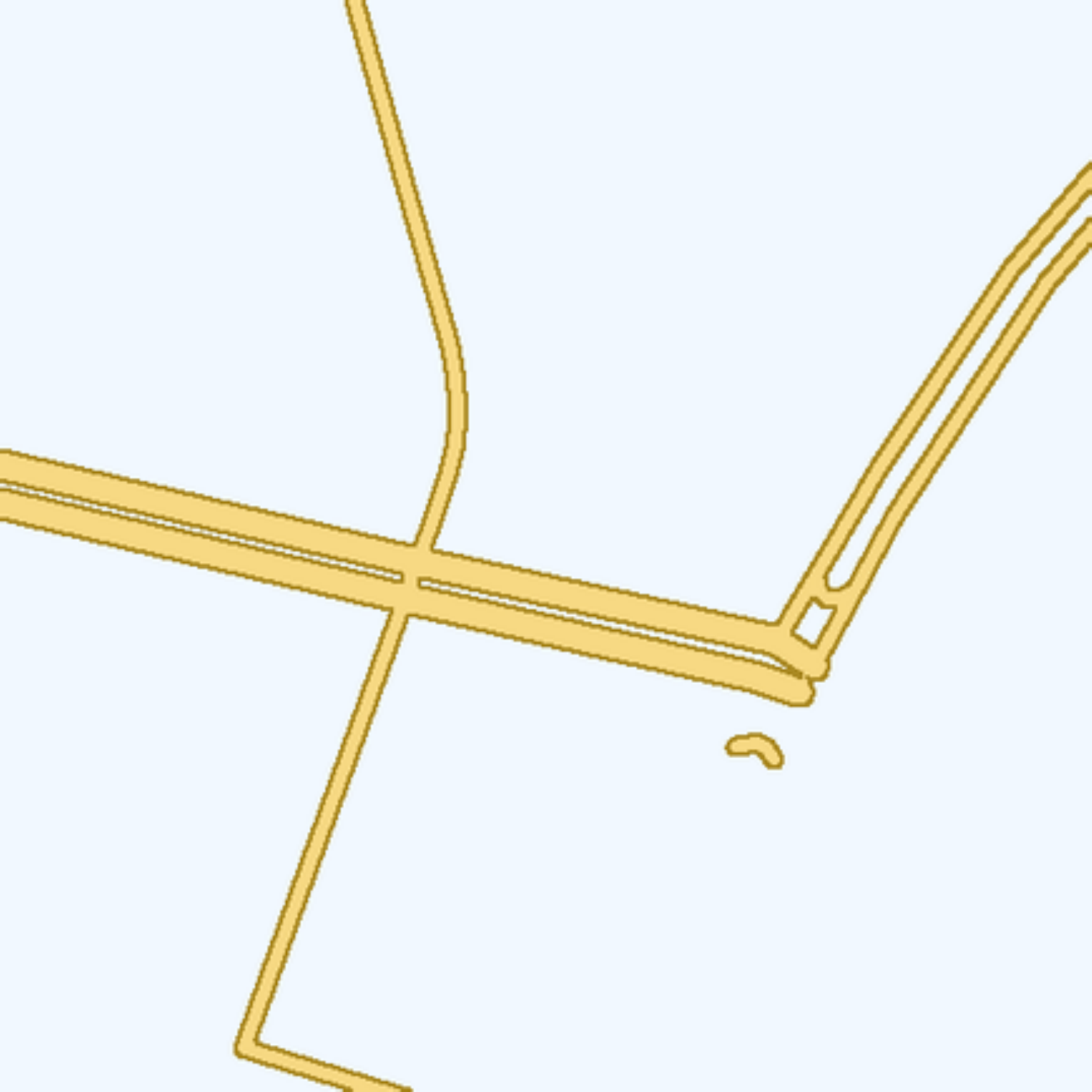}
			& 
\includegraphics[width=0.12\textwidth]{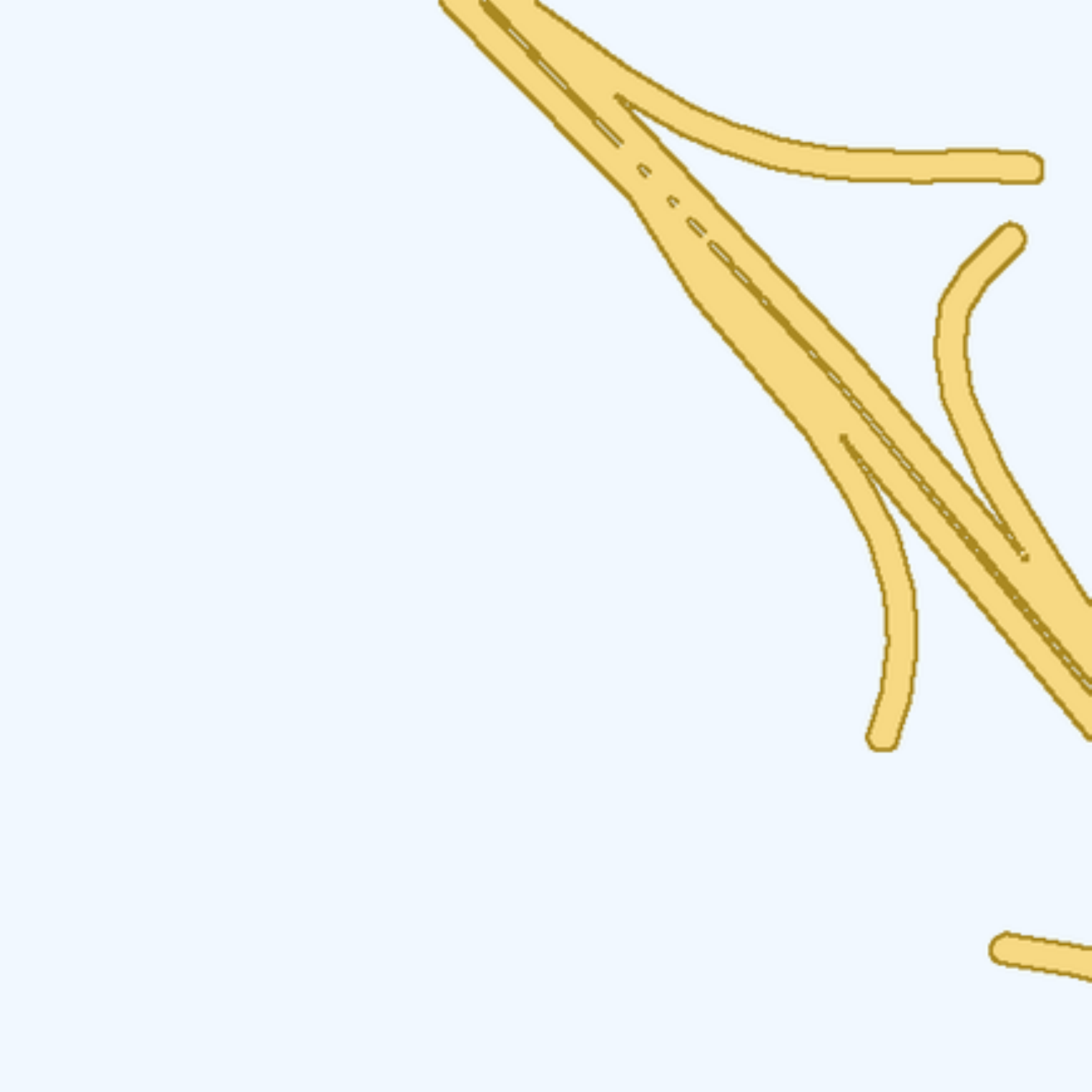}
			& 
\includegraphics[width=0.12\textwidth]{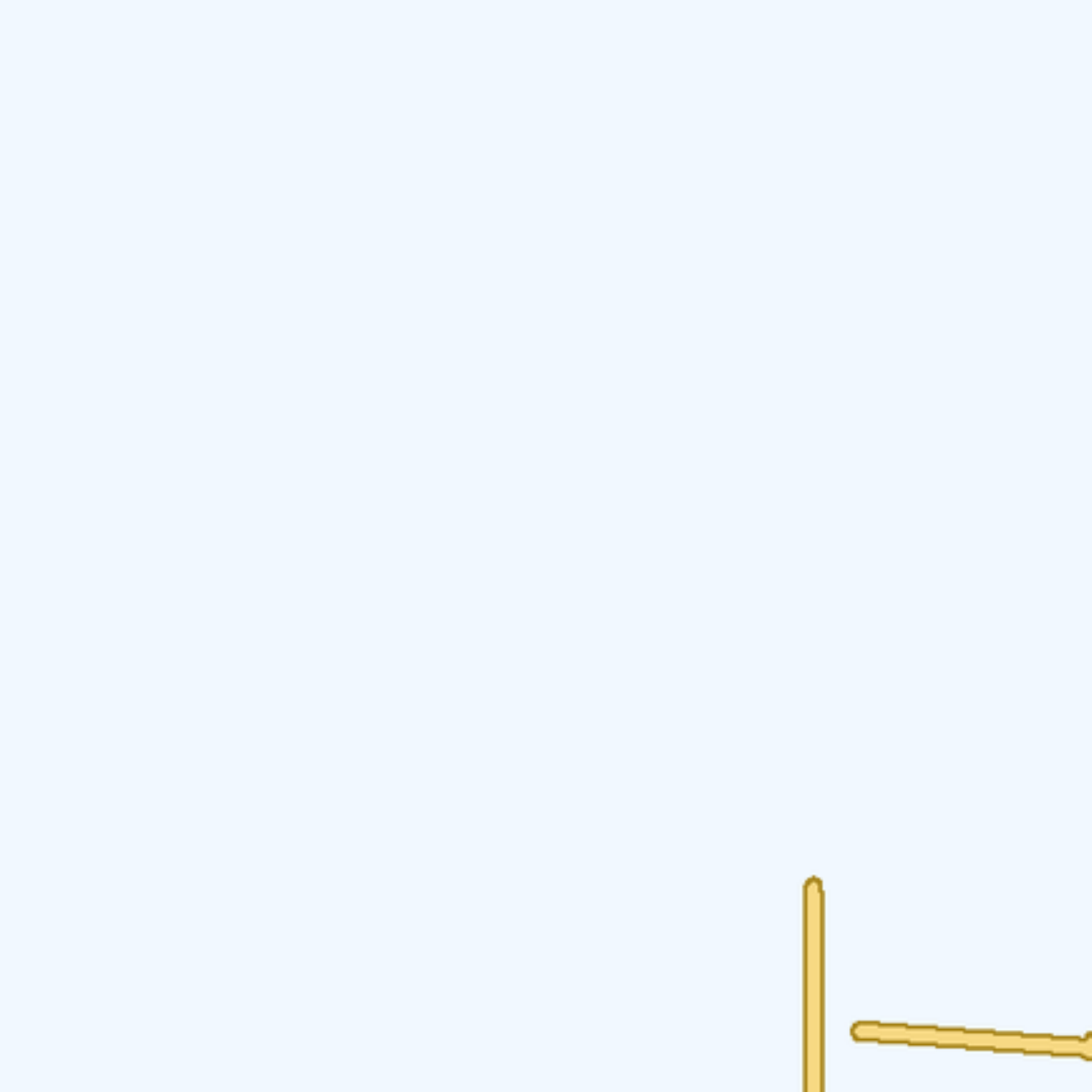}
			& 
\includegraphics[width=0.12\textwidth]{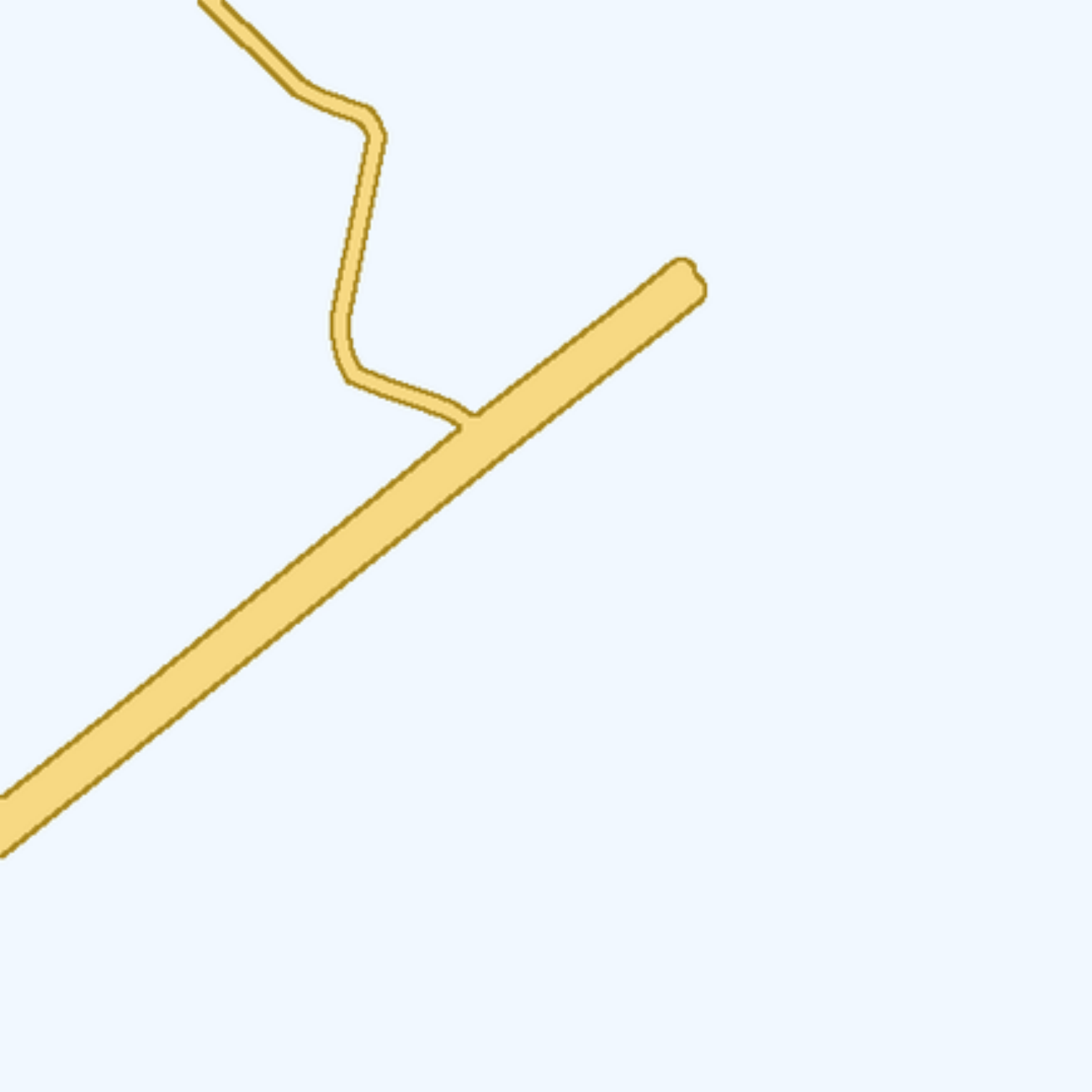}
		\\
Prediction
            &
\includegraphics[width=0.12\textwidth]{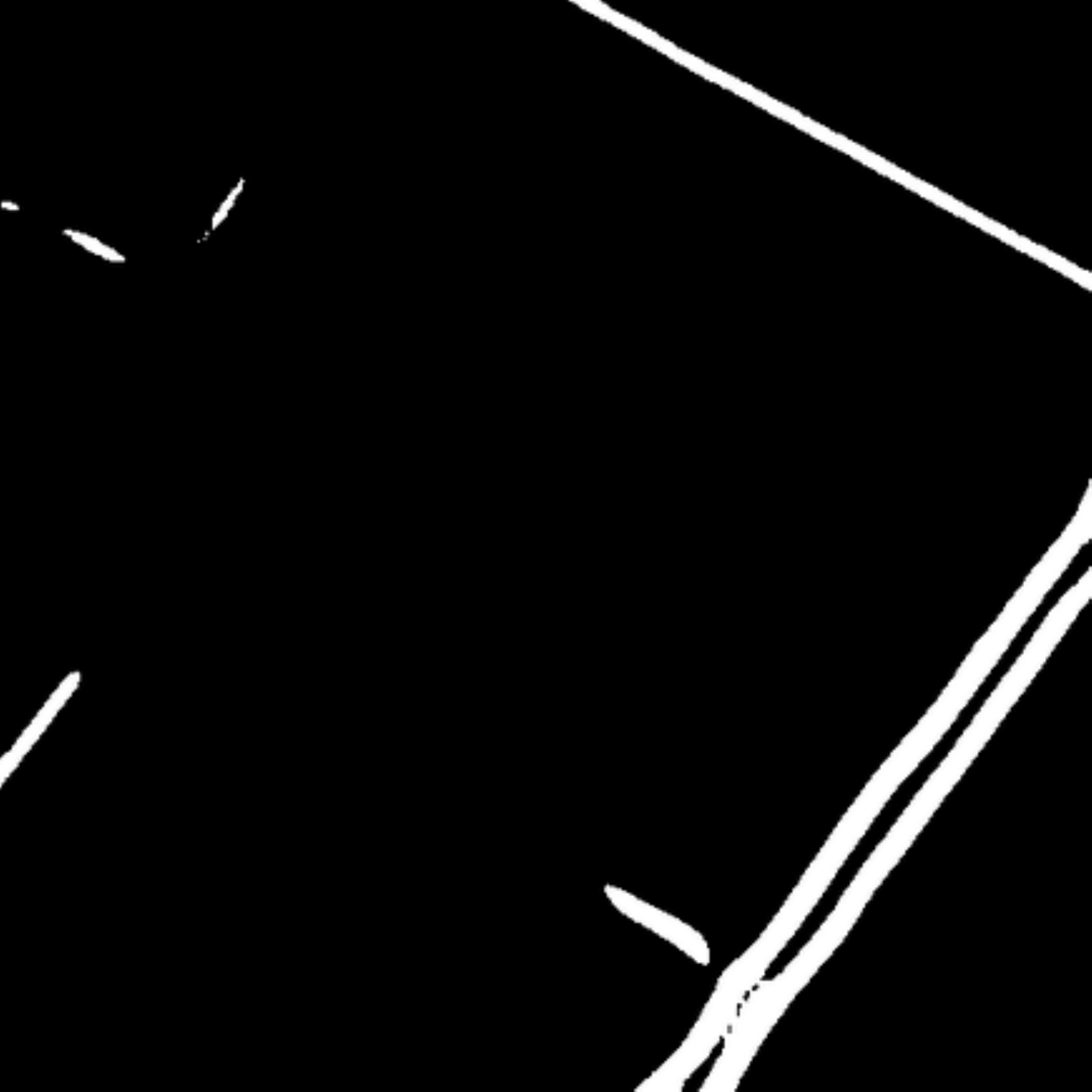}
			& 
\includegraphics[width=0.12\textwidth]{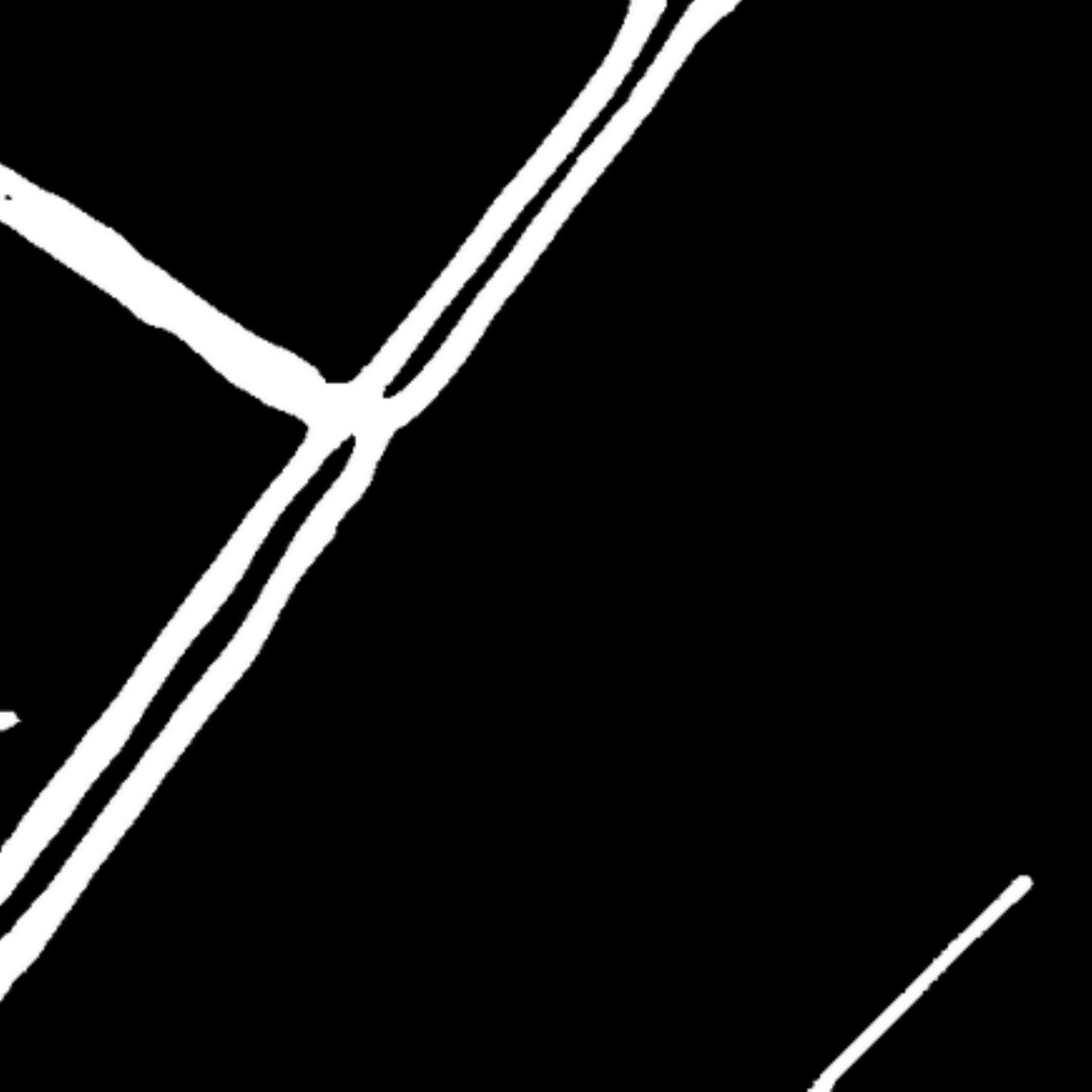}
			& 
\includegraphics[width=0.12\textwidth]{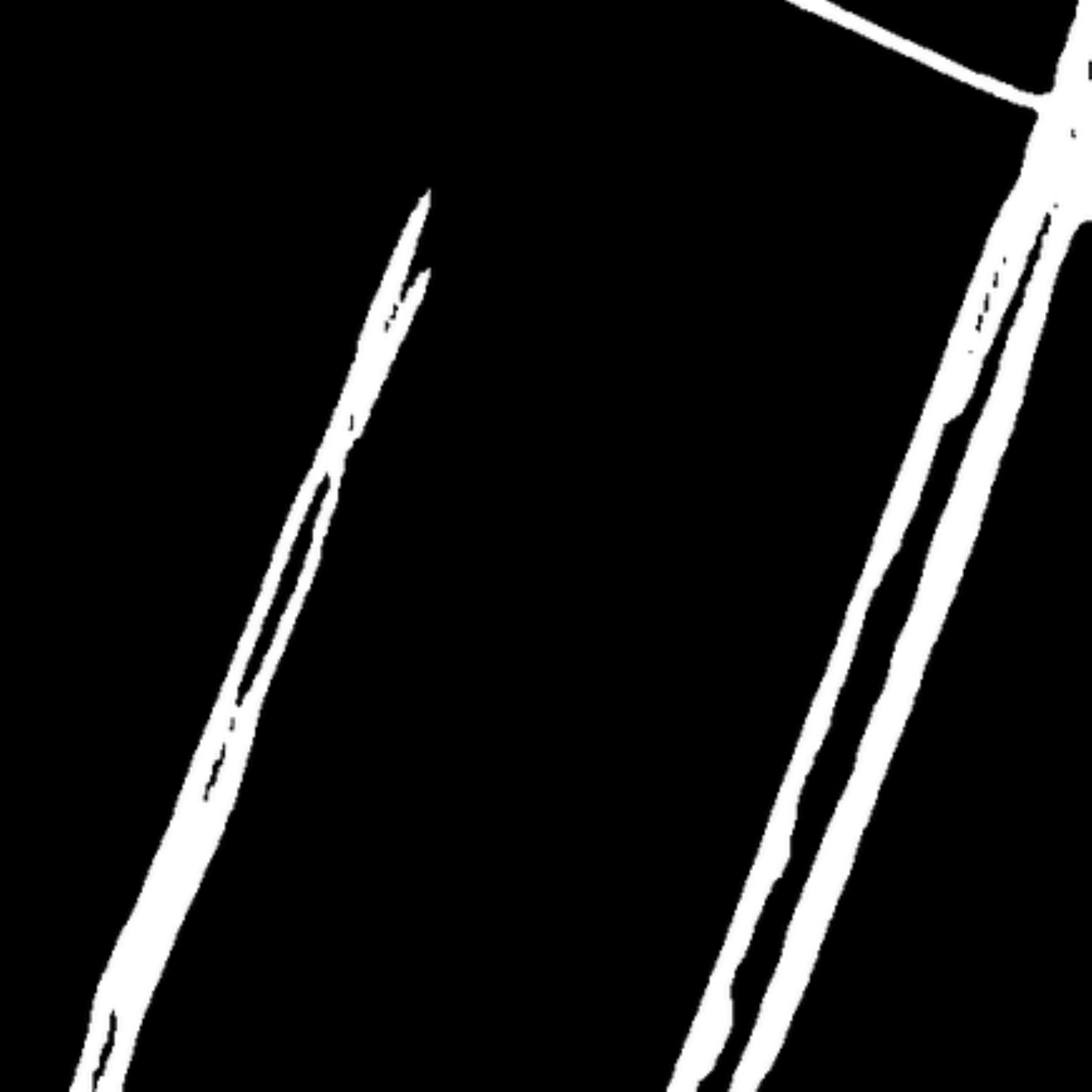}
			& 
\includegraphics[width=0.12\textwidth]{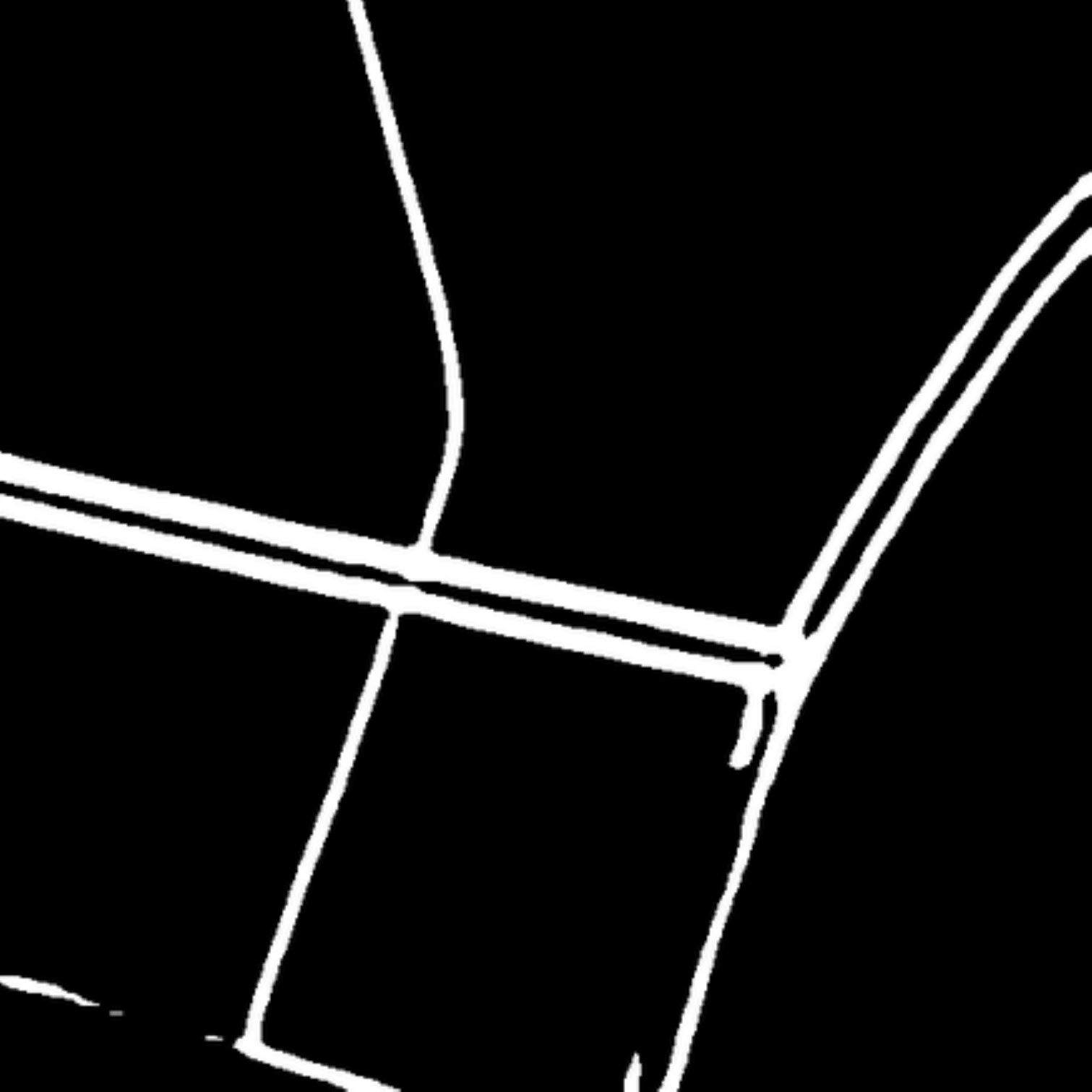}
			& 
\includegraphics[width=0.12\textwidth]{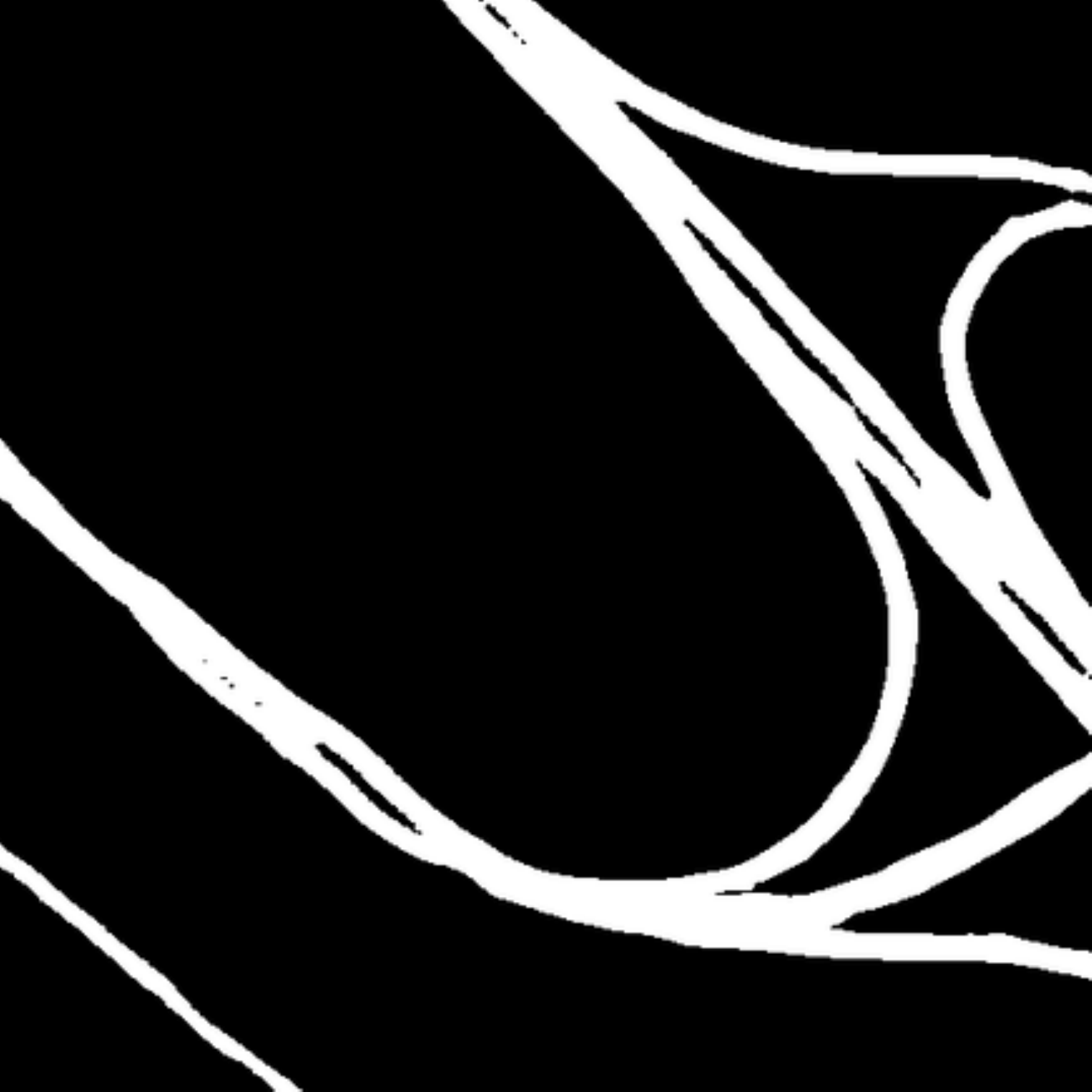}
			& 
\includegraphics[width=0.12\textwidth]{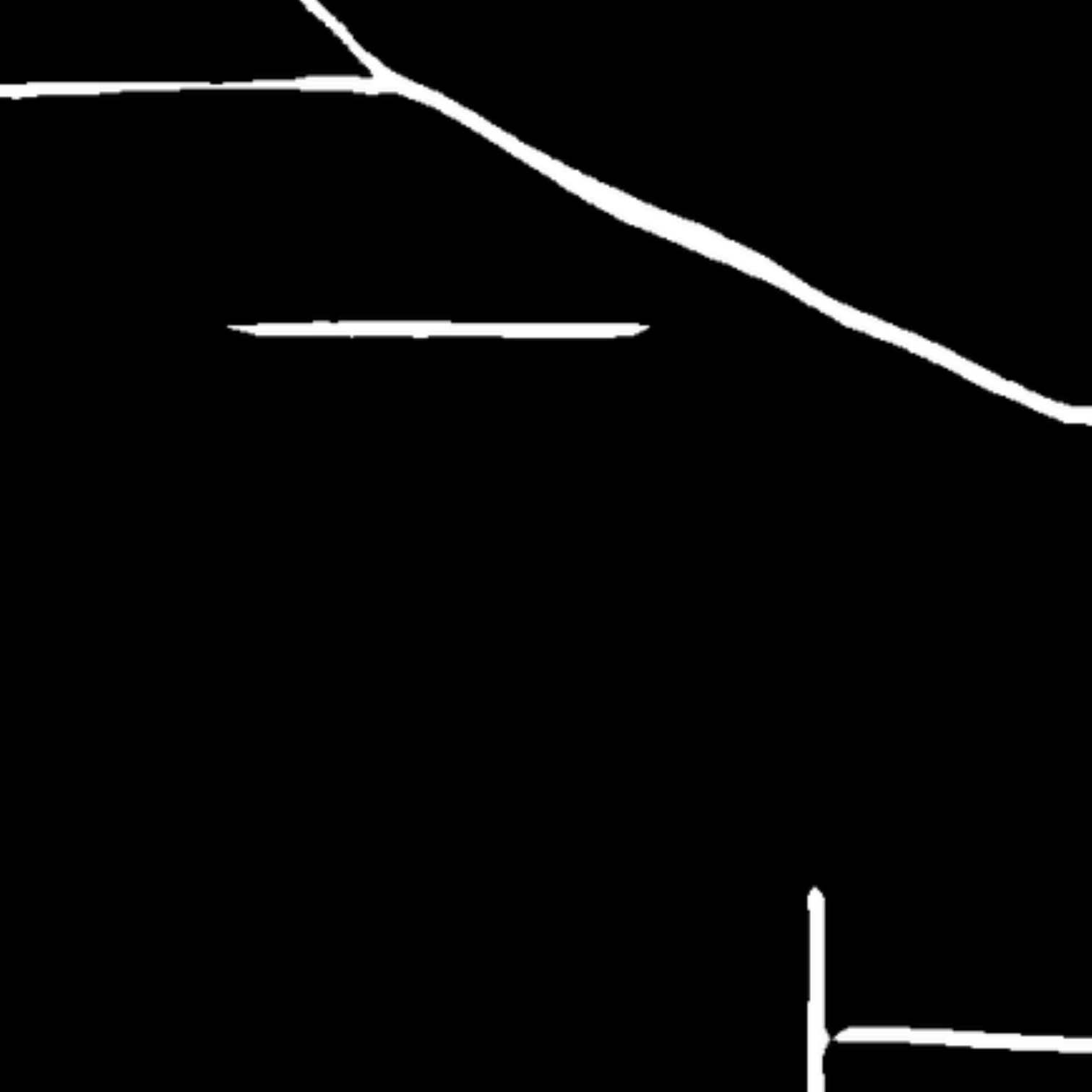}
			& 
\includegraphics[width=0.12\textwidth]{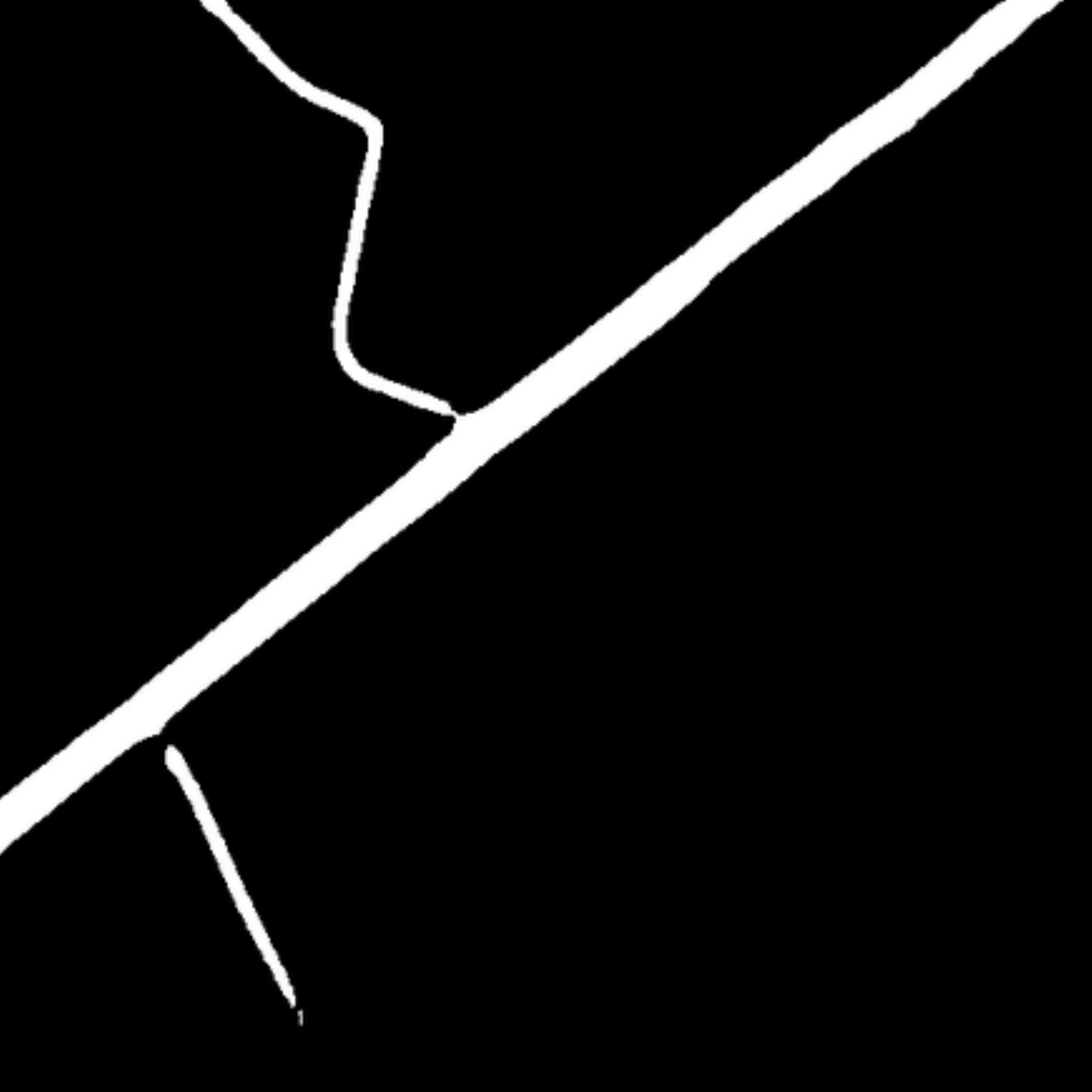}
		\\
\makecell*[c]{Overlap \\ (changed in red)}
            &
\includegraphics[width=0.12\textwidth]{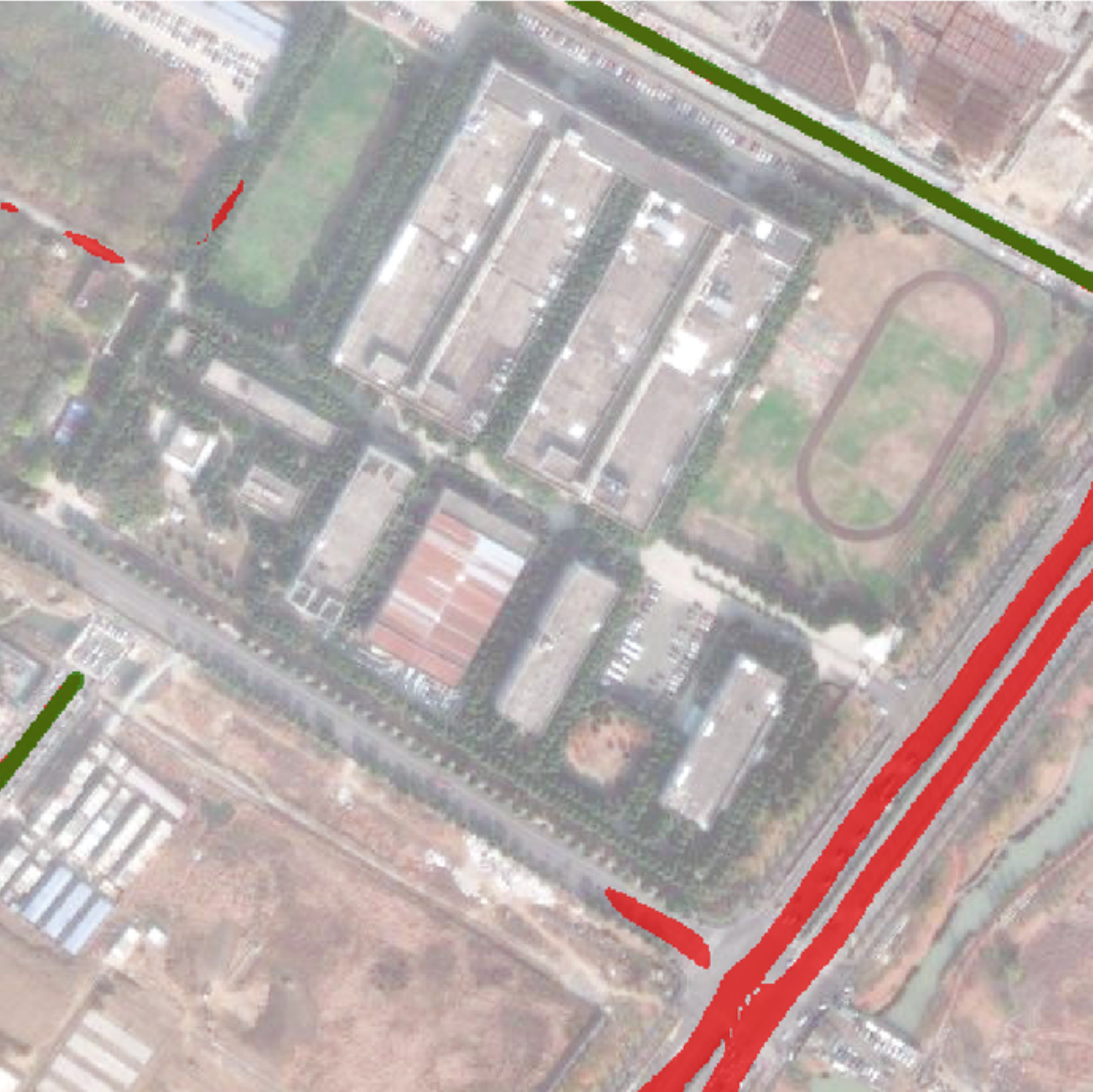}
			& 
\includegraphics[width=0.12\textwidth]{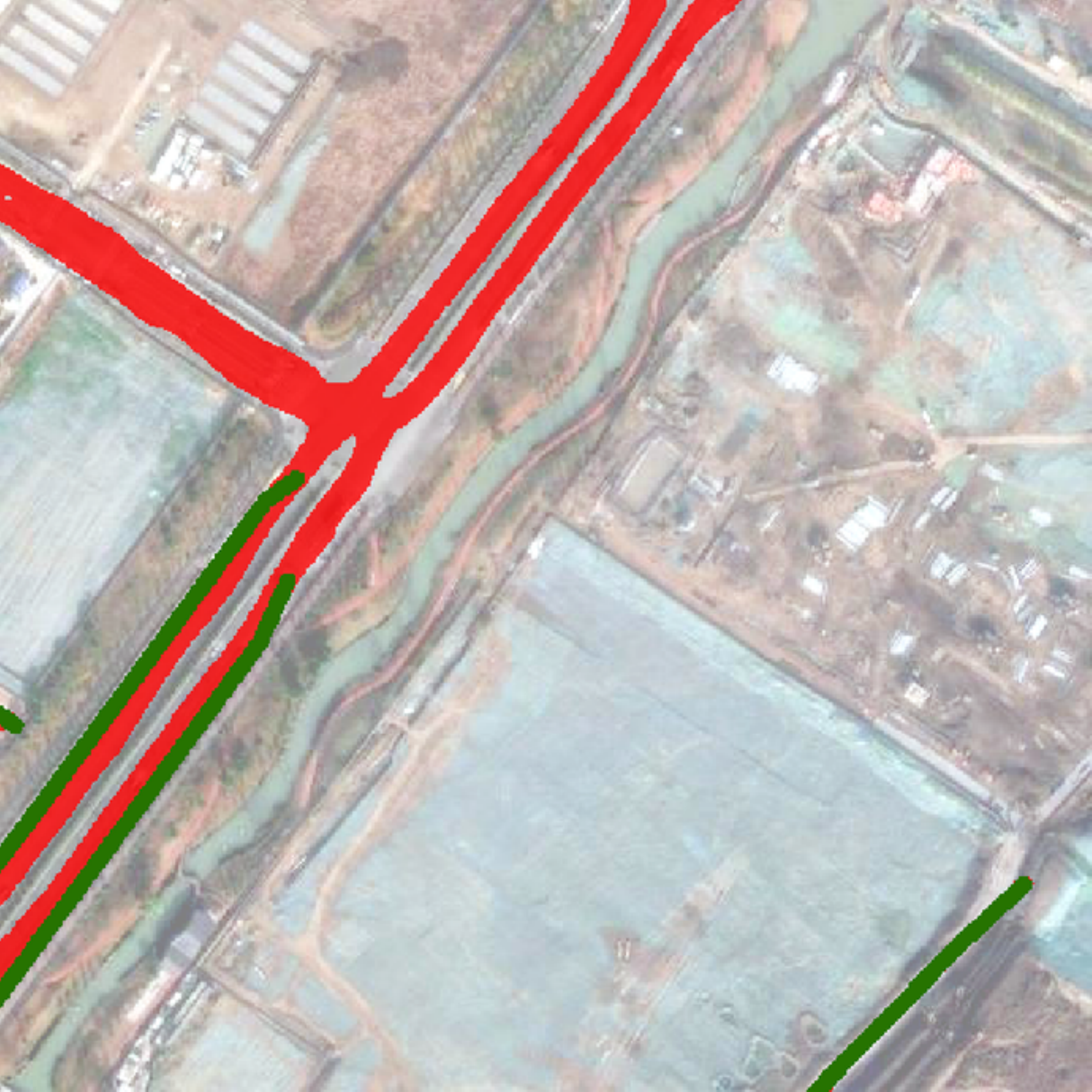}
			& 
\includegraphics[width=0.12\textwidth]{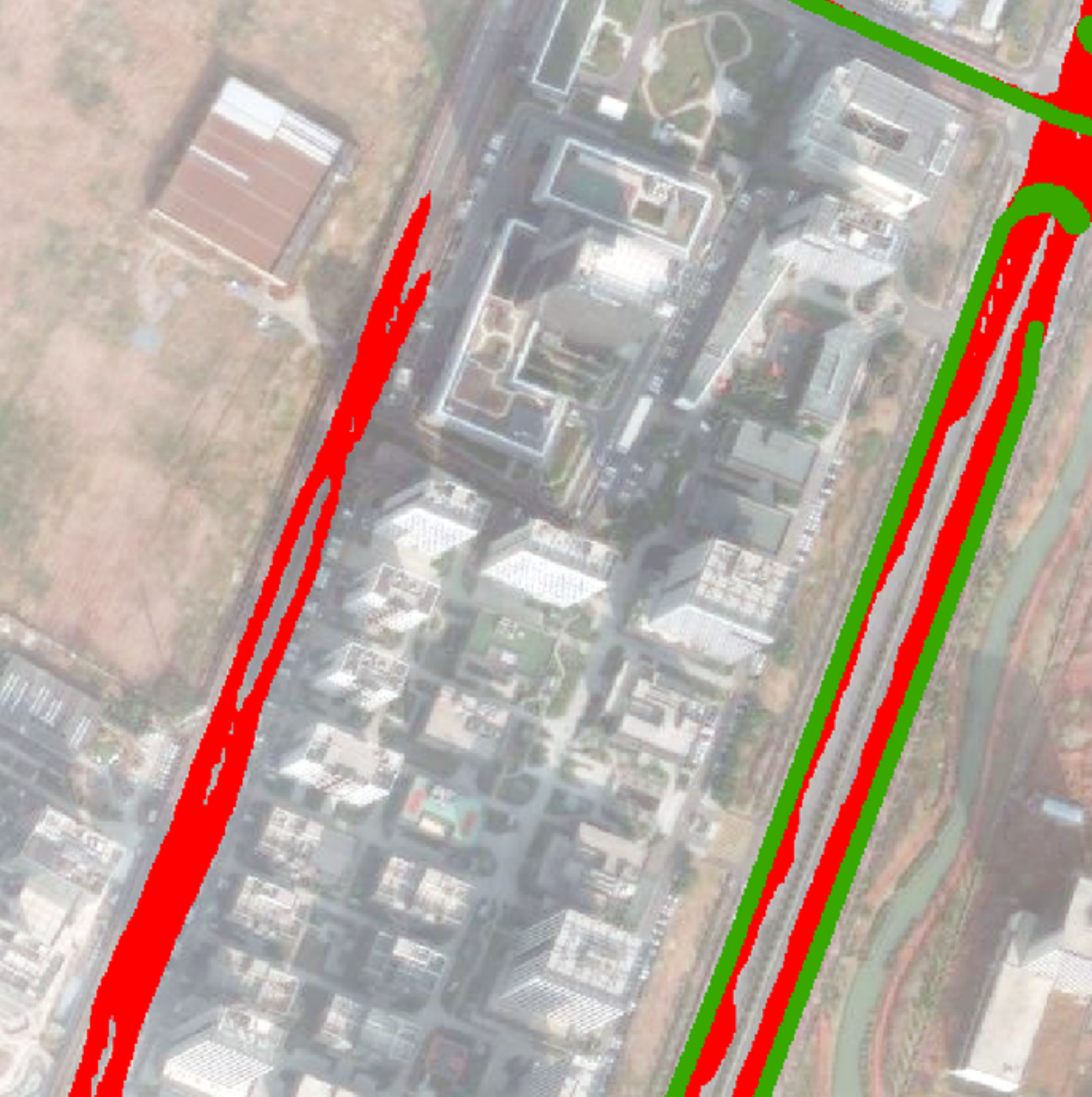}
			& 
\includegraphics[width=0.12\textwidth]{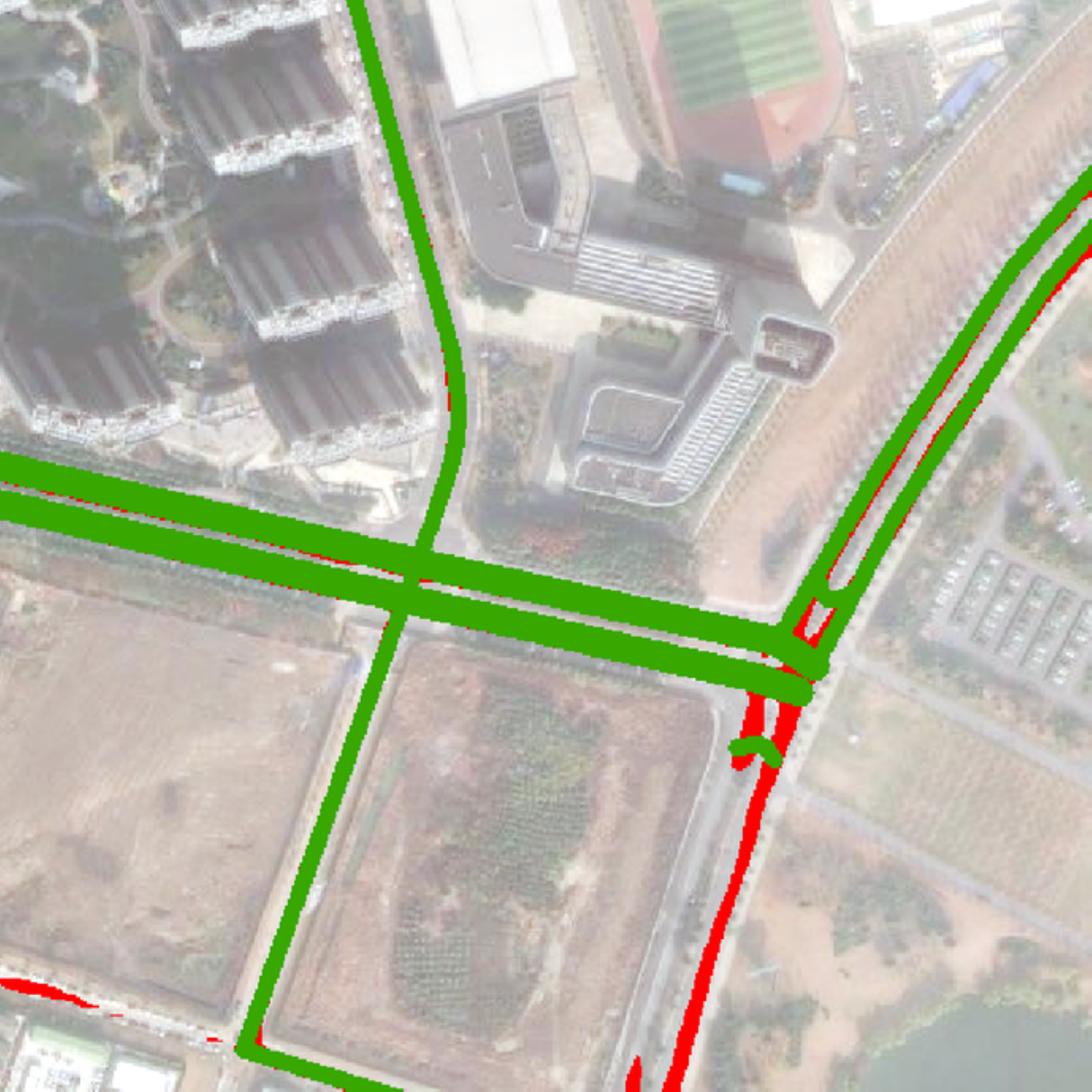}
			& 
\includegraphics[width=0.12\textwidth]{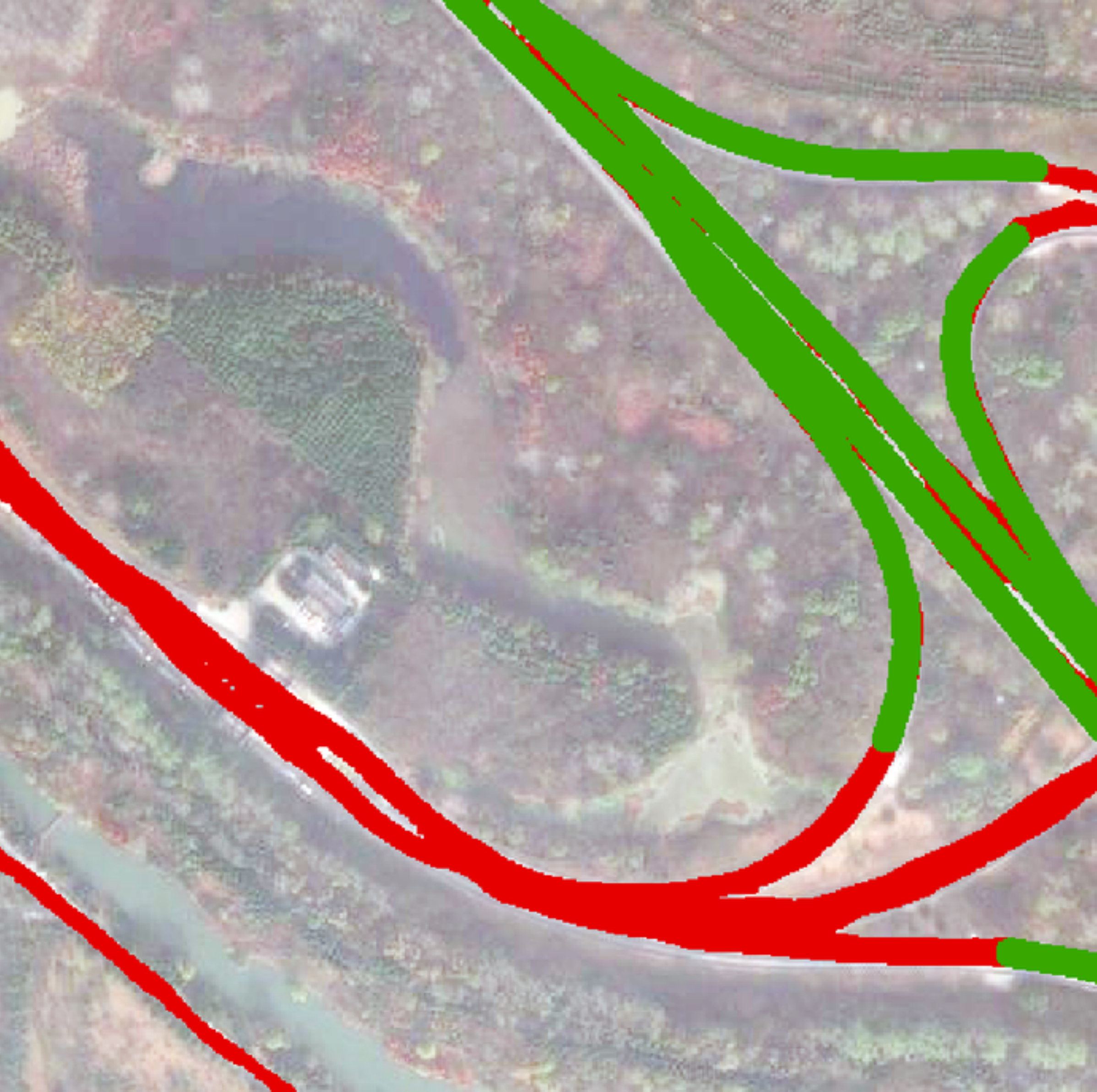}
			& 
\includegraphics[width=0.12\textwidth]{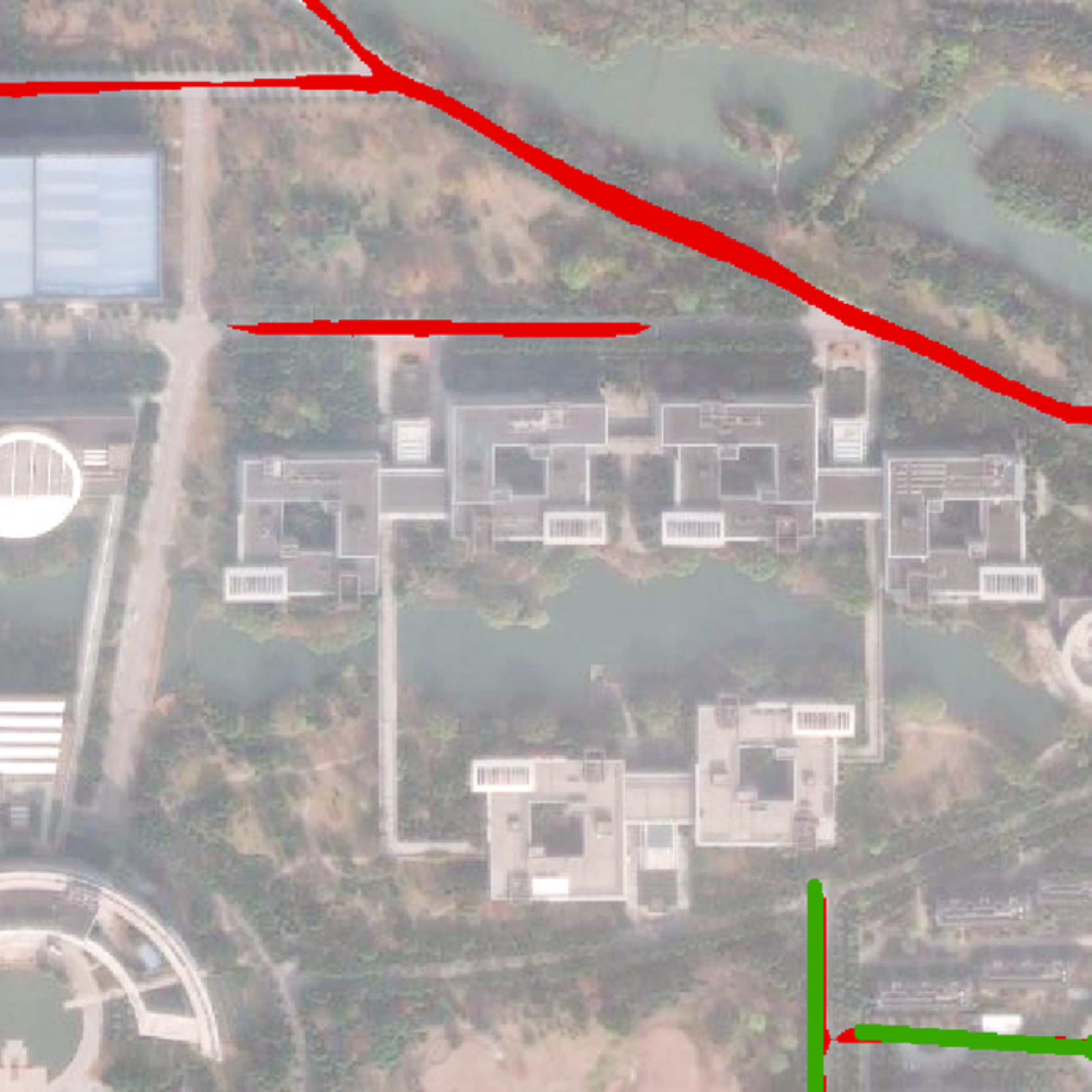}
			& 
\includegraphics[width=0.12\textwidth]{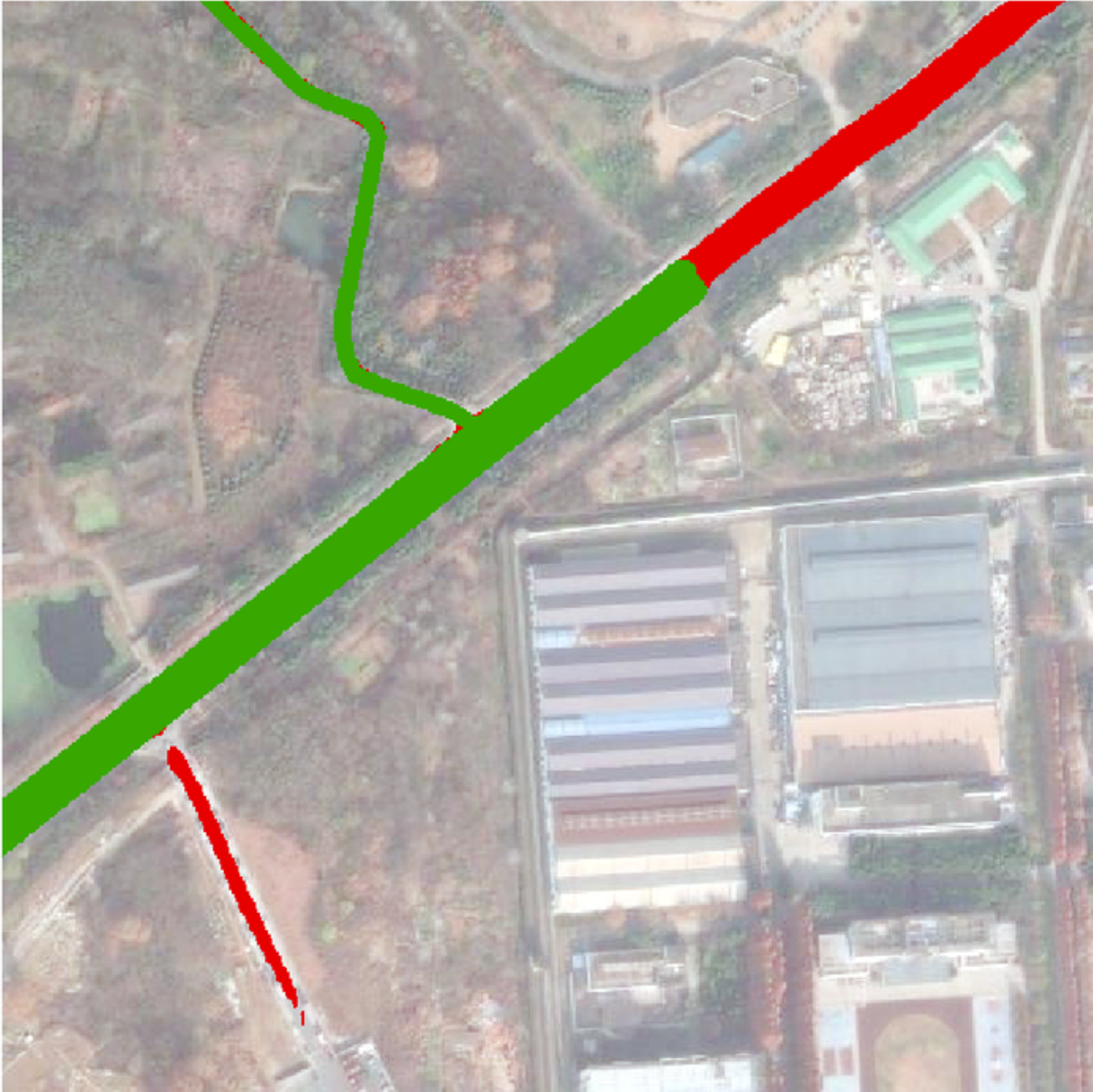}
		\\
 &  \multicolumn{7}{c}{(a) Images from Nanjing, Jiangsu Province}
        \\
Image
            &
\includegraphics[width=0.12\textwidth]{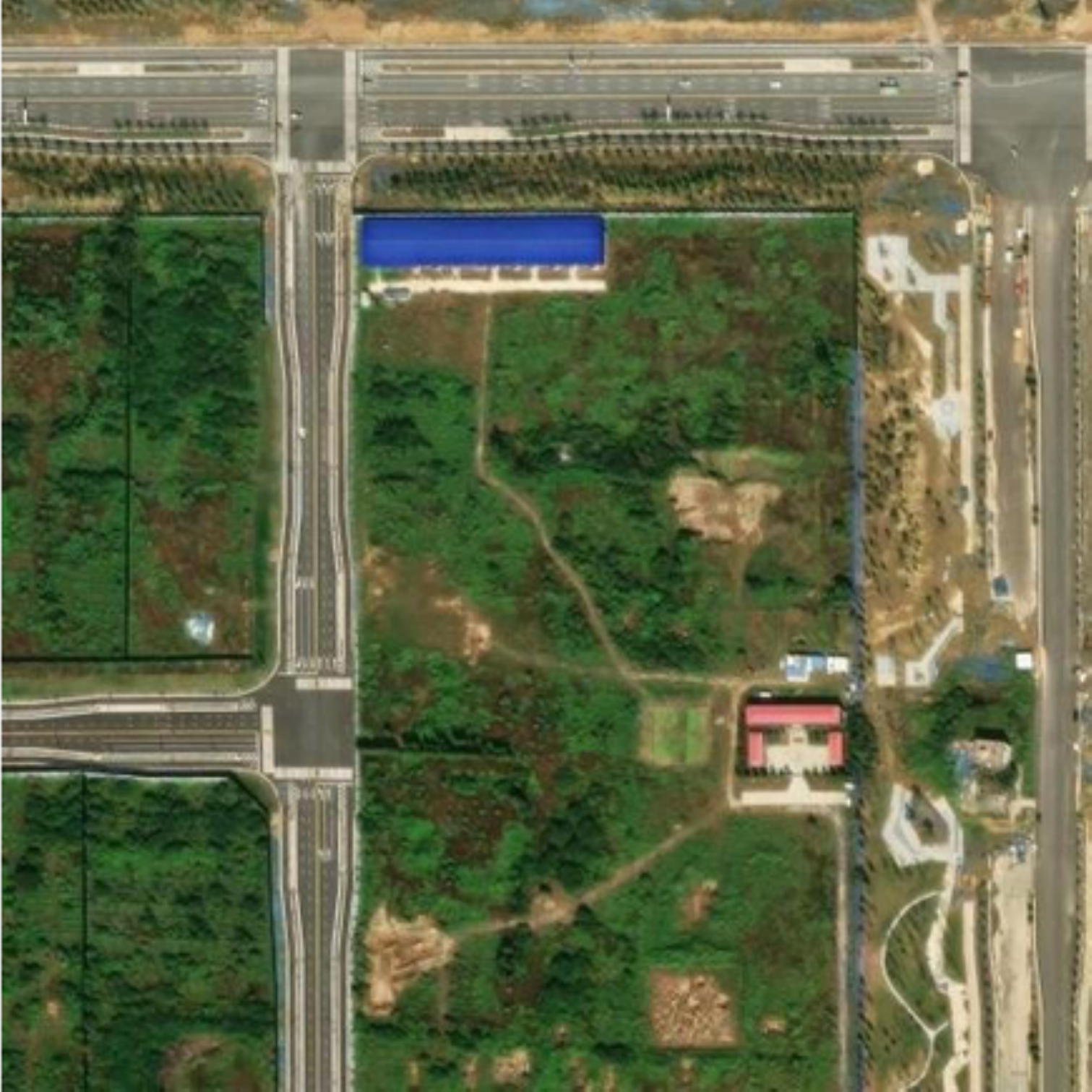}
			& 
\includegraphics[width=0.12\textwidth]{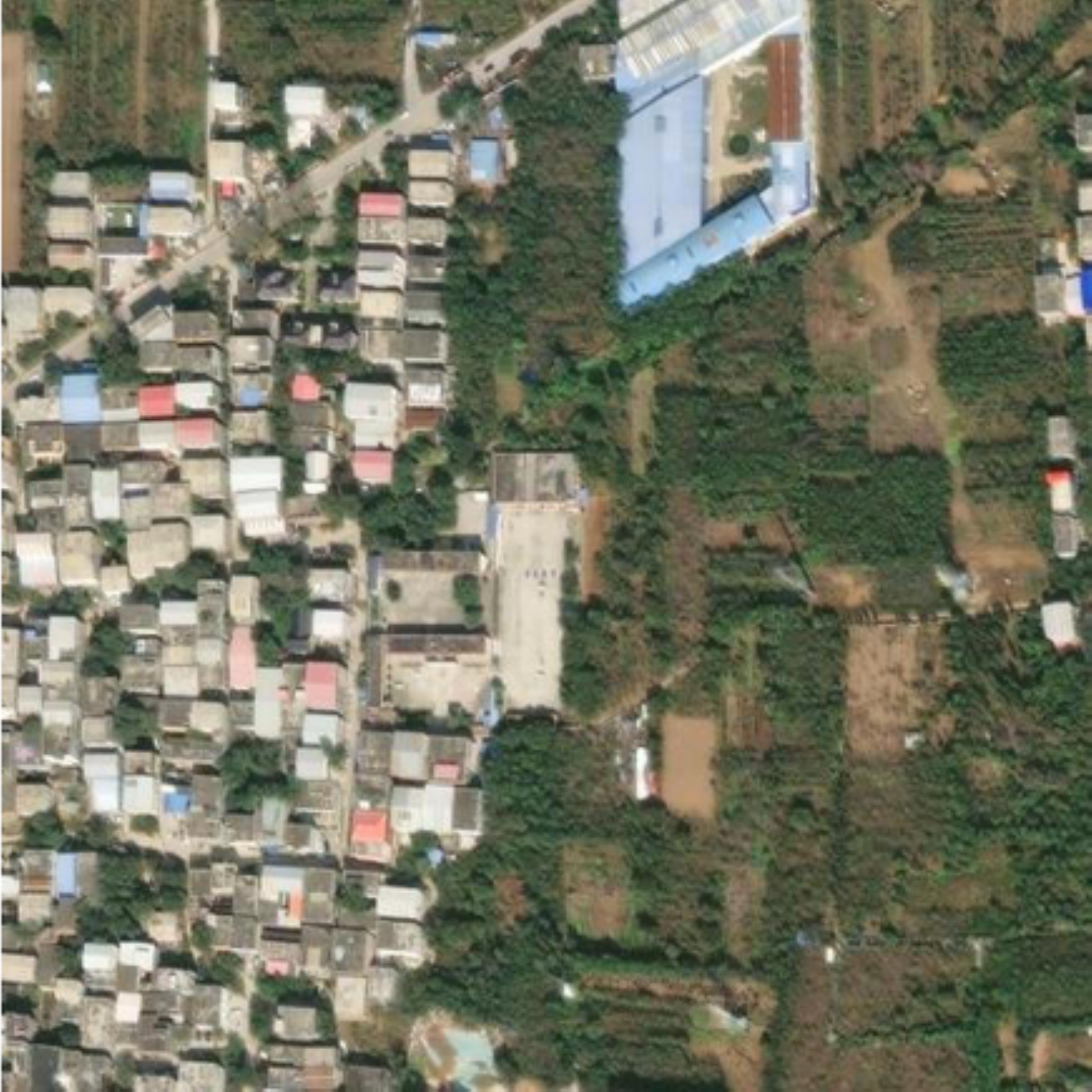}
			& 
\includegraphics[width=0.12\textwidth]{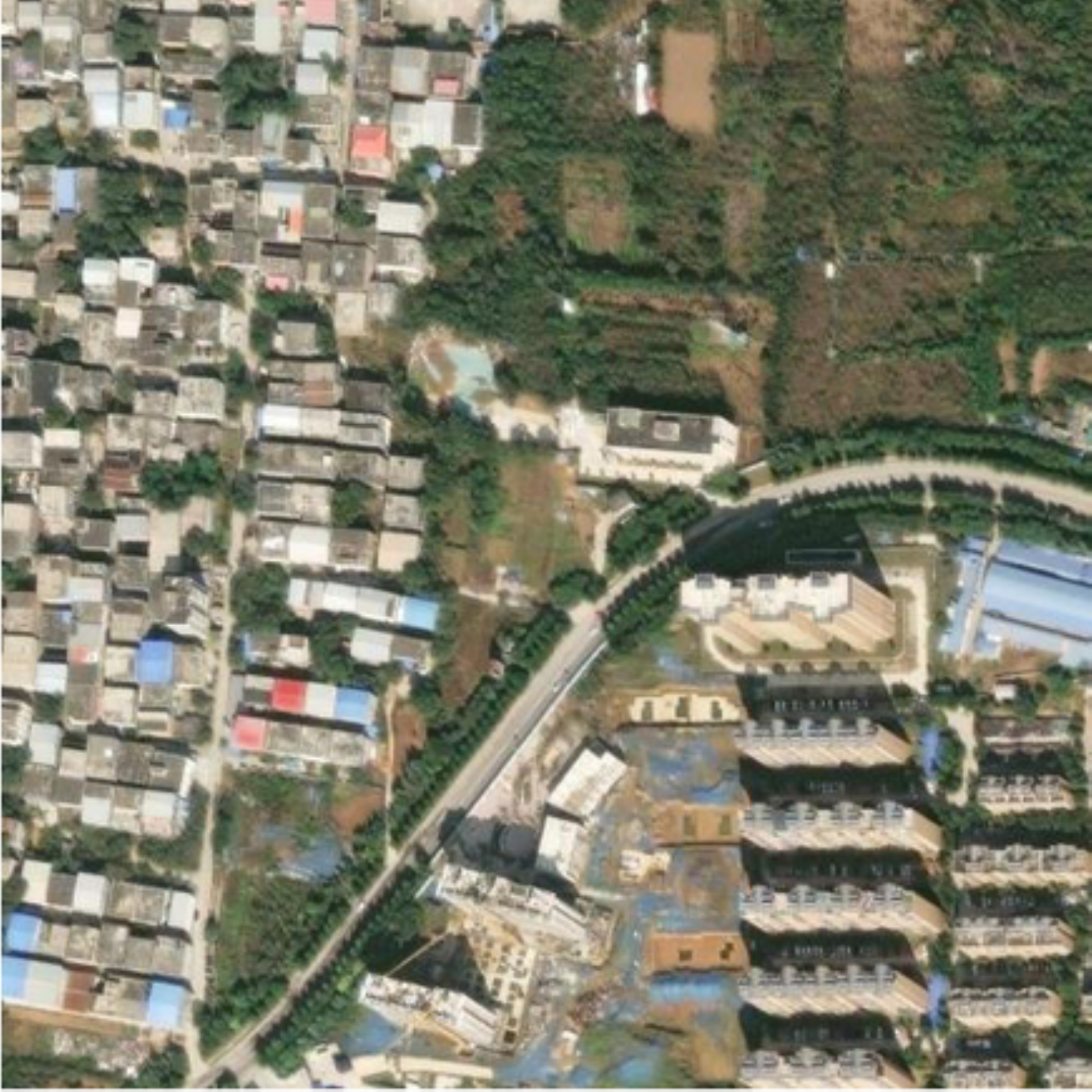}
			& 
\includegraphics[width=0.12\textwidth]{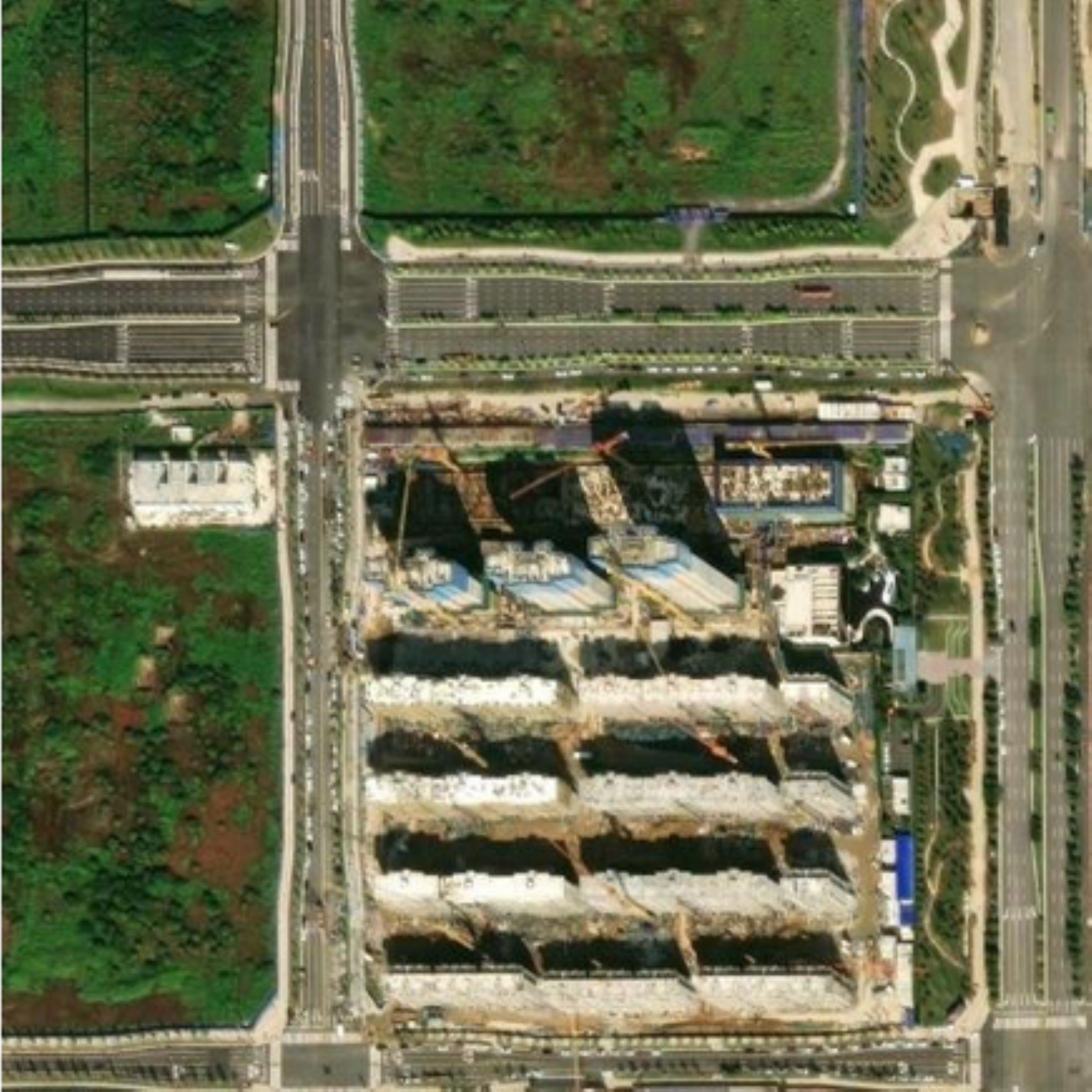}
			& 
\includegraphics[width=0.12\textwidth]{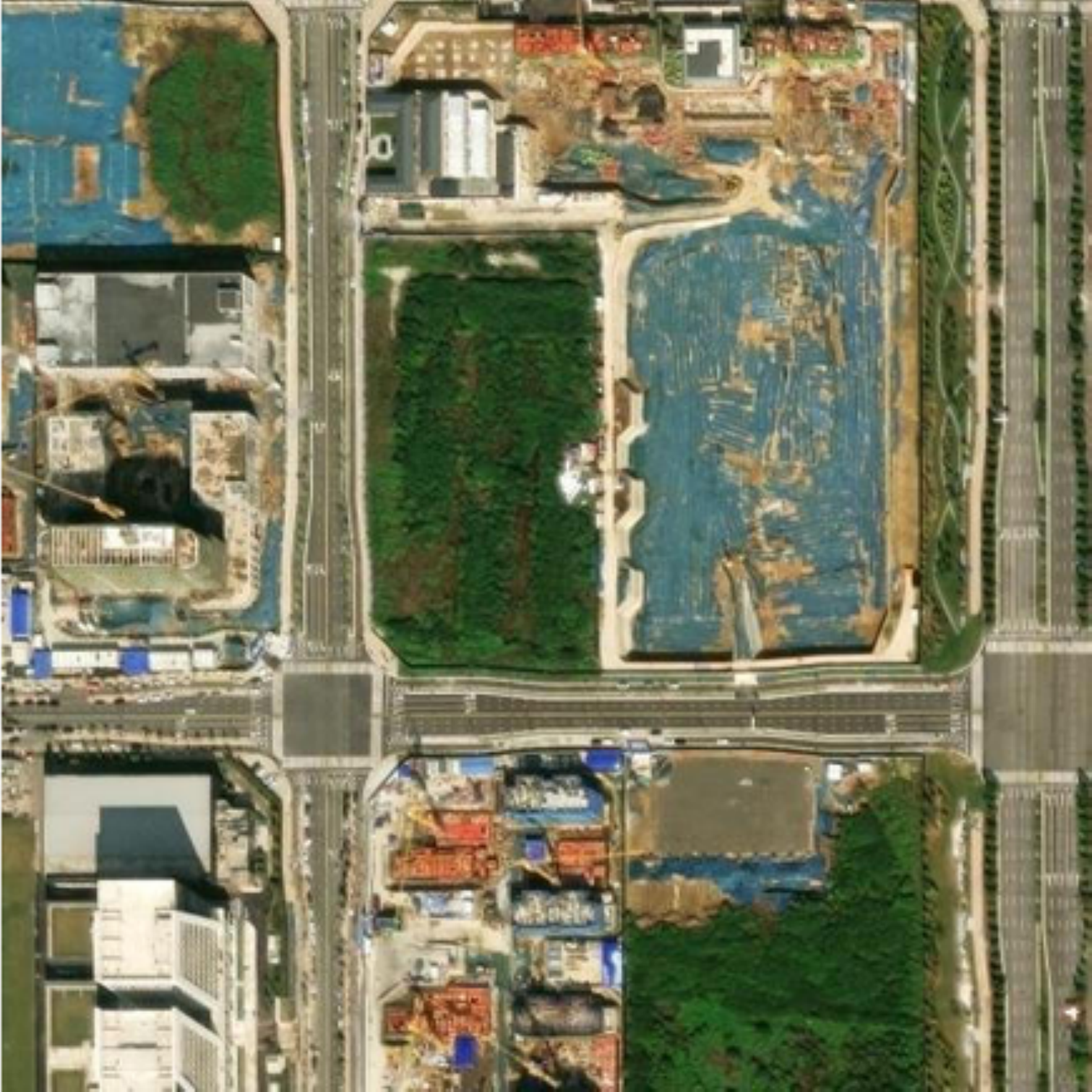}
			& 
\includegraphics[width=0.12\textwidth]{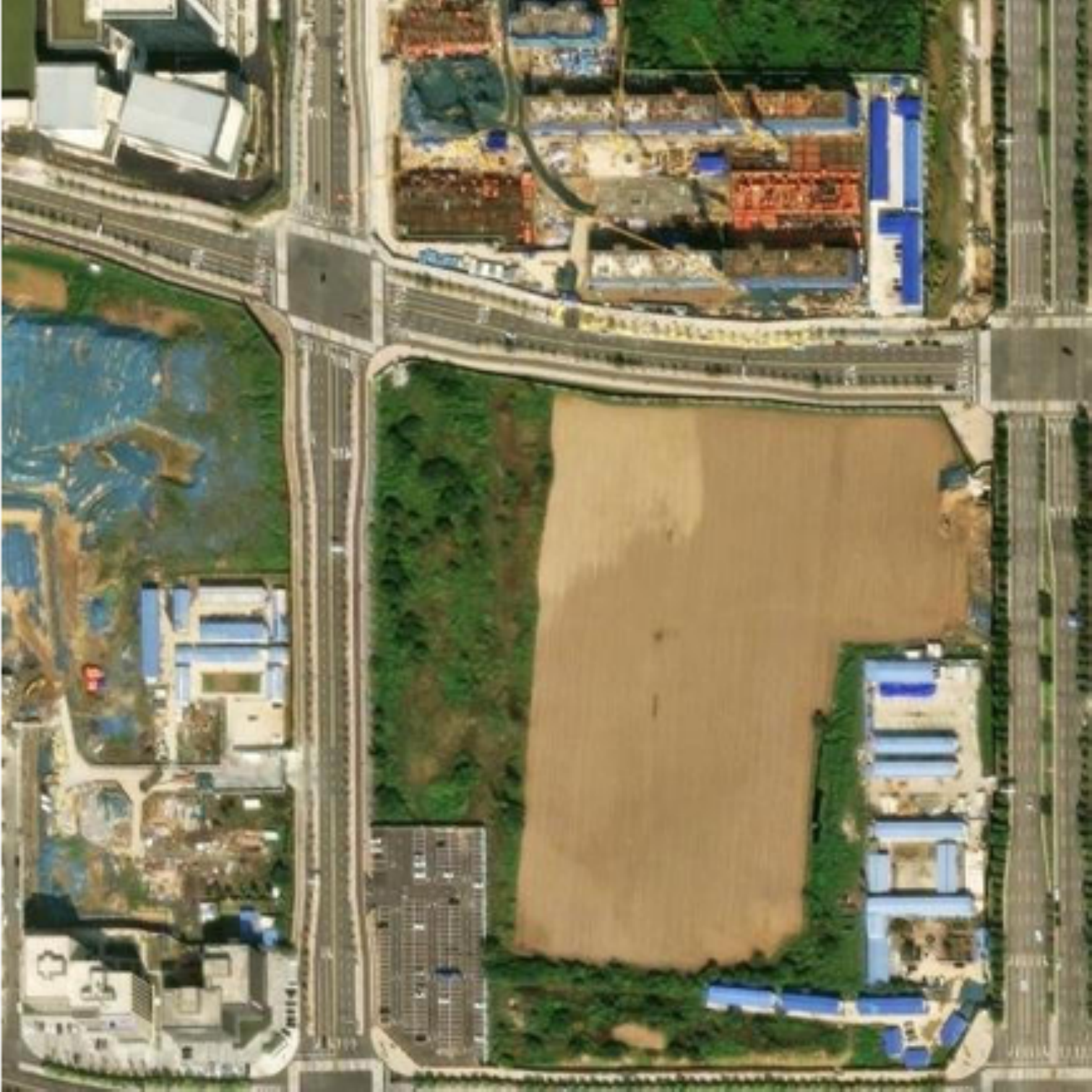}
			& 
\includegraphics[width=0.12\textwidth]{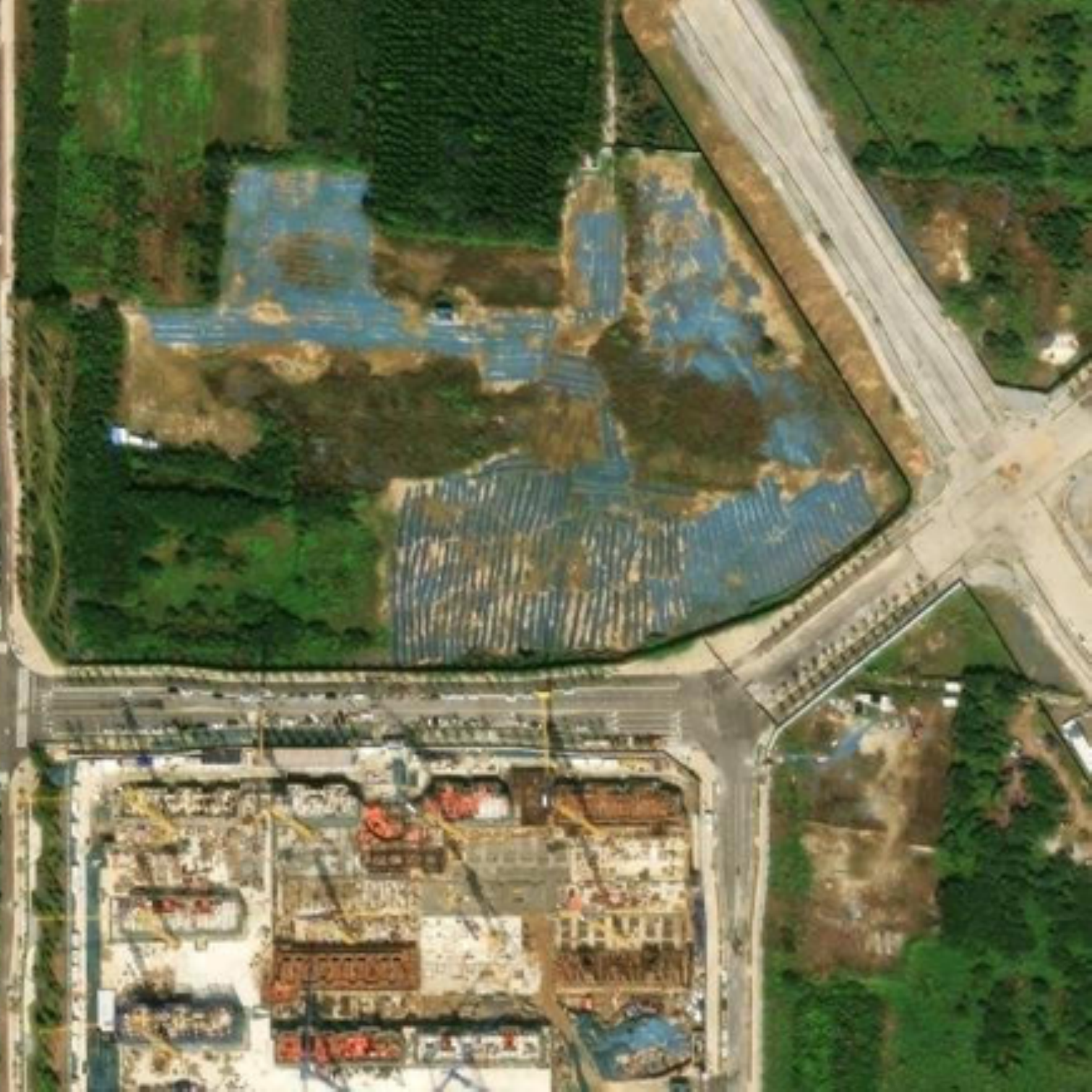}
		\\
\makecell*[c]{Historical \\ Map}
            &
\includegraphics[width=0.12\textwidth]{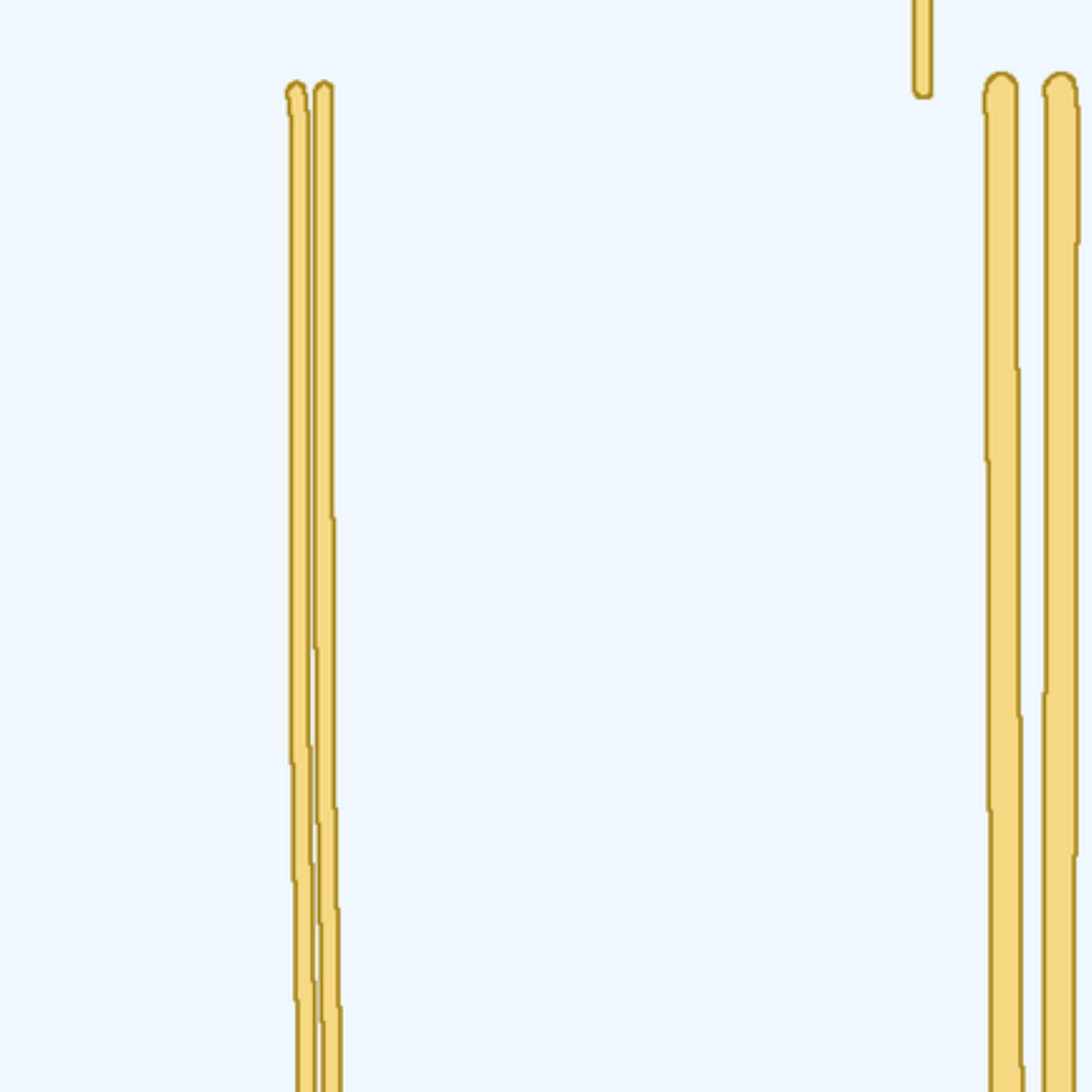}
			& 
\includegraphics[width=0.12\textwidth]{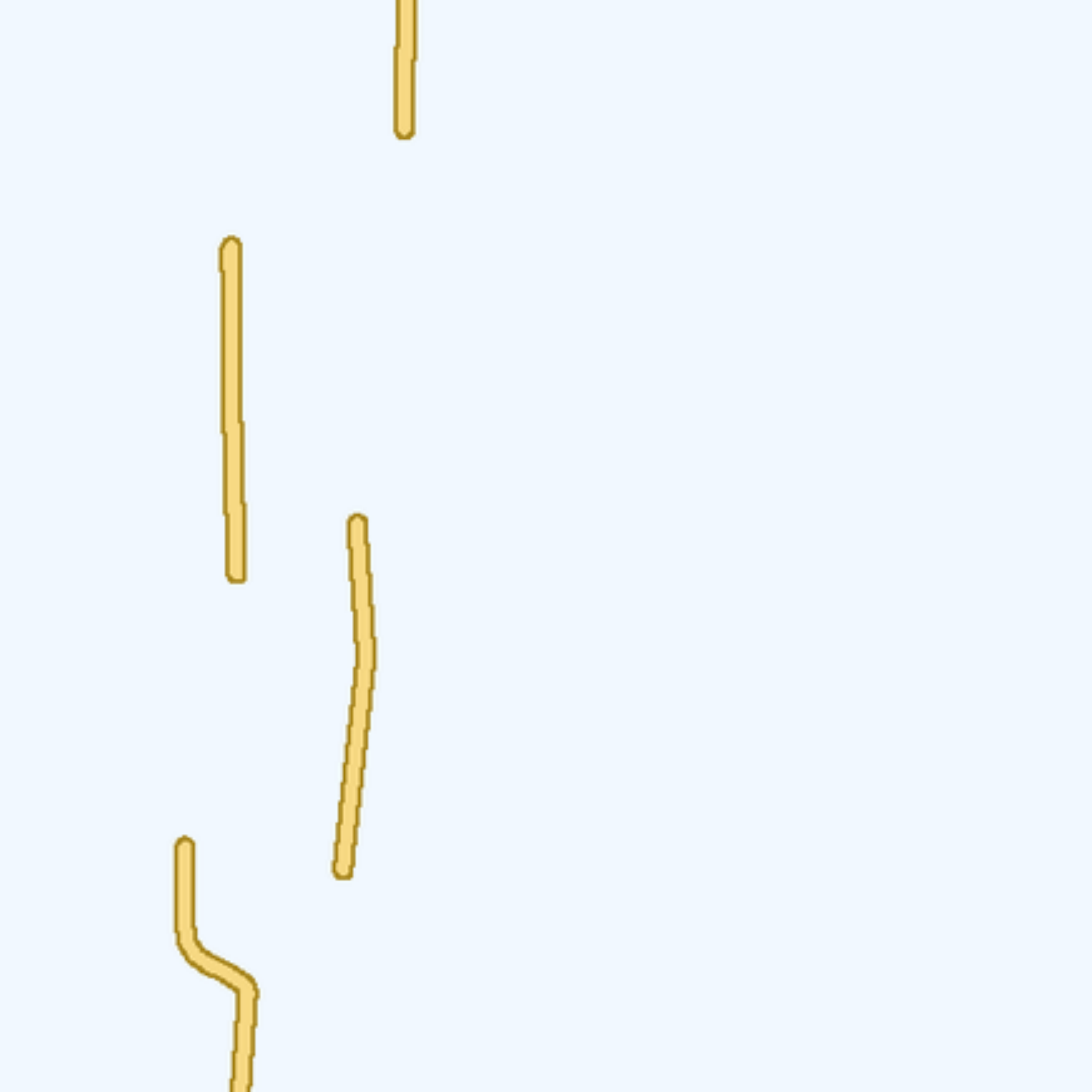}
			& 
\includegraphics[width=0.12\textwidth]{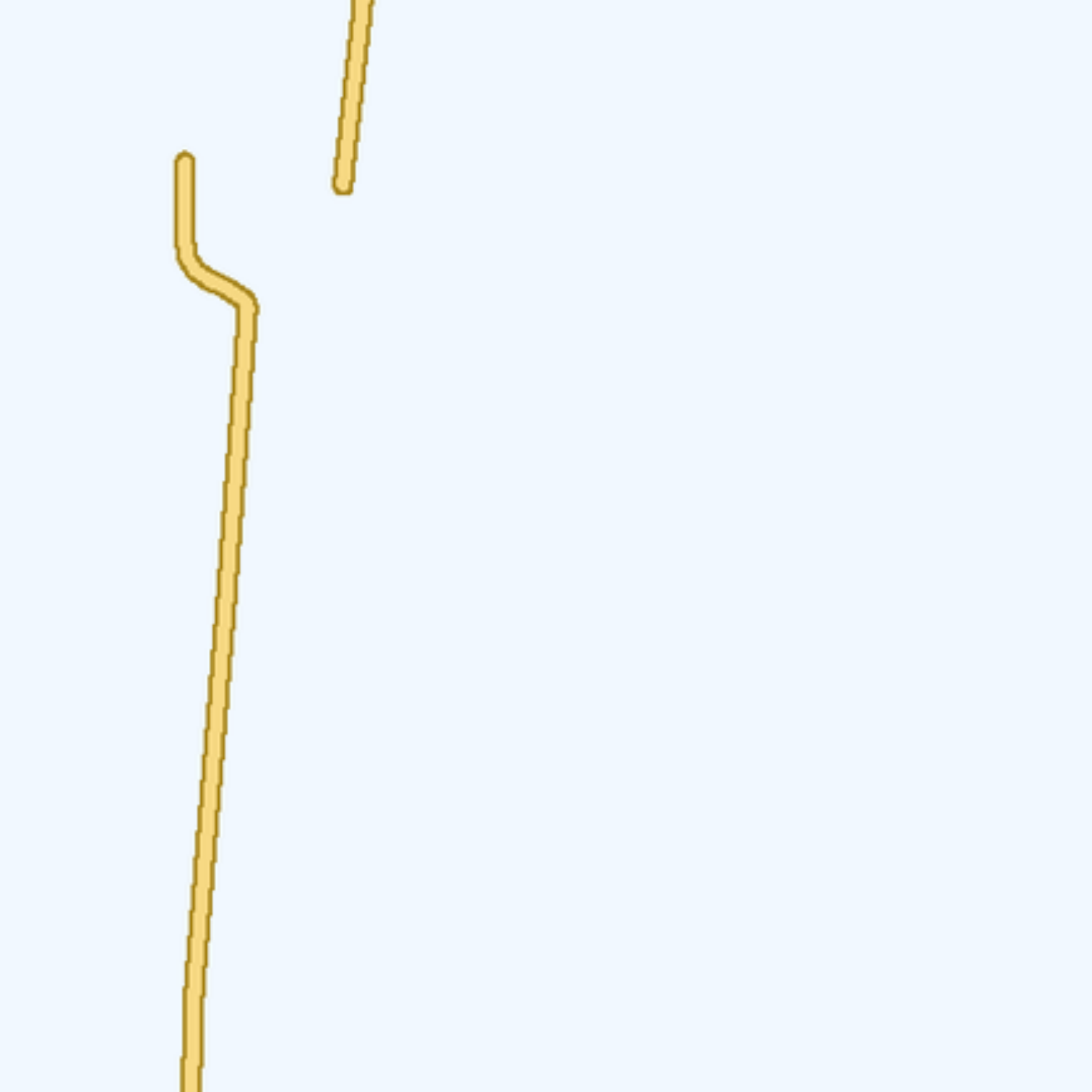}
			& 
\includegraphics[width=0.12\textwidth]{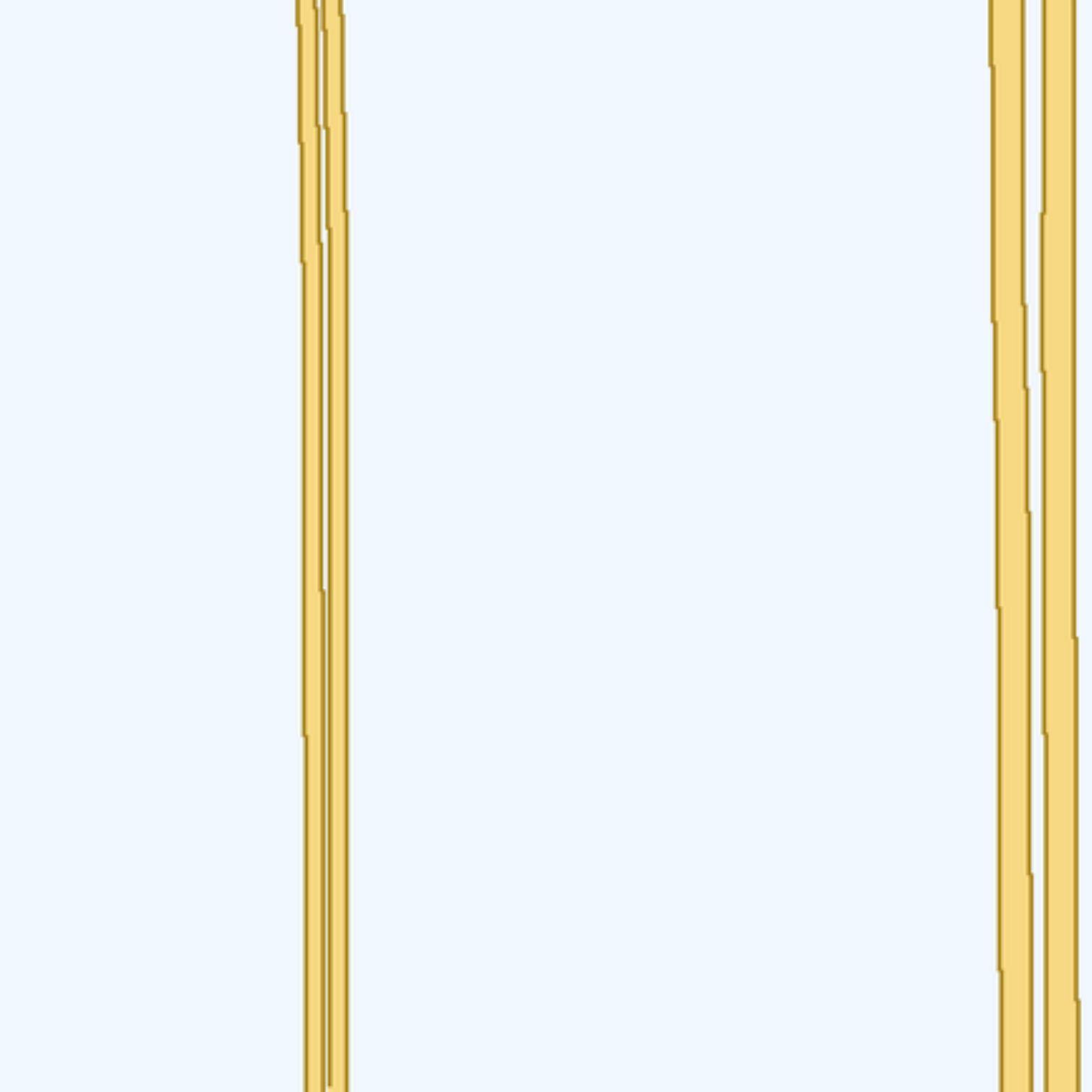}
			& 
\includegraphics[width=0.12\textwidth]{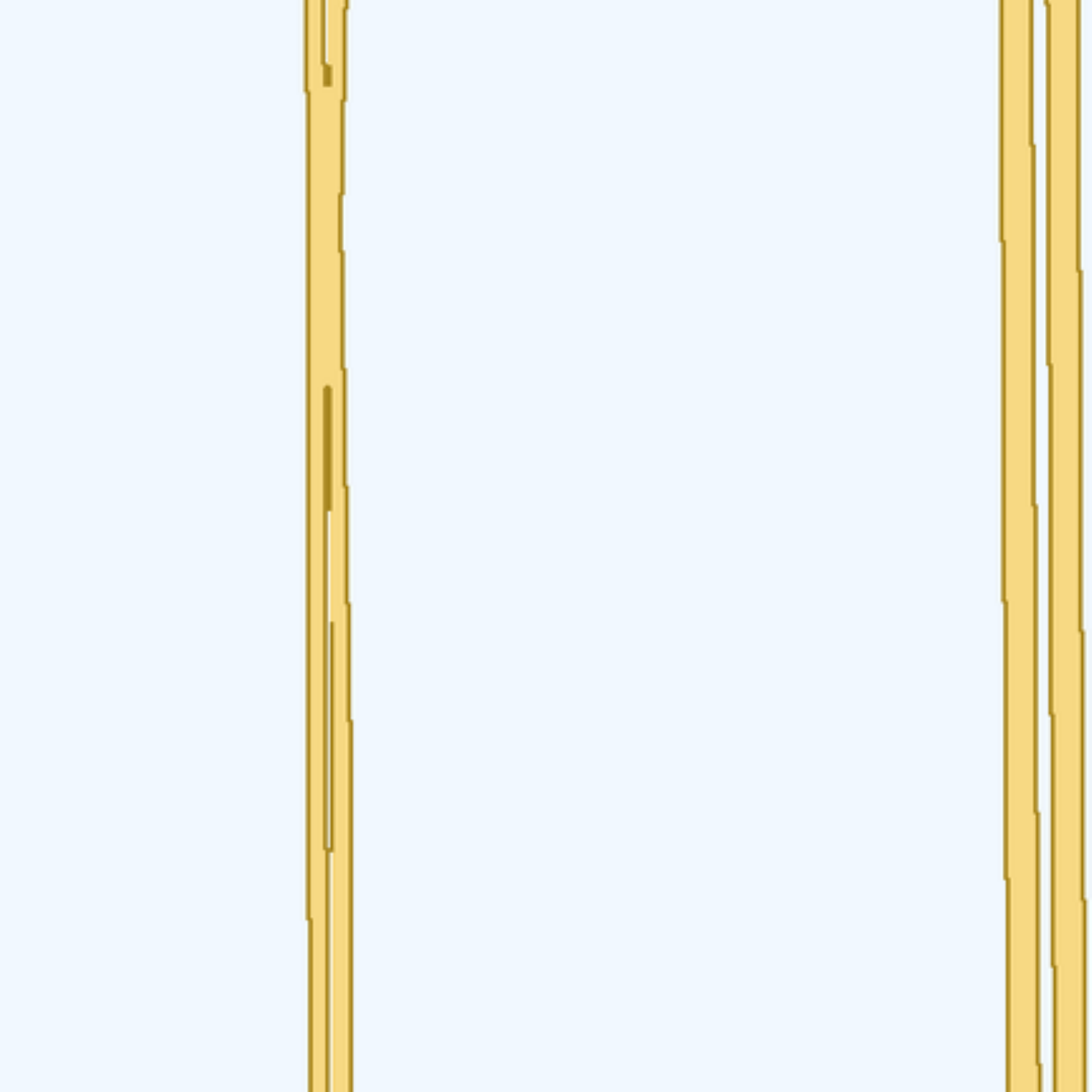}
			& 
\includegraphics[width=0.12\textwidth]{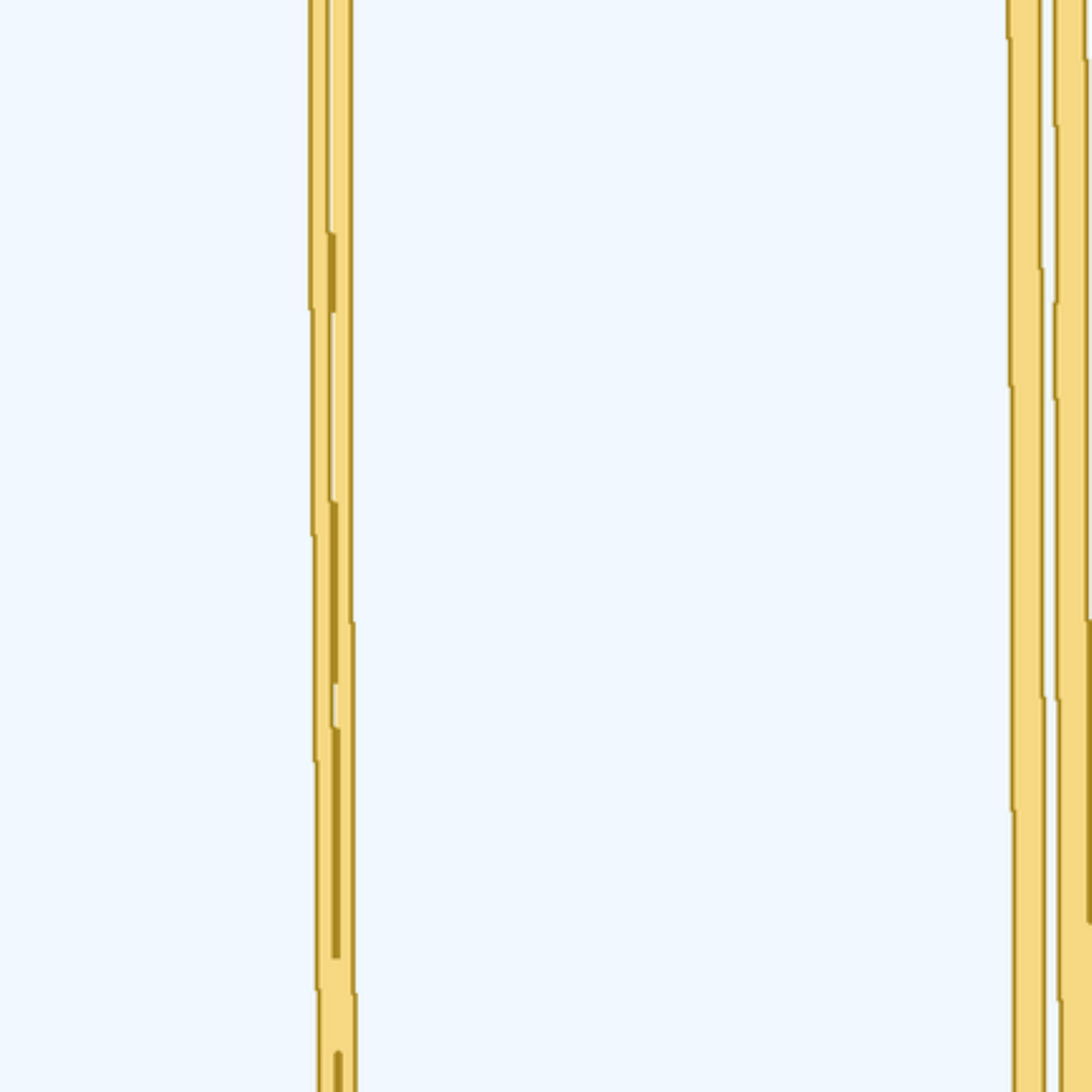}
			& 
\includegraphics[width=0.12\textwidth]{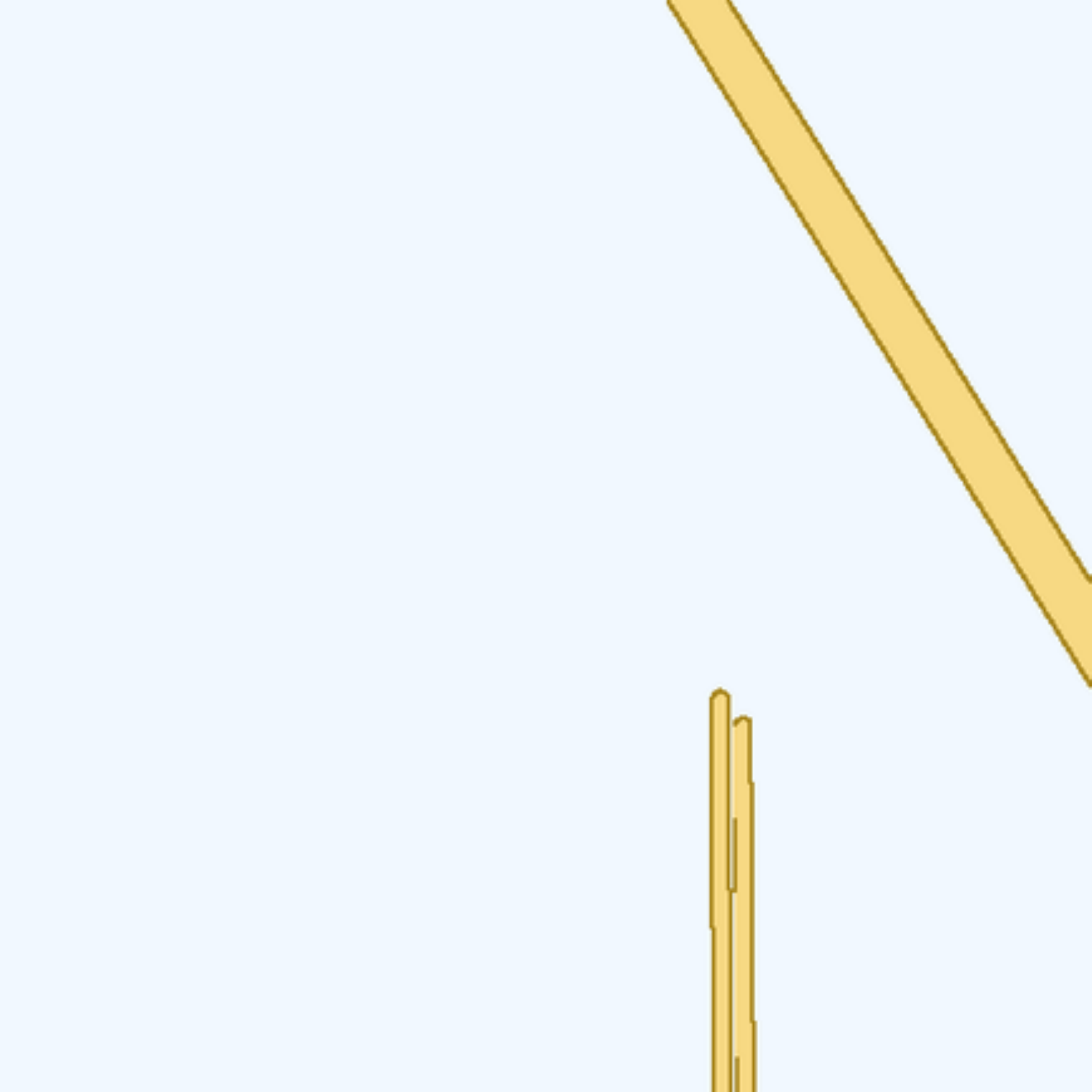}
		\\
Prediction
            &
\includegraphics[width=0.12\textwidth]{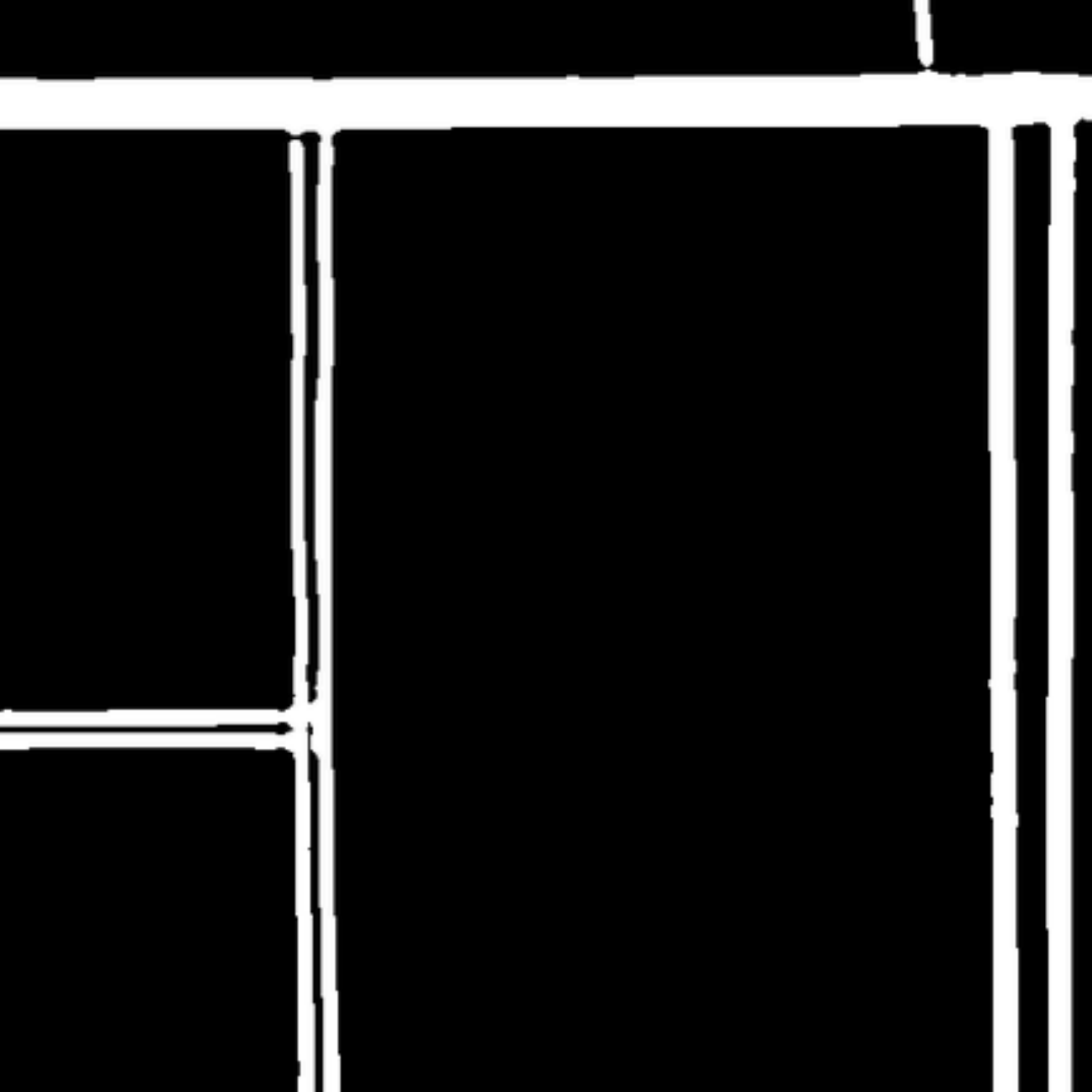}
			& 
\includegraphics[width=0.12\textwidth]{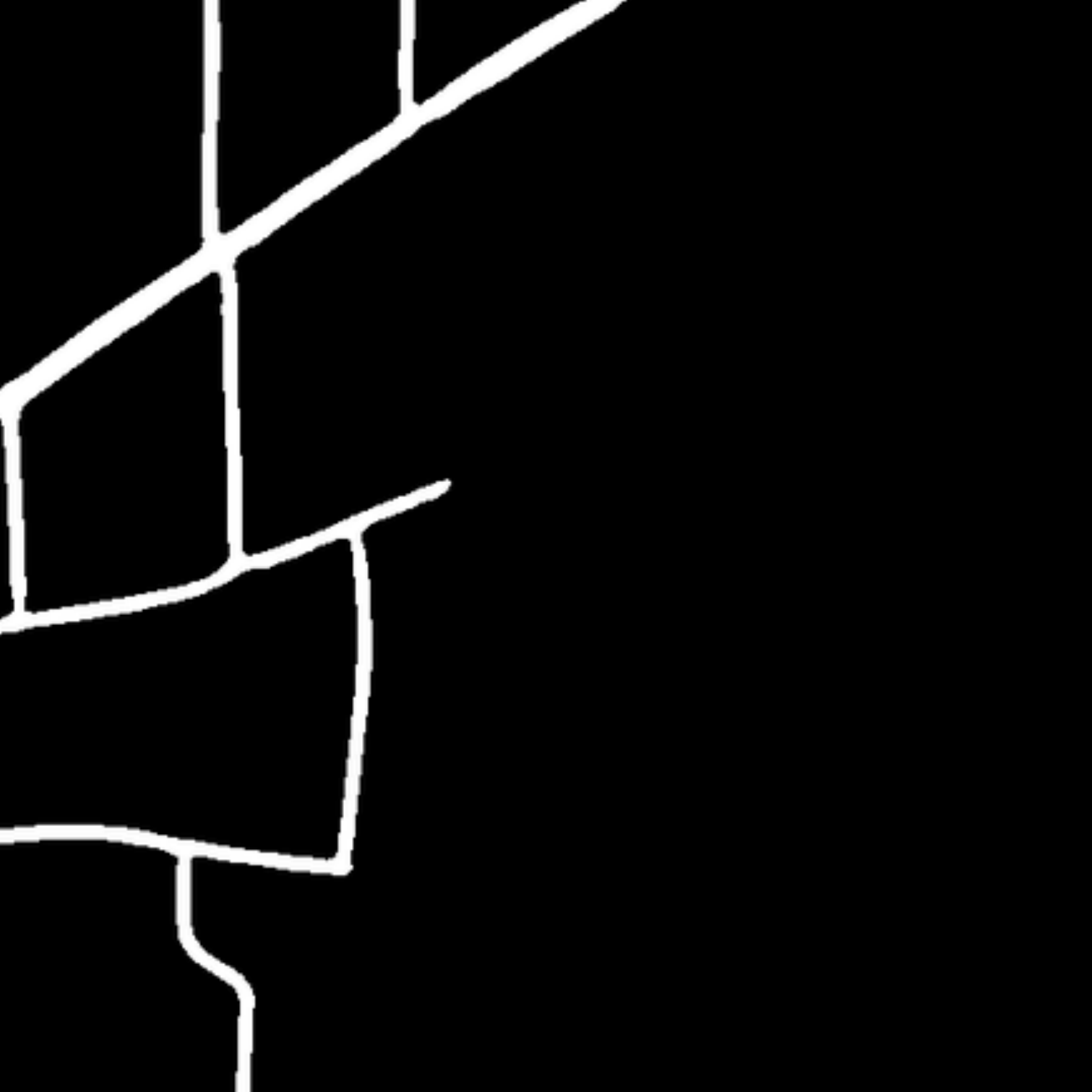}
			& 
\includegraphics[width=0.12\textwidth]{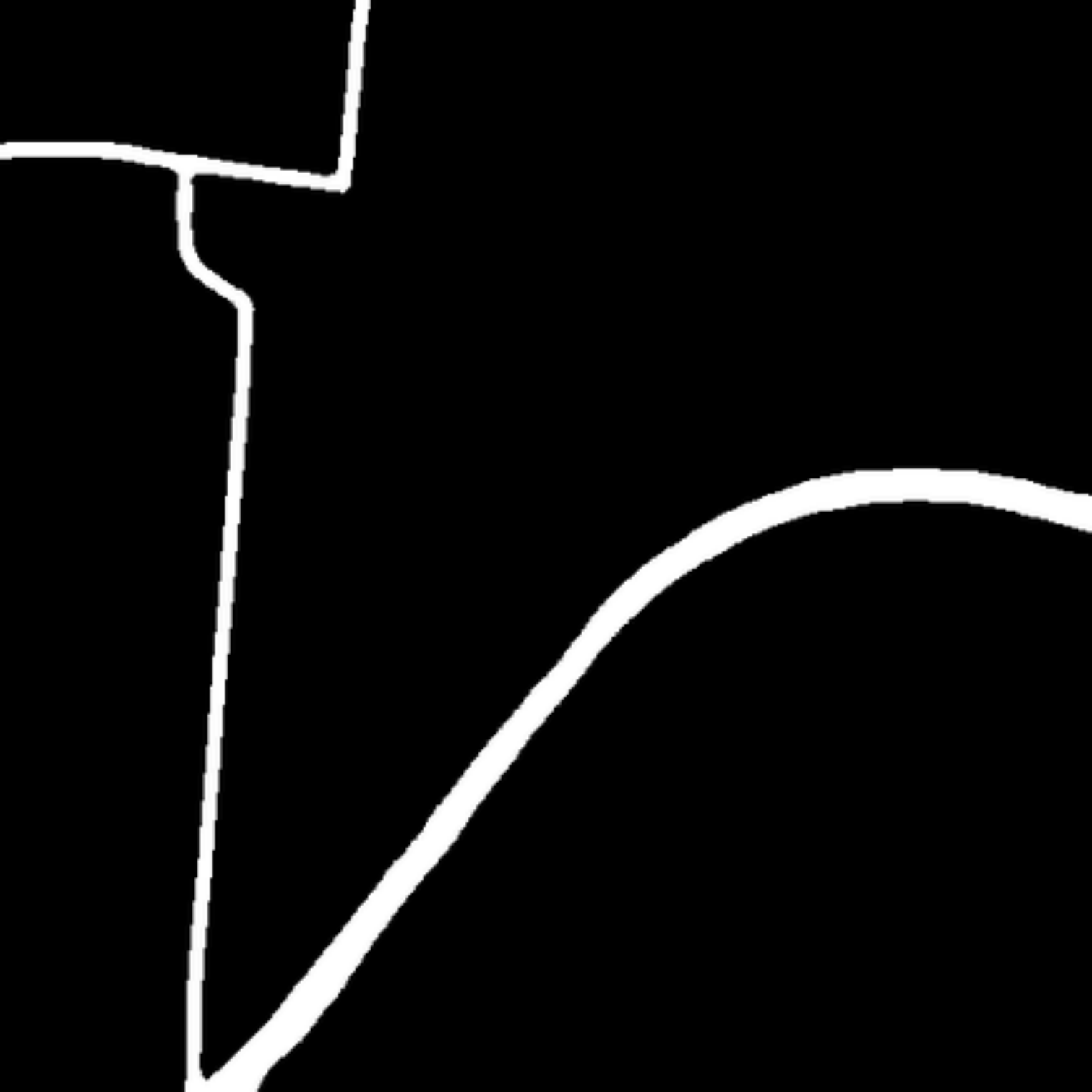}
			& 
\includegraphics[width=0.12\textwidth]{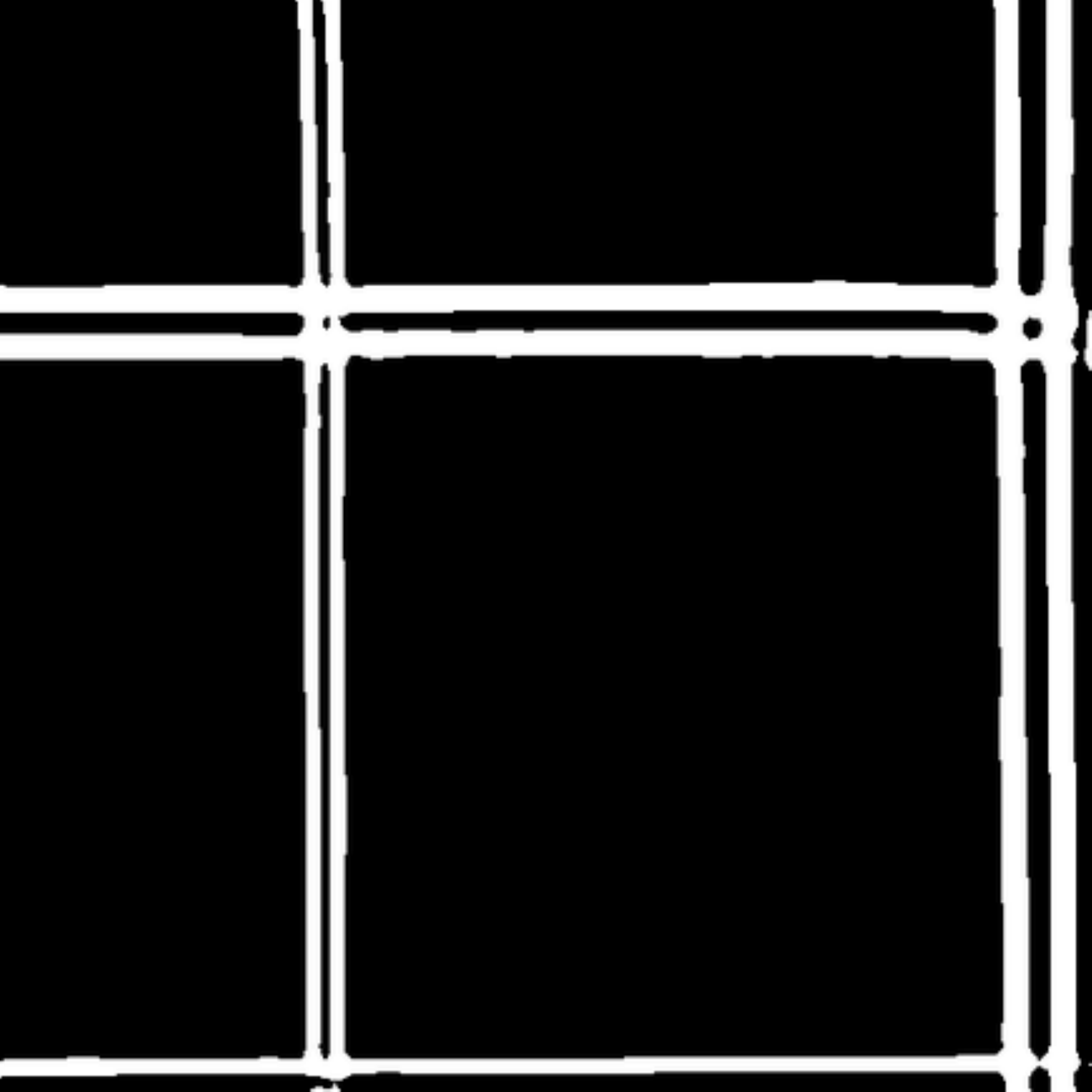}
			& 
\includegraphics[width=0.12\textwidth]{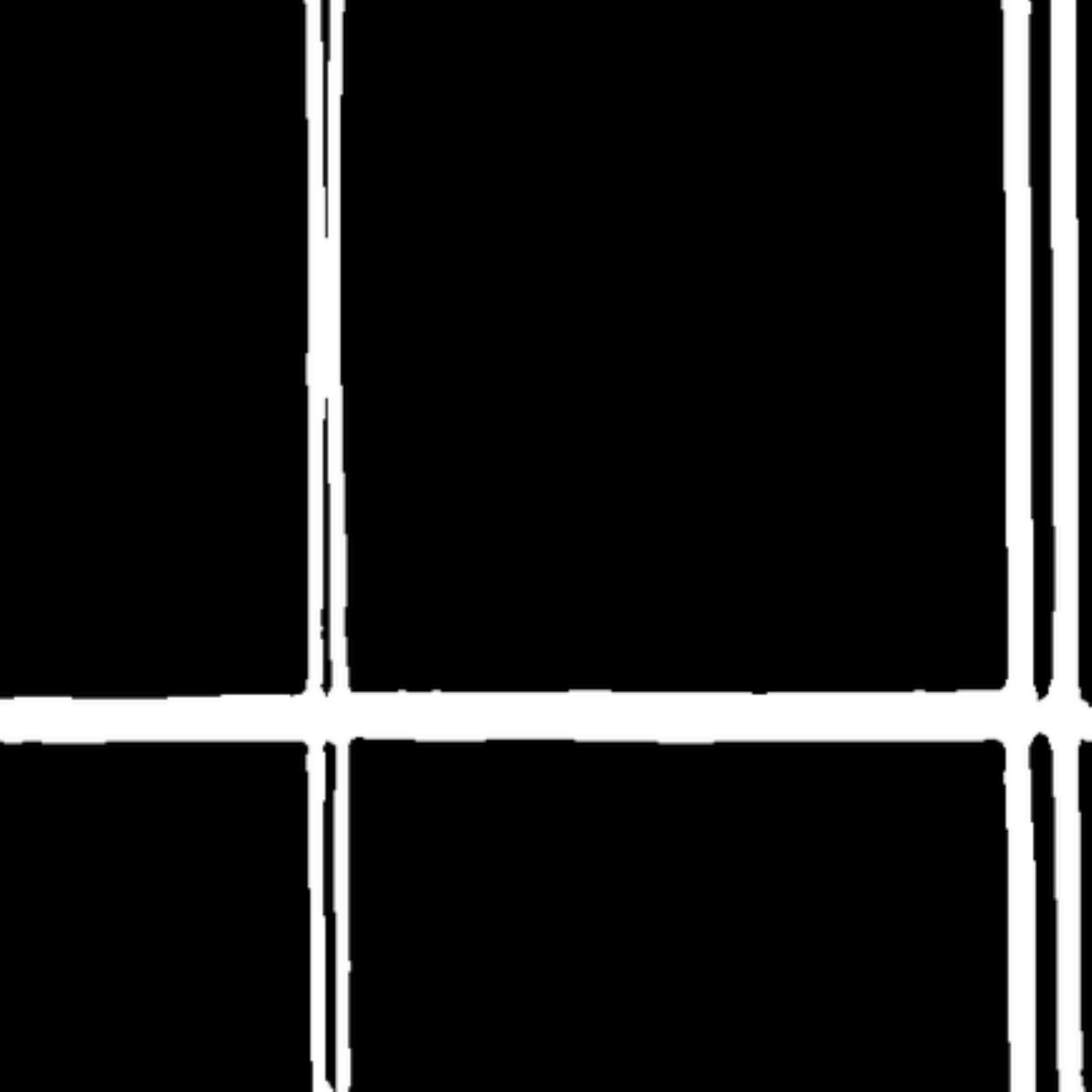}
			& 
\includegraphics[width=0.12\textwidth]{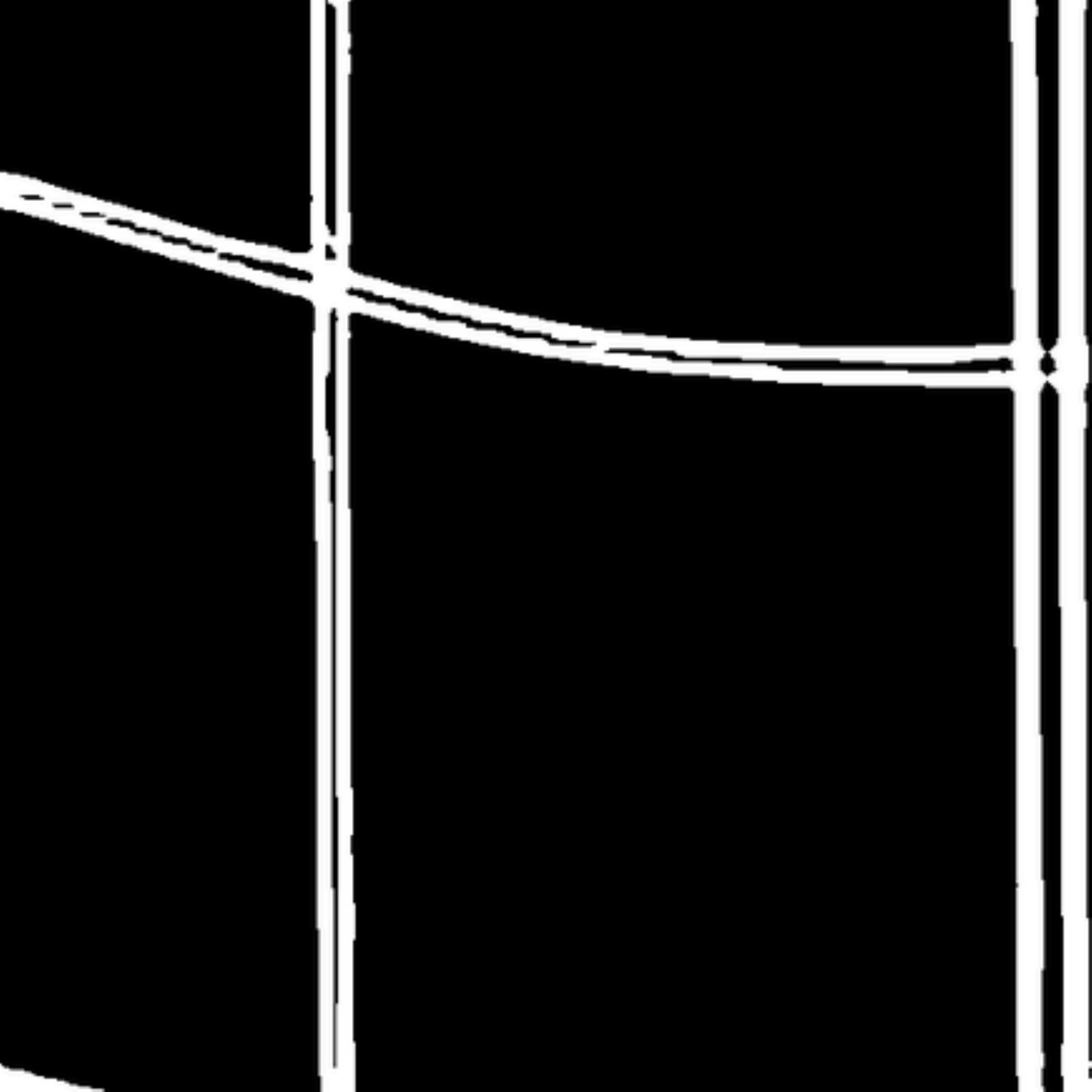}
			& 
\includegraphics[width=0.12\textwidth]{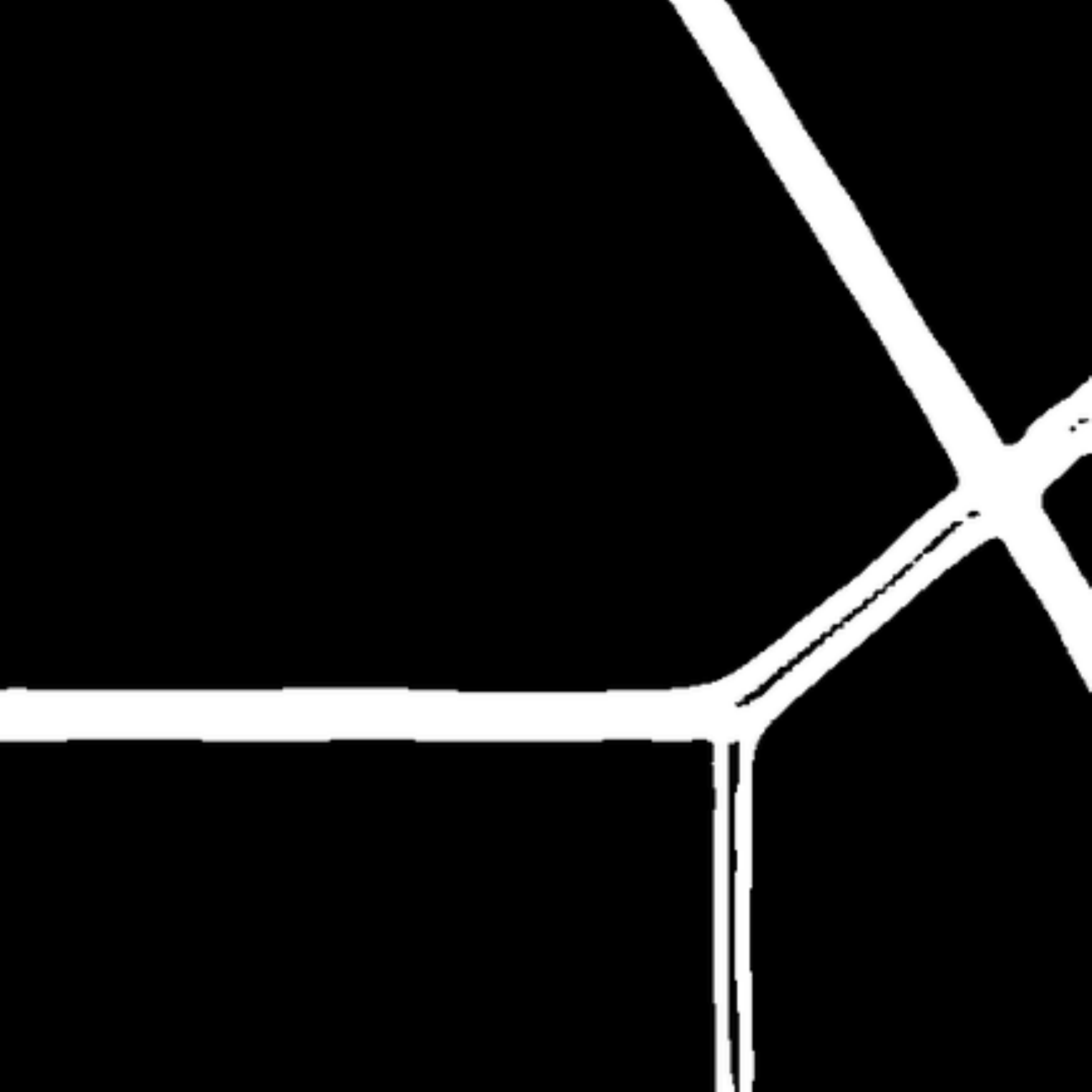}
            \\  
\makecell*[c]{Overlap \\ (changed in red)}
            &
\includegraphics[width=0.12\textwidth]{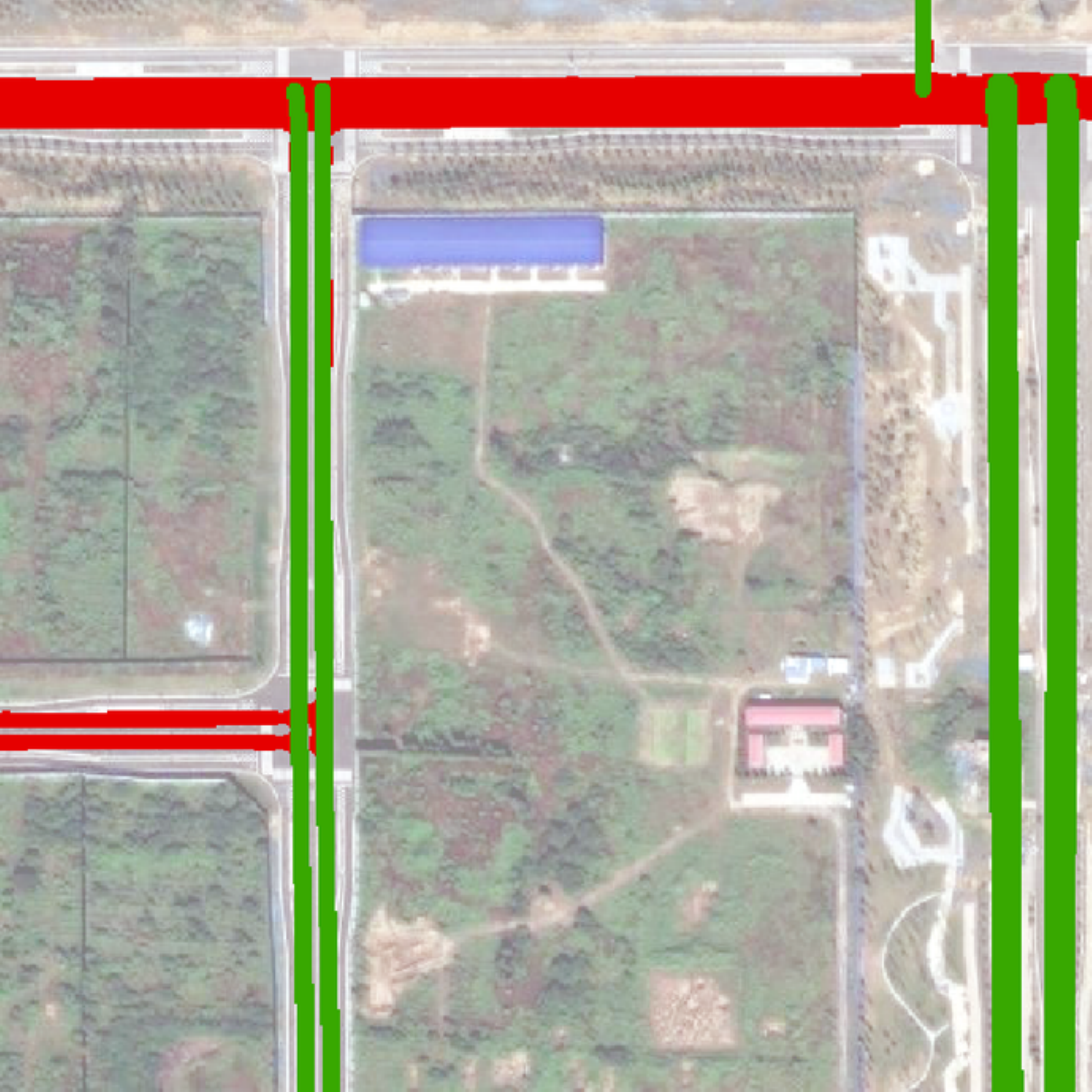}
			& 
\includegraphics[width=0.12\textwidth]{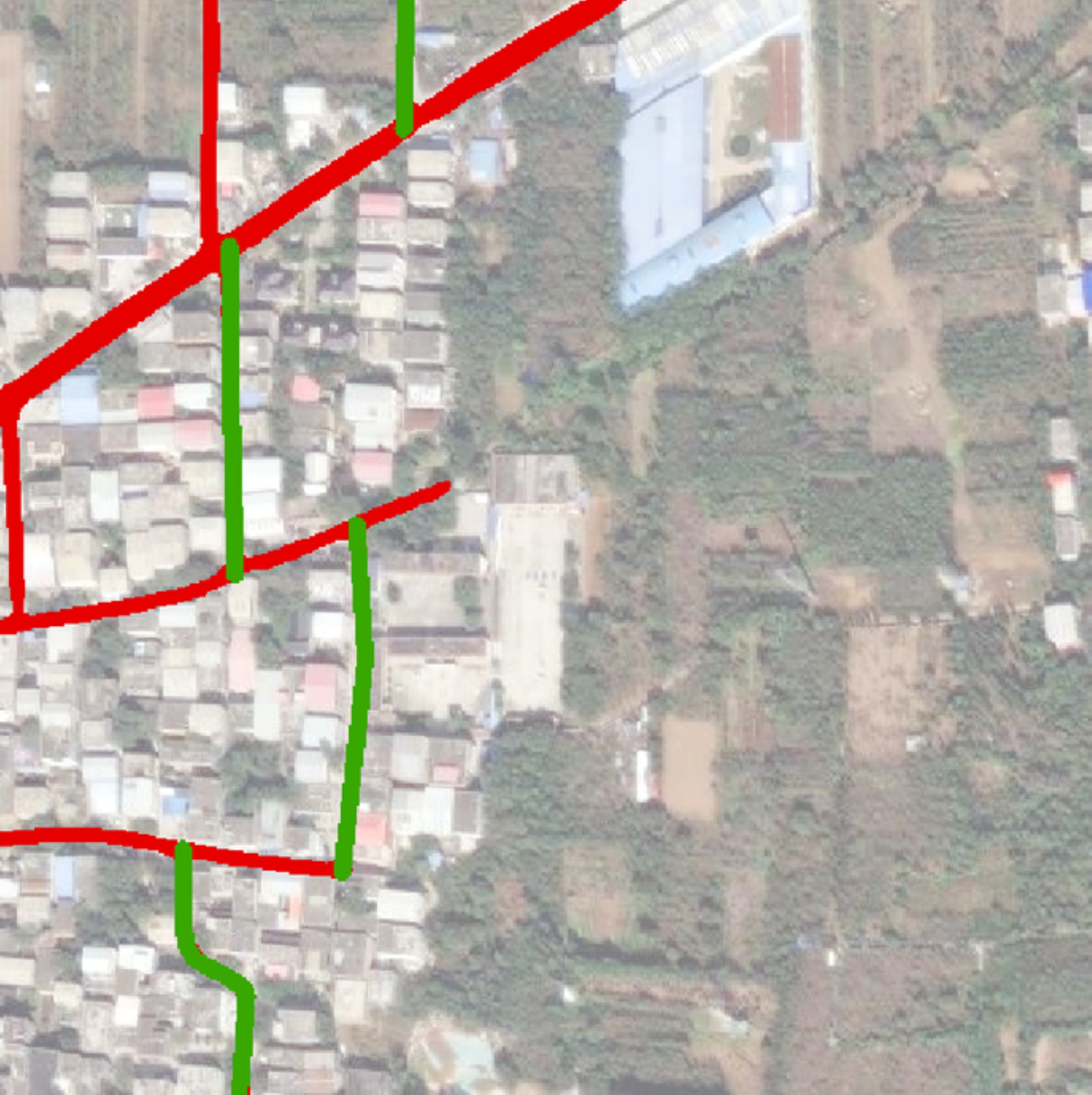}
			& 
\includegraphics[width=0.12\textwidth]{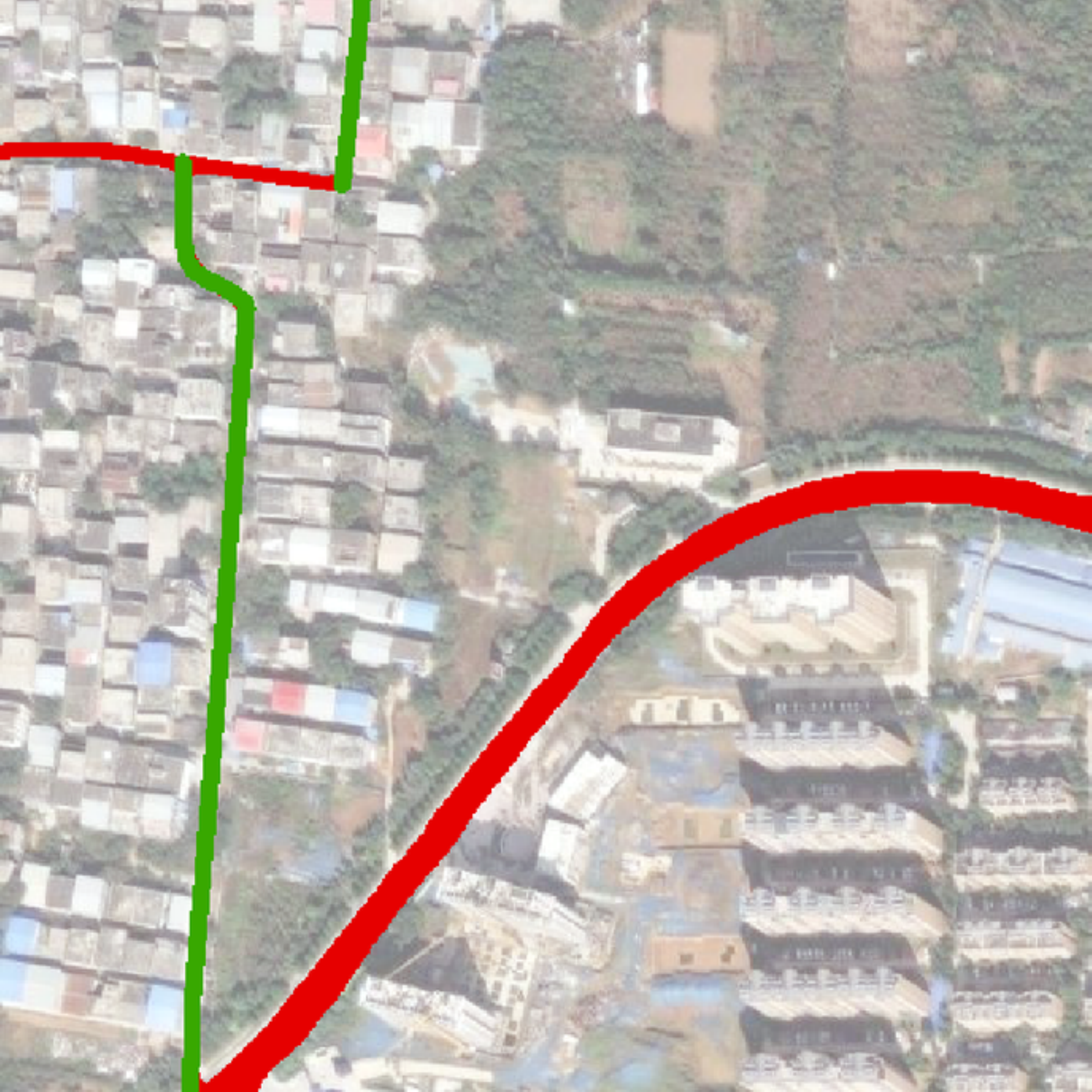}
			& 
\includegraphics[width=0.12\textwidth]{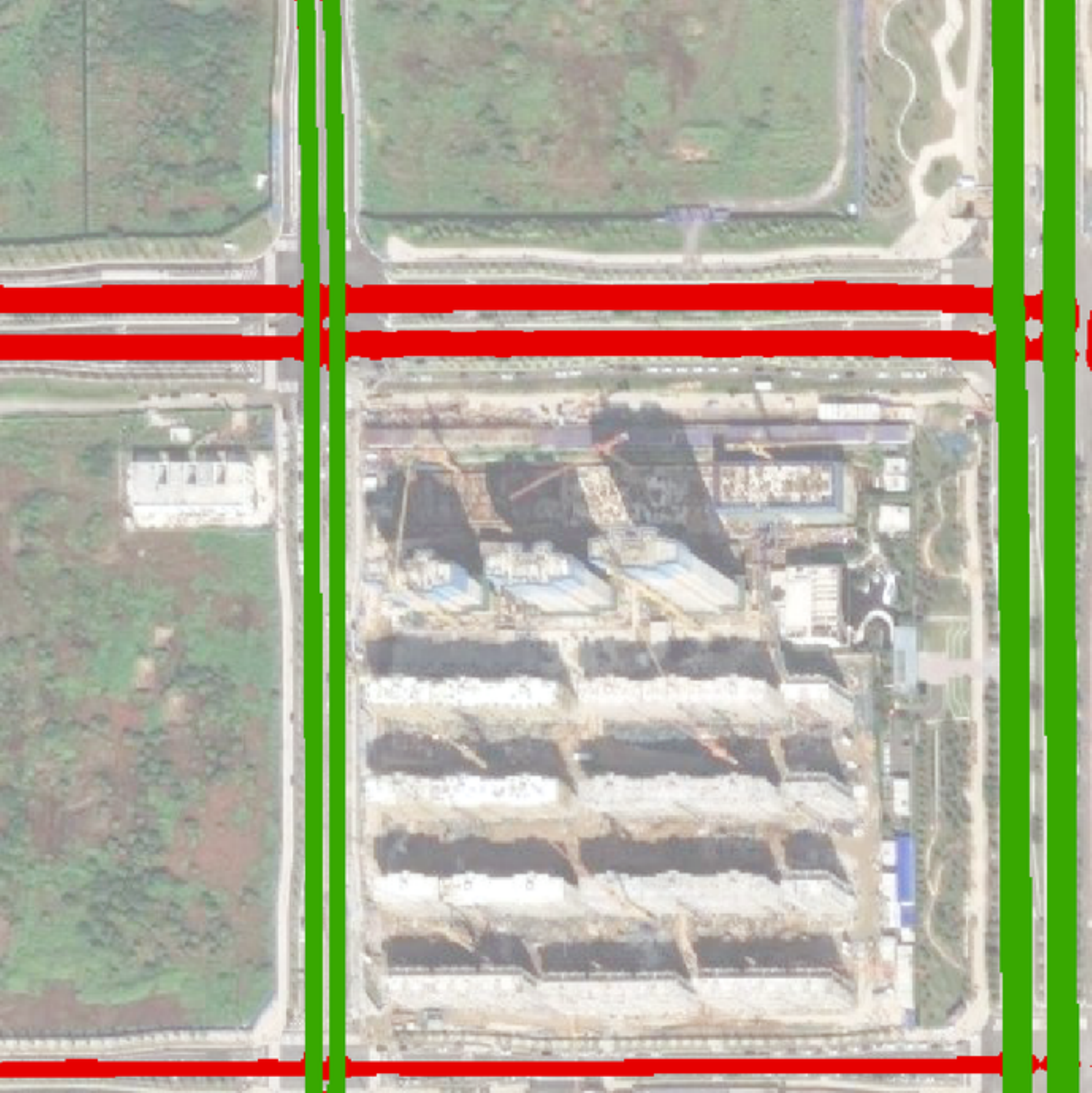}
			& 
\includegraphics[width=0.12\textwidth]{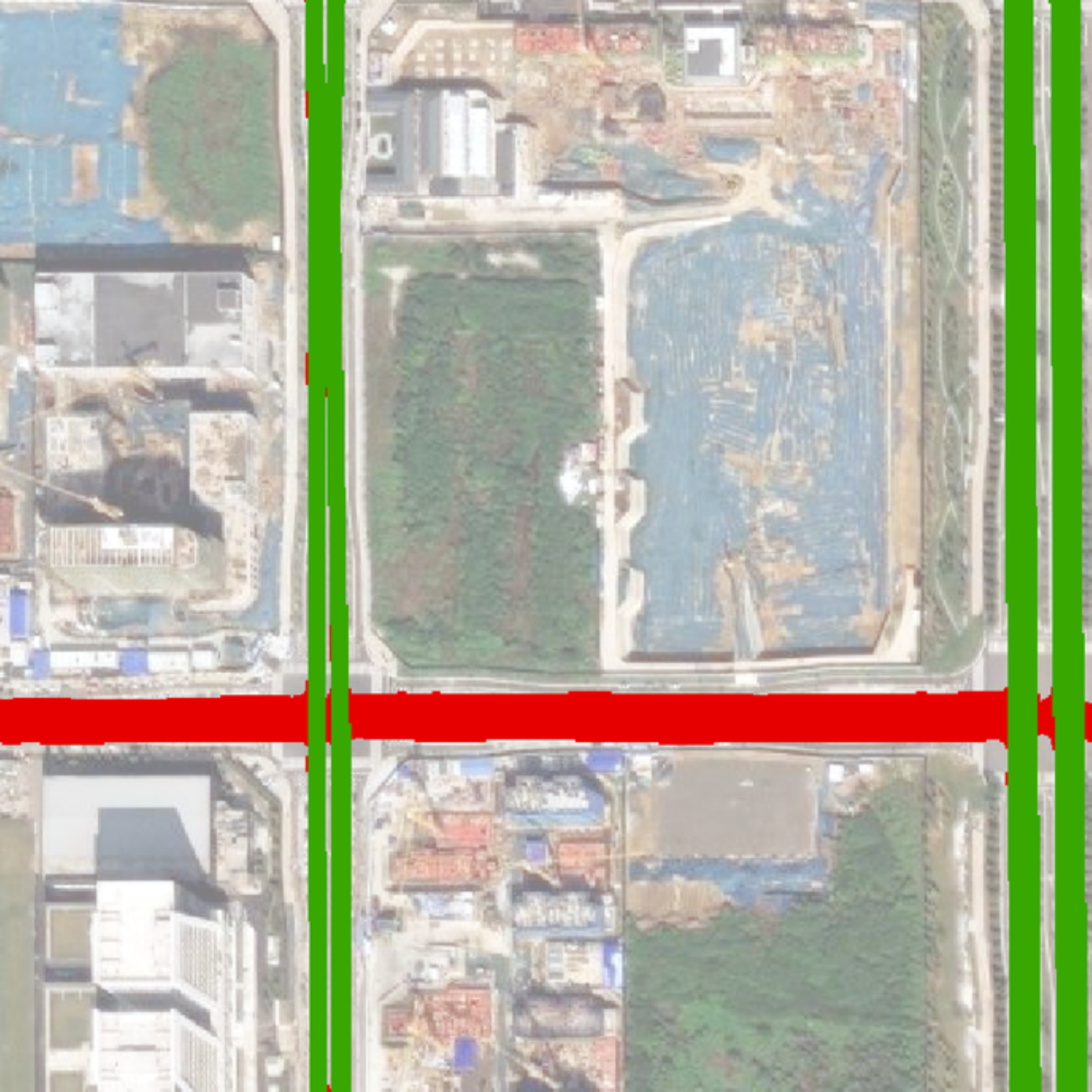}
			& 
\includegraphics[width=0.12\textwidth]{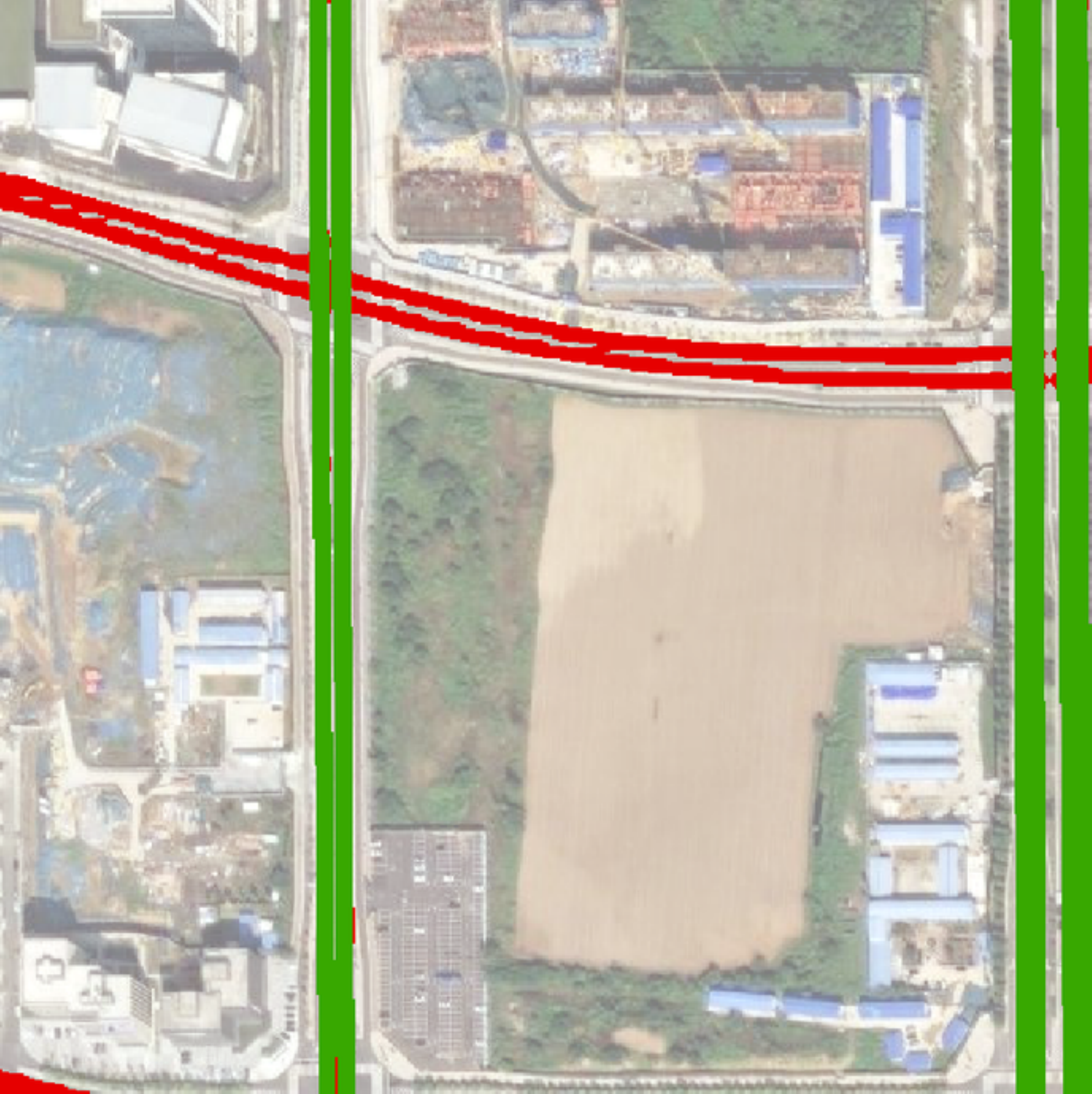}
			& 
\includegraphics[width=0.12\textwidth]{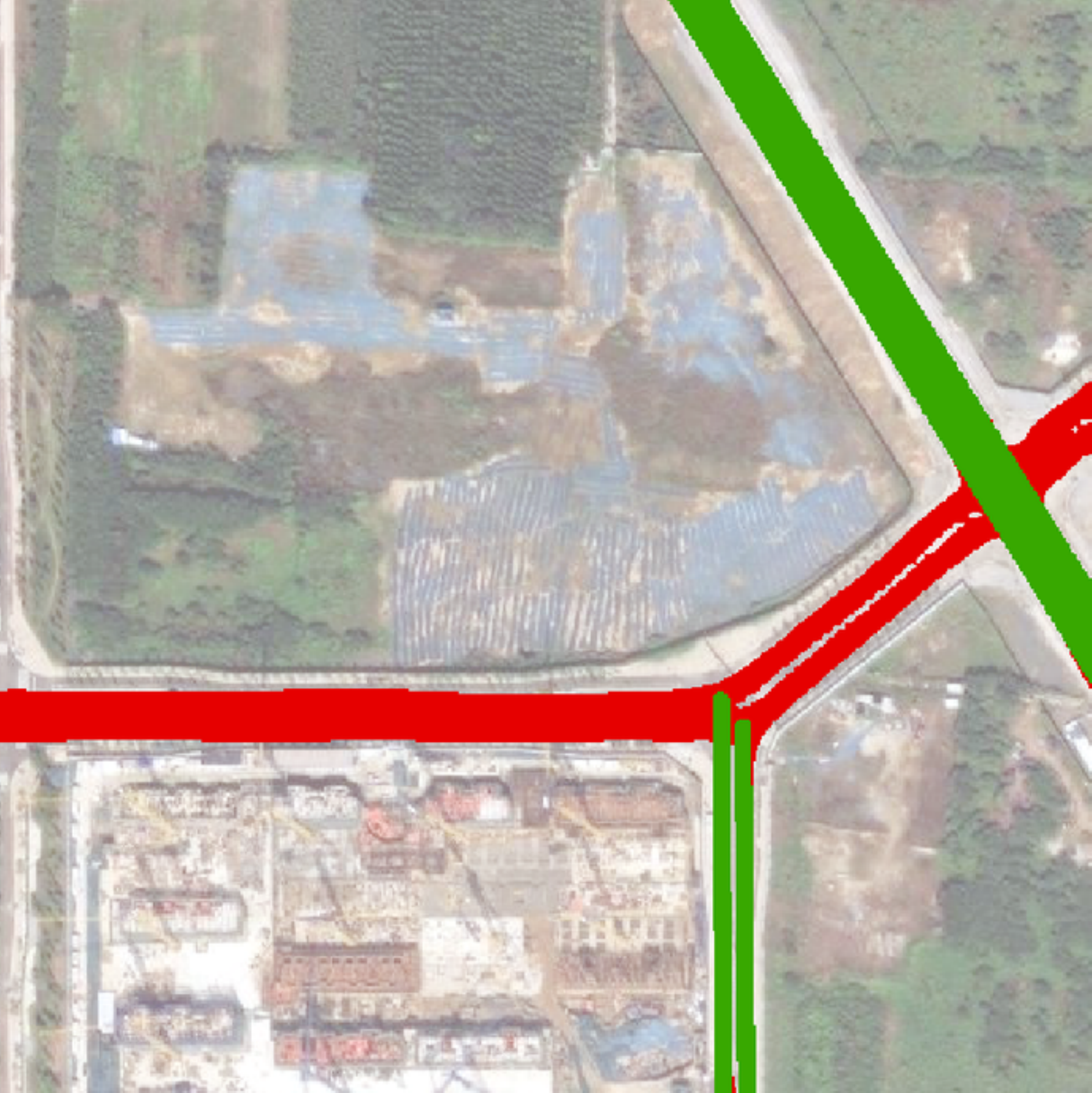}
		\\
 & \multicolumn{7}{c}{(b) Images from Zhengzhou, Henan Province}
\end{tabular}%
\caption{Visual comparison of historical maps and predicted results. Results from GA-Net are more complete and have less road segments. False negatives are marked in red, and false positives are marked in blue.} %
\label{update_results_images} 
\end{figure*}

To validate the effectiveness of the proposed road-updating method in practical applications, we visualized and reported the results using a self-constructed dataset. We first compared the performance of SRUNet with that of other state-of-the-art methods and analyzed the overall road update results on the latest imagery from Nanjing and Zhengzhou, as shown in \figref{update_results_large}. We further compared the road extraction results with a historical map and reported the road updating results in \figref{update_results_images}.

From \figref{update_results_large}, we observed that roads are diverse across cities. Visually, the roads in Nanjing are complex and that the road density is uneven in different areas, whereas the roads in Zhengzhou are more coherent and straight. As shown in \figref{update_results_large}, the images in the left column show the prediction results of the proposed SRUNet, and the images on the right are the close-up views. Intuitively, more true-positive pixels were found to be obtained using SRUNet, and the road fragments were more continuous compared with those from the other methods. A more representative example can be found in the right column, where the results of Adv and S4l are fragmented and a serious missing-classification problem exists. In contrast, the results of ReCo and SRUNet are much more complete, where high-class roads have a higher detection integrity. Moreover, in some special areas such as intersections and viaducts, SRUNet provides clearer prediction details. This illustrates that our method can cope with large-scale imagery and outperforms other SOTA methods; thus, it can be applied in practice. However, the proposed SRUNet showed a poor performance and contained omissions and false predictions in areas where the road features were not significant or were obscured. This indicates that semi-supervised methods cannot fully detect all road areas and have limited reasoning abilities.

The updated and comparative results are shown in \figref{update_results_images}. To verify whether the proposed method can effectively detect the changed road information compared with the that in the historical map, the updated and comparison results are shown in \figref{update_results_images}. Seven $512\times 512$ pixel images were selected from both Nanjing and Zhengzhou. In \figref{update_results_images}, the first row shows the up-to-date remote sensing images, the second row shows the historical road maps from the geodatabase, the third row shows the up-to-date road information extracted by the proposed SRUNet, and the last row shows the overlapping results of historical maps and up-to-date road information, where green and red represent the unchanged and changed road areas, respectively. The results show that our model can adaptively maintain prior knowledge of unchanged areas and detect changed areas to update roads. The unchanged areas of the predicted results remain the same as those in the historical map, which means that our model can perceive the historical information and retain the correct information. However, the increased roads have good continuity with the existing roads and the inaccurate roads are modified (as shown in the third row of \figref{update_results_images} (a); the parallel roads on the right are not connected to the road above; however, they are modified in the updated road map), which means that our model can adaptively update the historical map information. Moreover, the results of our model had accurate and clear boundaries, possibly because of the design of BEM and semi-supervised training methods. Finally, it is worth mentioning that omission occurs in the urban–rural fringe and rural areas, and we presume that the improvement in SRUNet is mainly attributed to the introduction of historical information. Thus, historical information can effectively enhance the reasoning ability for unlabeled data. Thus, for regions with little prior knowledge, the model easily misses typical road regions. 

Considering that the road is stored in a database in the form of vector data in practical application, we further conducted postprocessing operations, including removal of isolated spots, vectorization, topology construction, consistency processing, and so on. The results are shown in \figref{update_results_vec}, where the thick green line represents the old data and the thin red line represents the latest state of the road network. For unchanged areas, the updated road data maintained good consistency with the historical data. For the changed areas, the increased roads continued and maintained good continuity with the other roads. In general, SRUNet can obtain stable prediction results for complex images; therefore, the prediction results are up-to-date, reliable, and suitable for a wide range of road renewal tasks.

\section{Conclusion}\label{5}
To address the problem of updating road information from geographical databases, this study proposes a solution based on semi-supervised learning and proposes a road update method that considers the characteristics of road edges. This method takes historical database data as the object to be updated and uses new images and a small amount of manual annotation information as the update information to semi-automatically extract road patches. First, the two types of data are processed simultaneously and independently through a parallel branching structure, and the ReCo model is introduced to improve the representation ability of the model. Second, an edge feature enhancement module is designed to improve the model performance in terms of road continuity and integrity. Finally, the predicted patches are further fine-tuned and corrected based on the output prediction results to obtain a better prediction output. Experiments show that the proposed method outperforms existing semi-supervised methods in terms of IoU values for road extraction, and that the proposed modules can improve the model performance to some degree. The model obtained ideal road detection results on the experimental datasets of Zhengzhou and Nanjing, demonstrating the good generalization capability of the method for actual operational data. Additionally, the method considers the details and morphology of roads, facilitating subsequent vectorization and road data warehousing operations.

During the experimental process, the model did not perform well in the case of poor historical data quality and limited labeled data, particularly when the labeled data had a significant impact on the model results. Two aspects of subsequent improvements were considered. One is to improve the learning ability of the unsupervised part of the model on road features and introduce pre-training models such as MAE to improve the generalization ability of the model on unlabeled data. The second is to further explore the relationship of the quantity and quality of labeled data with the prediction results. Thus, manual intervention can be further reduced in update tasks by optimizing the quality of labeled data and existing labeling modes.

\section{Acknowledgement}
This study was supported by the National Natural Science Foundation of China (42101458, 41801388, and 42101455) and Fund Project of Zhongyuan Scholar of Henan Province (202101510001).

\section{Declaration of Competing Interest}
The authors declare that they have no competing financial interests or personal relationships that may have influenced the work reported in this study.

\bibliography{refs.bib}

\end{document}